\pdfoutput=1

\documentclass[phd,ilcc,twoside, logo]{infthesis}
\usepackage[utf8]{inputenc}
\usepackage{times}
\usepackage{latexsym}
\usepackage{microtype}
\usepackage[table]{xcolor}
\usepackage[export]{adjustbox}

\usepackage{palatino}

\usepackage{mathptmx}

\usepackage[np, autolanguage]{numprint}
\usepackage{booktabs}
\usepackage{xspace}
\usepackage{pifont}
\usepackage{arydshln}
\usepackage{url}
\usepackage{helvet}
\usepackage{censor}

\usepackage{csquotes}
\usepackage{epigraph}

\usepackage{algorithm}
\usepackage{algorithmic}

\usepackage{multirow}
\usepackage{enumitem} %
\usepackage{arydshln} %
\usepackage{booktabs} %
\usepackage{rotating}
\usepackage{tablefootnote} %
\usepackage{nicematrix}

\usepackage[mode=buildnew]{standalone}
\usepackage{tikz}
\usetikzlibrary{shapes,positioning,fit,calc, matrix}

\usepackage{graphicx}
\usepackage{caption}
\usepackage{subcaption}
\usepackage{wrapfig}

\usetikzlibrary{arrows}

\usepackage{multicol}
\usepackage{float}
\usepackage{environ}
\usepackage{scalefnt}
\usepackage{nicefrac}       %
\usepackage{ifthen}
\usepackage{marginnote}
\usepackage{tipa}
\usepackage{framed}
\usepackage{soul}
\usepackage{mathtools}
\usepackage{amsmath,amsfonts,amssymb,amsthm,bm}

\soulregister\cite7
\soulregister\citep7
\soulregister\citet7

\soulregister\ref7

\colorlet{shadecolor}{blue!20}

\usepackage[style=apa,backend=biber,minbibnames=3,natbib=true]{biblatex}
\bibliography{tidy.bib}

\usepackage[detect-weight=true, detect-family=true]{siunitx}
\usepackage{bm}
\usepackage{qtree}
\usepackage{tikz}
\usepackage{arydshln}
\usepackage{tabularx}

\usepackage{makecell}

\usepackage[T1]{fontenc}
\usepackage{pgfplots}
\pgfplotsset{compat=1.18} 

\usepackage{hyperref}
\definecolor{darkgreen}{RGB}{23, 59, 36}
\hypersetup{colorlinks=true,citecolor=darkgreen, linkcolor=darkgreen, urlcolor=darkgreen, bookmarks=false}
\usepackage{lmodern}

\newcommand*\circled[1]{\tikz[baseline=(char.base)]{
            \node[shape=circle,dashed,draw,inner sep=2pt] (char) {#1};}}

\usepackage{amsmath,amsfonts,bm}

\def\eqref#1{equation~\ref{#1}}

\def\1{\bm{1}}

\DeclareMathAlphabet{\mathsfit}{\encodingdefault}{\sfdefault}{m}{sl}
\SetMathAlphabet{\mathsfit}{bold}{\encodingdefault}{\sfdefault}{bx}{n}

\newcommand{\param}{\boldsymbol \theta}
\newcommand{\update}{\text{Update}}

\newcommand{\loss}{\mathcal{L}}

\newcommand{\sBatch}{\mathcal{B}}

\newcommand{\comment}[1]{}

\def\|#1|{\mathid{#1}}
\newcommand{\mathid}[1]{\ensuremath{\mathit{#1}}}
\def\<#1>{\codeid{#1}}
\protected\def\codeid#1{\ifmmode{\mbox{\smaller\ttfamily{#1}}}\else{\ttfamily
		#1}\fi}

\usepackage[detect-weight=true, detect-family=true]{siunitx}
\usepackage{bm}

\makeatletter
\newsavebox{\measure@tikzpicture}
\NewEnviron{scaletikzpicturetowidth}[1]{%
  \def\tikz@width{#1}%
  \begin{lrbox}{\measure@tikzpicture}%
  \BODY
  \end{lrbox}%
  \pgfmathparse{#1/\wd\measure@tikzpicture}%
  \BODY
}

\tikzset{every picture/.style={font issue=\footnotesize},
         font issue/.style={execute at begin picture={#1\selectfont}}
        }

\newcommand{\mbig}{$large$ }
\newcommand{\mmid}{$medium$ }
\newcommand{\msmall}{$small$ }

\newcommand{\round}[1]{\num[round-mode=places,round-precision=2]{#1}}
\newcommand{\longround}[1]{\num[round-mode=places,round-precision=4]{#1}}

\newcommand{\acc}[2]{\round{#1}&\ifthenelse{\equal{#2}{}}{}{\tiny ${\scriptstyle \pm}$\round{#2}}}
\newcommand{\ac}[1]{\longround{#1}}

\newcommand{\graycellcolor}{\cellcolor{gray!25}}

\newcommand{\gacc}[2]{\graycellcolor\round{#1}&\graycellcolor\ifthenelse{\equal{#2}{}}{}{\tiny ${\scriptstyle \pm}$\round{#2}}}

\newcommand{\ACC}[2]{\textbf{\round{#1}}&\ifthenelse{\equal{#2}{}}{}{\textbf{\tiny ${\scriptstyle \pm}$\round{#2}}}}

\newcommand{\GACC}[2]{\graycellcolor\textbf{\round{#1}}&\graycellcolor\ifthenelse{\equal{#2}{}}{}{\textbf{\tiny ${\scriptstyle \pm}$\round{#2}}}}

\newcommand{\mc}[2]{\multicolumn{#1}{c}{\textbf{#2}}}

\newcommand{\origtrain}{\mathcal{T}}

\newcommand{\mtrainBatch}{\mathcal{B}_{t}}
\newcommand{\mtestBatch}{\mathcal{B}_{g}}

\newcommand{\sround}[1]{\num[round-mode=places,round-precision=4]{#1}}

\newcommand{\noserif}[1]{{\small \fontfamily{phv}\selectfont #1}}
\newcommand{\nosbold}[1]{\textbf{ \fontfamily{phv}\selectfont #1}}
\newcommand{\aaacc}[2]{\small \makesans \round{#1} \small ${\footnotesize \pm}$\sround{#2}}

\newcommand{\aagacc}[2]{\graycellcolor\round{#1}&\graycellcolor\ifthenelse{\equal{#2}{}}{}{\tiny ${\scriptstyle \pm}$\round{#2}}}

\newcommand{\aaACC}[2]{\textbf{\round{#1}}&\ifthenelse{\equal{#2}{}}{}{\textbf{\tiny ${\scriptstyle \pm}$\round{#2}}}}

\usepackage{color}

\newcolumntype{Y}{>{\centering\arraybackslash}X}

\newcommand{\drawline}[0]{\par\noindent\rule{\textwidth}{0.4pt}\\[3ex]}
\newcommand{\thinline}[0]{\par\noindent\rule{\textwidth}{0.4pt}\\}

\title{Information Structure in Mappings: An Approach to Learning, Representation, and Generalisation}

\author{Henry Coxe Conklin}

\abstract{

\noindent Mappings relate two different spaces, transforming things of one kind into another; they are ubiquitous across the sciences and the world around us. Mathematical functions map between a domain and range, digital phone systems map waveforms to binaries, ribosomes map DNA sequences to proteins as part of a larger mapping between genotypes and phenotypes. Telegram operators map back and forth between text and morse code, artificial neural networks map inputs to vector representations, and language allows us to map our thoughts to sentences that express them. The structure of these mappings differs widely, having conformed either to the selection pressures of their environment or the concerns of their architects.

Despite the remarkable success of large large-scale neural networks in recent years, we still lack unified notation for thinking about and describing their representational spaces. We lack methods to reliably describe how their representations are structured, how that structure emerges over training, and what kinds of structures are desirable. This thesis introduces quantitative methods for identifying systematic structure in mappings between spaces, and leverages them to understand how deep-learning models learn to represent information, what representational structures drive generalisation, and how design decisions condition the structures that emerge. To do this I identify basic kinds of system-level structures present in a mapping, along with information theoretic quantifications of each of them. I use these to analyse learning, structure, and generalisation across multi-agent reinforcement learning models, sequence-to-sequence models trained on a single task, models trained with meta-learning objectives, and Large Language Models. I also introduce a novel, performant, approach to estimating the entropy of vector space, that allows this analysis to be applied to models ranging in size from 1 million to 12 billion parameters. 
 
The experiments here work to shed light on how large-scale distributed models of cognition learn, while allowing us to draw parallels between those systems and their human analogs. They show how the structures of language and the constraints that give rise to them in many ways parallel the kinds of structures that drive performance of contemporary neural networks.

}

\begin{document}

\begin{preliminary}
    \maketitle
	\makesans
    \begin{laysummary}
    The world is made up of systems that convert one type of information into another - much like a translator changes words from one language to another. Your phone turns your voice into digital signals, your body turns genetic code into physical traits, and your brain turns thoughts into spoken words. These transformations can be found everywhere in nature and technology, each shaped by different needs and purposes. 

\vspace{3mm}

\noindent In recent years, artificial intelligence systems called neural networks have become incredibly powerful at processing information, but we don't have ways to understand how they organise and structure this information internally. In part because they represent it as huge lists of numbers, that we as humans have trouble reasoning about. This research develops new mathematical tools to  look inside these AI systems and understand how they learn to represent information. 

\vspace{3mm}

\noindent These tools work by looking for structure in the relationship between what we show the AI system, and the numbers it transforms them to. The tools look for the same kinds of structure we see in the way human language transforms our thoughts into the things we say. 

\vspace{3mm}

\noindent By applying these tools to various AI systems - from those that play games together to those that process language - this work reveals patterns in how these systems organise information and which patterns help them solve new problems. The research also introduces a new method to measure how efficiently these systems store information, which works on both small and enormous AI models. 

\vspace{3mm}

\noindent Apart from helping us better understand artificial intelligence, these findings also show parallels between how AI systems and human language structures information. This suggests there may be some universal principles in how both natural and artificial systems learn to represent information effectively.

    \end{laysummary} 

    \begin{acknowledgements}
    {\makesans

\noindent I have not done this of my own accord, and so have people in need of thanks.

\vspace{3mm}

\noindent Kenny Smith my primary supervisor, to whom I owe some substantial debt of gratitude. Thank you for agreeing to supervise my undergraduate dissertation in 2018, and for suggesting I apply for a PhD position a year later, and for making the time to meet with me for an hour every week for the past 5 years. You taught me how to talk about research, and how to pin the big picture down to something testable. I would not have pursued any of this were it not for you, so thank you. Ivan Titov (my secondary supervisor), thank you for encouraging my interest in information theory, for always welcoming me into your group meetings, for introducing me to Bailin, and for supervising our work together. Paul Smolensky (my internship supervisor), thank you for teaching me the elegance of an outer product.

\vspace{3mm}

\noindent To my PhD Friends; George Carter for their appreciation of Edinburgh's lesser known parks and sad smudges. Tom Hosking, for being such an excellent discussion partner. Seraphina Goldfarb-Tarrant, for the time spent by any and all fires. Annie Holtz, for your appreciation of bakeries, coffee, and bad contemporary art. Laurie Burchell for the many laps of Inverleith park. Rohit Saxena, for not panicking as I cut across 8 lanes of traffic. Matthias Lindemann, for putting up with questions about grammars and linear algebra. Bailin Wang for putting up with questions about meta learning and tensor2struct. Verna Dankers, for putting up with questions about interpretability and for giving excellent feedback. Stella Frank, for helping me get started. Shira and Tomer, for dinner. I'd also like to thank a number of other people for making the PhD time what it was; Marc Meisezahl, Vlad Nedelcu, Elizabeth Pankratz, Aislinn Keogh, Maisy Hallam, Juan Guerrero Montero, Lauren Fletcher, Irene Winther, Dan Wells, Paul Soulos, Anna Kapron-King, Tamar Johnson, Marianne de heer Kloots, Nik 

\vspace{3mm}

\noindent My Edinburgh family; thank you Abby Jackson for the film nights, and politics --- and Roddy McDermott for the tunes. Mel Philips and Craig Methven, thank you for the art, and the art of the pal smash. Anna Stewart and Marty McLennan, thank you for fringing. Celia Dugua for the aesthetics --- Roxy Cook for dinner. Pedro Leandro and Macleod Stephen for discussion. Izzy Moulder and Caz Elms for the fortress. Eric and Josie Geistfeld for scroobin.

\vspace{6mm}

\noindent Amy Sheahan for seeing friendship as the serious business it is. Kat Knoerl, go team. 

\vspace{3mm}

\noindent And to my family; thank you Dana Catharine and Susan Buckley for guiding me through the seas of moral turpitude. James Kirby Rogers, thank you for introducing me to Swensen's, contemporary dance, and the Roy St. Parking structure. AKC, for your relentless enthusiasm for my continued existence. PAK \& PMC --- none of this would have happened without you.
}

    \end{acknowledgements}

    \standarddeclaration

    \dedication{\makesans for paul.\\ \scriptsize{and everyone else who left the party early}}

    \tableofcontents 

\end{preliminary}

\chapter{How to Represent Information}
\label{chapt:intro}

{\makesans
\emph{
in Mappings, Language, \& Artificial Neural Networks} \\[1em]
}

{\makesans
\begin{quote}
\censor{[.........................................................]} to count leaves is not less meaningful than to count the stars \censor{[.................................]} It would help them to know whether the world is finite. I discovered one tree that is finite. \censor{[.........................]}\flushright{-David Ignatow}
\end{quote}
}

\drawline

\noindent Mappings relate two different spaces, transforming things of one kind into another; they are ubiquitous across the sciences and the world around us. Mathematical functions map between a domain and range, digital phone systems map waveforms to binaries, ribosomes map DNA sequences to proteins as part of a larger mapping between genotypes and phenotypes. Telegram operators map back and forth between text and morse code, artificial neural networks map inputs to vector representations, and language allows us to map our thoughts to sentences that express them.

In each case, some information from one space needs to be transferred to another. Morse code represents 26 letters and 10 digits as a sequence of binary values - dots or dashes; the spanish alphabet uses 27 letters to encode the 19 consonant phonemes used in the language\footnote{the exact number of phonemes being dependent on dialect}. While in simple cases like these the scope of what information needs to be represented is well defined, what happens when what we need to encode is much larger: like all of the text on the internet combined with the text of every book written in the past 100 years? As what we need to describe increases in complexity, it becomes less and less clear how to map it to a representation which preserves the structure of the original. How do complex mappings represent information - and are there structural properties that are shared across representational systems that do this effectively?

\section{Deep Learning}
This question is of particular importance when it comes to artificial neural networks\footnote{Throughout I use the terms artificial neural networks, neural networks, deep-learning, and connectionist models broadly interchangeably.}. These are models trained to map their inputs to high-dimensional vector representations which preserve enough relevant information from the input for the model to succeed at a task. This is difficult in and of itself, but making it more challenging is that these models are usually trained on only a subset of the space they will need to encode - like a sample of sentences, rather than every sentence possible in a language\footnote{Given language contains a functionally infinite number of possible sentences, training on all of them is intractable.} - making it difficult to know what information in the data they see is part of larger generalisations across the unattested space. 
Depending on the task this can mean learning to drive on any road - mapping video input to actions like turn/accelerate/brake - from video footage of only a few thousand \citep{yurtsever2020survey}, or learning to map any sentence in French to English having been trained on a selection of websites and news articles \citep{kalchbrenner2013recurrent}.

Despite their success in recent years, large-scale neural networks often fail to learn a mapping which generalises systematically - often failing to learn a representation of their training data that can generalise far outside it. This becomes clear when models are evaluated on a different data distribution than the one they were trained on. In linguistic tasks this can mean their performance degrades substantially when words they have seen before appear in novel contexts, or appear in sentences longer than they see during training \citep[e.g.][]{lake_generalization_2018, keysers_measuring_2020, kim_cogs_2020}. Vision models can struggle with tiny changes to a small subset of pixels which doesn't change the image to the human eye \citep{goodfellow2014explaining} or have difficulty identifying a horse standing on anything other than grass, given the frequency of pastured-horses in training data \citep{dagaev_too-good--be-true_2021}. 

In response data scaling \& augmentation has become the prevailing strategy - making a model's training data sufficiently large that it is unlikely to encounter something it hasn't seen before - reducing the amount that it needs to generalise. But models struggle even when trained on more data than a human hears in 200 lifetimes \citep{griffiths_understanding_2020, furrer_compositional_2021}. A dataset can never cover the entire space of possible examples; there are an infinite number of grammatical sentences that have yet to be said \citep{chomsky1969quine} --- ultimately data underspecifies for the generalisations that produced it \citep{goodman1955new}. Despite this, humans reliably learn their native language from a small fraction of the data shown to a large language model.
What is missing from the representations learned by models trained orders of magnitude more data than we are, that affords them some generalisation but no more?  %

Building models that generalise robustly out-of-distribution remains a core goal of machine learning \citep{bishop_pattern_2006},  despite this we continue to have a limited understanding of what kinds of representational structures are needed enable generalisation. Some work identifies the kinds of shallow heuristics models learn instead of the underlying structure of the data \citep{mccoy2019right}, but these are rarely \emph{intrinsic} measures --- they're not based on models' internal representations but rather their (\emph{extrinsic}) downstream performance. It proves challenging to relate representational properties to behaviours,  in fact in some contexts it's been shown that no existing intrinsic measures can predict model behaviour \citep{goldfarb-tarrant_intrinsic_2021}.

\section{The Problem of Interpretability}
The challenge of understanding structure in deep-learning models is driven in no small part by their scale. The past two decades have seen a striking shift in the tractability of training neural architectures with gradient descent. Early models performed digit recognition with 9760 trainable parameters \citep{lecun1989backpropagation}, or learned sentence dependency structures with only 50 \citep{elman1990finding}. By contrast the `small' model used in chapter 3 of this thesis uses just over 1,000,000 parameters, and the large language models used in chapter \ref{chapt:llm} have 12,000,000,000 (even these are comparatively small by current standards - with state-of-the-art LLMs exceeding 400 billion parameters \textcite{dubey2024llama}). As they scale up these models represent information as increasingly high-dimensional vectors, something about which humans tend not to have strong intuitions. This makes it hard for us to decipher or reason about how a model represents information, how it learns to do so, or predict which representations might be best.

 We lack a clear understanding of what kinds of representational structures are desirable, or if there are domain-general, quantifiable properties of a representational system that enable systematic generalisation. 

\section{How to Interpret Representation Spaces}

 This lays out a core set of problems - 
 
  \begin{itemize}
	\item Models fail to represent their training data in a way that allows them to generalise systematically
	\item Their representations are high-dimensional vectors that are hard for us to interpret
	\item We lack a framework for defining how those representations are structured that lets us understand what kinds of structures drive behaviours like generalisation.
\end{itemize}
 
The remainder of this thesis works to address these issues, introducing a framework for thinking about  representation spaces grounded in existing work in cognitive science and information theory. First though, it is worth considering what it means to interpret a representation space, and by extension what any successful approach should do. I break this question into two parts, \textbf{what phenomena an approach to interpretability should give an account for}, and \textbf{the properties that approach should have.} At a minimum a framework for interpreting deep-learning models needs to be able to give an account of

 \begin{itemize}
        \item how representations are structured
	\item how representations change over the course of training
	\item how different design decisions (e.g. hidden size, choice of optimiser, learning rate, dropout ...) affect representation space
	\item what kinds of representation structures generalise best
\end{itemize}

These requirements already constrain some properties our desired approach can have. To characterise representation structure an approach needs to deal with representations directly, rather than making inferences about them based on downstream performance. Additionally to give an account over the timecourse of training an approach ideally needs to be sufficiently fast and resource efficient that it can be run at each training step, rather than only once as a post-hoc analysis. More than that, it is worth remembering that interpretability has an audience: humans. As such it is not enough to provide just any account of the above phenomena, it needs to be an account that is intuitive for us - relating the structures found here to things we already have an understanding of, like existing work on representations in other areas of science. In summary, our desiderata for an approach to interpretability are that it:

\begin{itemize}
        \item deals directly with representations rather than their downstream effects
	\item is efficient enough to leverage throughout training
	\item accounts for representational structure in models in a way that can be clearly related to work on representations in existing areas of science.
\end{itemize}

While existing approaches to interpretability have shed light on a variety of empirical phenomena in deep-learning models, they often meet only some of the criteria laid out above. A prominent existing set of approaches leverages behavioural evidence, treating models as akin to psycholinguistic subjects \citep{futrell2019neural, futrell2018rnns}. By treating model outputs as behaviours, experiments enable conclusions about what kinds of information a model may have learned. Like looking at whether models assign higher probability to grammatical sentences, to reason about whether their representations encode syntactic information \citep{marvin_targeted_2018, warstadt_blimp_2019, hu_systematic_2020}. While valuable, this line of work is removed from the models' representations themselves - characterising downstream behaviours rather than characterising the representational structures that drive them.

Probing represents another form of interpretability with closer ties to representational structure \citep{hupkes_visualisation_2018, pimentel2020information, muller2023subspace}. It relies on training a probe --- a smaller model, like a linear classifier --- to predict certain properties from representations. If a model can take a representation for a sentence, and predict the correct part of speech labels, or constituency parses for that sentence, it acts as some evidence that that information is encoded in those representations \citep[e.g.][]{voita_information-theoretic_2020}. Although this, again, does not directly characterise the structures in representation space but the information that can be predicted from them - and in some cases how complex a classifier is required for that prediction. Additionally as it requires training a secondary model, its computational complexity can limit the contexts where it is applied.

Mechanistic interpretability \citep{elhage2021mathematical}, tries to offer explanations more tightly tied to what happens model internally. However it often relies on training unsupervised probes \citep[termed \emph{sparse auto encoders}][]{elhage2022toy}. This enables some analysis of which parts of a model correspond to different words or concepts from the training data \citep{bricken2023towards}, but again this relies on training a secondary model, meaning it can have similar compute cost to other forms of probing and gives a limited understanding of how representations are structured or how those structures relate to work on representations in other areas of science. 

Having discussed what an approach to interpretability needs, and the limited ways in which this is addressed by existing work, I now introduce the approach taken here. This thesis introduces quantitative methods for identifying systematic structure in mappings between discrete and continuous spaces, and leverages them to interpret how neural networks learn and when \& why they generalise successfully. These methods are predictive of downstream performance, grounded in information theory, and fast to compute enabling analysis of even large models throughout training.

\section{An Analogical Approach}

As stated above, a goal here is to develop an approach to interpretability that lets us leverage intuitions from other disciplines. In particular disciplines with existing work on what representation structures are likely to be learnable, expressive, and can enable generalisation. To do this we can start by identifying a representational mapping, which is well studied, and which bears some resemblance to representation systems learned by the models studied here. By defining measures of structure applicable to both we can draw analogies between the exemplar and the structures that emerge in a deep learning model. This kind of analogical interpretability helps us understand something novel, through how it relates to something similar that is well understood.

In our case an ideal exemplar would have examples of  structures that enable the kinds of sample efficient learning and generalisation which neural networks likely need to succeed in the general case. It would also be one about which we have strong intuitions for what kinds of structures are desirable - unlike high-dimensional vector spaces - and for which there is an substantive body of work analysing \& describing those structures. Given these desiderata, the obvious choice is natural language.
 
At its core language is a mapping - relating objects, concepts, and events, to words, constructions, and phrases which refer to them \citep{saussure_course_1916}. While many natural communication systems fit this bill, language is unique amongst them \citep{Hockett1960}. It's acquired, rather than being built-in from birth. It generalises readily to novel concepts and contexts, instead of containing a finite repertoire of calls - we can readily interpret sentences we have never heard before\footnote{e.g. "At the airport I smiled myself an upgrade" ---\citep{goldberg_constructions_2006}}. Its units are meaningful despite being arbitrary, with its systematic structure providing us a representation system simple enough to be learned by children, but complex enough to describe the universe.

A growing body of work presents an account of how structures in language may result from language evolving to conform to domain-general cognitive constraints, and the dynamics of transmission and use rather than reflecting properties of some innate language faculty \citep[e.g.][]{Christiansen1994InniteLF, kirby_spontaneous_2001, zuidema_2002, brighton_language_2005, smith_cultural_2008, kirby_cumulative_2008, chater2009restrictions, smith_learning_2011, fedorenko2014role, kirby_iterated_2014, kirby_compression_2015, culbertson_simplicity_2016, wehbe2021incremental}. Given this, the structures present in language, and how we think they originated, may have explanatory power domain-generally, giving us an exemplar of how mappings become structured in response to their environment.

We expect neural networks to learn a mapping from inputs to representations from a finite sample of data, and generalise to examples not seen during training, by learning to encode structural properties of the world from which its training data is drawn rather than relying on heuristics. These expectations are in parallel to design features of language, which suggests if we develop sufficiently general ways of quantifying structures that underpin language we may be able to assess whether or not those structures are present in other domains - like vector spaces internal to models. While human language is clearly distinct from the representations inside a deep-learning model, their teleological similarities - and arguments about the domain-generality of language - make it a reasonable exemplar for our approach to interpretability. Building an understanding of deep-learning representations by drawing analogies between their structures and the structure of natural language.

\section{For Our Purposes, What is Structure?}

If we're going to look for structure in mappings, and draw analogies with language in the process, we need to be clear what we mean by structure. 
Part of what makes language unique as a mapping is its systematic structure (sometimes referred to as systematicity). Structure that exists not just at the item level - like at the level of an individual word or sentence - but across the entire language. In this thesis we consider two inter-related notions of structure present in language: compositionality and regularity. 

\subsection{Compositionality}

Compositionality describes how language builds the meaning of a whole as a product of the meaning of its parts \citep{Hockett1960, chomsky1969quine, cann_formal_1993, partee1995lexical}. It allows us as speakers and learners to make `infinite use of finite means' \citep{von1863humboldt}, in that from a finite set of words and constructions we can generate or understand a potentially infinite number of sentences. By composing together known morphemes, words, or sentences in novel combinations we can produce new words, sentences, and paragraphs, where the meaning of the larger construction is a predictable function of the meaning of the parts and they way they are combined. Compositionality represents a core building block of the syntactic system, as exemplified by the minimalist programme, which boils the innate component of syntactic knowledge down to primarily a merge operation \footnote{Many instantiations of minimalism also rely on a slightly broader assortment of operations, like agreement, and transfer.} which takes two arguments and composes them \citep{chomsky1995language,chomsky2014minimalist}. The origins of compositionality in language have been the subject of  study in linguistics \citep[e.g.][]{bickerton1984language, kirby_spontaneous_2001, kirby_cumulative_2008, kirby_compression_2015}, evolutionary biology \citep{nowak_evolution_2000} and more recently in mutli-agent reinforcement learning \citep[e.g.][]{lazaridou_emergence_2018, resnick_capacity_2020, chaabouni_compositionality_2020}. 

\subsubsection*{in deep learning}
A major driver of the continued interest in compositionality across disciplines is how essential it seems to generalisation; if we want a system to generalise, it's difficult to conceive of how it could do so non-compositionally\footnote{There's an argument to be made that some generalisation may be iconic instead, but such systems are unlikely to empower generalisation at the same order of magnitude as compositionality - that is enable generalisation to tens of thousands of unseen examples rather than a handful.}. How can we understand something novel, which we haven't encountered before, except by breaking it into parts we already know? Whether or not neural networks are capable of learning compositional representations, and generalising compositionality is the subject of longstanding debate, most notably with \citet{fodor_connectionism_1988} arguing that artificial neural networks are structurally unable to compose their representations. More than that, \citet{fodor1975language} argues that human thought is compositional \& symbolic to its core - neural networks operate in continuous vector space making symbolic processing impossible \citep[see][for discussion]{symons2014systematicity}. This criticism is oft repeated, even now, despite the fact that in the years immediately following, \citet{smolensky_tensor_1990} showed how compositional, symbolic structures can be losslessly embedded in vector space, \citet{elman1990finding} introduced the recurrent neural network which has explicit composition, and \citet{chalmers1993connectionism} pointed out structural flaws in Fodor and Pyshlyn's argument.

Today, a substantive body of work still questions the compositional abilities of contemporary neural networks, and large language models \citep{hupkes_compositionality_2019, keysers_measuring_2020, akyurek_compositionality_2022, andreas_good-enough_2020, csordas_devil_2021}, with many introducing benchmarking datasets \citep[e.g.][]{lake_generalization_2018, kim_cogs_2020} intended to evaluate models' compositional abilities, by testing them under \emph{distributional shift}, where a model is trained on data sampled from one distribution, and evaluated on data sampled from another. In practice what it means for distributions to differ can depend on the domain, but on text-based tasks 
data is usually synthetic, generated by a context free grammar, with training and evaluation splits created by subsampling different portions of the entire space that grammar covers. As a result any evaluation split would be trivial for the model had it learned the underlying grammar used to generate the data. In reality contemporary architectures like LSTMs and Transformers perform well when training and evaluation splits are sampled at random (referred to as in-distribution generalisation, or independent and identically distributed \emph{i.i.d.}), but where this is not the case - as when evaluation contains longer sentences as mentioned above- those same architectures perform remarkably poorly. Large language models, like T5 \citep{raffel2019exploring}, pretrained on vast amounts of data then finetuned on these tasks fare substantively better than standard architectures but still below ceiling \citep{furrer_compositional_2021}.

\subsubsection*{if not compositional then what?}
This work often paints a confusing picture, asserting that a model's limited ability to generalise out of distribution, to examples generated by the same underlying grammar as the training data, provides evidence of non-compositionality; evidence that models have in fact induced heuristics via statistical learning rather than inducing the rules of the grammar. But this fails to reckon with the fact that these models \emph{can} generalise, to tens of thousands of examples provided those examples and the training data were sampled i.i.d.. Short of an explanation for how generalisation of that scale can happen non-compositionality we instead need an explanation for how a compositional system can generalise systematically to some examples but not others.

\subsection{Regularity (\& Variation)}
Unlike compositionality, regularity is a property of language that has received less attention outside of linguistics. Compositionality enables predictable mappings between meanings and forms by building wholes out of reusable parts - like words - which makes it so that  human's best friend will reliably be referred to as a \emph{dog} regardless of the context. However language is also rich with \emph{variation} \citep{weinreich_empirical_1968}, affording us as speakers an enormous variety of ways to express ourselves dependent on the given context. Regularity - in the form of predictable system-level structures  -  underpins language's generalisability, but is interwoven with \emph{variation} which gives us robust tools for conveying meaning, ambiguity, \& intention in context - making ourselves clear with precision. It allows our collective best friend to sometimes be a \emph{dog}, but also to be a \emph{canine, pup, good boy, good boi} or \emph{mutt} as the situation demands. In language regularity refers to how predictable realisations of the same property are across a system. This is the inverse of variation with respect to a given property which, somewhat intuitively, describes how much that property varies
A substantive body of work looks at how humans regularise their input during learning \citep{senghas1997argument, hudson_kam_investigating_2009}, and how languages undergo regularisation over time (\textcite{Reali2009}, \textcite{smith_eliminating_2010}, see \textcite{ferdinand_cognitive_2019} for review) often quantifying it probabilistically. For our purposes we'll say that:

\thinline
\begin{quote}
{
	\makesans
	{\flushright \textbf{\textit{\underline{Definitions 1 \& 2}\\[2em]}}}
	\textbf{Regularity:} How predictable realisations of the same property are across a system\\[1em]
	\textbf{Variation:} How much realisations of the same property vary in that system dependent on context
}
\end{quote}

\thinline

At a computational level compositionality is a binary property - either a merge operation is performed somewhere in a system or it isn't \footnote{\citet{martins2019language} make a case that the binary analogy may not extend to the hardware level, but I leave arguments about the origins of merge to other work, and here for the sake of argument consider systems where merge exists.}. I argue in chapter \ref{chapt:comp_reg}, that regularity and variation are quantities best suited to assessing whether a system is structured, in no small part because they are naturally graded - offering degrees of variation - rather than binary. Although the two concepts are intrinsically related, I point out that what is often discussed as compositionality is in reality compositionality + maximal regularity\footnote{You can also think of it as regularity reflecting the predictability or frequency of merge operations in a system, which is likely what is implied when work discusses the `degree of compositionality'. However the predictability of compositions is distinct from their existence.} likely predicated on an assumption that variation impairs generalisation.

 Variation is not an accident. It proves ubiquitous across languages because it enables us to express effectively the kinds of complex context-dependent information we encounter every day. In extremity variation can impair generalisation - imagine having a different word for dog depending on what it holds in its mouth, like a bone or a tennis ball. This would allow us to be maximally efficient and precise in the rare context of picking one out of a lineup of dogs each holding a different object. But becomes problematic in the far more likely scenario that we encounter a dog holding something novel, like a smartphone, or the collected works of David Foster Wallace. 
 
 While we could describe a word without separate parts referring to [dog] and [what is held] as non-compositional, this elides the fact that the resulting word can go on to be used compositionally in a sentence --- \emph{a lack of compositionality at the item-level does not preclude it at the system-level.} Similarly a set of synonyms (e.g. best, baddest, leading, terrific) are often best suited to subtly different contexts (best: a blogpost about what coffee maker to buy; baddest: a TikTok about the best coffee maker to buy; leading: the description of the coffee maker on the manufacturer's website; terrific: the review you leave of the coffee maker). The fact that each of these conveys both a notion of `excellence' along with contextual information doesn't undermine the compositionality of language. Were we to stipulate that a truly compositional system represent everything context independently the result would be substantively removed from the realities of human language. 
In chapter \ref{chapt:information} I introduce methods for quantifying regularity, variation, and disentanglement (a particular kind of variation), that can in principle be applied to any discrete-discrete or discrete-continuous mapping that I apply in the remainder of the thesis to a variety of artificial neural network models.

\subsection{Information Structure}
Across domains information theory \citep{shannon_communication_1949} is a tool of choice for analysing how information is packaged and mapped - finding explanatory power from genetics \citep{schneider2010brief, vinga2014information}, to cognitive science \citep{chater_simplicity_2003, smith_eliminating_2010} and machine learning \citep{MacKay2004InformationTI}. 
Additionally as a discipline it rests on a similar analogy to language as the one I make here, with Shannon introducing the field as a mathematical model of communication. In the general case Information Theory considers the mapping between a message from an information source and a signal that represents it, and presents quantitative methods for describing the relationships between spaces and the mapping that relates them. In this thesis I build on this analogy using basic information theoretic quantities to quantify regularity, variation, and disentanglement in a mapping between spaces. These linguistic concepts are intuitively related to basic information theoretic quantities, links I make explicit in chapter \ref{chapt:information}. 

There are three basic kinds of structure I consider in a mapping between two spaces: one-to-one, one-to-many, and many-to-one --- related to regularity, variation, and disentanglement respectively. In reality, at a system level, a mapping can be comprised of a combination of these three structures, so we quantify the prevalence of each of them probabilistically; defining quantitative measures reflecting the probability of each basic structure across a system. An approach first introduced in chapter \ref{chapt:information}, then built on in chapters \ref{chapt:learning} and \ref{chapt:llm}. To distinguish this approach to understanding representational structure from previous work I refer to it as \emph{information structure}\footnote{This is unrelated to exiting linguistic notions of information structure, which focus on different ways of communicating the same information (e.g. to draw focus to a particular part). The name is adopted here to describe our information-theoretic approach to structure.}, given it aims to quantify structure in the way information is mapped between spaces. It's worth noting that this approach is not the first to emphasise the interplay between structure and probability with usage-based approaches to language \citep[e.g.][]{goldberg_constructions_1995, croft2001radical, tomasello2005constructing}, eroding the binary distinction between grammar and lexicon in favour of constructions that unify meaning and form and are learned probabilistically on the basis of experience \citep{goldberg_constructions_2003}. The relationship between usage-based approaches and an information structure approach are discussed further in chapter \ref{chapt:meta}.

\section[Leveraging Work on Language to Understand Mappings]{Leveraging Work on Language \\ to Understand Mappings}

We've discussed how making analogies with language can help us form a strong intuition about what a structured, learnable, generalising mapping looks like - formalising these intuitions quantitively is a core goal of this thesis. But by drawing this analogy we can also leverage intuitions from the cognitive sciences about the conditions needed for structure to emerge in a mapping, what kinds of structures can drive or hinder generalisation, and the constraints or pressures which condition the prevalence of those structures. Building on existing theories and intuitions is a major advantage of contextualising mappings found in neural networks in terms of existing areas of science - like language and information theory - rather than approaching them with methods and terminology which make them out to be something wholly alien.

\paragraph*{\emph{When Structure Emerges}} Structural regularities can emerge as a result of the meaning space speakers need to describe expanding such that substantively greater fitness is given to systems with structural regularity \citep{nowak_evolution_2000}. Or they can arise from repeated iterated chains of learning, applying a pressure for simplicity making the system easier to learn \citep{kirby_spontaneous_2001, kirby_cumulative_2008}.

\paragraph*{\emph{What Structures are Desirable}} Compositionality is essential for generalisation \citep{chomsky_aspects_1965, cann_formal_1993}, enabling predictable, regular structures that are recombinable. Variation is equally essential to enable expressivity, giving speakers sufficient fidelity to describe what they need to \citep{kirby_compression_2015}. Other structures, like homonymy, make a system more compressible but at the expense of introducing ambiguities that can be difficult interpret \citep{piantadosi_communicative_2012}. Often languages mitigate these ambiguities by collapsing over concepts that are contextually mutually exclusive and unlikely to co-occur \citep{winters2018contextual}.

\paragraph*{\emph{What Conditions Structure}} As mentioned above, pressures relating to the needs of learners and speakers have major effects, with speakers introducing pressure for variation to express themselves, and learners introducing pressure for regularity to aid acquisition \citep{kirby_compression_2015}. These pressures can also be introduced to the system via population dynamics, with prevalence of second language learners, number of speakers, and geographic spread of a population having potentially regularising effects \citep{lupyan2010language, dale_understanding_2012}. Competing needs of speakers and listeners can drive the mapping from meanings to forms to become more efficient in an information theoretic sense, with languages often evolving to be optimally compressed for the amount of information they encode \citep{zaslavsky_efficient_2018, kemp2018semantic}. More general cognitive constraints are also thought to be a major driver of regularity, with limitations like our finite memory having a regularising effect by placing an upper bound on the amount of variation we can faithfully remember and reproduce \citep{griffiths_understanding_2020, lieder_resource-rational_2020}. \\[0.5em]

\section{Capacity's Role in Shaping Structure}

Across chapters of this thesis particular attention is paid to the role capacity plays in how structure develops. A wide array of work has looked at how the finite memory of human learners can drive the kinds of regularities ubiquitous across languages \citep[e.g.][]{smith_eliminating_2010, newport_maturational_1990, ferdinand_cognitive_2019, griffiths_understanding_2020}. In part because we are likely limited in the number of low probability forms we can recall \citep{hudson_kam_regularizing_2005}. In work on humans however, it can to be difficult to directly modulate the capacity of learners as an independent variable. In models, by contrast, this is a hyper-parameter we set.

Each experimental chapter looks at the effect capacity has on measures of representational structure. Chapter \ref{chapt:comp_reg}, varies the size of representational spaces\footnote{I also look at dropout and l2 regularisation as ways of modulating model capacity.} learned by agents in an emergent communication model, showing populations with smaller agents develop more regular systems. Chapters \ref{chapt:learning} and \ref{chapt:llm} look at continuous representations inside models ranging from 1 million parameters to 12 billion --- finding larger models can accommodate more contextual variation.  Finally Chapter \ref{chapt:meta} introduces a way to manipulate a model's capacity via optimisation, using a meta-learning objective instead of manipulating a model's parameter count. Models trained with this objective generalise more robustly out of distribution, in line with expectations from work in cognitive science.

The through-line of capacity here allows us to consider how general effects of capacity are on representational structure --- looking at the degree to which conclusions from work on humans, can apply to learners in general.
Throughout we show that reduced capacity almost always has a regularising effect of some kind, in accordance with expectations from existing work. However, in the experiments looking at model-internal representations, the story becomes more complicated - we analyse regularity with respect to both words and the context they occur in. Larger models are less regular at the word level but can accommodate far greater regularity with respect to context, and it is this contextual regularity that proves predictive of model performance. This nested, multi-level approach to regularity (introduced in chapter \ref{chapt:learning}) potentially offers a way for thinking about certain linguistic phenomena - like iconicity - where languages seem to exhibit regularities with respect to event structure.

\subsection*{A note on other approaches in this direction}

It's important to point out that I'm not the first to notice the potential for language to help us understand other complex systems. In fact much of early cognitive science leverages analogies with language in discussion of other aspects of cognition \citep[e.g.][]{lashley1951problem, miller1951language}. To focus on a few examples of previous approaches, of relevance to the work presented here: \citet{fodor1975language} asserted that human thought is best understood as a language, with our cognition functioning as a system of signs mapping between the world and our thoughts about it. \citet{smolensky_tensor_1990} showed how vector spaces, particularly those learned by early connectionist models, can be understood in terms of a generative grammar. \citet{Beckner2009} draws parallels between language and complex systems in physics. More recently analysis of multi-agent deep-learning models has leveraged tools from emergent communication (\cite{brighton2006understanding} used in \cite{lazaridou_multi-agent_2017}). In evaluating different deep-learning models for vision classification \citet{lu_expressiveness_2022} instantiate a method for measuring quantities related to \citet{kirby_compression_2015}'s notions of expressivity and learnability. The long history of work along these lines, makes clear the utility of alluding to language in understanding representational systems.

It's also important to point out that the approach taken here differs from previous work. I make a point of looking at, discussing, and quantifying structure in \emph{mappings}. As discussed in the next chapter this is a fairly general set of functions, at a relatively high level of abstraction. I argue that language is a mapping and can be used as an exemplar against which to make analogies about structures in other mappings like connectionist models -- which is distinct from claiming that other mappings are themself a language. This may seem like a needlessly fine hair to split but its an important one. I make no assertions that representations in a mapping need to be symbolic \citep[a la][]{fodor1975language}, nor do I focus on embedding discrete structures in representations space like \citet{smolensky_tensor_1990}. I also explicitly quantify structures (e.g. regularity) in a mapping using general-purpose methods, instead of looking at behavioural properties like learnability which is necessarily relative to a learner, I focus on clear, self-contained formalisations that are computationally efficient. Some of my engagement with linguistics at only a high level is also out of respect for the complexity of language and awareness that talking about it in terms of regularity, and variation abstracts much of that away.

\vspace{10mm}

\noindent Part of the focus on mappings in the abstract is that, to me, some of the beauty found in language's domain generality, is that our understanding of language can underpin our understanding of the world, in general: the information in it, and the structures that define it.

\vspace{15mm}

\thinline

\section{Thesis Outline}

To business. This thesis falls across 7 chapters, and broadly revolves around three core themes.

\begin{enumerate}
	\item Structural properties found in language are domain-generally useful for understanding mappings that need to be learned, structured, and generalise
	\item Quantifying information structure in the mapping learned by a neural network can allow us to describe their learning process, and when and why they generalise
	\item Capacity's effect on the emergence of structure in neural networks
\end{enumerate}

\thinline

To summarise the chapters below

\begin{enumerate}
	\item \textbf{\makesans How to Represent Information:} A general introduction to the core concepts of this thesis
	\item \textbf{\makesans Information Structure:} Introduces 3 basic structures present in a mapping between two spaces and relates them to information theoretic quantities. The remainder of the chapter provides a brief introduction to discrete information theory.
	\item \textbf{\makesans What We Talk About When We Talk About Compositionality:} This chapter discusses challenges in quantifying structure, and looks at the relationship between compositionality and regularity. I introduce methods for quantifying variation in a discrete $\to$ discrete mapping, showing how previous measures of compositionality implicitly assess regularity. This distinction allows us to make sense of previous results suggesting compositionality isn't related to generalisation. Finally I vary model capacity showing how capacity to have a regularising effect in line with what's predicted by work in linguistics. Work in this chapter is based around \citet{conklin2022compositionality}. 
	\item \textbf{\makesans Regularity and Variation in Vector Space}: I use the structural quantities defined in chapter 2 to understand what happens when training a neural network. Transformer models trained on a sequence-to-sequence task go through distinct patterns of expansion, compression, and disentanglement. Based on quantifications introduced in this chapter I can predict how well a model will generalise out of distribution, laying out the kinds of structures that seem critical for generalisation. Work in this chapter is based around \citet{conklin2024representations}.
	\item \textbf{\makesans Information, Generalisation and Scale, in Large Language Models}: Here I apply the information structure analysis from earlier chapters to large language models. Showing how they follow a similar training trajectory to their smaller counterparts. As with models trained on a single task we find correlations between particular representational structures and downstream performance, further showing what representational structures drive generalisation.
	\item \textbf{\makesans Biasing Representational Structure with Meta-Learning}: In a final chapter, I look at how to bias representational structure using a meta-learning objective. Optimizing a model's update steps to be beneficial to similar examples, and showing that this improves out-of-distribution generalisation ability. This chapter also starts with some discussion of how the information structure framing used throughout the thesis relates to more behavioural properties like memorisation and generalisation. Experiments presented here are based around \citet{conklin_meta-learning_2021}.
	\item \textbf{\makesans Conclusion:} Here we revisit the core themes of the thesis, highlighting common threads between the preceding chapters and laying out directions for future work.
\end{enumerate}

\paragraph*{Reproducibility}
Unless stated otherwise, code and data are available at  \url{https://github.com/hcoxec/h}.

\chapter{Information Structure}%
\label{chapt:information}
{\makesans
\emph{
a primer on information theory for quantifying structure} \\[1em]
}

{\makesans
\begin{quote}
\censor{[.............................................]} feeling herself change
painfully cell by cell
into a shadow, a laurel, you, a constellation. \\[0.1em]
\flushright{\emph{- James Richardson}}
\end{quote}}

\begin{figure*}[!h]
	\includegraphics[width=0.5\textwidth]{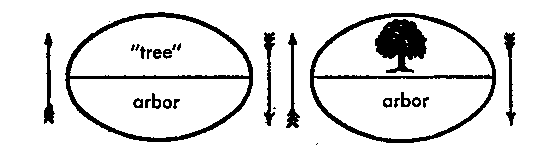}
\end{figure*}

\drawline

\noindent Mappings relate two spaces - alphabet/morse code, input/encoding, meaning/form - and their structures can vary widely depending on where and how they're used. We want a general-purpose way to describe their structure quantitatively, so we consider three kinds of primitive structure present in a mapping: one-to-one, one-to-many, and many-to-one. By assessing each of these quantities continuously, we can describe a mapping in terms of how much of each structure is present. Each of our primitive structures relates intuitively to basic information theoretic quantities, the majority of this chapter is a primer on information theory (in the discrete case) and the quantities relevant to the chapters that follow. Before that, I give a quick overview of the primitive mapping structures we look at later. The next chapter presents initial experiments quantifying specific linguistic structures in the discrete-to-discrete mapping learned by a multi-agent model. Chapters \ref{chapt:learning} \& \ref{chapt:llm} build on this, introducing the more general framework for thinking about structure described briefly in the next section and applying it to discrete-to-continuous mappings learned by Neural Networks.

\section{Structural Primitives}\label{sec:struct_prim}

Consider 3 basic kinds of structure that can exist in a relational mapping between two spaces: one-to-one, one-to-many, and many-to-one. These are depicted visually for a mapping between spaces $X$ and $Z$ in figure \ref{fig:basic_mapping}, along with the information theoretic quantities we relate them to later on. At a high-level these are: %

\begin{figure}
	\centering
	\input{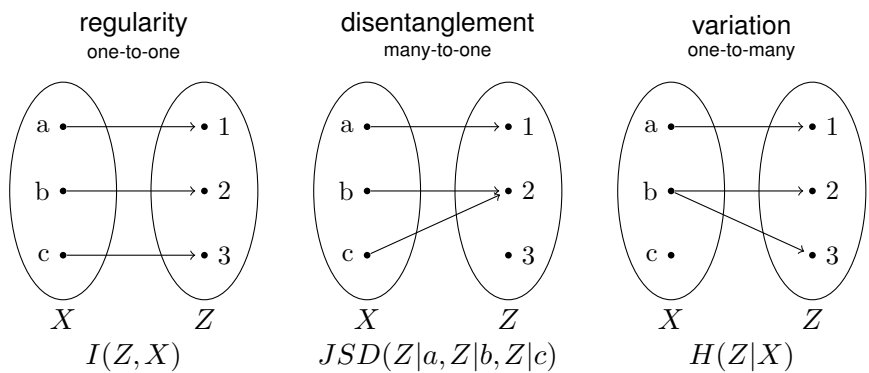}
	\caption{Three basic kinds of mapping structure we consider here, labelled with their linguistic analog, and the information theoretic quantity we introduce to measure them in chapter \protect\ref{chapt:learning}. Note that we show part of the mapping ($a\to 1$) as regular in all cases because the mappings we consider exhibit a combination of all 3 structures. As such we assess the \emph{degree} of each structure, not whether or not it exists. Variation (one-to-many) is possible here because our $X$ contains instances of the same label (like a word) in different contexts (like sentences), meaning $b\to 2$ and $b\to 3$ reflects b in different contexts which are not shown here for brevity.}
	\label{fig:basic_mapping}
\end{figure}

\drawline

\subsection*{One-to-one (Regularity)}

Maximised when each unique $x_i \in X$ maps to a unique $z \in Z$ and each $x_i^k \in x_i$ maps to the same $z$ regardless of context. This reflects how predictable a mapping between spaces is, or the degree to which the spaces $X$ and $Z$ are monotonically aligned. If we were to think about this in terms of a function $f$ that maps $f(x) \to z$ regularity is related to how \emph{injective} $f$ is. A regular mapping is structure preserving - we can recover the input $x$ from the corresponding output $z$

\subsection*{Many-to-one (Entanglement\footnote{Note that we often quantify disentanglement, rather than entanglement because it better aligns with the information theoretic divergences used to measure it. For our purposes these are the same quantity, inverted. })}

 Maximised when all $x \in X$, map to the same $z_i$, regardless of which $x$ it is or the context in which it occurs. Reflects the degree of ambiguity in the mapping - how hard it is to infer which input $x$ has been mapped to a given $z_i$. Given $f(x) \to z$ this is related to how \emph{non-injective} $f$ is. Linguistically this is most clearly related at a lexical level to \emph{homonymy} where multiple meanings have the same surface form - an analogy discussed at length in chapter \ref{chapt:comp_reg}. A entangled mapping is not structure preserving - a given output $z$ could have many corresponding input $x$s.

\subsection*{One-to-many (Variation)}

Maximised when each input has a different representation for each context where it occurs --- i.e. each $x_i^k$ maps to a unique $z_i$. This is the inverse of regularity, reflecting the degree of contextual variation in the mapping. A function which maps the same input $x$ to different outputs $z$ violates the general definition of a function. But we can think of this as a kind of reciprocal of entanglement, and say given $g(z) \to x$ this quantity reflects how non-injective the mapping from z's to x's is. This highlights that entanglement and variation are virtually identical except for their directionality (one-to-many vs. many-to-one). Lexically this is related to \emph{synonymy} in natural language, where the same meaning has multiple different realisations in form often dependent on context. Structurally we can relate this to word-order freedom, or the degree of variability in the mapping between semantic roles and linear order in form. 

\vspace{5mm}

\noindent These are the structures of interest in brief and, while basic, later chapters show that evaluating them at different levels of abstraction can give substantial insight into the structure present in a wide array of mappings. In order to quantify these we need to be able to quantify the information present in each space, and what information is preserved or compressed as we map between them. For a quantitative approach to information, we turn to information theory.

\drawline
\vspace{-15mm}

\section{Discrete Entropy}

Information theory describes relationships between spaces, and is built upon a `mathematical theory of communication' \citep{Shannon1948AMT}. Shannon considers an information source producing messages that are encoded in a signal by a transmitter, to be later decoded back to the original message by a receiver; built on a broad analogy to language where a speaker encodes their thought in a sentence decoded by a listener. It's worth noting this is similar to the analogy I make throughout this thesis, relating different kinds of mappings to language as a point of reference. Originally concerned with how to optimally map messages to signals, information theory has found explanatory power across a wide array of disciplines from genetics \citep{lezon2006using}, to neuroscience \citep{paninski_estimation_2003} and machine learning \citep{MacKay2004InformationTI}. In later work Shannon looks at ambiguity in encoding schemes, quantifying the allowable degree of ambiguity in a mapping from messages to signals that still allows the structure of the original message to be recovered \citep{shannon_communication_1949} - this work eventually forms the basis of lossy compression, and means this area of mathematics has extensive tools for thinking about information quantitatively. At it's core information theory considers data probabilistically, and so like probability itself has different instantiations for discrete and continuous cases. Here we introduce the discrete case, which is easier to reason about intuitively, and is used in the next chapter. Later in Chapter \ref{chapt:learning} we formalise the same quantities for continuous cases.

\subsection{Quantifying Information}
How can we tell how much information is in a sample of data? Not how many gigabytes it is, or how many entries it has, but how much \emph{information} it contains. Let's say for a moment that we had two datasets, one of which contains all 26,145 words of the full text of King Lear, the other containing 26,145 words comprised of just "king" and "lear" repeated over and over again. Clearly the former contains far more information despite the fact that both are identical in size - what we want is a way to reliably quantify this difference. We can do this using information entropy \citep{Shannon1948AMT} which quantifies information by looking at data probabilistically. It follows from the intuition that the more frequent something is, the less informative it is. Consider the 5 most frequent words in English \emph{the, be, to, of, and}, most of which have a primarily grammatical function, conveying little semantic content of their own; this becomes even more clear were we to take a passage

\begin{itemize}
    
    \item[(a)] We went outside. He adjusted the shutter. He told me where to stand, and we got down to it. We moved around the house. Systematic. Sometimes I'd look sideways. Sometimes I'd look straight ahead. "Good," he'd say. "That's good," he'd say, until we'd circled the house and were back in the front again. "That's twenty. That's enough." "No," I said. "On the roof," I said. - \citep{carver1981love}
\end{itemize}

\noindent and remove any of the 100 most frequent words in English according to the Oxford English Corpus \citep{stevenson2010oxford}.

\begin{itemize}
    \item[(b)] \censor{We went} outside. \censor{He} adjusted \censor{the} shutter. \censor{He} told \censor{me} where \censor{to} stand, \censor{and we got} down \censor{to it. We} moved around \censor{the} house. Systematic. Sometimes \censor{I'd look} sideways. Sometimes \censor{I'd look} straight ahead. \censor{"Good," he'd say.} \censor{"That's good," he'd say,} until \censor{we'd} circled \censor{the} house \censor{and were back in the} front again. \censor{"That's} twenty. \censor{That's} enough \censor{." "No," I said. "On the} roof \censor{," I said.}
\end{itemize}

\noindent Which immediately makes the text ungrammatical, but we can still recover quite a lot of what the passage is about - taking pictures outside (`outside adjusted shutter') while circling a house (`moved around house.. until circled house'). By contrast, removing any words not in 100 most frequent

\begin{itemize}
    \item[(c)] We went \censor{outside.} He \censor{adjusted} the \censor{shutter.} He \censor{told} me \censor{where} to \censor{stand,} and we got \censor{down} to it. We \censor{moved around} the \censor{house. Systematic. Sometimes} I'd look \censor{sideways. Sometimes} I'd look \censor{straight ahead.} "Good," he'd say. "That's good," he'd say, \censor{until} we'd \censor{circled} the \censor{house} and were back in the \censor{front again.} "That's \censor{twenty.} That's \censor{enough} ." "No," I said. "On the \censor{roof} ," I said.
\end{itemize}

\noindent we end up with text, where the meaning of the original is essentially unrecoverable. We can still piece together that two or more people (`we went' `he' `me') are doing something that goes well ("Good," he'd say. "That's good," he'd say,") but nothing more. While there is information in both edits of the sentence, there is considerably more in the version that retains lower-probability words.

\subsection{Self Information/Surprisal}
Armed with this intuition, that the amount of information in a piece of data is related to how likely it is, Shannon quantifies the amount of information in an event as the \emph{self information}, also termed surprisal. Because we're talking about information in terms of probability, given some data we need a probability distribution that describes it. For text data there are a number of ways of describing it probabilistically - for simplicity we create random variable $\mathcal{X}$ that describes our data at the word level, where each event in the distribution $x_i$ is a word that occurs in the text, and its probability refers to the frequency of that word. Given this, the self-information of each word is the negative log of its probability.

\begin{equation}\label{eq:suprisal}
    s(x_i) = -\log p(x_i)
\end{equation}

\noindent When the probability of an event $p(x_i)$ is 1.0 its log is 0, and as $p(x_i)$ approaches 0 $s(x_i)$ monotonically increases (shown figure \ref{fig:suprisal}
 left). This definition and its use of a logarithm are intended to satisfy:

\begin{figure}[!h]
\centering
\input{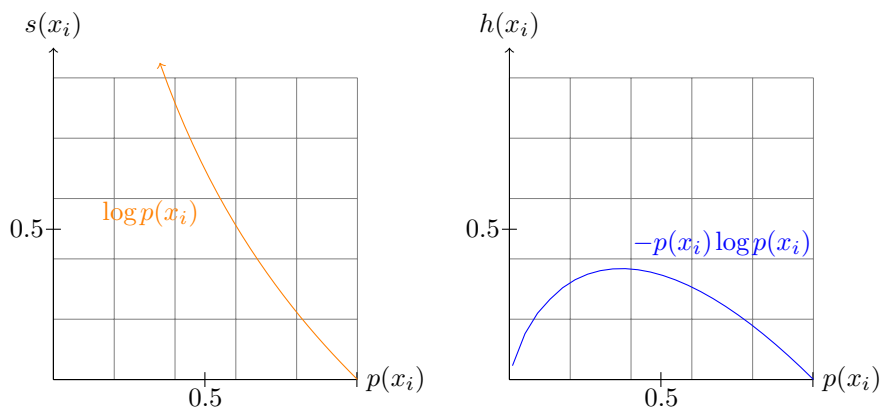}
\caption{Plots showing suprisal (left) and entropy (right) values for a single event. The x axis is the event probability, while the y axis shows the infromation theoretic quantity. Note that suprisal continues off the plot approaching infinity as $p(x_i)$ approaches 1.}
\label{fig:suprisal}
\end{figure}

\thinline
\begin{itemize}
\makesans
    \item[i.] A constant event, with probability 1.0, conveys no information; it's completely unsurprising
	\item[ii.] An unattested event, with probability 0.0, is infinitely suprising
    \item[iii.] The less likely an event, the more information it contains
\end{itemize}
\drawline

\noindent Note that the `unit' of entropy depends on the base of the logarithm used. In what follows I use the natural logarithm unless otherwise noted, which means all entropies are in \emph{nats} (for bits use base 2, for dits use base 10).

\subsection{Entropy: Expected Information}

To get the amount of information contained in the whole distribution $\mathcal{X}$, rather than just one event, we aggregate the self information of the events it contains --- for instance, we might aggregate across the information contained in each word in a vocabulary for text data. Information Entropy aggregates using the expected value operator, a weighted mean where the weight is determined by the probability of the event. Figure \ref{fig:suprisal} (right) shows this quantity for a single event. 

\begin{equation}\label{eq:shannon_entropy}
    h(\mathcal{X}) = \sum_{x_i\in \mathcal{X}} -p(x_i)\log p(x_i)
\end{equation}

\noindent Compared with self-information, entropy assigns proportionally less information to less likely events. In practice this can be useful in cases with many low probability events - whose self-information will approach infinity which can make estimates numerically unstable.

At the distribution level, entropy describes how peaked a distribution over events is. As a single event in the distribution becomes increasingly probable the overall entropy decreases. This is shown below for a random variable with four events. The uniform distribution achieves highest entropy, which decreases as the distribution becomes more peaked.

\begin{figure}[!h]
\centering
\input{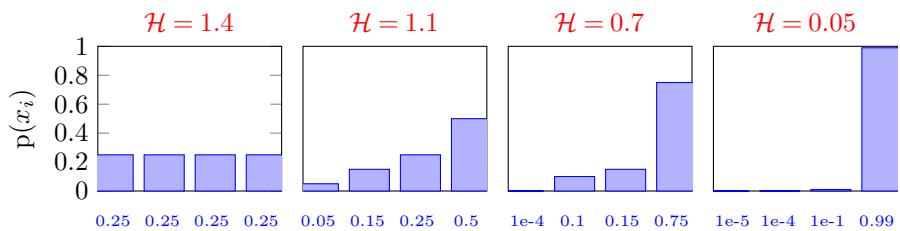}
\caption{A distribution with 4 events is shown above in 4 different versions. On the x axis is the probability of each event, also labelled below each bar. Above each plot is the entropy of the distribution. As events become less uniformly distributed - more peaky - entropy decreases.}
\label{fig:entropy_peaks}
\end{figure}

\noindent We can also see it as reflecting the number of samples needed from a distribution in order to tell its shape and the probability of the events it contains. When one event occurs 100\% of the time, and entropy approaches 0, we hardly need any samples at all. But as the distribution becomes more uniform we need more and more samples in order to have all possible events be attested, and to get a good estimate of their probability. 

This leads us to a final intuitive way of thinking about entropy most relevant to later chapters: as the amount of variation in data. If $\mathcal{X}$ is a distribution over words, then $\mathcal{H}(\mathcal{X})$ is minimised when only one word is used - meaning there's no variation in word-choice. As the words used in the text vary more and more, the distribution over them becomes more uniform, and entropy increases.
To summarise the perspectives, entropy reflects:

\thinline
\begin{enumerate}
\makesans
	\item the amount of information in a random variable
	\item the expected level of surprise from any sample from a distribution
	\item how peaked a distribution over events is
	\item the relative number of samples needed from a distribution in order to estimate it
	\item \textbf{the amount of variation in the data a distribution describes}
\end{enumerate}
\thinline

\begin{figure}
\centering
\input{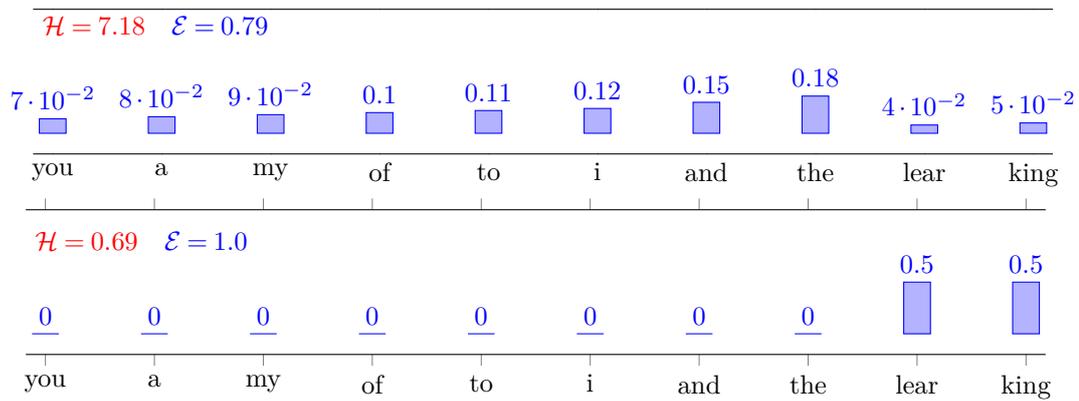}
\caption{The probabilities of the 10 most frequent words from both documents plotted, along with their respective entropies and efficiencies. As shown the repeated document has considerably lower entropy than the full text, but higher efficiency as it is more uniformly distributed over events it contains. Note that the lower distribution is only comprised of the two words with non-zero probability, non-existent events are shown for continuity with zero probability but do not factor into the efficiency calculation.}
\label{fig:entropy:lear_top_k}
\end{figure}

\noindent With this in mind we can return to the task we started with - telling apart our two documents, one with the text of king lear, the other with the words `king' and `lear' repeated for the same number of words. We build a vocabulary for each document containing all words that occur in it, then create a random variable for each of them with the events in the distribution reflecting the probability of a given word (summarised in figure \ref{fig:entropy:lear_top_k}). With these distributions we can say the entropy of the full text of kind lear is 7.18 nats, while the document of the same length with two repeated words is only 0.69 - the full text has more information.

\subsection{Efficiency: Normalised Entropy}
	
\begin{quote}
\makesans
	Degree of non-uniformity, independent of distribution size. Bounded $0 < \mathcal{E}(\mathcal{X}) < 1$: the more uniform it is, the closer its efficiency is to 1. The more peaked it is, the closer to 0.
\end{quote}

\noindent An issue in interpreting entropy values is that the quantity itself is, in principle, unbounded. You could have longer, and longer documents with greater complexity so entropy is bounded $0 < \mathcal{H}(\mathcal{X}) < \inf$. This can make it difficult to compare entropy values for different distributions - a uniform distribution with 10 events will have lower entropy than a uniform distribution with 20 events. While this makes sense given the intuitions we've discussed so far - 20 equiprobable events encode more information than 10 - often  we want a relative quantity that can be compared across distributions of different sizes. To get this we focus on the degree of non-uniformity in a distribution, or peak-iness, by normalising the entropy of a random variable by the entropy of a same sized uniform distribution - remember that a uniform distribution represents the highest possible entropy. Helpfully, the entropy of a uniform distribution is equivalent to the logarithm of the number of events it contains. 

\begin{equation}
	\mathcal{E}(\mathcal{X}) = \frac{\mathcal{H}(\mathcal{X})}{\log(|\mathcal{X}|)}
\end{equation}

\noindent The resulting quantity is called efficiency and is bounded between 0 and 1, such that an efficiency of 0 indicates a one-hot distribution, and 1.0 indicates a uniform. Shannon terms this efficiency because it reflects what proportion of a distribution's maximum possible entropy is actually used. 

\begin{figure}[!ht]
\centering
\input{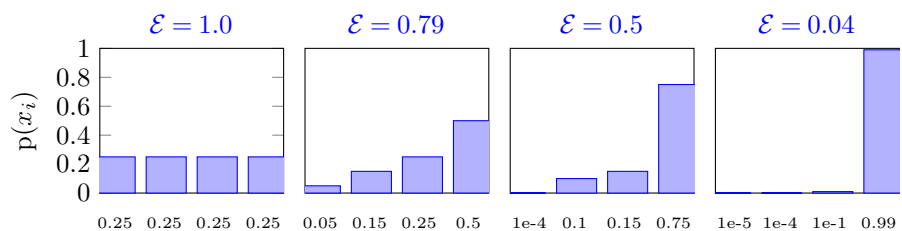}
\caption{The same distributions shown in figure \protect\ref{fig:entropy_peaks} but now labelled with the efficiency of each distribution.}
\label{fig:efficiency_peaks}
\end{figure}

\noindent Referring back to figure \ref{fig:entropy:lear_top_k}, note that the document with just the words `king lear' repeated has lower entropy than the full text of the play, but higher efficiency. The two words present in the document are perfectly uniformly distributed so the repeated document scores an efficiency of 1.0, higher than the full text's efficiency of 0.79. The repeated text contains as much information as is possible for a document with only 2 words. Even if a random variable contains less information than another, it may be more efficient by virtue of being more uniformly distributed over the events it has.

\subsection{Conditional Entropy}

\begin{quote}
\makesans
	The amount of variation with respect to a single feature in data. Analogous to the entropy of a distribution when a certain feature is true -- it tells us how much information about that feature is contained in a distribution
\end{quote}

\noindent Conditional entropy builds on top of conditional probability. If we know that our data exhibits certain features we can calculate the probability distribution over events, when a feature is true. For a distribution $X$ that describes word frequencies in text data which is a mixture of fiction and non-fiction documents, the conditional distribution $P(X|\text{fiction})$ gives word probabilities using only the fictional portion of the data. Accordingly the entropy $\mathcal{H}
(X|\text{fiction})$ tells us how much information is in the fiction data alone, separate from non-fiction. The entropy of a conditional distribution is a conditional entropy:

\begin{equation}
    \mathcal{H}(X|\text{label}) = \sum_{x_i\in \mathcal{X}} -p(x_i|\text{label})\log p(x_i|\text{label})
\end{equation}

\noindent Where \emph{label} refers to a known feature label for the data. This quantity is useful because it allows you to understand how parts fit together into a whole -- it's a key building block of the approach taken in later chapters. To get a better intuition of how it works, let's say that we have a library containing selected texts from the following authors:

\begin{itemize}
	\item William Shakespeare (1564-1616): \emph{King Lear, Hamlet, Titus Andronicus, Twelfth Night, As You Like It}
	\item Christopher Marlow (1564-1593): \emph{Doctor Faustus, Tamburlaine}
	\item Raymond Carver (1938-1988): \emph{What We Talk About When We Talk About Love}
\end{itemize}

\noindent We want a single distribution that describes the entire library.
Again, there are many ways to do this, for simplicity we opt to create a distribution reflecting word-level information, where each event is a word and its probability reflects word frequency. We'll call this distribution for the entire library $P(words)$.

The entropy $\mathcal{H}(words)$ is 7.2 nats, its efficiency is 0.76 (shown in table \ref{tab:conditional_entropies}). This reflects the degree of variation in word use across the entire library. If all the words in the library were more uniformly used efficiency would be closer to 1, if only a few of the words were reused over and over the efficiency would be closer to 0. Computing the conditional entropy of words given each each of the individual books $\mathcal{H}(words|title)$ reflects how much information is encoded in each title, or how much word use varies in a given title. As shown in table \ref{tab:conditional_entropies} the entropy of each title is less than the overall $\mathcal{H}(words)$, but not by much - with the lowest entropy text $\mathcal{H}(words|\text{what we talk about...}) = 6.457$. This tells us that most word frequency information is shared across texts, which makes sense given these are all in the same language and we wouldn't necessarily expect words like \emph{the, be, a, and} to appear dramatically less often in any of them. But the fact that the overall entropy is higher than any individual text tells us that there is information that they don't share which is added by combining the texts together. Importantly we also know more than one thing about the source documents - for example, we know their authors. We can just as easily compute the conditional entropy $\mathcal{H}(words|author)$. This groups together the 5 Shakespeare plays into a single distribution and tells us how much word choice varies in the collection of plays we have by them for each author, or how much information is that collection. 

\drawline

\noindent In the general case we can define sets of labels that describe different values for a feature in the data a distribution describes. These are used in conditional entropies, and we take the entropy of a set as the average entropy across the constituent labels. 

\begin{equation}
    h(\mathcal{X}|\text{set}) = \frac{1}{|\text{set}|} \sum_{\text{label}\in \text{set}} \mathcal{H}(\mathcal{X}|\text{label})
\end{equation}

\noindent We can use this to look at variation in data at different levels of abstraction based on what we know about the texts that comprise the library. Some examples of sets of labels we could consider:

\begin{itemize}
	\item Title: \emph{King Lear, Hamlet, Titus Andronicus, Twelfth Night, As You Like It, Doctor Faustus, What We Talk About When We Talk About Love}
	\item Author: \emph{Shakespeare, Marlow, Carver}
	\item Genre: \emph{Tragedy, Comedy, History}
	\item Century: \emph{16th, 20th}
\end{itemize}

\thinline

\begin{table}[t]
    \centering
    \rowcolors{2}{white}{blue!3}
    \begin{tabularx}{\linewidth}{l|XXX}                        
    
    \nosbold{Library} &  $\mathcal{H}(words)$ & $\mathcal{E}(words)$ &  \\
    \midrule
    
    \scriptsize{library} & 7.1591 & 0.7685 &  \\

    \midrule
    \nosbold{Title} & $\mathcal{H}(words|title)$ & $\mathcal{E}(words|title)$ & $I(words, title)$ \\
    \midrule
	\scriptsize{King Lear} & 6.4703 & 0.6946 & 0.6888 \\
	\scriptsize{Hamlet} & 6.4269 & 0.6899 & 0.7322 \\
	\scriptsize{Titus Andronicus} & 6.4310 & 0.6904 & 0.7280 \\
	\scriptsize{Twelfth Night} & 6.3090 & 0.6773 & 0.8501 \\
	\scriptsize{As You Like It} & 6.2584 & 0.6718 & 0.9007 \\
	\scriptsize{Doctor Faustus} & 6.3583 & 0.6826 & 0.8008 \\
	\scriptsize{Tamburlaine} & 6.3995 & 0.6870 & 0.7596 \\
	\scriptsize{What We Talk About ...} & 6.0382 & 0.6482 & 1.1209 \\
	\midrule
    \nosbold{Author} & $\mathcal{H}(words|author)$ & $\mathcal{E}(words|author)$ & $I(words, author)$ \\
    \midrule
    \scriptsize{William Shakespeare} & 6.4603 & 0.6935 & 0.6987 \\
\scriptsize{Christopher Marlow} & 6.4683 & 0.6944 & 0.6908 \\
	\scriptsize{Raymond Carver} & 6.0382 & 0.6482 & 1.1209 \\
    \bottomrule
  \end{tabularx}
  \caption{Entropies, efficiencies, and mutual informations for our library of texts. Conditional entropies are shown for two different kinds of conditioning labels, title and author. }
  \label{tab:conditional_entropies}
\end{table}

\section{Mutual Information}

\begin{quote}
\makesans
	How much variation we can explain in terms of a property of the data. Reflecting the reduction in entropy when a given label is true, or how much knowing a label tells us about data.
\end{quote}

\noindent Mutual information is related to both entropy and conditional entropy. It tells us how much we reduce the overall entropy by knowing a conditioning label. It's computed by taking the difference between the overall entropy and the conditional.

\begin{equation}
    I(\mathcal{X}, \text{label}) = \mathcal{H}(\mathcal{X}) - \mathcal{H}(\mathcal{X}|\text{label})
\end{equation}

\noindent This tells us the relationship between a distribution and its subset. Given we can look at conditional entropy as reflecting the degree to which a property varies, mutual information quantifies the inverse - how regular the data is with respect to a property, or how aligned a distribution is with respect to a label. In the library example the distribution contains word level information, so a mutual information $I(words, shakespeare)$ reflects how predictable Shakespeare's word choice is. $I(words, shakespeare)$ would be maximised if Shakespeare used only one word in all his plays, meaning just knowing the play was by Shakespeare would tell us everything there is to know about which words it contains. When maximised the author label \emph{Shakespeare} would be monotonically aligned with a single word, with degree of alignment decreasing as the author's word choice varies more. 

In practice mutual information for all 3 authors in our library is relatively low, indicating somewhat unsurprisingly that they each use well more than one word in their writing. Raymond Carver has higher mutual information $I(words,carver)$, than the other authors indicating that his word choice is more predictable (less variable) than the library in general. This could reflect something stylistically about modernist writing of the later 20th century being less verbose than iambic verse from 1608. Alternately it could be driven by the fact that the library is predominantly texts from 400 years before Carver. As a result $P(words)$ likely reflects Shakespeare and Marlow's use of words like \emph{anon, assay, dost, doth, hark, thee}, and the lower $I(words,carver)$ may reflect Carver being more aligned with a subset of words still in use in the late 20th century.

\begin{figure}
	\centering
	\input{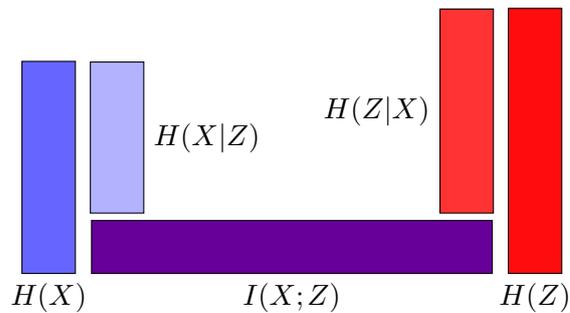}
	\caption{Relationships between basic information-theoretic quantities. Given two spaces $X$ and $Z$, their entropies $H(X)$ and $H(Z)$ are the information each of them contains. Mutual information $I(X,Y)$ reflects the information they share - how monotonically aligned they are. The conditional $H(X|Z)$ reflects the information unique to $X$ not contained in $Z$ and $H(Z|X)$ is the information unique to $Z$.}
	\label{fig:basic_h_quantities}
\end{figure}

\section{Relationships Between Entropy, Conditional Entropy, and Mutual Information}

These three quantities fit together to describe how two distributions relate to each other. Shown in figure \ref{fig:basic_h_quantities}, for two distributions $P(X)$ and $P(Z)$, or in the current example a distribution over words based on the entire library $P(words)$, and a distribution over authors $P(author)$ where each event is an author's name and probability reflects the number of words in the library written by that author. $H(words)$ and $H(author)$ describe the amount of information in each of them. $I(words,author)$ describes the amount of information they share - how predictive knowing the author of a text is of the words it contains. The conditional $H(words|author)$ - discussed above - reflects the variation in each author's word choice, or - as a reciprocal to mutual information - the amount of information about an author's word choice we can't determine just by knowing the author. We can also compute a condition in the other direction $H(author|words)$ which tells us for a given word the variation in which author uses it - maximised when that word is equiprobably used by all authors.

We can look at this as an example of a simple mapping, between authors names and the words they write. Using these basic concepts we can tell how much information is preserved moving between spaces (mutual information), and how much information is unique to each space (conditional entropy). In some cases though we want to be able to tell how much information is unique to each in a set of labels - like how different the word choices are for different authors - without computing the conditional $H(author|words)$. For this we can use a divergence.

\section{Jensen-Shannon/Lambda Divergence}

\begin{quote}
\makesans
	How separable a set of distributions are from each other. How much the information in different distributions overlap.
\end{quote}

\noindent Often we have a number of different distributions and we want to know how similar they are to each other; if their information overlaps or is fully separable. There are a number of ways of assessing this; here because we need to tell apart a number of distributions we opt for the Multivariate Jensen Shannon Divergence, sometimes called the Lambda Divergence. This computes a mixture of the distributions $M$ by taking a weighted mean of the distributions we're comparing. In the general case laid out above for a set of labels this is the sum of the conditionals $P(Z|\text{label})$ each weighted by the probability of the label $P(\text{label})$.

\begin{subequations}
	\begin{equation}
		M \propto  \sum_{\text{label}}^{\text{set}} P(\text{label})P(Z|\text{label})
	\end{equation}

\noindent Given this mixture we try to explain the information it contains $H(M)$ in terms of the information in each component of the mixture $H(Z|\text{label})$ weighted again by how much that component contributed $P(\text{label})$ --- shown below. The resulting quantity is bounded by the entropy of the mixture $H(M)$ and so can be bounded to lie between 0 and 1. As values approach 1 component distributions overlap less, and as the divergence approaches 0 the component distributions become identical. This is implicitly the mutual information between the mixture distribution $M$ and the weights used to combine distributions $P(\text{label})$. If the components of the mixture don't overlap then the mixture weights will explain all the information in the mixture - if components do overlap then $M$ will be less aligned with the weights used to compute it. Two example cases are visualised in figure \ref{fig:js_div}.

\begin{figure}
	\centering
	\input{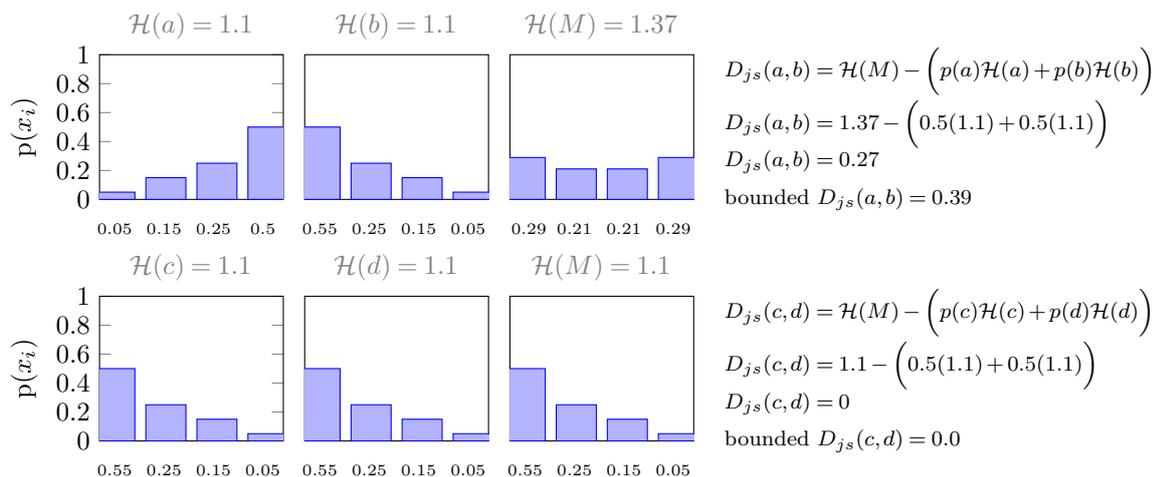}
	\caption{Jensen Shannon divergences computed for two sets of distributions \{a,b\} and \{c,d\}. In both cases we compute a mixture distribution $M$ by taking an unbiased mean --- as noted in body text, this mixture can be weighted by the probability of each of the component distributions, we opt to weight them equally here for simplicity. The JS divergence then looks to see if we can explain the information in the mixture distribution in terms on the component distributions. a and b, overlap somewhat but are distinct - as a result the JS divergence is 0.27, or 0.39 when bounded to lie between 0 and 1. By contrast c and d are identical, and so their JS divergence is 0.0.}
	\label{fig:js_div}
\end{figure}

	\begin{equation}
	    D_{JS}(Z || \text{set}) = H(M) - \sum_{\text{label}}^{\text{set}} P(\text{label})H(Z|\text{label})
	\end{equation}
\end{subequations}

\noindent It's worth noting that this computes the divergence between the component distributions and their mixture, rather than comparing the distributions individually. We can use this to tell how separable sets of labels are from each other, taking the example above we can look at how separable the word frequency distributions are for each author. $D_{JS}(Z || \text{[shakespeare, marlow, carver]}) = 0.534$, indicating the distributions for the 3 authors are relatively separable. Note that in this case because the conditionals include all the data in the library the mixture ends up equalling the library distribution $P(M)=P(words)$. If we instead estimate $D_{JS}(Z || \text{[shakespeare, marlow]}) = 0.5263$ or $D_{JS}(Z || \text{[shakespeare, carver]}) = 0.5845$ we can see word choice is more similar for Shakespeare and Marlow, than for Shakespeare and Carver.

These are the 4 information theoretic quantities we need for the remainder of the thesis, so to summarise:

\thinline
\begin{itemize}
\makesans
	\item[i] \textbf{Entropy:} The amount of variation in data
	\item[ii] \textbf{Conditional Entropy:} The amount of variation in a subset of data
	\item[iii] \textbf{Mutual Information:} The reciprocal of Conditional Entropy, reflecting how much less variable a subset is than overall. By extension how predictable a subset of data is.
	\item[iv] \textbf{Jensen Shannon Divergence:} How separable subsets of data are from one another. The degree to which their information does not overlap (1 indicating no overlap).\end{itemize}

\drawline

\noindent I have introduced these quantities here in the discrete case; chapters \ref{chapt:learning}, and \ref{chapt:llm}, consider information theory for continuous spaces. First though, the next chapter draws explicit parallels between different conditional entropies in a discrete-to-discrete mapping and different kinds of linguistic variation.

\chapter[What We Talk About When We Talk About Compositionality]{What We Talk About \\ When We Talk About Compositionality}
\label{chapt:comp_reg}

{\makesans
\emph{
Variation in Discrete $\to$ Discrete} \\[1em]
}

{\makesans
\begin{quote}
\censor{Doctor,} you say there are no haloes around the streetlights \censor{in Paris and} what I see is an aberration \censor{caused by old age, an affliction.} I tell you it has taken \censor{me.......} all my life \censor{to arrive at the vision of gas lamps as angels,} to soften and blur and finally banish the edges you regret I don't see \censor{to learn that} \\
\flushright{\emph{- Lisel Mueller}}
\end{quote}}

\drawline

\noindent Neural Networks are known for finding solutions to complex problems, but not always the solutions we'd expect. %
A model trained to predict whether or not a patient had pneumonia based on their chest x-ray appeared to do so with remarkable accuracy, until a meta-analysis \citep{zech2018variable} noticed that each x-ray has information in it indicating which scanner and which hospital it came from (most notably from a metal tag radiographers place on the patient's shoulder). In the training data different hospitals had different prevalences of pneumonia, meaning you can predict whether or not a patient had the disease with relatively high fidelity based on where the x-ray was performed. Rather than learning to identify if a patient's lungs had damage consistent with pneumonia, the model learned a much simpler solution: identify what hospital performed the scan. Deep-Learning models are most often optimised with back-propagation of error via gradient descent \citep{rumelhart1986learning}. This tries to minimise the model's error with respect to an objective - like classifying scans as healthy or diseased - but provides no supervision for how the model solves that problem. As a result it's often difficult to work out if a model's behaviour is reflective of it having learned a mapping that identifies and preserves the necessary information from its input, or it having found some simpler solution that can mimic that behaviour. Models can easily rely on heuristics - like the co-occurence probability of different input features - to perform well on a task without learning the properties of their training data we'd expect them to \citep{mccoy2019right}. Understanding what representational information drives models' behaviour remains a major challenge - across domains - when trying to draw conclusions from experiments with deep-learning.

As training large-scale neural networks became more tractable in the past decade, a series of papers started using them to replicate earlier work on the origins of human language \citep{lazaridou_multi-agent_2017, kottur_natural_2017, lazaridou_emergence_2018, choi_compositional_2018, mordatch_emergence_2018}. Given that language leaves behind no fossil record, linguists often turn to computational simulations to study how linguistic systems can emerge in a population. Previously, simple probabilistic models gave an account of how structural properties of language like compositionality can emerge in response to the dynamics of transmission and use rather than natural selection on the language faculty \citep[e.g.][]{kirby_spontaneous_2001, brighton_language_2005} or by processes of biological evolution \citep{nowak_evolution_2000} or gene-culture co-evolution \citep{smith2003complex}. \citet{lazaridou_emergence_2018} implemented a multi-agent model with two neural networks playing a signalling game where a sender network maps a meaning to a discrete signal, a receiver network then tries to map this signal back to the original meeting, in a high-level analogue to communication. Both are then optimised for `communicative success' - to have the receiver's reconstruction match the original meaning as often as possible. Using this setup senders and receivers could reliably converge to near-perfect communication on both the examples they saw during training, and on thousands of unseen examples.

Despite this, the mappings that emerged showed limited evidence of compositional structure -- an essential component of generalisation in natural language. %
Qualitative analyses \citep{choi_compositional_2018, havrylov_emergence_2017}, try to identify certain subunits of of signals that corresponded to specific features in the input space - but it's difficult to manually look for structure in thousands of strings of characters. Quantitative assessments of compositionality showed the mappings scored well below any `idealised' compositional systems used for reference (\textcite{brighton2006understanding} used in \textcite{lazaridou_emergence_2018}, and \textcite{resnick_capacity_2020}), and that `degree of compositionality' had no correlation with generalisation performance \citep{chaabouni_compositionality_2020}. This raises a real question as to whether language-like structure had emerged in the sender's mapping from meaning's to signals, or the model had found a heuristic solution to the communicative task.

Implicitly, when you look for structure in a system, you make some assertions about what structure is and what it should look like. Each method instantiates its own definition of structure which makes proving a null-result difficult: does a system lack compositional structure or does your quantification look for something else? The remainder of this chapter looks at existing methods for identifying compositional structure in models of language emergence, and shows that they actually assess the degree of regularity in a system, not whether or not the system is compositional. As a result they discount mappings with variation as being non-compositional, and by extension indicative of a in-human approach to communication, despite the fact that variation is typologically ubiquitous.

This chapter represents initial experiments in quantifying structure in a mapping. It approaches the problem with less generality than later chapters, focusing on models designed to be directly relatable to human language. As a result it introduces information theoretic measures of four specific kinds of linguistic structure, two structural and two lexical. These are essentially specific instantiations of the one-to-many (variation) and many-to-one (disentanglement) kinds of structure mentioned at the start of the previous chapter (section \ref{sec:struct_prim}). By starting with a discrete-to-discrete mapping in a model of language emergence we can draw clear parallels between the structures we quantify and their analogs in linguistics, which prove useful when we apply these measures to discrete-to-continuous mappings in the next chapter.

\thinline
{
\makesans
\noindent The remainder of this chapter is based around a paper \textbf{Compositionality with Variation Reliably Emerges in Neural Networks} that appeared at the International Conference on Learning Representations in 2023. Authors are myself and Kenny Smith - I conceived of and ran experiments myself, and wrote the paper - Kenny gave writing feedback prior to submission to the conference. The paper is presented here minimally changed from the conference version that underwent peer-review. Changes are largely related to formatting to make the content more readable outside of the original conference paper template.
}
\drawline

\section{Compositionality with Variation Reliably Emerges in Neural Networks}\label{sec:comp_v_reg:introduction}
Compositionality is a defining feature of natural language; the meaning of a phrase is composed from the meaning of its parts and the way they're combined \citep{cann_formal_1993}. This underpins the powerful generalization abilities of the average speaker allowing us to readily interpret novel sentences and express novel concepts. 

Robust generalization like this is a core goal of machine-learning: central to how we evaluate our models is seeing how well they generalize to examples that were withheld during training \citep{bishop_pattern_2006}. Deep neural networks show remarkable aptitude for generalization in-distribution \citep[][]{dong_language_2016, vaswani_attention_2017}, but a growing body of work questions whether or not these networks are generalizing compositionally \citep[][]{kim_cogs_2020, lake_generalization_2018}, highlighting contexts where models consistently fail to generalize \citep[e.g. in cases of distributional shift;][]{keysers_measuring_2020}. 

Recent work has looked at whether compositional representations emerge between neural networks placed in conditions analogous to those that gave rise to human language \citep[e.g.][]{kottur_natural_2017, choi_compositional_2018}. In these simulations, multiple separate networks need to learn to communicate with one another about concepts, environmental information, instructions, or goals via discrete signals - like sequences of letters - but are given no prior information about how to do so. A common setup is a `reconstruction game' modelled after a Lewisian signalling game \citep{lewis2008convention}, where a sender network describes a meaning using a signal, and a receiver network needs to reconstruct that meaning given the signal alone. The resulting set of mappings from meanings to signals can be thought of as a language. 

Previous work has shown that in this setup models reliably develop a language that succeeds not only in describing the examples seen during training but also successfully generalizes to a held-out test set, allowing accurate communication about novel meanings. Despite this capacity to generalize, which is a product of compositionality in natural languages, existing analyses of those emergent languages provide little evidence of reliable compositional structure \citep[see][for a review]{lazaridou_emergent_2020}, leading some to suggest that compositionality is not required in order to generalise robustly \citep{andreas_measuring_2019, chaabouni_compositionality_2020, kharitonov_compo_2020}.

\paragraph*{If not compositional, then what?}
This interpretation leaves us with a major puzzle: if the languages that emerge in these models are non-compositional, how do they allow successful communication about thousands of unseen examples \citep[e.g.][]{havrylov_emergence_2017, lazaridou_emergence_2018}? If the meaning of a form is arbitrary rather than being in some way composed from its parts there should be no reliable way to use such a mapping to generalize to novel examples \citep{brighton_compositional_2002}. Here we provide an answer to this question showing that emergent languages are characterised by {\em variation}, which masks their compositionality from many of the measures used in the existing literature.
Existing measures take regularity as the defining feature of a compositional system, assuming that in order to be compositional separate semantic roles need to be represented separately in the signal \citep{chaabouni_compositionality_2020}, or that symbols in the signal must have the same meaning regardless of the context they occur in \citep{kottur_natural_2017, resnick_capacity_2020}. Alternately they expect that each part of meaning will be encoded in only one way, or that the resulting languages will have a strict canonical word order (\textcite{brighton2006understanding} used in \textcite{lazaridou_emergence_2018}). However, natural languages exhibit rich patterns of variation \citep{weinreich_empirical_1968, goldberg_constructions_2006}, 
frequently violating these four properties: forms often encode multiple elements of meaning (e.g. fusional inflection of person and number or gender and case), language is rife with homonymy (where the meaning of a form depends on context) and synonymy (where there are many ways of encoding a meaning in form), and many natural languages exhibit relatively free word order.

This offers us a different explanation of previous results: compositional systems may emerge, just with variation. If so that doesn't necessarily undermine their compositionality, natural languages show us that systems can have considerable variation while retaining the generalizability that makes compositionality so desirable. We focus on explicitly assessing variation independent of compositionality and illustrate how emergent languages can generalize robustly even with substantial variation. %
Our core contributions are as follows:
\begin{itemize}
    \item We introduce 4 measures of natural language-like variation
    \item We show that the languages which emerge tend to exhibit a high degree of variation which explains why previous metrics would classify them as non-compositional.
    \item We find that a language's degree of regularity correlates strongly with generalization early in training, but as the emergent language becomes \emph{regular enough} to generalize reliably this correlation goes away.
    \item We reduce the capacity of our models by reducing the size of the hidden layers, and show that lower capacity models develop more regular languages, as predicted by accounts linking cognitive capacity and regularity in natural language
\end{itemize}

\section{Variation, Regularity, \& Compositionality}

Variation and compositionality in language are related but distinct. We look at them separately, taking a language's generalization performance as an indication of whether or not it is compositional \citep[in line with][]{brighton_compositional_2002, kottur_natural_2017}. Linguistic regularity - the absence of variation - has been studied in broad array of contexts \citep[see][for discussion]{ferdinand_cognitive_2019}. At a high-level it describes how predictable a mapping from meaning to form is; if there's only one way of encoding a meaning that mapping is highly-regular \citep{smith_eliminating_2010}. Conversely if there's a variety of different ways of encoding a meaning that mapping likely has high variation (low regularity). 
In our context - mapping meanings to discrete signals - regularity is maximized by a language of one-to-one mappings. For example where each position in the signal encodes one part of the meaning – position 1 $\rightarrow$ Subject – and each character in that position refers to only one possible subject – $A$ in position 1 $\rightarrow$ Subject: Ollie – and is the only character ever used to refer to that subject. A maximally regular language encodes the same (part of) meaning with the same (part of) form every time, rather than affording a speaker a variety of ways to encode a meaning.

This kind of maximally regular system is intuitively compostional, given the meaning of a signal would be composed from the parts of meaning its characters map to and the position they're in (in line with \citeauthor{cann_formal_1993}, \citeyear{cann_formal_1993}) but it's by no means the \emph{only} kind of compositional system. To better characterise the space of possible languages in section \ref{sec:measures} we introduce four kinds of variation - drawn from kinds of variation attested in natural language - and ways of quantifying each of them individually. 
Then in section \ref{sec:prior-measures} we look at some of the most relevant existing measures of `compositionality' and discuss how they could be interpreted in terms of regularity. Results from a standard emergent communication model in section \ref{sec:results} show that every run results in a highly-generalizing (and therefore compositional) language but with varying degrees of variation. To better understand the relationship between variation and generalization we look over the time-course of training and find regularity is a strong predictor of how well a language generalizes early on but this effect goes away as the models approach ceiling i.i.d. generalization. We take this as an indication that while a language needs to be more regular than a random mapping in order to generalize,  it doesn't need to minimize variation in order to do so – a point made clear by natural languages. At the end of training when the emergent languages have become sufficiently regular for the task at hand, whether one is more regular than another doesn't necessarily correspond to better generalization.

In a final set of experiments we look at how to decrease the amount of variation in an emergent language. Limitations on humans' memory and cognitive capacity are thought to be a driving force in the emergence of compositional structure and regularity in natural language \citep{kirby_spontaneous_2001, hudson_kam_investigating_2009, smith_eliminating_2010}. Learners with less memory are believed to regularize their input because they are more constrained in their ability to store low-frequency forms \citep{newport_maturational_1990, ferdinand_cognitive_2019}. We reduce the capacity of our models by reducing the size of the hidden layers, and show that lower capacity models develop more regular languages, as predicted by accounts linking learner capacity and regularity in natural language and in line with previous work in this area \citep{resnick_capacity_2020}.

\subsection{Quantifying Variation}\label{sec:measures}

We quantify four kinds of linguistic variation two lexical Synonymy \& Homonymy and two structural Entanglement \& Word-Order Freedom. This is not intended to be an exhaustive list, but offers a starting point for thinking about linguistic variation in this context. Each of these measures is bounded between 0 and 1, where 0 indicates a perfectly regular language with no variation, and 1 represents a maximally variable language. For comparison we generate a maximally regular compositional language which scores near 0 across our measures, and maximally irregular non-compositional language (where each meaning maps to a unique randomly-generated signal) which scores near 1, as shown in table \ref{table:core-results}. Our task (described fully in section \ref{sec:methods}) asks models to map meanings to signals. With meanings comprised of roles - e.g. Subject, Verb, and Object - and semantic atoms which can occur in each role (e.g. Subject: \emph{Ollie, Isla ...} Verb: \emph{loves, hates, ...}). Prior work in this area sometimes refers to these as attribute-value pairs \citep[see][for a review including some mention of attribute-value pairs, p. 11]{lazaridou_emergent_2020}. Similarly signals are comprised of positions (indices), and the character that occurs in each.  We can frame linguistic concepts of variation in terms of how semantics (roles \& atoms) map to signals (positions \& characters). 

\begin{table}
\centering
\setlength{\tabcolsep}{6pt}
\resizebox{\textwidth}{!}{
\begin{tabular}{lrlrlrlrlrlrlrlrlrl}
\textbf{Model}             &  \mc{2}{i.i.d. acc} & \mc{2}{o.o.d. acc} & \mc{2}{synonymy} & \mc{2}{entanglement} & \mc{2}{freedom} & \mc{2}{homonomy} & \mc{2}{variation} & \mc{2}{topsim} & \mc{2}{posdis}\\
\toprule
\ \emph{ideal} &   &   &    &    & \acc{0.000000}{}  & \acc{0.000456}{} & \acc{0.000000}{} & \acc{0.117480}{} & \acc{0.03}{} & \gacc{0.6198330080221915}{} & \gacc{0.998593807220459}{} \\
\ \emph{random} &   &   &    &    & \acc{0.988728}{}  & \acc{0.999410}{} & \acc{0.991180}{} & \acc{0.991339}{} & \acc{0.99}{}& \gacc{-0.0006382553260635669}{} & \gacc{0.0003377813845872879}{}\\

\hline
\msmall & \ACC{97.5375}{0.493245}   & \ACC{72.860741}{7.069696} & \ACC{0.455210}{0.024911} & \ACC{0.535158}{0.03336} & \ACC{0.487803}{0.027775} & \ACC{0.528932}{0.025642} & \ACC{0.501776}{0.026813} & \GACC{0.207710}{0.012232} & \GACC{0.236053}{0.025485}\\
$\Delta$ \emph{best o.o.d.} &   &   &    &    & \acc{-0.202355}{0.025362} & \acc{-0.422817}{0.037430} & \acc{-0.190417}{0.027775} & \acc{-0.183017}{0.027270} & \acc{-0.249651}{0.027272} & \gacc{0.123945}{0.013292} & \gacc{0.217608}{0.028349}\\

\hline
\mmid   & \ACC{97.728123}{0.591060}   & \ACC{82.134232}{3.620366} & \ACC{0.516574}{0.046178} & \ACC{0.600457}{0.069451} & \ACC{0.544073}{0.048955} & \ACC{0.581572}{0.047360} & \ACC{0.560669}{0.051873} & \GACC{0.192656}{0.022502} & \GACC{0.189808}{0.047213}\\
$\Delta$ \emph{best o.o.d.} &   &   &    &    & \acc{-0.127696}{0.046083} & \acc{-0.353916}{0.068404} & \acc{-0.123496}{0.048934} & \acc{-0.125160}{0.049155} & \acc{-0.197348}{0.049150} & \gacc{0.104226}{0.024074} & \gacc{0.170235}{0.043923}\\

\hline
\mbig   & \ACC{97.528122}{0.520974}  & \ACC{81.332218}{3.031506} & \ACC{0.632521}{0.024338} & \ACC{0.796620}{0.041677} & \ACC{0.657835}{0.021945} & \ACC{0.691419}{0.018170} & \ACC{0.694599}{0.024987} & \GACC{0.143740}{0.011139} & \GACC{0.075220}{0.018811}\\
$\Delta$ \emph{best o.o.d.} &   &   &    &    & \acc{-0.008768}{0.032690} & \acc{-0.127469}{0.053858} & \acc{-0.008296}{0.029815} & \acc{-0.027689}{0.017633} & \acc{-0.081847}{0.017633} & \gacc{0.032526}{0.016290} & \gacc{0.045650}{0.021401}\\

\bottomrule
\end{tabular}
}
\setlength{\tabcolsep}{4pt}
\caption{Mean accuracy and variation with 95\% confidence interval across 20 runs, taken from the epoch with the best o.o.d. generalization performance, along with the change in measures $\Delta best$ between the least regular language that occurs between epochs 1 and 10 and the best generalizing one. Also included are the variation measures applied to a perfectly regular and a maximally variable language one as well as an average across all 4 variation measures. Two measures of regularity from previous work (topsim and posdis) are included in the grey cells.
}
\label{table:core-results}
\end{table}

 All four measures start with a probability table that describes the mapping between meanings and signals probabilistically, in terms of a distribution over characters in each signal position given a semantic atom in a role. This encodes, for example, how likely character `A' is in signal position 1 given that `Ollie' is in the subject role of the signal's meaning. We can quantify this as a straightforward conditional probability using maximum likelihood estimation, shown in equation \ref{eq:co_prob}. We estimate this for every atom ($\forall atom_{r,i} \in A_r$) in every role ($\forall r \in R$), looking at every character ($\forall char_{p,j} \in C$) in every position of every signal ($\forall p \in P$). 

\begin{equation}\label{eq:co_prob}
    \mathbb{P}(char_{p, j} | atom_{r, i}) = \frac{count(char_{p, j}, atom_{r, i})}{count(atom_{r, i})}
\end{equation}

\noindent The resulting tensor describes how often each letter occurs in a position, given a certain atom in a role in the meaning (like Subject: Ollie)\footnote{Here we use semantic roles given the meanings are sentences, this can be generalised to any analogous attributes a dataset exhibits.}. This tensor has dimensions semantic roles $\times$ semantic atoms $\times$ max signal length $\times$ characters\footnote{For all experiments reported here these values are 3 $\times$ 25 $\times$ 6 $\times$ 26}, where the last axis is a probability distribution over all possible characters in a given position - here denoted by $\mathbb{P}(char_p | atom_{r,i})$. 

\begin{figure}[tp]
    \centering
    \makesans
    \includegraphics[width=0.7\textwidth]{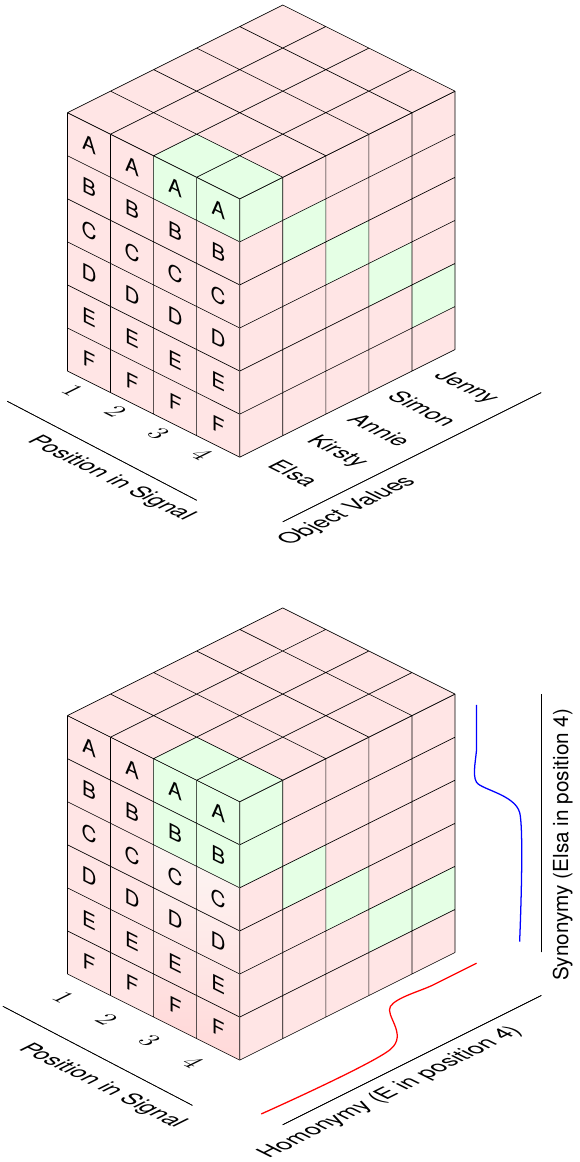}
        \caption{A depiction of the probability tensor built with equation \protect\ref{eq:co_prob} where $r = Object$. Green indicates high probability and red low. \textbf{(Top:)} A perfectly regular language, Elsa is always encoded by `AA' in the final two positions, Kirsty by `BB' etc. \textbf{(Bottom:)} The same cube is shown (object labels removed) for a language with basic synonymy (Elsa can be encoded by `A' or `B') and homonymy (Jenny and Simon are both encoded by `E'). We quantify the degree of synonymy by taking the entropy of each column (equation \protect\ref{eq:synonymy_pt_2}) and the degree of homonymy by taking the entropy of each row (equation \protect\ref{eq:homonomy_pt_2})}
    \label{fig:synonymy}
\end{figure}

\paragraph*{Synonymy \& Homonymy:}
Synonymy is minimised when each atom in a meaning maps to a single character in a position. Homonymy is minimised when each character in a position maps back to a single atom \citep{hurford2003synonymy}. While a perfectly regular compositional language minimises these, natural language is rife with both synonymy and homonymy (e.g. `loves', `adores', `fancies' all map to approximately the same concept; the homonymous `bank' maps to a financial institution, the act of turning a plane, and the land at the side of a river). One-to-many mappings (synonymy) aren't a problem for compositionality, as each different synonym can still be composed with the rest of a signal.  Similarly many-to-one mappings (homonymy) can be used compositionally, with meaning disambiguated by context. Homonyms in natural languages also tend to be contextually mutually exclusive meaning the ambiguity they introduce at a system level tends not to impair communication (thinking of `bank' - it's rare to cash a check while piloting a plane into the side of a river).

In our setting synonymy is how many different characters can refer to an atom in a role. For example when $r=Subject$ and $atom_{r,i}=Ollie$ how many characters have non-zero probability in each signal position? A perfectly regular language where `Ollie' is always encoded by `A' in position 1 would have a probability of 1.0 on `A' in position 1. A maximally variable language would have a uniform distribution over all characters. We can take the entropy over characters in a position $\mathcal{H}(char_p|atom_{r,i})$ as a measure of synonymy in that position (illustrated in figure \ref{fig:synonymy}). We take the position with the lowest entropy as a lower-bound estimate of synonymy for that $atom_{r,i}$.
We bound this measure dividing it by the entropy of a same-sized uniform distribution $\log(n_{char})$. The resulting quantity is an efficiency \citep{Shannon1948AMT} where a uniform distribution scores 1.0 and a one-hot distribution scores 0.0.
A language with no synonymy where each atom is encoded by a single character in a position achieving close to 0, and maximal synonymy where any character can refer to each atom achieving close to 1 (shown empirically in table \ref{table:core-results}).
The synonymy of an entire language $(\mathcal{L})$ is obtained by averaging across all atoms in a role, then across all roles. 

\begin{equation}\label{eq:synonymy_pt_2}
    \textit{Synonymy}(atom) = \frac{\mathcal{H}(char|atom)}{\log (n_{chars})}
\end{equation}

\noindent We measure homonymy in a similar way, looking at how many semantic atoms a character in a position can refer to. As depicted in figure \ref{fig:synonymy} this is akin to applying the synonymy measure to a different axis of the probability tensor $\mathbb{P}$. We estimate $\mathbb{P}(atom_{r} | char_{p,j})$ to get a distribution over atoms given characters in a position\footnote{For simplicity we re-normalize $\mathbb{P}$ to create a probability distribution over atoms in a role which is equivalent to directly computing $\mathbb{P}(atom_{r} | char_{p,j})$ see appendix for further discussion.}. To get a lower-bound estimate of language-level homonymy we take the position with the lowest entropy over atoms, again bound between 0 and 1, then average across all characters and roles. When the resulting value is close to 1 each character maps to every atom. Approaching 0 each character uniquely refers to a single atom. 

\begin{equation}\label{eq:homonomy_pt_2}
    \textit{Homonomy}(char)=\frac{\mathcal{H}(atom| char)}{\log (n_{atoms})}
\end{equation}

\paragraph*{Word Order Freedom} is minimized when each role in the meaning is always encoded in the same position(s) in the signal, resulting in a single canonical word order. Looking at a language like Basque we see that a compositional language can support a number of different grammatical word orders \citep{laka1996brief}, with at least two equivalently valid translations of `Ollie saw Ernest:' \emph{Ollie Ernest ikusi zuen, Ollie ikusi zuen Ernest.} Even in English which has relatively strict word order we see processes like topicalization that result in alternate orders that are equally acceptable \emph{Let's go down to the lake for some fun; For some fun, let's go down to the lake}, or even more commonly dative alternations \citep{chomsky1957logical} like \emph{Ollie gave Orson a book; Ollie gave a book to Orson}. While many languages have some constraints on word-order, even when there is maximal word order freedom the resulting language can still be clearly compositional, with characters encoding the meaning and their order conveying little information.

\begin{figure}[tp!]

\centering
    
        \includegraphics[width=0.7\linewidth]{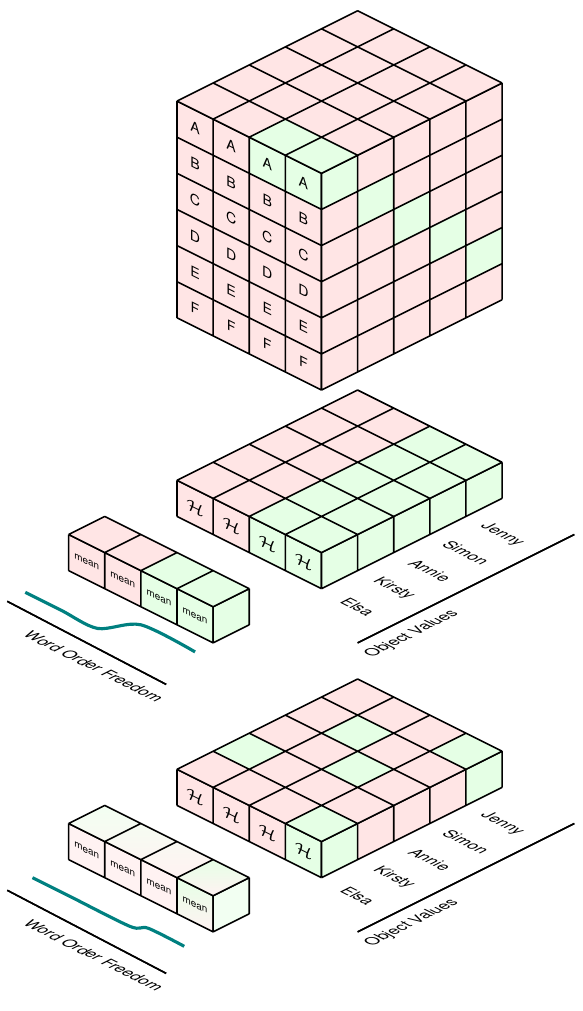}
        
     \caption{
        The Freedom measure applied to a regular and variable language. At top the cube from figure \protect\ref{fig:synonymy} red indicates low probability (high entropy), and green high probability (low entropy). Directly below it is the entropy of each column - when we take the mean of these column entropies across atoms in a role a non-uniform distribution indicates semantic roles are disentangled.
        }
    \label{fig:freedom}

\end{figure}

A language with free word order is equally likely to encode any $role \in R$ in any position, while a maximally regular language always encodes atoms from the same role in the same position(s). If a given $atom_{r,i}$ is not encoded in a position we expect its distribution over characters to be roughly uniform. So we can take the entropy for each position $(\mathbb{P}(char_p|atom_{r,i}) ) : \forall p \in P )$ (also computed as part of equation \ref{eq:synonymy_pt_2}), and average across all atoms in that role $\forall i \in A_r$. If all the atoms in a role are encoded in the same position the distribution resulting from the mean will be non-uniform, with some positions having lower mean entropy than others (illustrated in figure \ref{fig:freedom}).

\begin{subequations}
\begin{align}\label{eq:freedom_pt_1}
    \mathbb{F}(role_r) &= \frac{1}{|r|}\sum_{i=1}^{|r|}\mathcal{H}(char|atom_{r,i})
\end{align}

 \noindent To get a lower-bound estimate of the language-level word-order freedom we take the minimum from the mean distribution $\mathbb{F}(role_r)$ and bound between 0 and 1, then average across all roles:

\begin{equation}\label{eq:freedom_pt_2}
    \textit{Freedom}(\mathcal{L}) = \frac{1}{|R|}\sum_{r=1}^{|R|} \frac{min \, \mathbb{F}(role_r)}{\log(n_{char})}
\end{equation}
\end{subequations}

\begin{figure}[tp!]

\centering
    
 \makesans
        \includegraphics[width=0.7\linewidth]{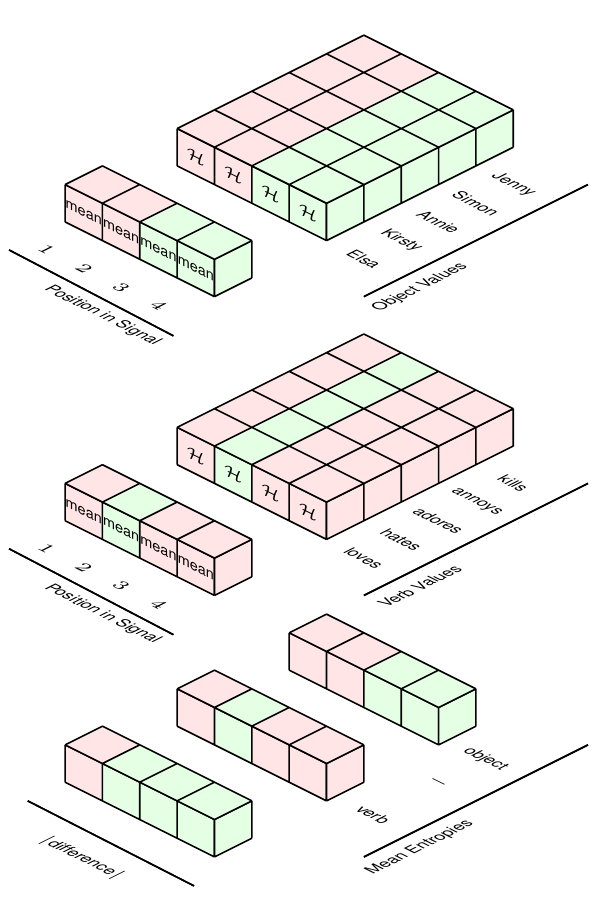}
        \vfill\null
     \caption{
        Column entropies for two different roles, object and verb. The mean of both are taken and then compared by taking the absolute value of the difference of the two averaged entropies. If two roles are encoded in the same part of the signal then their difference will be close to zero: they both have consistently similar in the same parts of the signal. The maximum of the difference is then divided by the maximum of the mean distribution for the respective roles before they were subtracted - this bounds the measure between 0 and 1 and makes the quantity reflect the proportion of overlap between roles.
    }
    \label{fig:entanglement}

\end{figure}

\paragraph*{Entanglement} is minimised when each role is encoded in different positions in the signal. While a dis-entangled language is likely compositional, consider the English past tense form of `go.' `Went' is irregular, encoding action and tense together, in contrast to the hypothetical regular form `goed' where action and tense are encoded in separate parts of form \citep[][p. 55]{anderson_-morphous_1992}. Despite this we can go on to use the entangled form `went' compositionally in a sentence: \emph{Ollie went down to the shore} \citep[for discussion][p. 105]{odonnell_productivity_2015}. While maximal entanglement where every role is encoded in every position would be non-compositional, the existence of even a high degree of entanglement does not preclude compositionality, given the entangled forms can be straight-forwardly recomposed with others. We can quantify this by seeing if two roles are consistently encoded in the same (or different) positions. We compare the means $\mathbb{F}(role)$ from equation \ref{eq:freedom_pt_1} for each possible pair of roles $r_i, r_j \in {}^RC_2$ by taking the magnitude of their difference, if two roles are encoded in the same position the result will be close to zero. If the roles are maximally disentangled then the result will be close to the $max(\mathbb{F}(role_i), \mathbb{F}(role_j))$ for that position. To get a lower bound estimate of two roles' entanglement we take the maximum of the difference and divide by the pre-difference max. When the resulting value approaches 0 all roles are mapped to different parts of the signal, as it approaches 1 all roles are encoded in the same positions (illustrated in figure \ref{fig:entanglement}).

\begin{equation}\label{eq:entanglement_pt_1}
    \textit{Entanglement}(\mathcal{L}) = 1 - \frac{1}{|{}^RC_2|}\sum_{r_i, r_j}^{{}^RC_2}\frac{max\Bigl(|\mathbb{F}(role_{i}) - \mathbb{F}(role_{j})|\Bigl)}{max\Bigl(\mathbb{F}(role_i), \mathbb{F}(role_j)\Bigl)}
\end{equation}

\subsection{Existing Measures}\label{sec:prior-measures}

\paragraph*{Topographic Similarity} (Topsim) \citep{brighton2006understanding} has been used as a measure of compositionality in a wide array of contexts \citep[e.g.][]{smith2003complex, kirby_cumulative_2008, lazaridou_emergence_2018}. It assumes that in a compositional system where a whole signal is composed from reusable parts, similar meanings will map to similar signals. This can be assessed by measuring the correlation between pairwise meaning-distances and edit distances between their associated signals: a perfectly regular compositional language without variation achieves a correlation score close to 1, while a non-compositional (random) mapping between meanings and signals achieves a correlation close to 0. While languages that score highly are likely to be compositional synonymy and word order freedom can reduce the score for this measure, as they can result in similar meanings having dissimilar signals. Synonymy can mean two meanings with the same subject encode it with different characters. Freedom can mean signals for similar meanings with different word orders have high edit-distance despite containing many of the same letters. 

\paragraph*{Posdis \& Residual Entropy} \citep{chaabouni_compositionality_2020} \& \citep{resnick_capacity_2020} provide entropy-based measures of `compositionality.' Posdis captures the extent to which each position of the signal univocally refers to a role in the meaning (e.g. subject, object, verb) and looks for each signal position to refer to only one role. This is similar to what our entanglement measure assess (though computed differently). 
Similarly, residual entropy assesses the degree to which a sub-string of the signal encodes a single atom in a role (e.g. Ollie in the Subject) and is minimized when a sub-string refers to only one atom in a role. This requires there to be minimal homonymy and entanglement in a subset of the signal (across 1 or more positions), with each unique sub-string in those positions referring to only one atom in a role. As discussed above natural language shows us that even a high degree of homonymy and entanglement in a language doesn't preclude its compositionality.
We show empirically in table \ref{table:core-results} that a maximally regular language maximizes topsim and posdis while minimizing residual entropy (for brevity residual entropy results are deferred to appendix \ref{appendix:residual-entropy}). Like topsim, languages that score highly on these measures are very likely to be compositional - the issue is that they take some kinds of variation as evidence of non-compositionality.

\section{Methods}\label{sec:methods}

\paragraph*{Models} We implement a reconstruction game with a sender network and receiver network. The overall architecture used is intentionally similar to \citet{chaabouni_compositionality_2020, resnick_capacity_2020} and \citet{guo_expressivity_2021} to allow comparison of results.
The sender network is comprised of an embedding layer, linear layer, and a GRU \citep{Cho2014OnTP} - the receiver architecture is the inverse. A linear layer is used as the input is of fixed length, so can be presented at once as a one-hot encoding - while a GRU spells out the variable-length signal a character at a time. The maximum signal length used here is 6, with 26 characters available to the model in each position. 
The sender is optimized using REINFORCE \citep{williams1992simple} due to the discrete channel, while the receiver is optimised using ADAM \citep{kingma2014adam}. Models are implemented using pytorch \citep{paszke2019pytorch}, and make use of portions of code from the EGG repository \citep{kharitonov_egg_2019}. Full hyperparameters for the experiments presented here can be found in appendix \ref{appendix:hyperparams}.

\paragraph*{Data}
The sender is shown examples drawn from a meaning space of two place predicates (e.g. \emph{Ollie loves Osgood}) generated using a context free grammar, with three roles: subject, verb, and object and 25 atoms per role, resulting in a total of 15625 examples. This is equivalent to the attribute, value setup used in previous work \citep{resnick_capacity_2020, chaabouni_compositionality_2020}.
Data is divided into 4 splits for training: 60\%, validation 10\%, i.i.d. testing 10\%, and o.o.d. testing 20\%.  %

\paragraph*{O.O.D. Evaluation} Previous emergent communication work typically evaluates generalization on an in-distribution held out test-set. In order to better align our findings with the broader literature on compostional generalization in neural networks \citep[e.g.][]{lake_generalization_2018, kim_cogs_2020} we implement a version of the maximum compound divergence (MCD) algorithm from \citet{keysers_measuring_2020}, and report results for both in-distribution generalization, and out-of-distribution generalization to an MCD split. Additionally we use an O.O.D. split because models often converge to ceiling i.i.d. performance, which potentially makes it difficult to look for correlations between generalization performance and attributes of the language, like regularity. For our \msmall model i.i.d. performance $95\%$ confidence intervals are $\pm 0.49\%$ while o.o.d. performance is $\pm 7.07\%$, allowing a broader range of values with which to look for correlations (we include the same analyses on i.i.d. performance in appendix \ref{appendix:iid-correlation} and in practice i.i.d. and o.o.d. results are very similar).

\paragraph*{Capacity}

 We look to see if models with less capacity arrive at more regular languages than their larger counterparts as predicted by work in natural language \citep[e.g.][]{hudson_kam_investigating_2009}. We vary model capacity by varying the size of the hidden layers used by the model 
reporting and comparing results of three different model sizes \emph{small, medium,} and \emph{large} with hidden layer sizes 250, 500, and 800 respectively. 

\begin{figure}[t!]
    \centering
    \includegraphics[width=1.0\textwidth]{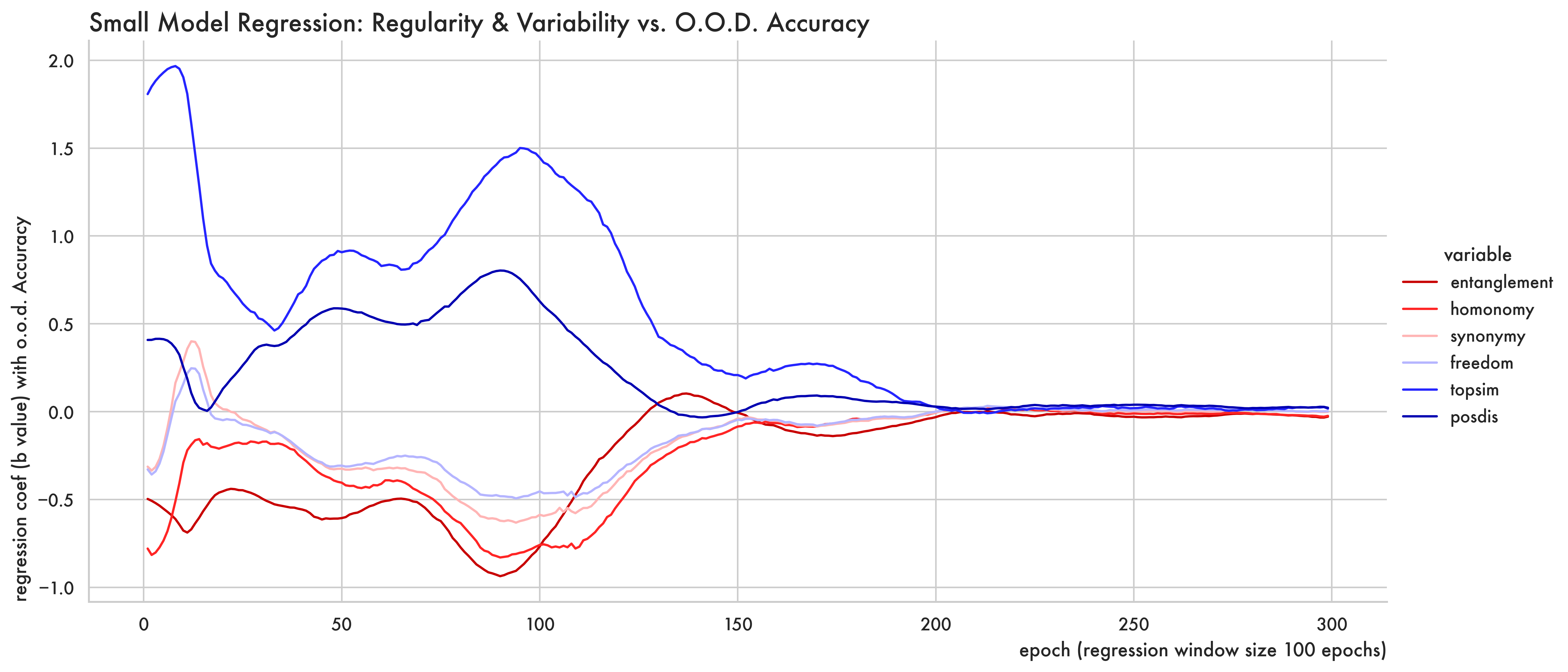}
    \caption{A model is fit to a sliding window of data from 100 epochs at a time across 20 initializations between o.o.d. accuracy and each measure of variation. Shown are the regression coefficients (b values) of our four measures of variation, and two previous measures of regularity (topsim and posdis) with o.o.d. generalization accuracy for the \msmall model for each window. 
    }
    \label{fig:correlations}
\end{figure}
\section*{Results and discussion} \label{sec:results}

Our results for all model sizes are summarised in table \ref{table:core-results}. As stated in section \ref{sec:comp_v_reg:introduction} a language must be compositional in order to generalize, in line with previous work in this area \citep[][]{kottur_natural_2017} and in linguistics \citep{brighton2006understanding}. All versions of our model get near ceiling i.i.d generalization and robust o.o.d. generalization indicating a compositional system. Compositionality and variation are related, but distinct; while a system needs to be more regular than a completely random mapping in order to generalize compositionally it does not need to be perfectly regular. Natural languages show us that a system can support a high-degree of variation while remaining compositional. In line with this in all conditions of our model the language that emerges is substantially more regular than a random mapping, but more variable than a perfectly regular language of one-to-one mappings.

\paragraph*{The relationship between regularity and generalization}
We use linear mixed effect models to evaluate the relationship between our four measures of variation and o.o.d. performance, fitting a model on rolling windows covering the time course of training (implementation details in appendix \ref{appendix:mixed-effects}). The resulting regression coefficient (b value) for a window indicates how strong a predictor our measures are of generalization performance over that period of training. As shown in figure \ref{fig:correlations} early on a language's regularity is a strong predictor of how well it generalizes, but later in training this effect goes away. This is consistent with the idea that some regularity is needed for generalization, but maximal regularity is not required. Later in training, as a language emerges that is \emph{regular enough} to succeed at the task (achieving ceiling i.i.d. generalization performance), the relationship between regularity and generalization trends toward non-significance. Supporting this we see languages become more regular over time with a negative relationship between training step and variation ($b=-0.038, p<1e-10$) – in table \ref{table:core-results} we also see that in every condition the model decreases the variation in its language between early training and the best generalizing epoch indicated by a negative value for $\Delta$ \emph{best o.o.d.}. A limitation of these results is that the language for every run is still highly-variable (with the lowest mean variation score of any run being 0.43), possibly because the task here is quite simple in comparison to compositional generalization datasets in other domains \citep[e.g.][]{kim_cogs_2020}. As languages approach maximal regularity, regularity may again be a strong predictor of generalization performance - but given none of our models approach minimal variation this remains an open question (further discussion of these results in appendix \ref{appendix:oodregression-discussion}).

\paragraph*{How can a variable language still be compositional?} Figure \ref{fig:freedom_raw_data} helps us to understand what these highly variable but robustly generalizing languages look like. It visualizes the word order for the run of our \msmall model with the \emph{highest} word order freedom - meaning all other runs of that model exhibit even stricter word order. It shows that while the language is still much more variable than a perfectly regular one (this language has $freedom = 0.57$, a compositional language with fixed word order has $freedom = 0$), it nonetheless exhibits a high degree of word order regularity, with verbs most likely encoded at the start of the signal, subjects in the middle, and objects at the end, but with each individual atom sometimes being encoded slightly differently. Given compositionality requires the meaning of a whole to be a function of its parts the pattern seen here where each role is encoded in part of the signal appears to meet that threshold despite its high variation.

\begin{figure}[tp]
    \centering
    \includegraphics[width=1.0\textwidth]{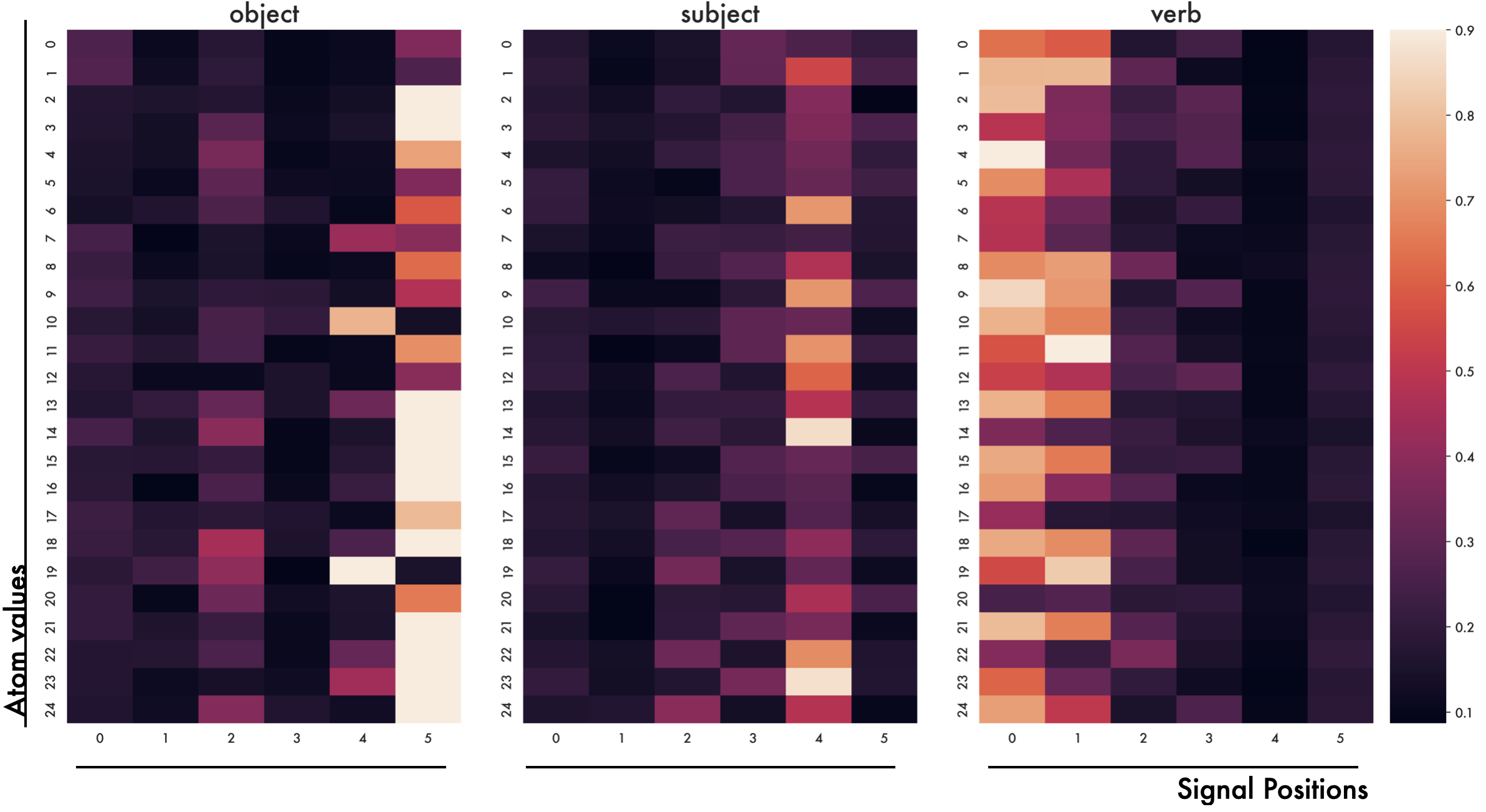}
    \includegraphics[width=1.0\textwidth]{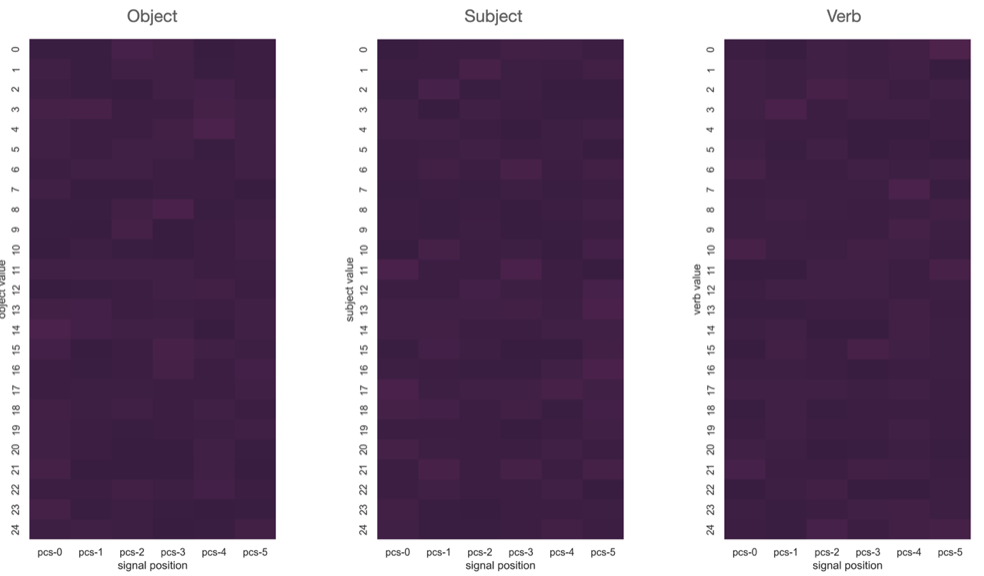}
    \includegraphics[width=1.0\textwidth]{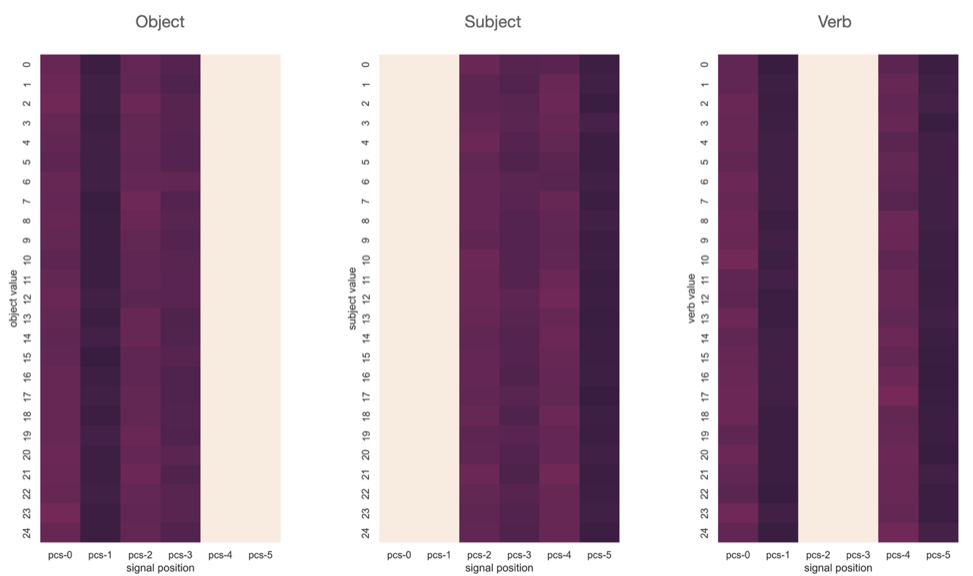}
\caption{\emph{Caption Opposite}}

\end{figure}
\addtocounter{figure}{-1}
\begin{figure} [t!]
  \caption{Plots showing the max from the distribution over characters for each atom in each position, with a plot for each separate role (object, subject, verb). x axis: positions, y axis: id for each $atom_i \in A_{r}$. Shown to the left are these plots for the synthetic ideally regular compositional language (with SVO order), and the maximally variable random mapping. The large plot shows data from the run of the \msmall model with the \emph{highest} variation
    This run's variation scores: $freedom=0.57, entanglement=0.61, homonymy=0.61, synonymy=0.51, topsim=0.28, posdis=0.26$}
    \label{fig:freedom_raw_data}
\end{figure}

\paragraph*{Capacity effects regularity}
with an increase in the number of trainable parameters resulting in an increase in variation across all measures with \mbig arriving at significantly more variable languages than \msmall or \mmid ($p<0.05$). Spearman correlations show model size does not correlate significantly with o.o.d. accuracy ($p=0.24$) but correlates with synonymy ($r=0.67$), word order freedom ($r=0.69$), entanglement ($r=0.68$), and homonymy ($r=0.68$) indicating larger models develop more variable languages (all of which are significant $p<0.00001$). 
This result is in line with work that points to constraints on human cognition as a key driver of regularization in natural language, suggesting that similar factors shape the regularity of emergent communication in neural networks. Previous work studying the effect of network capacity on emergent languages \citep{resnick_capacity_2020} found that while most model sizes could generalize well, larger models could do so using a `non-compositional code' indicated by a higher residual entropy measure (which has similarities to our measures of homonymy and entanglement). This is consistent with our results, although we believe this indicates that larger models develop a language characterised by greater variation rather than non-compositionality (residual entropy scores for our model can be found in appendix \ref{appendix:residual-entropy}).

\paragraph*{Framing prior results in terms of regularity} Existing measures (topsim and posdis) correlate negatively with model size ($r=-0.63$, $r=-0.71$) strongly suggesting that rather than tracking compositionality these measures implicitly track the degree of regularity in a language, especially given that the magnitude of their correlation coefficient is similar to that of our measures that explicitly assess variation. This helps us to interpret results suggesting compositionality doesn't correlate with generalization: if these measures assess regularity instead we know a wide array of languages can be regular enough to generalize well without needing to maximize regularity to do so.

\section{Conclusion}
Neural networks reliably arrive at compositional languages when natural language-like variation is taken into account. 
Previously these languages' compositionality has been assessed on the basis of their regularity, but 
natural languages show us a system can be rich with variation while retaining the generalizability that makes compositionality so desirable. 
Similar to natural language the capacity of learners is a key driver of the degree of regularity that emerges. By accounting for variation we can see striking similarities between the structure of the languages that emerge and structures in natural language.

\thinline

\section{Epilogue}

The experiments in this chapter lay out a couple of key premises that are built on moving forward. Namely that

\begin{itemize}
	\item Conditional entropies can describe variation in a mapping
	\item Applying conditional entropies at different levels of abstraction, like lexical or structural, can shed light on how a mapping is structured.
	\item Capacity has a regularising effect on model behaviour, with smaller models producing more regular languages.
	\item Generalisation performance isn't directly related to regularity - a system doesn't need to minimise variation in order to generalise.
\end{itemize}

\noindent The model in this chapter is intended to have a clear relationship to communication in humans. As a result the structures discussed are introduced in terms of their linguistic analogs. While the remainder of the thesis focusses on mapping structure in a more general way, I think starting with linguistic examples arms us with intuitions for what different structural properties are good for. Synonymy increases the complexity of a language, by increasing the number of forms it contains. However this also increases the complexity that a language can faithfully describe. By having different forms for different contextualisations of a concept we end up being able to convey fine-grained contextual information. Homonymy makes the system as a whole more compressible, by reducing the number of forms - but does so with a cost of ambiguity. Reusing the same form for different meanings requires reconstructing which meaning a form maps to from context. While this isn't an issue if homonyms are contextually mutually exclusive - often the case in natural language - when homonyms appear in the same context this ambiguity can be a problem. In the next chapter synonymy and homonymy are generalised to \emph{Variation} and \emph{Disentanglement}\footnote{Homonomy is directly related to entanglement, but as noted previously to better align the measures with the divergence they use we opt to quantify disentanglement, which for us is equivalent to 1-entanglement.}, but at a high-level intuitions about the utility and costs of each kind of structure broadly hold.

Having started with an illustrative set of experiments, using a small model and dataset with discrete mappings, we turn now to mappings less clearly related to human language. The next chapter takes a step towards studying mappings learned by deep learning models, instantiating information theoretic measures for studying relationships between discrete and vector spaces that can be applied to models trained on more complex tasks than used in this chapter. Despite the shift in domain we'll continue to look at whether linguistic intuitions about different mapping structures hold, and the role that capacity plays in conditioning them.

\chapter{Regularity, Variation \& Disentanglement in Vector Space}
\label{chapt:learning}

{\makesans
\emph{
Discrete $\to$ Dimension-Wise Continuous} \\[1em]
}

{\makesans
\begin{quote}

The Era of Having Famous Painter Parents \censor{To where............................} of Making a Myth of Oneself \censor{...................}
of Patenting International Klein Blue \censor{To where..........................................................} of Needing More and More of a Crowd \censor{To where.....................................}
of One's Friends' Suspicion That One Had Arranged to Vanish Not
         Actually Died \censor{To where..........} \\
\flushright{\emph{- Anne Carson}}
\end{quote}}

\drawline

\noindent Chapter \ref{chapt:intro} discussed how deep-learning models put up impressive performance across a broadening array of tasks. Despite this we have a limited understanding of how they learn to represent information, and what information structures are desirable. In the experiments in this chapter, we look at models trained on large-scale semantic parsing tasks (like mapping 100,000 questions to SQL queries that answer them) and measure structure between their inputs and representation spaces. We want to see if a model's vector representations develop the same kinds of system-level structures present in the language data on which they're trained. While there's a growing body of work on \emph{interpretability}, this area tends not to focus on models' representation spaces directly. Instead, approaches often focus on behavioural evidence using a model's outputs to reason about what may be happening internally --- like looking at when models assign higher probability to grammatical sentences \citep{marvin_targeted_2018}. Another popular approach is probing \citep{veldhoen_diagnostic_2016, hupkes_visualisation_2018} - also called diagnostic classification - which trains a classifier to predict properties based on representations from a trained model. If it's possible to predict part-of-speech tags for a sentence from a model's representation of that sentence, it suggests the model has learned something about syntax. MDL probing \citep{voita_information-theoretic_2020} extends this line of work, by also quantifying how complex of a classifier is required to predict the diagnostic labels. While this kind of interpretability has yielded valuable insights it remains removed from directly assessing structure in representation space - relying on model performance on downstream tasks, or the performance of a diagnostic classifier. %

A major reason for the limited work on representations themselves, is that representations are high-dimensional vectors about which humans tend not to have strong intuitions. In their work on probing \citet{hupkes_visualisation_2018} note that earlier work would visualise models' hidden states and look for patterns in an effort to make representation spaces more intuitive to analyse, it's an approach that proves difficult to scale:
{

\makesans
\begin{quote}
[...] studying hidden layer activations is an interesting puzzle and can – especially for relatively low dimensional network such as ours – give pointers to which aspects should be studied in more depth.  However, it is hard to draw definite (and quantitative) conclusions, and the usefulness of the method decreases with higher dimensionality of the networks. [...] Disentangling the behaviour of networks through visual inspection of activations is searching for a needle in a haystack.\\ \flushright{ --- \citet*{hupkes_visualisation_2018} | p.918}
\end{quote}

}

\noindent What we need is a method for talking about representations directly, quantitatively, that scales to the kinds of models in use today whose latent spaces easily exceed 1000s of dimensions. The next two chapters propose methods for doing exactly this, information theoretically, treating a model as a mapping between inputs and representations, and quantifying the degree of our 3 base structural properties in a mapping --- regularity, variation, and disentanglement (one-to-one, one-to-many, many-to-one).

The major challenge in this approach is quantifying entropy in vector space. Entropy estimation for continuous spaces is a notoriously challenging problem \citep{paninski_estimation_2003}. In Chapter \ref{chapt:information}  I looked at discrete entropy, where each event in a categorical distribution refers to the probability of something discrete, like the probability of a word. Now though, we want a probabilistic interpretation of vector space, where probabilities refer to how likely it is for representations to occur in a certain region of space. Shannon proposed differentiable entropy as the continuous analog of discrete entropy, swapping the summation over events with an integral - unfortunately as \citet{Jaynes1957InformationTA} notes this quantity is not in-fact the true continuous analog of discrete entropy and lacks many of the properties which make entropy desirable\footnote{According to \citet{Jaynes1957InformationTA} Shannon merely assumed use of an integral was the correct continuous analog of discrete entropy (it is not), and did not derive Differential Entropy from Discrete Entropy. A rare moment of fallibility from Shannon who introduced Information Entropy in his masters thesis.}. Most critically, differential entropy $\mathcal{D}(Z)$ is unbounded ($-\inf < \mathcal{D}(Z) < \inf$) making it difficult to interpret, and is not invariant to linear transformations, meaning two uniform distributions can have substantively different differential entropies depending on what region of space they cover. There are also practical challenges in estimating the quantity given the difficulty of integration.  

Chapter \ref{chapt:llm} engages with this problem at length, discussing limitations of \emph{differential entropy} and proposing a new theoretically grounded approach to estimating entropy of continuous space. The remainder of the current chapter takes initial steps in that direction, opting to discretise representations, cutting attested space into bins and using bin probabilities to estimate discrete entropy. This approach is used in previous work looking at smaller networks trained on simpler tasks \citep{shwartz-ziv_opening_2017, saxe2019information, goldfeld2018estimating}, and is instantiated similarly here. While this approach does allow meaningful quantification of structure in continuous space it has a number of drawbacks discussed in the next chapter. Most relevant here is that it is exceedingly memory intensive, limiting it's applicability to contemporary models. In this chapter we circumvent this by performing dimension-wise discretisation; discretising and analysing dimensions one at a time before aggregating across them. While this limits our ability to track cross-dimensional dependencies, it does allow the approach to be applied to 256 dimensional spaces, and enables direct comparison between models of different dimensionalities. It also allows us to relate the analysis here to the analysis performed in the previous chapter. 

The previous chapter looked at structure in a mapping between meanings and signals, where signals were a sequence of symbols. Signals have a maximum length (how many symbols can occur in sequence) and an alphabet (how many different symbols are possible at each position). Dimension-wise discretisation treats a representation as a sequence of random variables, where number of dimensions is analogous to maximum signal length, with number of bins as the alphabet size. The approach in this chapter therefore acts as a middle ground between the preceding chapter which considers structure in discrete signals, and the next chapter which works to quantify entropy while remaining more faithful to continuous space. In all three cases I show how structure develops over the course of training, and that representational systems end up looking - in many regards - a lot like natural language: regular enough to generalise, but retaining substantial variation to represent the contextual complexity of the world their training data describes.

\thinline
{
\makesans
\noindent The remainder of this chapter is based around a paper \textbf{Representations as Language: An Information Theoretic Framework for Interpretability} that appeared at the International Meeting of the Cognitive Science Society in 2024. Authors are myself and Kenny Smith - Kenny and I conceived the experiments together which I then implemented and wrote up - Kenny gave writing feedback prior to submission to the conference. The paper is presented here minimally changed from the conference version that underwent peer-review. Changes are largely related to formatting to make the content more readable outside of the original conference paper template, and to make notation consistent across different chapters. Here I've also added some visualisations of this chapter's measures of structure applied to example distributions in 2-dimensions to help give an intuition for how they work. This addition comes just after the measures are defined and added text is marked, like here, by horizontal lines above and below.
}
\drawline

\section[Representations as Language]{Representations as Language: An Information Theoretic Framework for Interpretability}

Deep-Learning models achieve remarkable performance across a broad range of natural-language tasks \citep{vaswani_attention_2017}, but we still have a limited understanding of the learning process they undertake, and how they come to represent information so effectively. This is in part because these models are black-boxes \citep{tishby_deep_2015, shwartz-ziv_opening_2017}. They learn representations of their training data that are high-dimensional vectors, gigantic lists of numbers that are hard to interpret. While there is a growing body of work on interpretability, offering techniques for predicting what is encoded in a model's representations \citep[e.g.][]{voita_information-theoretic_2020, pimentel2020information}, there's still lack of clarity about how representations themselves are structured, how that structure emerges, and what kinds of structures are desirable. 

\begin{figure}
    \centering
    \includegraphics[width=0.6\textwidth]{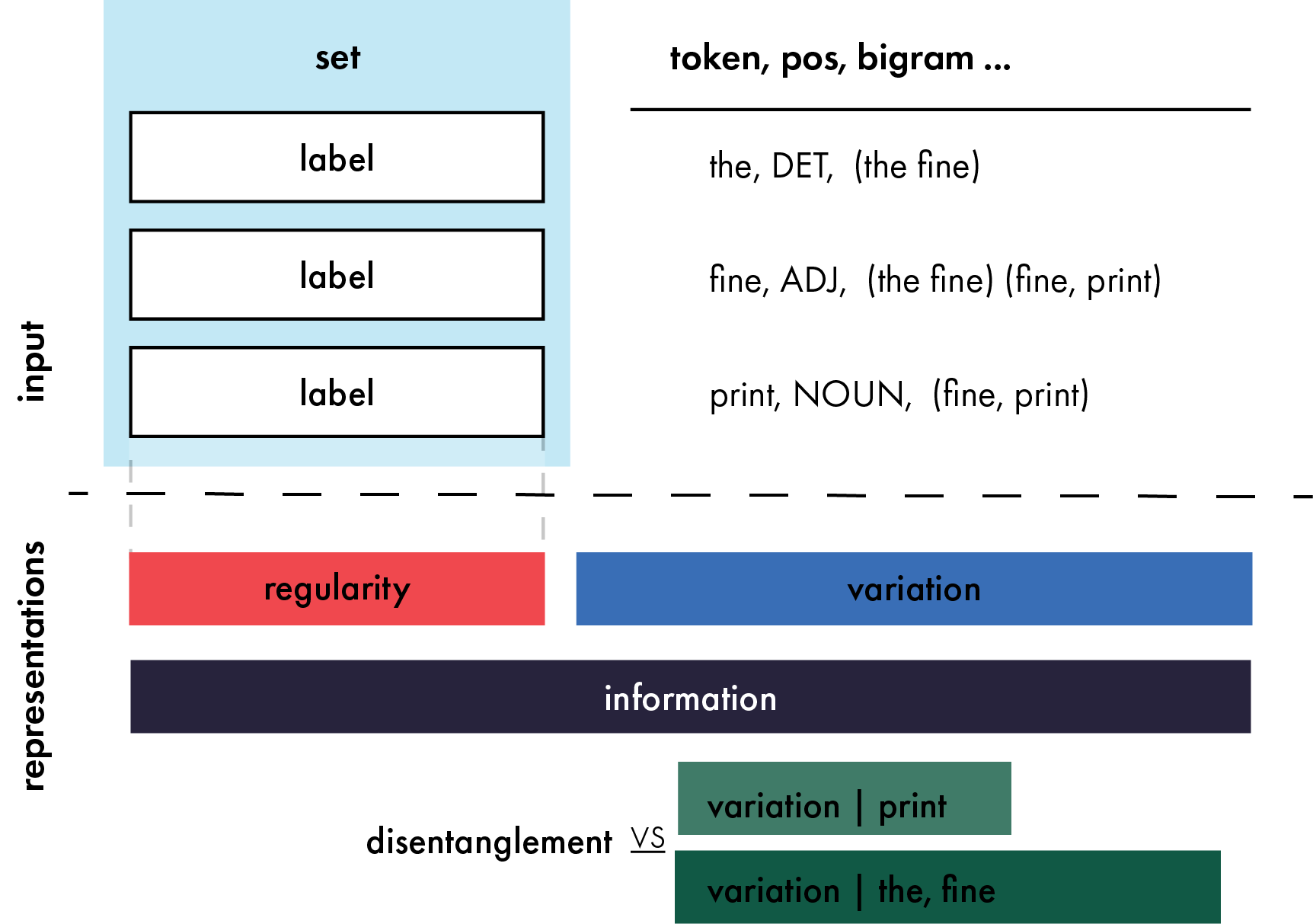}
    
    \label{fig:basic_overview}
    \caption{a depiction of basic quantities we measure and how they relate to each other. We measure structure in the mapping between labels for the dataset, and latent representations inside of a transformer. Here some example labels are given for the sentence "the fine print."}
\end{figure}

Central to language's ability to generalise is its regularity, exemplified by syntactic structure \citep{partee1995lexical}, which allows predictable \& regular encoding of meanings across the entire system. Languages are also rich with variation which can make them more expressive \citep{hurford2003synonymy} and structured ambiguities that can make them more compressible \citep{piantadosi_communicative_2012}.
Do the representations learned by a transformer model \cite{vaswani_attention_2017} exhibit similar \emph{system level-structures}? To answer this we look at the representations that emerge over the course of training as a kind of language in their own right.  At a high-level we can think of language as a mapping between spaces, like between meaning and form \citep{saussure_course_1916}. A multi-layered neural model needs to learn to map a sentence to a vector representation that later layers can successfully map to the output; encoder-decoder models \citep{cho2014learning} even more explicitly use separate parts of a model to map in and out of vector space. We draw an analogy between these two mappings, quantifying different kinds of regularity and variation in a model's mapping between inputs and representations. While there has long been interest in the kinds of representations learned by deep-models \citep{bengio_representation_2013, locatello_challenging_2019}, there has been little work quantifying systematic structure in the representations learned by transformers or relating them to the kinds of structures that characterise natural language. It's worth noting our approach is in contrast to some existing work that draws parallels between model weights and formal languages (like trying to infer functions or `source code' from model weights; as in \textcite{elhage2021mathematical}) --- we think an approach grounded in natural language is more scalable and better suited to characterising the kinds of systematic structure, variation, \& ambiguity that can emerge in deep-learning models, especially those trained on data from natural languages. 

We introduce a novel information-theoretic framework for assessing whether the representations learned by a model are systematic. In order to do this we first discretise vector representations into a sequence of symbols, then quantify 4 properties of the learned mapping from sentences to symbols: the degree of compression, regularity, variation and disentanglement. By doing this at different levels of abstraction we show when lexical and syntactic information are learned. We  identify two clear phases of training, the first characterised by the model rapidly learning to disentangle and align representations with token and part of speech information, the second (far longer) phase of training characterised by representations becoming more robust to noise. During this second phase models compress their representations, with larger models compressing considerably more; at the same time, generalisation performance begins to slowly improve, showing a link between robustness to noise and generalisation. Finally we discuss what kinds of representational structure are desirable, using our measures to predict which models will perform best on a generalisation set.

\section{Methods}

Our experiments use a Transformer \citep{vaswani_attention_2017} encoder-decoder model, with a two layer encoder, and single layer decoder. The model's encoder maps each input sentence to a vector representation, then the decoder uses this representation to generate the corresponding output, in our case a formal semantic representation of the input sentence. We look at the encoder's mapping between sentences and representations, and quantify the degree to which it exhibits systematic structure. We train each model (from scratch) on two different semantic parsing datasets, designed to evaluate a model's ability to systematically generalise: SLOG \citep{li_slog_2023} where the task is to generate lambda expressions for a sentence, and CFQ-MCD \citep{keysers_measuring_2020} where questions about movies need to be mapped to SQL queries that answer them. Both of these datasets come with an out-of-distribution generalisation set containing examples generated by the same grammar as the training data, but purposefully designed to be challenging. We also look at whether the capacity of a model affects the kinds of structure that emerges, training three different model sizes (with hidden dimensions of 64, 128, or 256).

\subsection{Estimating Entropy in Vector space}
Shannon Entropy describes the amount of information contained in a random variable \citep{Shannon1948AMT}. While methods exist for estimating entropy of continuous variables, these approaches are difficult to compare across representational spaces and often require strong assumptions about the underlying probability distribution \citep{Jaynes1957InformationTA}. Instead we discretise the hidden representations into a sequence of random variables, enabling us to directly estimate the Shannon entropy of our latent space. Our method is analogous to converting each vector into a sequence of discrete symbols, with a symbol for each dimension of vector space. Previous information theoretic analyses of deep learning have performed a similar estimation (e.g. \textcite{shwartz-ziv_opening_2017}, \textcite{saxe2019information}), although it's been noted this approach is more reflective of non-uniformity, like clustering behaviour, than it is of the true entropy of the space \citep{goldfeld2018estimating}. For our purposes identifying the degree to which a variable is uniformly distributed, or tightly clustered is sufficient to draw substantive conclusions.

For a given vector in a set of vectors $v_i \in V$ with dimensions $d \in D$ we cut each dimension $V_d$, into $N$ equal-width bins between the attested maximum and minimum values of that $V_d$. This enables a straightforward maximum-likelihood estimate of the entropy of $V$ by counting the frequency of each bin and normalising by number of representations in $V$. Resulting bin probabilities $p(V_{dn})$ are used to estimate the entropy of each dimension, then averaged across dimensions to give us an overall estimate of the dimension-wise entropy of $V$. 

\begin{equation}
    \label{eq:H}
    H_{dw}(V) = \frac{1}{|D|}\sum_{d}^{D}\sum_{n}^{N} -p(V_{dn})\log(p(V_{dn}))
\end{equation}

\noindent On the right in \ref{eq:H} is the equation for shannon entropy \citep{Shannon1948AMT}, as this estimate is an approximation we also use the Miller-Meadow correction in order to smooth the estimate based on sample size and improve its accuracy \citep{Miller1955NoteOT}. No method of estimating discrete entropy in continuous spaces is perfect (see \textcite{paninski_estimation_2003} for extensive discussion), but our estimator is invariant to linear transformations while making minimal assumptions about the underlying distribution. Note that while in the results presented here we estimate entropy per dimension we can in principle just as easily estimate entropy per pair or set of dimensions (akin to modelling at the unigram vs n-gram level); in practice the memory demands of the full-discretisation approach used here and in previous work make this intractable. Our use of a dimension-wise estimate simplifies our analysis but limits its ability to track cross-dimensional dependencies. Although analysis at the dimension level allows us some insight into the role of different subspaces of representational space, by letting us break estimates down dimension by dimension.

\subsection{Measuring Structure}\label{measures}
We are interested in whether a model's representations become systematically structured during training, reflecting the system-level structures of the data they're trained on. Using our entropy estimator we assess 4 different quantities at different levels of abstraction, which allow us to describe the degree to which the representations a model learns are structured with respect to structure in a given dataset. Here we walk through our measures for describing the representational system that emerges over the course of training, quantifying the amount of Information, Variation, Regularity, and Disentanglement. 

\paragraph*{Information (Entropy):} We have a model $f$ that maps a set of sentences $X$ to representational space $Y$. For each sentence $S^k \in X$, the model takes as input a sequence of tokens --- usually words --- $t_a^k, t_b^k, t_c^k ... \in S^k$ and returns a sequence of vectors $v_a^k, v_b^k, v_c^k ... \in V^k$ where $v_a^k$ is the vector corresponding to token $a$ when it occurs in sentence $k$. While each sequence $V^k$ is of variable length, the individual vectors are the same size. We can therefore create a list $Y$ of all token representations from all sentences in the dataset

\begin{equation}
    Y = [v_a^k : \forall v_a^k \in f(S^k) : \forall S^k \in X]
\end{equation}

\noindent and calculate its dimension-wise entropy. The result gives us a measure of the average amount of information encoded in each dimension of the representation, $H_{dw}(Y)$. Given that the amount of information the model needs to encode is constant (the dataset doesn't change during training) this also tells us how compressed the model's representations are. As the dimension-wise entropy goes down, the model uses less of its available representational space. Information is minimised (i.e. compression is maximised) as all tokens are mapped to the same vector regardless of the token and sentence they correspond to, and information is maximised when token representations are spread out uniformly across representational space. To aid interpretation we normalise this measure, as well as Variation and Regularity, so that 1.0 indicates a uniform distribution and 0.0 is one-hot (this makes the quantity an efficiency).
Our estimator is \emph{invariant to linear transformations,} which means it ignores how numerically large a representational space is used. That is, this score is maximised if representations are spread out uniformly between the interval -2, 2 or -10,10 --- what matters is representations' uniformity, not their magnitude.

\paragraph*{Variation (Conditional Entropy)} captures how much a property varies in representation space. Given a class of labels, like tokens, or parts of speech, it reflects whether the model learns a single global representation of each label invariant to context, or if each representation is completely unique to the sentence it occurs in. We quantify this in terms of the conditional entropy of representations, given a label, creating a list of all instances of that label $Y|label$, across all contexts where it occurs

\begin{subequations}
\begin{equation}
    Y|label = [v_a^k \text{ \emph{if} } a=label : \forall v_a^k \in Y]
\end{equation}

\noindent Labels for the tokens fed into a model are virtually always known, so we can easily estimate the conditional dimension-wise entropy of $Y$ given a specific token $H_{dw}(Y|token)$. This is minimised when all instances of a token map to the same vector regardless of the sentence they occur in, and maximised when $H_{dw}(Y|token) = H_{dw}(Y)$ indicating instances of the same token are no more likely to be similar than two tokens chosen at random. The mean variation across the set $S$ of all tokens gives us a general sense of how much the model encodes context in its internal representations.

\begin{equation}
    variation(Y|Set) = \frac{1}{|S|}\sum_{label}^{S}H_{dw}(Y|label)
\end{equation}

\noindent We can also calculate variation with respect to any features we have a set of labels for. For example, if we know the part of speech for each of the input tokens $variation(Y|POS)$ could tell us if members of the same syntactic class share more information with each other than expected by chance. In the general case we just need a set of labels to condition on (e.g. part of speech, morphological case, tense etc.) when estimating $H_{dw}(Y|Set)$.

\end{subequations}

\paragraph*{Regularity (Mutual Information)} measures how structured a model's representations are with respect to a feature in the input --- in particular, whether the mapping between a label and its representation is monotonic (one-to-one).  The inverse of variation, Regularity quantifies how much knowing something about a token is going to tell us about its representation; quantifiable as the dimension-wise mutual information between a label and its representations.

\begin{equation}
    regularity(Y, Set) = \frac{1}{|S|}\sum^{S}_{label} H_{dw}(Y) - H_{dw}(Y|label)
\end{equation}

\noindent This is maximised when a label and its representations are monotonically aligned --- knowing the label tells us everything there is to know about the representation. As with variation we can quantify regularity with respect to individual labels in a set and mean across them to get a general notion of how aligned representations are with e.g. tokens, POS tags, or the bigrams a token is part of.

\paragraph*{Disentanglement (JS Divergence)} measures how separable different labels within a set are from one another, e.g. whether separate tokens are represented in distinct regions of representational space, rather than overlapping. We measure this by assessing the Jensen-Shannon divergence between $P(Y|label)$ and all other labels in the set $P(Y|Set \neg label)$; if tokens are distributed uniformly across a space their disentanglement will be 0, while if they are entirely separable it approaches 1.

\begin{equation}
    dis(Y, Set) = D_{JS}(P(Y|label) ; P(Y|Set \neg label))
\end{equation}

\noindent As with previous measures we aggregate this to get an assessment of how disentangled the class of labels is. This measure is related to previous assessments of entanglement \citep{chen_isolating_2018, conklin2022compositionality} but is implemented quite differently, and requires no pair-wise comparison of different labels.

\thinline
	{\flushright \makesans \textbf{\textit{\underline{Measure Visualisations}\\[2em]}}}

\begin{figure}[hp]
    
    \includegraphics[width=0.9\textwidth]{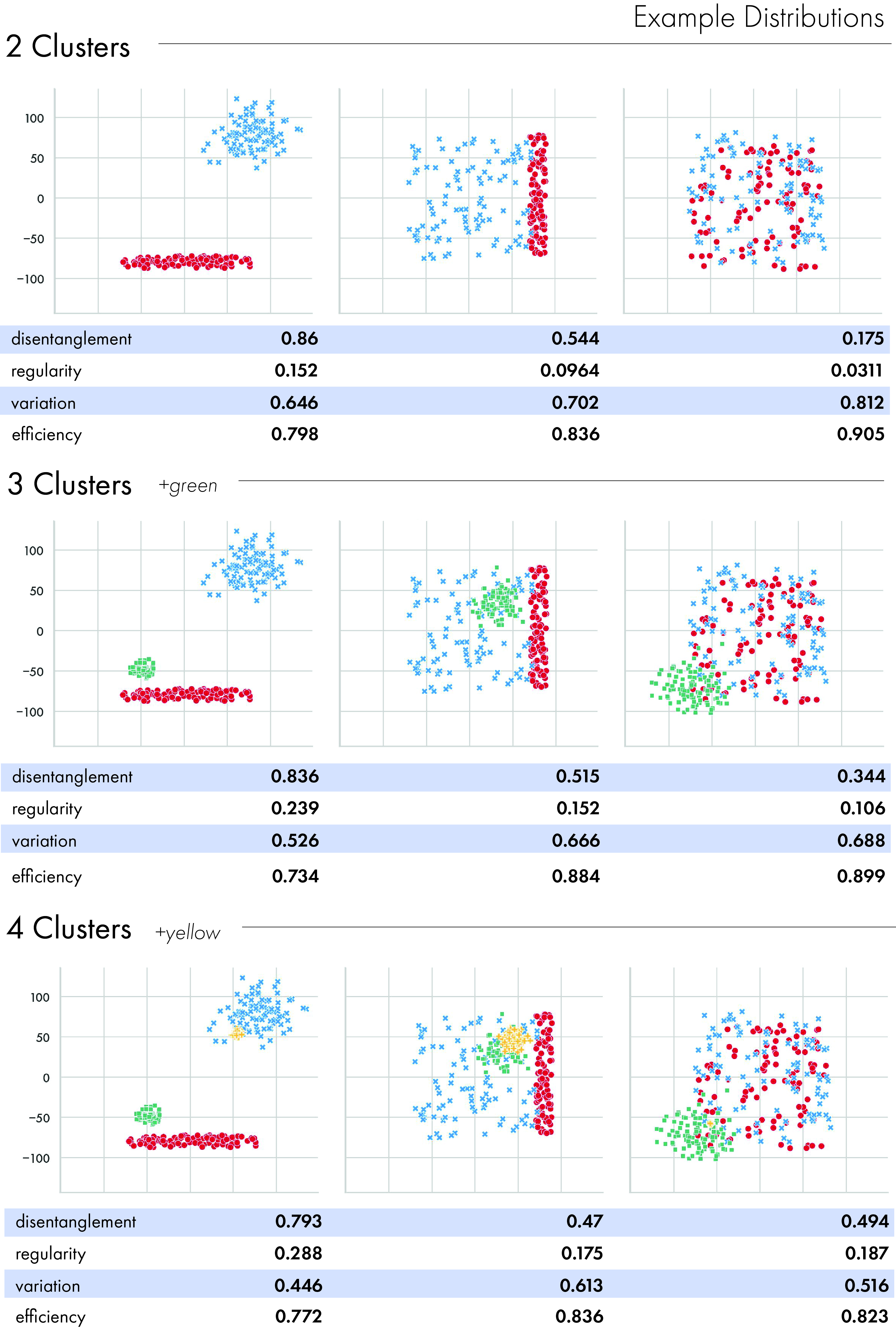}
    \caption{Examples of Measures defined here using the dimension-wise discretisation leveraged in this chapter. Each plot contains different clusters indicated by colour, with the scores across 4 measures below each facet. Disentanglement indicates cluster separability, regularity is the average mutual information with cluster label, variation is the average conditional entropy given a cluster label. Efficiency is the normalised entropy for the entire facet.}
    \label{fig:dimension_wise_measures}
\end{figure}

\noindent I have included visualisations for each of these measures, that were not included in the original publication in figure \ref{fig:dimension_wise_measures}. These help build an intuition for how different scores across the measures correspond to actual representations in vector space. Each facet contains 2 to 4 different distributions - each corresponding to a `label' in the analysis - indicated by colour. For each distribution either a uniform, or multivariate normal distribution is selected at random, then randomly parameterised. 100 samples are drawn from each distribution, and the 4 measures introduced above are applied to these samples. To enable straight-forward visualisation each distribution is 2 dimensional. Each figure includes 3 different examples (each column) with 2-4 distributions (each row), note that the rows are additive with each plot including the plots above in its column. These visuals help us to link each measure to properties of clusters in real-valued space.

\begin{itemize}
	\item Regularity: How predictable a region of space samples from a cluster will lie in.
	\item Variation: Cluster size - how much of the attested space is occupied by that cluster.
	\item Disentanglement: How separable clusters are from each other.
\end{itemize}

\section{Results}

We report results on two different datasets designed to assess compositional generalisation. Our measures allow us to characterise the trajectory of training, which we identify as having two distinct phases. We also compare models of different sizes to see how capacity changes representational space. Summary results are found in table \ref{tab:main_results}. 
It is worth noting that some of our results may be particular to the hyperparameters used for training. We use hyperparameters recommended by the authors of the datasets we use, or that the transformer was introduced with \citet{vaswani_attention_2017}. We believe this means that our design choices are representative of common ones for training sequence-to-sequence transformers, our code itemises all parameters used. We compute measures with respect to labels for Tokens, Parts of Speech, and Bigrams in the input and for brevity report the values with the clearest effect on model performance. We also focus discussion on results from the MCD CFQ dataset, as it's the larger of the two (100,000 training examples) and is a more realistic task --- mapping questions to SQL queries. We report results for the most challenging split of this dataset, known as MCD2. We include some discussion of SLOG, but an exhaustive listing of all results, across all datasets levels of analysis and model sizes can be found with the released code.

\subsection{Two Distinct Phases of Training}
We see 2 distinct phases of training, similar to \citet{shwartz-ziv_opening_2017}, despite using rather different methods (studying classification with a feed-forward network rather than a linguistic task with a transformer). This suggests some generality to this characterisation of deep-learning, though our results point to different analyses of each phase (particularly the second, much longer one), likely due in large part to the difference in model and domain. While overall trajectories are consistent across conditions when different model sizes move between phases differs, for clarity here we refer to specific steps in the training timeline for the mid-size model on CFQ.

\begin{figure}
	\centering

    \includegraphics[width=0.8\textwidth]{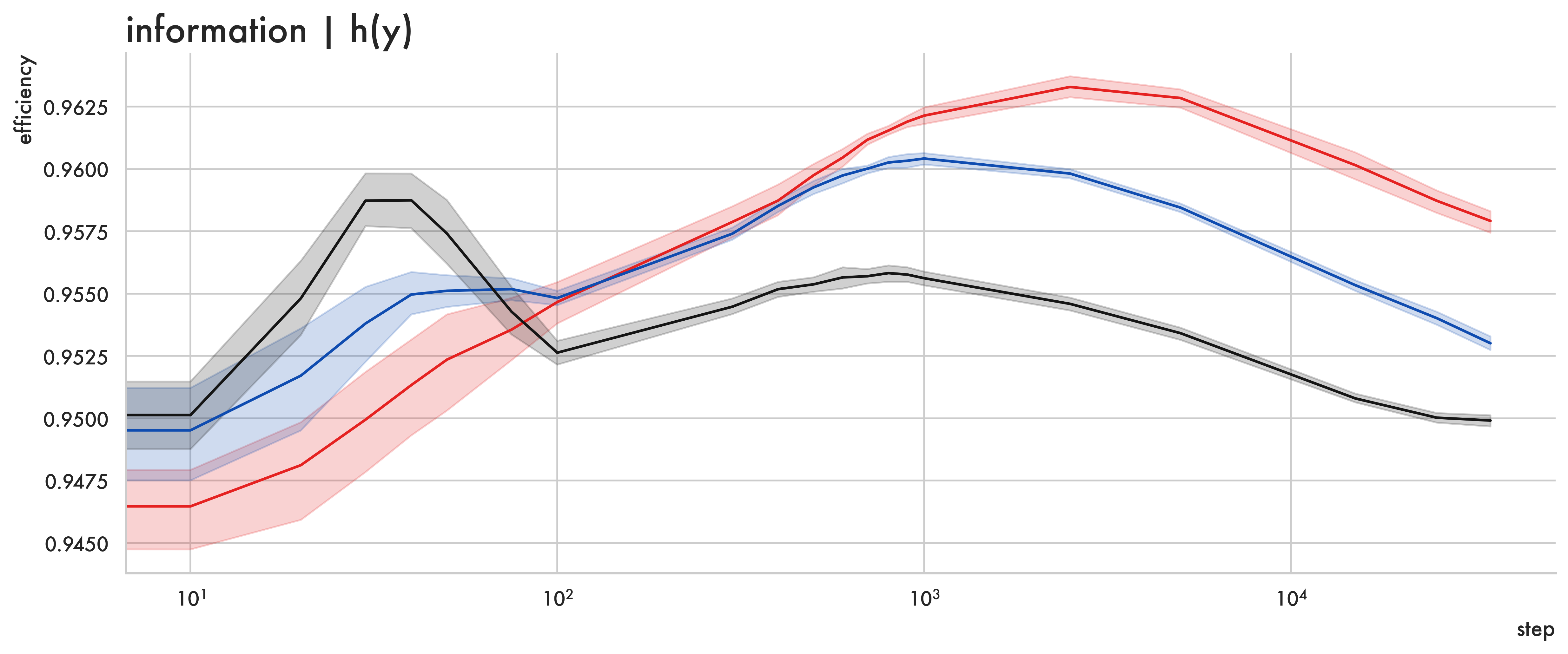}
    \includegraphics[width=0.8\textwidth]{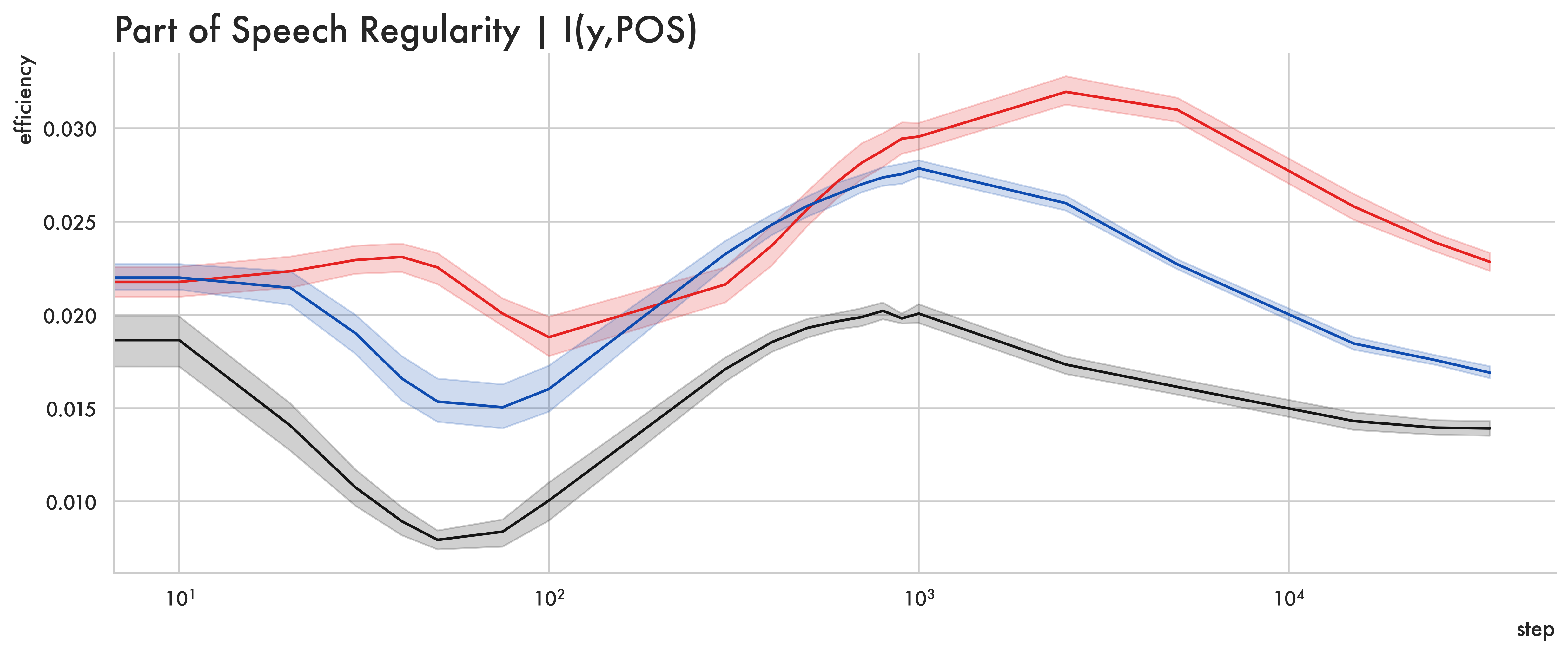}
    \includegraphics[width=0.8\textwidth]{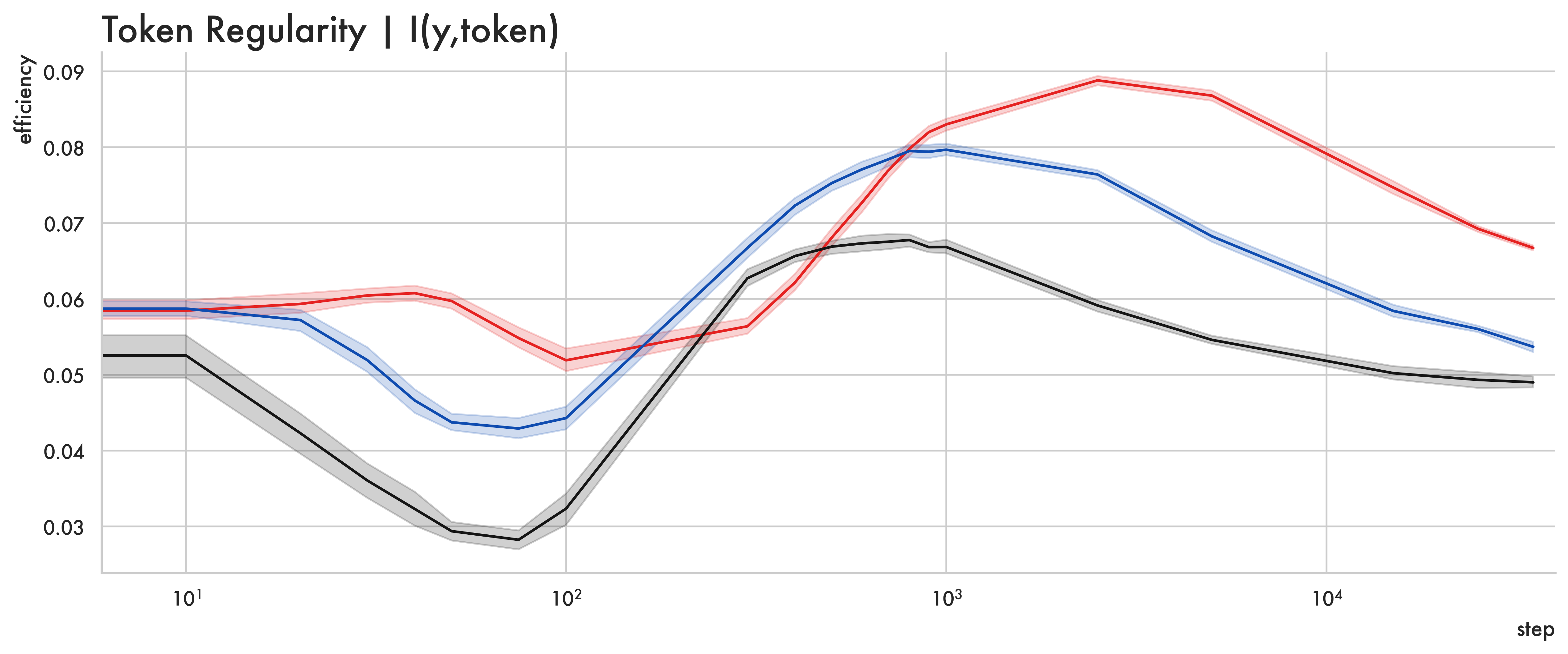}
    \includegraphics[width=0.8\textwidth]{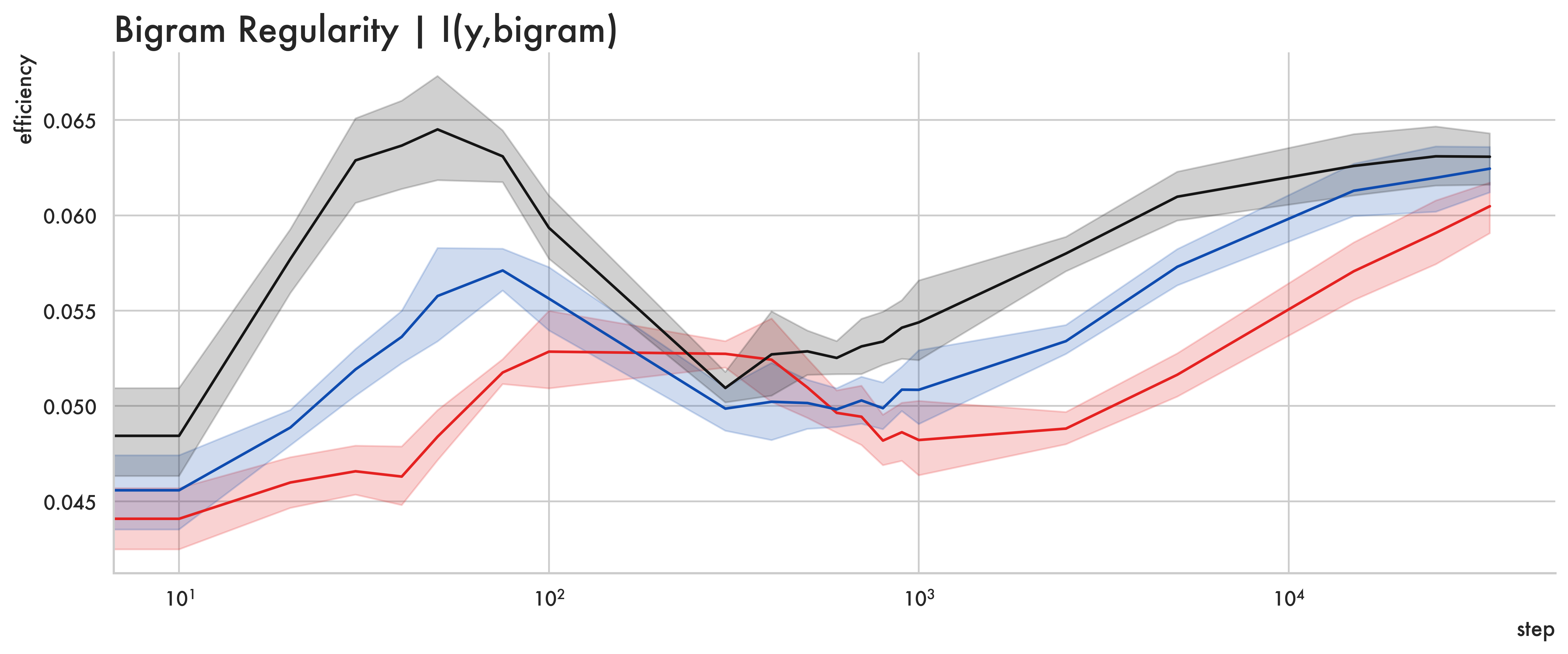}
	\includegraphics[width=0.8\textwidth]{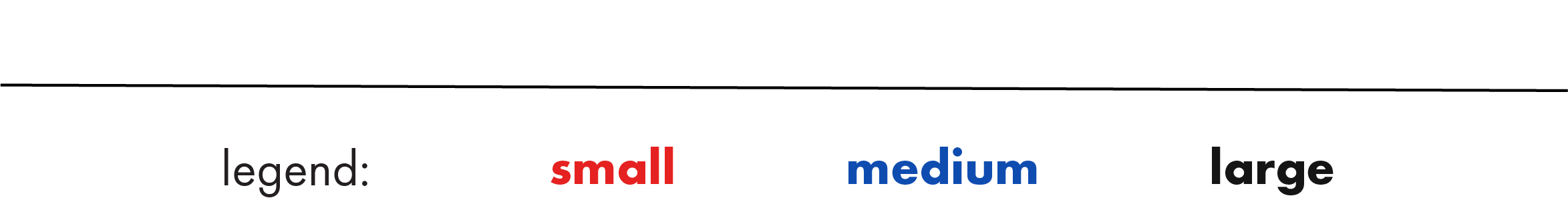}

    \caption[short]{Each facet shows a different measure (along the y axis) against training steps (log scaled). Lines and shading give means and 95\% CIs; line colours give results for 3 different model sizes. Values are calculated across the entire training set for 10 different random seeds. Efficiency (normalised entropy) is bounded  such that 1.0 indicates a uniform distribution and 0.0 one-hot.}

    \label{fig:slog_sizes}
    
\end{figure}

\begin{figure}\ContinuedFloat
	\centering

	\includegraphics[width=0.8\textwidth]{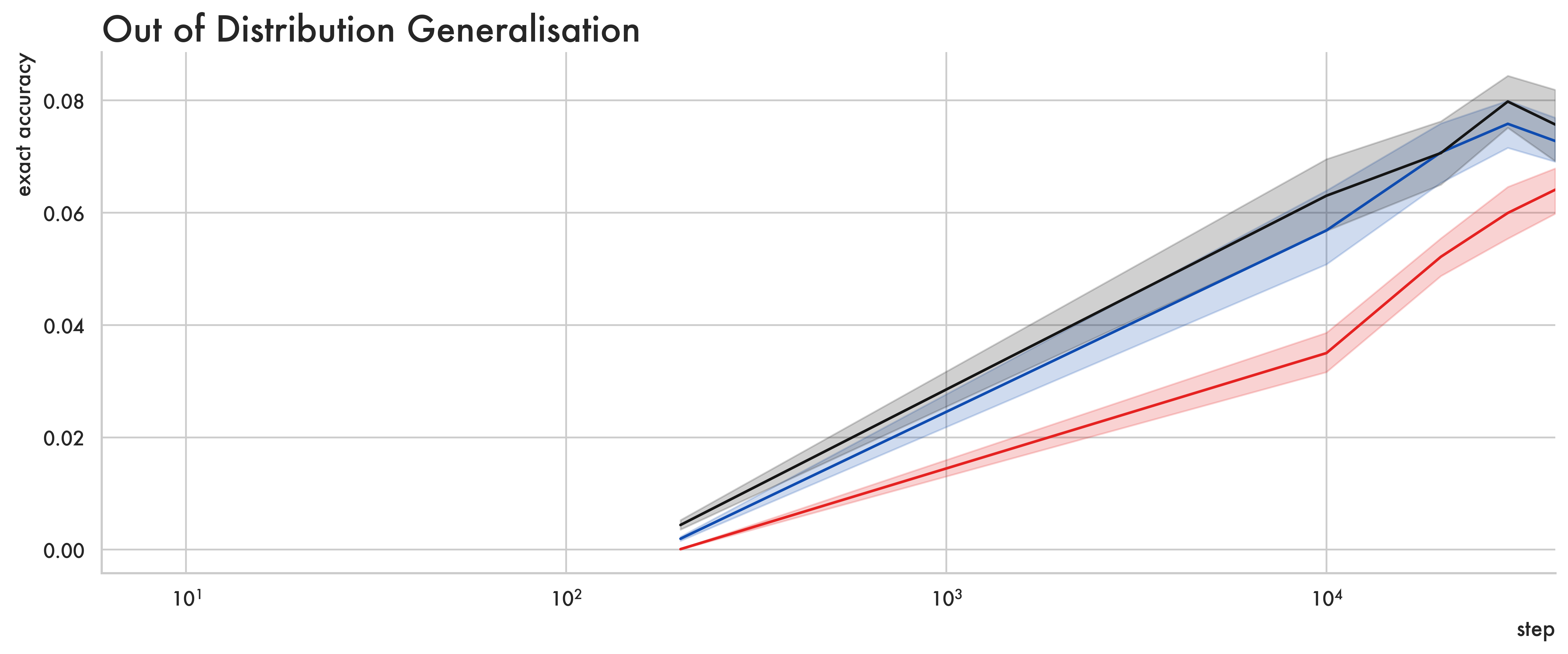}
    \includegraphics[width=0.8\textwidth]{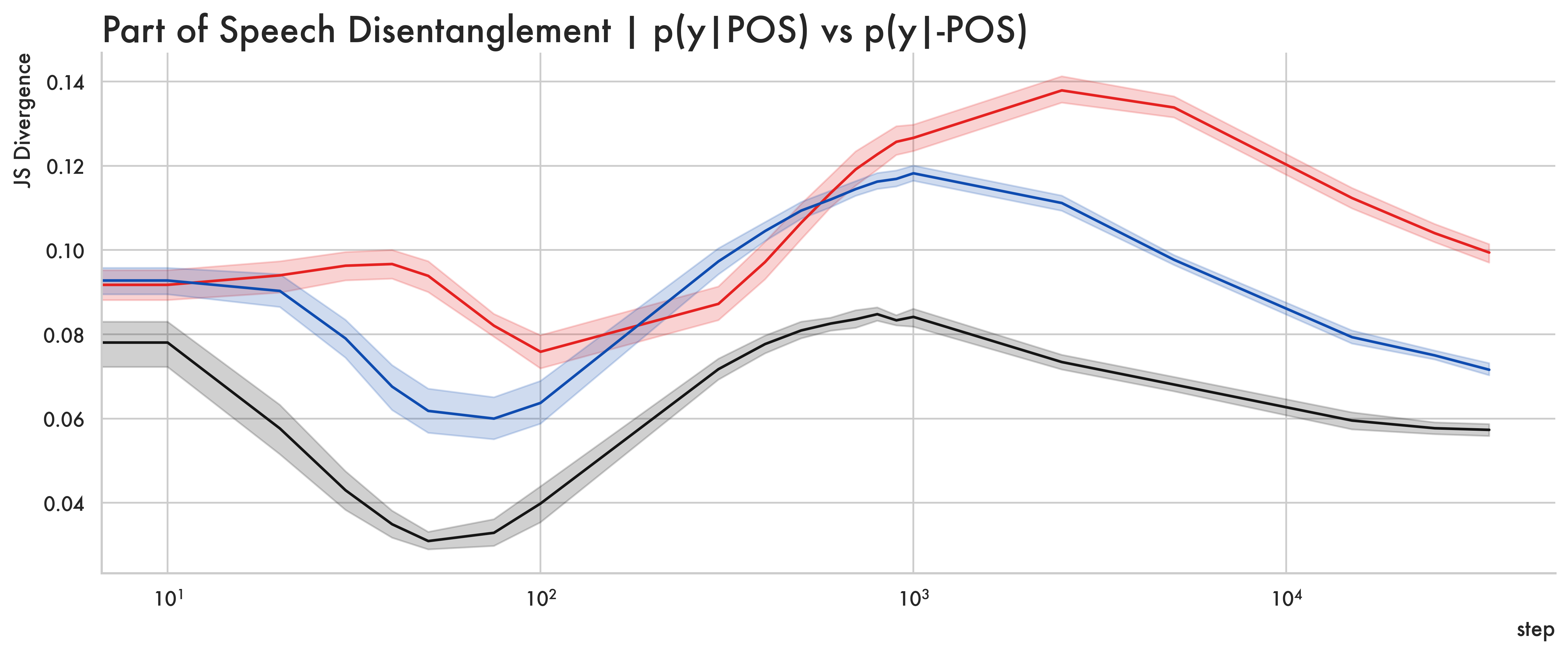}
    \includegraphics[width=0.8\textwidth]{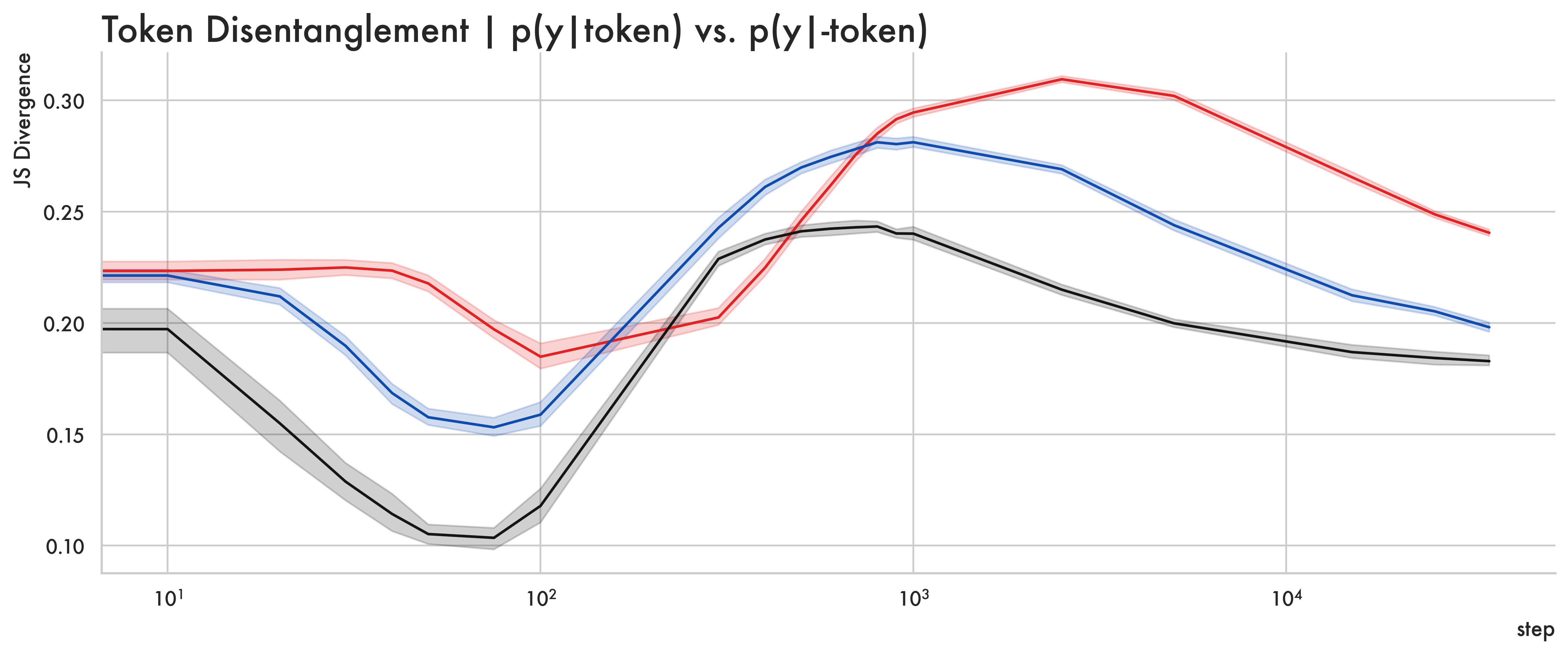}
    \includegraphics[width=0.8\textwidth]{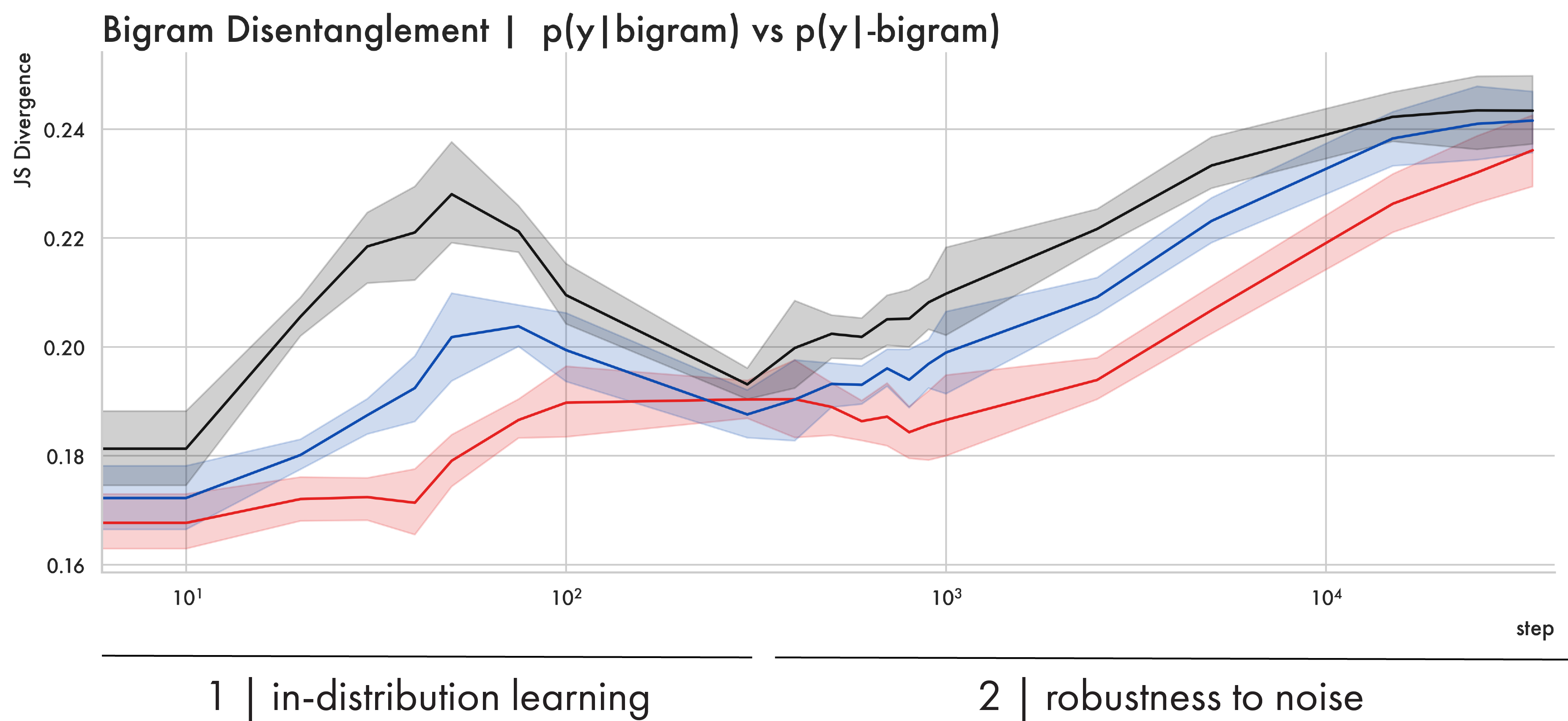}
    \includegraphics[width=0.8\textwidth]{chapters/chap-rep_as_lang/visuals/model_size_legend.png}
    \caption[short]{\textbf{(continued):} The top facet shows out of distribution generalisation performance. Note that this begins to increase as disentanglement increases across all levels of analysis. In particular though, the point at which generalisation performance increases closely aligns with when bigram regularity and disentanglement increase. Given the task requires models to interpret known words in new contexts, more disentangled contextual information (here bigram information acts as a proxy for contextual information) may allow the model to correctly decode the token in a broader range of sentences.}

    \label{fig:acc}

\end{figure}

\subsubsection{Phase 1 | In-Distribution Learning}
\paragraph*{Alignment \& Disentanglement.}
In Phase 1 the model achieves high in-distribution accuracy, climbing to ceiling performance on the training data by step 1,000. This increase in accuracy is driven by an increase in token and POS regularity between steps 100 and 1000 as representations become more monotonically aligned with the corresponding input token and its part of speech (fig ~\ref{fig:slog_sizes}, top left). This period also reflects an increase in POS and Token disentanglement indicating different tokens are represented in increasingly distinct regions of representational space. Conversely bigram regularity and disentanglement are reduced over the same interval as different token representations in the same bigram become more uniformly distributed over the support of $Y|token$.

\begin{figure}
    \centering
    \includegraphics[width=\textwidth]{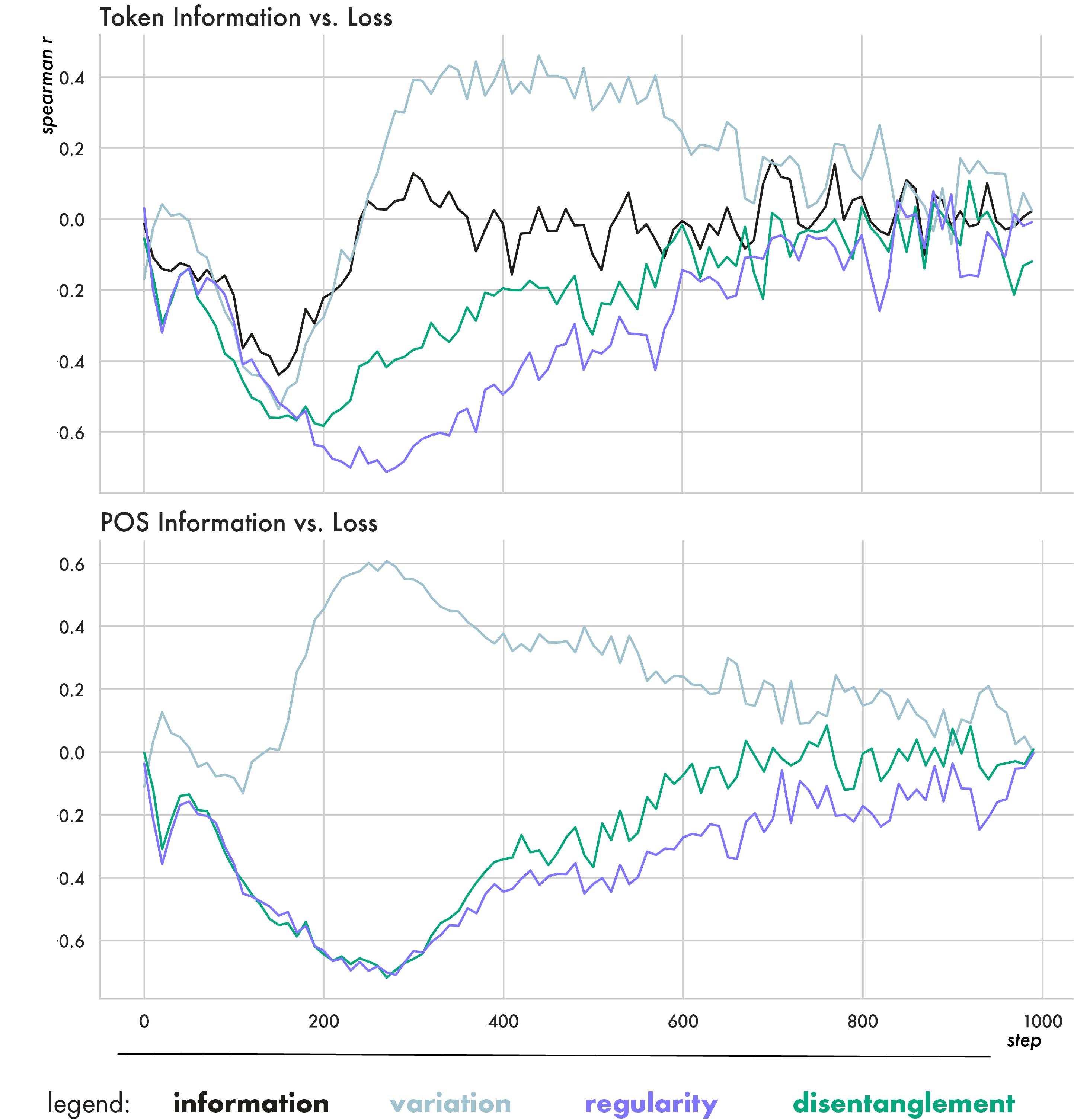}

    \caption{Spearman correlation coefficients between the loss minimised during training and our measures. Negative coefficients suggest the objective increases a quantity. Results for 100 runs of the medium model on SLOG. When exactly significance fades is noted in body text.  
    \textbf{Top:} measures conditioned on token labels. \textbf{Bottom:} measures conditioned on part of speech tags for the tokens.
    }
    \label{fig:loss_corr}
\end{figure}

\paragraph*{Does training select for structure?} During training the model tries to minimise a loss function, here the cross entropy between its predicted semantic representation for the input sentence and the correct one. During phase 1 we find a lower loss on the task (indicating better performance) correlates with our measures, suggesting the objective selects for certain structural properties in representation space. The timecourse of this is shown in figure ~\ref{fig:loss_corr}, with correlations between structure measures and the loss for 100 different runs of the medium model on SLOG. From steps 100 to 200 all four of our token-level measures correlate negatively with task loss ($p<0.001$), 
This dynamic shifts slightly from steps 200-600, where higher token disentanglement ($p<0.001$) and regularity (indicative of a more monotonic alignment) ($p<0.001$) continue to correlate with lower task loss but now with less variation ($p<0.001$, steps 280-600). 

\rowcolors{2}{white}{blue!3}
\begin{table}[t]
    \centering
    \begin{NiceTabularX}{\linewidth}{l|YYY}                        
    \CodeBefore
	  \rowcolor{blue!5}{1}
	  \rowcolors{2}{blue!5}{white}
	\Body
    \nosbold{SLOG} & \noserif{POS} & \noserif{Token} & \noserif{Bigram} \\
    \midrule
    
    \small{\makesans Information} & \aaacc{0.956647}{0.000303}& \aaacc{0.956647}{0.000303} & \aaacc{0.956647}{0.000303} \\
    \small{\makesans Variation} & \aaacc{0.920196}{0.000784} & \aaacc{0.849814}{0.000624} & \aaacc{0.754133}{0.000448} \\
    \small{\makesans Regularity} & \aaacc{0.036451}{0.000624} & \aaacc{0.106833}{0.000762} & \aaacc{0.096951}{0.000757}\\
    \small{\makesans Disentanglement} & \aaacc{0.147554}{0.002214} & \aaacc{0.353022}{0.001694} & \aaacc{0.373025}{0.001775} \\

    \multicolumn{4}{r}{\small{\makesans Accuracy:} \aaacc{27.8432}{0.7613}} \\

    \midrule
    \multicolumn{4}{l}{\nosbold{CFQ}} \\
    \midrule
    \small{\makesans Information} & \aaacc{0.952721}{0.000240} & \aaacc{0.952721}{0.000240} & \aaacc{0.952721}{0.000240} \\
    \small{\makesans Variation} & \aaacc{0.936019}{0.000327} & \aaacc{0.899654}{0.000468} & \aaacc{0.836738}{0.001703} \\
    \small{\makesans Regularity} & \aaacc{0.016702}{0.000292} & \aaacc{0.053066}{0.000540} & \aaacc{0.063160}{0.001236} \\
    
    \small{\makesans Disentanglement} & \aaacc{0.070673}{0.001215} & \aaacc{0.196321}{0.001453} & \aaacc{0.241247}{0.004246} \\

    \multicolumn{4}{r}{\small{\makesans Accuracy:} \aaacc{7.5936}{0.4130}} \\ 
    
    \bottomrule
  \end{NiceTabularX}
  \caption{Summary results for measures at the POS, Token, and Bigram level, across 10 runs of the medium model on both datasets, with 95\%CIs. Measures are computed at the last step of training across the entire training set. Models' accuracy \% reported on the held out generalisation set.}
  \label{tab:main_results}
\end{table}

Past this point all measures cease correlating with the task loss, which is also the point where empirical error begins to saturate --- as the model approaches ceiling performance on the training set, the loss asymptotically approaches its floor. Figure ~\ref{fig:loss_corr} also shows the correlation between loss and our measures conditioned on part of speech tags. Similarly, greater regularity and disentanglement with respect to part of speech labels and less variation correlate strongly with a better task loss from step 100 until 600 ($p<0.001$). The peak spearman coefficient for disentanglement reaches -0.71 indicating the objective optimizes more strongly for disentanglement of parts of speech than tokens (which peaks at -0.58 and fades from significance faster).

\subsubsection{Phase 2 | Robustness to Noise}
\paragraph*{Contextualisation \& Compression}
 This is the dominant dynamic of training, taking place from step 1000 onwards. During this period the representational space slowly compresses, with dimension-wise entropy decreasing. This is coupled with an increase in bigram regularity as clusters for different contextualizations become more distinct in representational space, respecting contextual structure forces the overall token regularity down as representations become less aligned with token information and more aligned with bigram information. These shifts happen slowly taking thousands of training steps.

\citet{shwartz-ziv_opening_2017} note that later in training, after the loss has reached its floor, the update steps the model takes begin to behave like `gaussian noise with very small means.' This aligns with what we see here, as measures of structure cease to consistently correlate with the task objective by phase 2. This suggests that a major dynamic of the latter period of training is representations becoming increasingly robust to noise. The model's mapping from sentences to representations needs to continue to encode the input, but do so robustly enough that the mapping won't be undermined by constant noisy updates, otherwise the task loss will begin to increase. Unlike previous work we note mutual information increases between inputs and representations later in training, just at higher level of granularity --- here, bigrams. 

 It's also worth noting that while the model achieves ceiling performance on the training and validation data during phase 1, it only begins to succeed on the more challenging out-of-distribution generalisation task 10,000 steps later (see figure \ref{fig:slog_sizes} top right). This means robust generalisation ability begins to appear only after a sustained period of representations becoming more robust to noise. This is related to the double descent phenomenon \citep{nakkiran2021deep}, where models begin to exhibit strong generalisation performance long after the initial learning of in-distribution data. \citet{voita_language_2021} also note that in machine translation a transformer starts by learning individual token probabilities before acquiring more complex sentential structure. Our results give a mechanistic account of how this may happen, with token alignment increasing first, then a much longer phase where representations become more contextualised. Though our task is simpler than large-scale translation, in future we aim to apply this analysis to that context.

\paragraph*{What kinds of representations generalise best?}
We also look within conditions to see if representational structure correlates with generalisation across different runs of the same model. We take the middle-sized model on CFQ and correlate across 10 runs at the final step of training. This analysis shows that runs with higher bigram disentanglement ($r=0.65$, $p=0.04$), and higher bigram regularity generalise better ($r=0.61$, $p=0.06$). The generalisation set of CFQ contains tokens seen during training as part of novel contexts. In order to do well our model needs to correctly encode tokens it has seen before, in contexts it hasn't. Higher bigram regularity and disentanglement indicates different contextualisations for the same token are more tightly clustered in space and that those clusters are more pure (being separable from other contextualisations of that token). More predictably and separably encoding different bigrams may help novel contextualisations of a token to be decoded correctly.
 
\subsection{Model Size Clearly Affects Representational Space}
While the overall phases of training are remarkably consistent across datasets and model sizes, there is a clear influence of model size on representational structure. Figure \ref{fig:slog_sizes} shows trajectories for our three different model sizes over the course of training. Smaller models are less compressed, and have greater regularity and disentanglement with respect to tokens and parts of speech. They also perform worse on both tasks than their larger counterparts. Larger models are more entangled at the POS and token level, but have more disentangled bigrams --- indicating larger models learn more pure clusters for different contextualisations of the same token.

\paragraph*{Why Models Compress \& Larger Models Compress More}
It's common to think of connectionist models as cognitive models, and expect them to be governed by similar constraints \citep{futrell2018rnns}. Humans may generalise robustly because constraints on our cognitive capacity force us to learn generalisable solutions rather than memorizing every possible outcome \citep{Griffiths2020UnderstandingHI, hahn_resource-rational_2022}. The fact that larger models (with greater capacity) compress more per-dimension, would seem at odds with this framing. 
While we agree that drawing cognitive parallels can be useful, on a representational level looking at models as a language can help us to reason about the effects of scale and the phases of training.

Specifically, our interpretation is that larger models are able to exploit their higher-dimensional internal representations to develop representations more robust to noise. An obvious analogy in communication is mapping an input to a discrete signal, where the signal space is defined by an alphabet of characters and a maximal signal length. If the signal length is low, a larger alphabet is needed to encode the input unambiguously. In contrast, if longer signals are allowed a smaller alphabet is required, the limiting case being a binary alphabet (like morse code) where sentences are encoded in comparatively long signals. Signals composed from a smaller alphabet are more resilient to noise \footnote{This is implicit in Shannon's definition of entropy, as the maximum uncertainty of a binary distribution is lower than one with more outcomes ($log(2)<log(3)$)} for instance, when an operator interprets morse code, at each point in the sequence they only need to differentiate between two possibilities, dot or dash, which is easier than distinguishing between e.g. 26 different outcomes, particularly on signals transmitted over copper wire. We have shown how, during the second phase of training, transformers compress their representations in response to noisy update steps. This is directly analogous to models using a progressively smaller vocabulary for each dimension of hidden space. Larger models have more dimensions, which in our analysis is akin to having a longer maximum signal length, enabling them to learn a mapping more robust to noise, like morse code, converging to a smaller alphabet but longer signal. 

\section{Conclusion}
We have introduced a linguistically-motivated approach to interpreting transformer models. By looking for system-level structure in the model's representations, we characterise two-distinct phases of training, and show how representational structure develops during those phases and how this explains model's ability to generalise. This is enabled by an efficient approach to estimating the entropy of transformers' latent space, that allows for non-parametric analysis of representational structure. Our findings help shed light on what the learning process looks like in deep-learning models, and makes a case that intuitions from linguistics and cognitive science about what makes for a `good' representation may meaningfully transfer here.

\chapter[Information Structure in LLMs]{Information Structure in Large Language Models with Soft Entropy}
\label{chapt:llm}

{\makesans
\emph{
A Continuous Approach to Entropy Estimation.} \\[1em]
}

{\makesans
\begin{quote}
Up the coast a few miles north \censor{in a lava reef under the cliffs} there are a lot of rock pools. You can visit them when the tide is out. Each pool is separate and different \censor{and you can, if you are fanciful, give them names} individual entities \censor{so you may think of a rock pool as an} throughout the day of the ebb tide, they know no other.
But \censor{that long day ends at last} the waters of the ocean come flooding \censor{over George and} Can they tell us, in any manner, about their journey? Is there, indeed, anything for them to tell-- except that the waters of the ocean are not really other than the waters of the pool? \\
\flushright{\emph{- Christopher Isherwood}}

\end{quote}}

\drawline

\noindent Chapter \ref{chapt:learning} studied how information structure emerges in a model trained on a single task. This chapter looks instead at large language models (LLMs), which are pre-trained on huge volumes of data scraped from the internet or from digitised collections of books \citep{raffel2019exploring}. Starting with BERT \citep{devlin_bert_2019}, LLMs have put up state of the art performance across a broad range of natural language tasks \citep[e.g.][]{devlin_bert_2019, brown2020language}, often generalising significantly more robustly than smaller models trained on a single task \citep{furrer_compositional_2021}. This chapter analyses large language models from an information structure perspective. 

While scaling up models has seemed to lead to greater performance, scale also makes analysis more challenging. The analysis from the previous chapter relies on binning representations in order to estimate entropy, but with the larger models studied in this chapter having a hidden dimension of 5120 this approach becomes intractable. To address this, this chapter also introduces a novel approach to entropy estimation --- \textbf{soft entropy}. Soft entropy splits the difference between discrete and differential entropy, estimating a quantity that behaves like discrete entropy but which is differentiable and works to respect properties of continuous space. This approach primarily relies on matrix multiplication, making it highly parallelisable which allows us to apply our analysis to models of arbitrary size.

Given large language model's significant improvements in performance compared with models trained on a single task, we study the information structures they converge to and the training dynamics that lead them there. Broadly the results in this chapter paint a similar picture to the training dynamics of the single task model in chapter \ref{chapt:learning}, namely early learning of lexical information followed by a slow contextualisation phase - suggesting some generality to this characterisation of deep-learning in the language domain.

\thinline
{
\makesans
\noindent The remainder of this chapter is a paper that is currently under review. Authors are myself and Kenny Smith - Kenny and I conceived the experiments together which I then implemented and wrote up. The paper is presented here minimally changed from the submitted version. Changes are largely related to formatting to make the content more readable outside of the original conference paper template, and to make notation consistent across different chapters.

}
\drawline

\section{Information Structure in Large Language Models}
Despite the remarkable performance of large language models \citep{brown2020language, dubey2024llama}, and their widespread use we still lack unified notation for thinking about and describing their representational spaces. We lack methods to reliably describe how their representations are structured, how that structure emerges over training, and what kinds of structures are desirable. This should be of concern to us for practical reasons - it makes it difficult to make design decisions when we don't have a clear picture of how they effect representational space - but also for broader  social reasons. Most people in the US and UK come into contact with an NLP system multiple times a day without realising \citep{kennedy2023public}. Given their increasing ubiquity  we should be able to account for the information they have learned and how that information is structured.

Our lack of tools for understanding representations in networks is in part because their representations are continuous, and we as humans tend not to have strong intuitions about high-dimensional vector spaces. Existing work interpreting large language models describes phases of training in terms of model behaviour \citep[e.g.][]{marvin_targeted_2018, blevins_deep_2018, dziri2024faith}, for example analysing when they begin to generalise robustly - or \emph{grok} \citep{power2022grokking, merrill2023tale}. Alternately work uses parametric methods like probing, leveraging a separate model to describe the first \citep{hupkes_compositionality_2019, voita_information-theoretic_2020, pimentel2020information}. 
We focus instead on giving a representational account of what training looks like, using information theoretic measures of representational space to quantify how structured representation spaces are in large language models, and what kinds of structure matter for generalisation. Ideally we need a way of thinking about deep-learning models in the general case that allows us to: 

\begin{enumerate}
	\item Describe structure in representation space, and what structures drive generalisation
	\item Clearly relate these to relevant work in linguistics and the cognitive sciences
	\item Quantify structure with methods that are efficient enough to apply the same analyses to models of any size, throughout training
	\item Meaningfully compare models of different sizes, trained with different objectives
\end{enumerate}

In an effort to do this, we look at deep-learning models as member of a more general class: mappings. Models map between their inputs and representational space, and are comprised of a sequence of linear and non-linear mappings. 

Here we quantify structure in the mappings learned by large language models while drawing parallels to a reference mapping about which we have strong intuitions for what structure looks like - unlike high-dimensional vector spaces - and which is related to the domain in which our models are trained: natural language. 

At its core language is a mapping - relating concepts, and complex propositions, to words, constructions, and phrases which refer to them \citep{saussure_course_1916}. While many natural communication systems fit this bill, language is unique amongst them \citep{Hockett1960}. It is learned from a finite sample, generalises readily to novel concepts and contexts, with \emph{system level} structures that provide us a system simple enough to be learned by children, but expressive enough to describe the universe. This parallels our desiderata for mappings in deep-learning models which need to be learned from finite data, able to generalise, and expressive enough to describe the world from which their training data is drawn. We look at whether \emph{system level} structures emerge in representation spaces learned by large language models; first introducing basic kinds of structure in a mapping, relating them to their analogs in linguistics, before quantifying each of them information-theoretically. 

\begin{figure}
	\centering
	\input{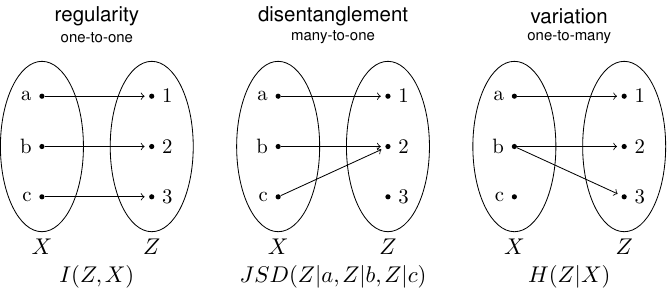}
	\caption{Three basic kinds of mapping structure we consider here, labelled with their linguistic analog, and the information theoretic quantity we introduce to measure them in section \protect\ref{sec:softH}. Note that we show part of the mapping ($a\to 1$) as regular in all cases because the mappings we consider exhibit a combination of all 3 structures. As such we assess the \emph{degree} of each structure, not whether or not it exists. Variation (one-to-many) is possible despite the fact that the networks mappings we examine are deterministic, because our $X$ contains instances of the same token in different sentences, meaning $b\to 2$ and $b\to 3$ reflects b in different contexts.}
	\label{fig:llm_mapping}
\end{figure}

We build on the framework for interpretability introduced in \citet{conklin2024representations}, redefining some of their measures, and extending it to large language models. To do this we also introduce a novel method for highly-parallelisable entropy estimation in vector space - \textbf{soft entropy}. This approach is similar to discretisation based methods used to analyse deep-learning \citep{shwartz-ziv_opening_2017, goldfeld2018estimating}, but fully differentiable, less affected by hyper-parameter settings, and dramatically more memory and compute efficient. Additionally the estimator can easily be applied at different levels of abstraction like model, layer, and subspace - this broken-down estimate enables direct comparison between different model sizes. We use soft entropy to quantify structure in language models ranging from 14 million to 12 billion parameters, looking at when system-level structure emerges during training, how scaling affects representation structure, and what kinds of structure drive generalisation. Our analysis is able to predict downstream performance on GLUE benchmarks based only on a models' representations at the end of pre-training (before 2 million steps of fine-tuning). To summarise our core contributions, this paper:

\begin{itemize}
	\item Frames structure in large language models in terms of related notions of structure from linguistics and information theory
	\item Introduces a novel method for entropy estimation of continuous spaces, that is fast, efficient and differentiable
	\item Shows how scaling a model's hidden dimension, or number of layers, affect representational structure
	\item Correlates representation structure at the end of pre-training with performance downstream after fine-tuning
\end{itemize}

\section{Related Work}
Our work is related to a long history of research in NLP which tries to identify correspondences between linguistic structures in training data and representations  or behaviours \citep{shi-etal-2016-string, belinkov-etal-2017-neural, marvin_targeted_2018, blevins_deep_2018, dziri2024faith}. It is particularly closely related to probing \citep{hupkes_visualisation_2018, pimentel2020information} which trains a classifier to predict labels from a larger model's representations. MDL probing \citep{voita_information-theoretic_2020} also includes a notion of regularity in terms of the complexity of the probe required to recover the labels. Given that we quantify structure in the mapping between labels and representations directly, our work represents a non-parametric approach to probing. The analysis here is also related to work in language emergence which looks at the languages that emerge between models in a multi-agent setting. A variety of quantifications of linguistic structure have been proposed for that domain that leverage similar intuitions to the ones used here \citep{brighton_language_2005, lazaridou_emergence_2018, resnick_capacity_2020, chaabouni_compositionality_2020, conklin2022compositionality} 

There is also existing work that tries to characterise training dynamics information theoretically \citep{tishby_deep_2015, shwartz-ziv_opening_2017, goldfeld2018estimating, saxe2019information}, however these are largely theoretical works and/or applied to feed-forward networks on tasks like digit classification. \cite{conklin2024representations} applies information theoretic methods to transformers trained on a single task - but uses dimension-wise discretisation, which is difficult to scale. Our approach to estimating entropy is similar to the limiting density of discrete points \citep{Jaynes1957InformationTA} and is related to kernel density estimation \citep{parzen1962estimation} in the way it relates discrete points to a continuous function.

\section{Identifying Structure in Mappings} 

We consider 3 basic structures in a mapping between two spaces: \emph{one-to-one}, \emph{many-to-one}, and \emph{one-to-many} (see Figure \ref{fig:llm_mapping}). These are related to linguistic concepts of regularity, disentanglement, and variation respectively. In a model we quantify these properties between labels for a model's input and the corresponding representations. Any labels for an input sentence can be used, experiments here use ones that come for free with any text data: token, bigram, and trigram. This enables analysis of lexical and contextual information in the model and shows the generality of the approach. Data labeled with parts of speech could show how syntactic information is represented --- given any set of labels for the input our analysis quantifies structure in representation space with respect to them.

\begin{subequations}

To formalise this in terms of transformer language models, consider mappings at the token level. Given a model $f$ that maps a set of sentences $X$ to representational space $Y$. For each sentence $x^k \in X$, the model takes as input a sequence of tokens $t_a^k, t_b^k, t_c^k ... \in x^k$ and returns a sequence of vectors $y_a^k, y_b^k, y_c^k ... \in Y^k$ where $y_a^k$ is the vector corresponding to token $a$ when it occurs in sentence $k$. While each sequence $Y^k$ is of variable length, the individual vectors are the same size. We can create a list $Y$ of all token representations from all sentences in the dataset, or a list of all tokens corresponding to a given label $Y|label$. 

\begin{equation}
    Y = [y_a^k : \forall y_a^k \in f(x^k) : \forall x^k \in X]
\end{equation}

   \begin{equation}
    Y|\text{label} = [y_a^k \text{ \emph{if} } a=\text{label} : \forall y_a^k \in Y]
  \end{equation}

\noindent We can apply the same approach to look at bigram or trigram information, where we label the representation $y_a^k$ with either bigram $(a,b)$ or trigram $(a,b,c)$. The next section explains how we estimate entropy in vector space; first we walk through the kinds of structure we measure. The estimation procedure gives us a categorical distribution which describes vector space $P(Y)$ used below.

\end{subequations}

\thinline

{\flushright \makesans \textbf{\textit{\underline{A Note on the Relationship with the Preceding Chapter}\\[1em]}}}

The formalisations presented here build on those introduced in the previous chapter, and for clarity we can look at where they overlap or are distinct. Measures of regularity and variation are the same in both. Regularity is the label-level mutual information aggregated across the entire set; Variation is label-level conditional entropy aggregated. In the previous chapter disentanglement used the Jensen-Shannon divergence between a label's distribution, and a distribution of all other labels in a set. This requires a computation of the divergence at the label level that then gets aggregated across all labels in a set. In this chapter it is instantiated as a multi-variate Jensen-Shannon divergence - the divergence between each individual label distribution and their mixture. This assesses a similar quantity as the previous disentanglement measure, but does so in a single Jensen-Shannon divergence rather than needing to aggregate across individual labels. The information proportion measure is new for this chapter, and represents a kind of normalised mutual information that reflects how regularity with respect to different sets of labels fit together in the model.

\thinline

\paragraph*{Variation} describes how much representations for a label vary in representation space. In the token case this reflects whether a model learns a single context independent representation of the token or a different representation for every sentence it occurs in. We can quantify this in terms of the conditional entropy of space given a label. The resulting quantity is related to intrinsic dimensionality \citep{levina2004maximum}, reflecting how much of representational space is used to represent a given feature in the input, but faster to compute given it requires no pairwise comparisons. In addition to the formalisation below we bound this and the regularity measure to lie between 0 and 1 to aid interpretation\footnote{Bounded by dividing by the entropy of a uniform distribution, converting entropy to efficiency.}.

\begin{equation}
    \text{\makesans variation}(Y, \text{set}) = \frac{1}{|\text{set}|}\sum_{\text{label}}^{\text{set}}H(Y|\text{label})
\end{equation}

\paragraph*{Regularity} reflects the amount of variation in representation space we can explain by knowing a label. It is bounded mutual information, and reflects the difference between overall variation in the space H(Y) and the variance in representations for a given label H(Y$|$\text{label}). It reflects how monotonically aligned representation space is with that label. In language regularity is often measured similarly \citep{smith_eliminating_2010, ferdinand_cognitive_2019} and is used to quantify how syntactically structured a system is.

\begin{equation}
    \text{\makesans regularity}(Y, \text{set}) = \frac{1}{|\text{set}|}\sum^{\text{set}}_{\text{label}} H(Y) - H(Y|\text{label})
\end{equation}

\paragraph*{Disentanglement} measures whether clusters corresponding to labels within a set are separable --- e.g. whether different tokens or bigrams are represented in different parts of space. We estimate this with a multi-variate Jensen-Shannon divergence.  This requires a mixture distribution $M$ computed by taking a mean of individual label distributions weighted by the probability of the label $M \propto  \sum_{\text{label}}^{\text{set}} P(\text{label})P(Y|\text{label})$. The divergence then looks to see if the entropy of the mixture $\mathcal{H}(M)$ can be explained in terms of the individual label distributions. The result is the mutual information between the mixture $M$ and the weights used to create it $P(\text{label})$ and so is bounded by the entropy of the weight distribution $\mathcal{H}(\text{label})$. We use this to normalise the measure so that as values approach 1 labels are maximally separable in space, and as it approaches 0 all labels in a set overlap. This is related to previous measures of entanglement  \citep{chen_isolating_2018, conklin2024representations} but is faster to compute, and allows labels to contribute proportionally to the estimate based on their probability.

\begin{equation}
    \text{\makesans disentanglement}(Y, \text{set}) = H(M) - \sum_{\text{label}}^{\text{set}} P(\text{label})H(Y|\text{label})
\end{equation}

\begin{figure}[t]
	\centering

	\includegraphics[width=\textwidth]{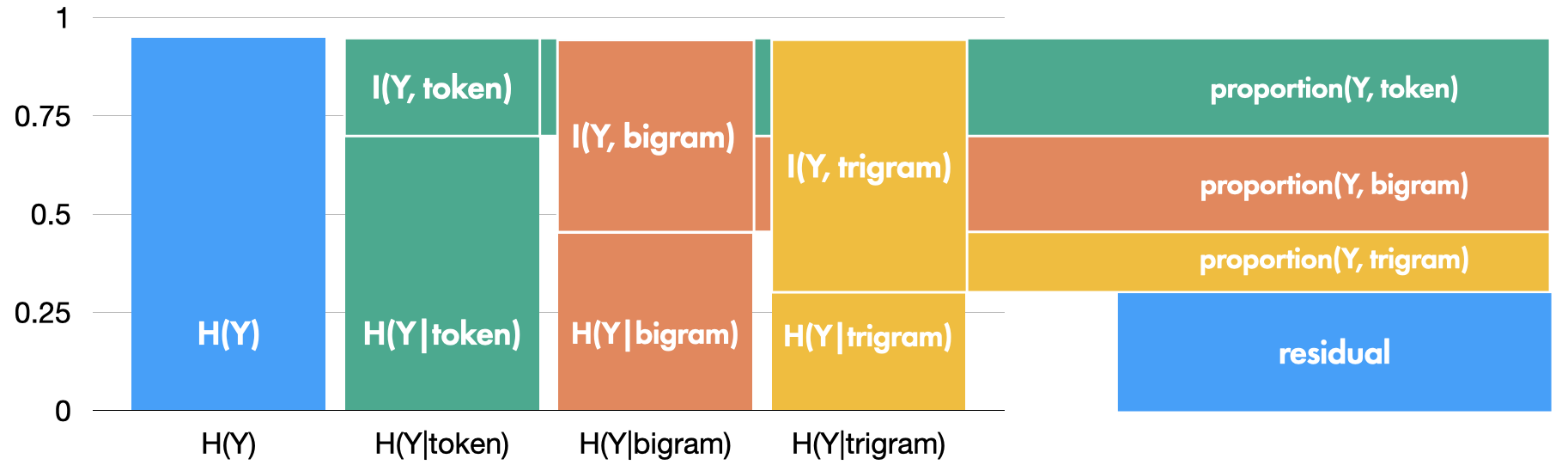}
	\caption{An exemplar showing how different information theoretic quantities relate to each other. x axis shows different sets of labels, y axis shows efficiency (entropy bounded to between 0 and 1). The lower portion of each bar is conditional entropy with respect to each label set. The top is mutual information with a given set $H(Y)-H(Y|\text{set})$. Information proportion is the difference between the mutual information of a set and its super set --- if it has one--- with residual information being information left in $H(Y)$ that cannot be explained in terms of any set of labels. Here the super set for bigrams are tokens ( given we analyse token representations occurring within a specific bigram), and the difference between their mutual information reflects the amount of information in representation space attributable to bigrams alone. Tokens have no super set, so their information proportion is equivalent to their mutual information (regularity) normalised by the entropy of the space $H(Y)$.}
	\label{fig:residual_relations}
\end{figure}

\paragraph*{Proportion} reflects the proportion of information in the model a set of labels can account for. Any information that can't be explained in terms of a label is considered `residual' - and is computed as a residual entropy \citep[see][for some discussion]{resnick_capacity_2020}. Recall that regularity describes how much variation in representations we can explain by knowing a label, or how much information in the model can be explained in terms of that label. To compute an information proportion we compute how much of the model's total information $\mathcal(H)$ is accounted for by regularity with respect to a specific label set $\text{\makesans regularity}(Y, \text{set})$. In experiments here though, the labels used nest inside each other: representations corresponding to a trigram label, are also part of a bigram label, which are also part of a token label. Which means regularity with respect to trigram information, includes regularity with respect to both bigram and token information. We separate this out by subtracting the regularity of the superseding label set $\text{\makesans regularity}(Y, \text{super set})$ if there is one and normalising by the entropy of the entire space $H(Y)$. If there is not a superseding set (as in the case of token level labels) then we simply normalise regularity by the entropy of the space. Relationships between variation (conditional entropy), regularity (mutual information) and the information proportion are shown in figure \ref{fig:residual_relations}.

\begin{equation}
    \text{\makesans proportion}(Y, \text{set}) = \frac{\text{\makesans regularity}(Y, \text{set})-\text{\makesans regularity}(Y, \text{super set})}{H(Y)}
\end{equation}

\noindent The residual, or remaining information in the model that cannot be explained in terms of a label set, is estimated by taking the label set with the highest regularity, and subtracting it from the entropy of the space. This leaves over the entropy that cannot be explained in terms of even the most regular set of labels (or any of the superseding ones). Normalising this by the entropy of the space gives us a proportion.
 
\begin{equation}
    \text{\makesans residual}(Y, \text{set}) = \frac{\text{\makesans regularity}(Y, \text{smallest set})}{H(Y)}
\end{equation}

\section{Soft Entropy Estimation}
\label{sec:softH}

There are few approaches to entropy estimation that are sufficiently fast and memory efficient to be applied to large language models. This is frustrating given information theoretic tools are well suited to quantifying complex structures in distributed systems. With soft entropy we introduce an approach that prioritises efficiency, while performing comparably to existing methods. It's worth noting that we focus on estimating the \emph{discrete} entropy rather than differential entropy. We draw inspiration from \citet{Jaynes1957InformationTA}, who notes differential entropy is not the true continuous analog of discrete entropy and proposes the limiting density of discrete points as an alternative. This takes entropy to be the divergence between a distribution and an invariant measure (usually a uniform distribution over the same support); it reflects how `non-uniform' a distribution is. Our method follows this intuition, sampling points uniformly across space, and comparing them with samples from the model. %

We define a mapping between real-valued space and information space, creating a categorical distribution that describes a model's representation space. Our estimator returns the entropy of the descriptor distribution, a quantity we call \emph{soft entropy} - distinct from the differential entropy of the space. This process is akin to `plug-in' estimation \citep[see][for review]{beirlant1997nonparametric}, where you first fit a distribution then estimate its entropy - except here the distribution we `fit' is categorical.  Existing approaches to estimating entropy of vector space often rely on discretisation with clustering \citep{sajjadi2018assessing}, or binning \citep{shwartz-ziv_opening_2017}, the approach described here can be seen as a differentiable relaxation of these methods. %

\subsection{Formalisation}\label{sec:h_estimation}
Given a set of representations $Y$ with dimensions batch size $bs$ by hidden size $h$ we take the euclidean norm, so they lie on the unit sphere (note the entire estimation process is depicted visually in figure \ref{fig:soft_h_sphere_visual}). We then sample points uniformly from the surface of the unit sphere, by drawing $n$ samples from a standard normal and taking their euclidean norm. The resulting points $S$ have dimensions $h \times n$ where $n$ is a hyperparameter controlling the number of points. To assess how close each representation is to each point we take the dot product between $Y$ and $S$. The result is a cosine similarity, which we pass through a softmax to get a distribution over points for each representation with dimensions $bs \times n$. By summing over the batch dimension and re-normalising we get a single categorical distribution that describes the space $P(Y)$ with dimensionality $1 \times n$. To get a binning based estimate we could treat each point as the center of a bin, and assign representations to the point they're closest to, rather than normalising distances.

    \begin{equation}
       \underset{1 \times n}{P(Y)} \propto \sum \text{softmax}\biggl(\underset{bs \times h}{\frac{Y}{|Y|}} \cdot \underset{h \times n}{\frac{S}{|S|}}\biggl)
    \end{equation}

\begin{figure}[hp]
\centering
\begin{subfigure}{\textwidth}
    \centering
    \includegraphics[width=0.3\textwidth]{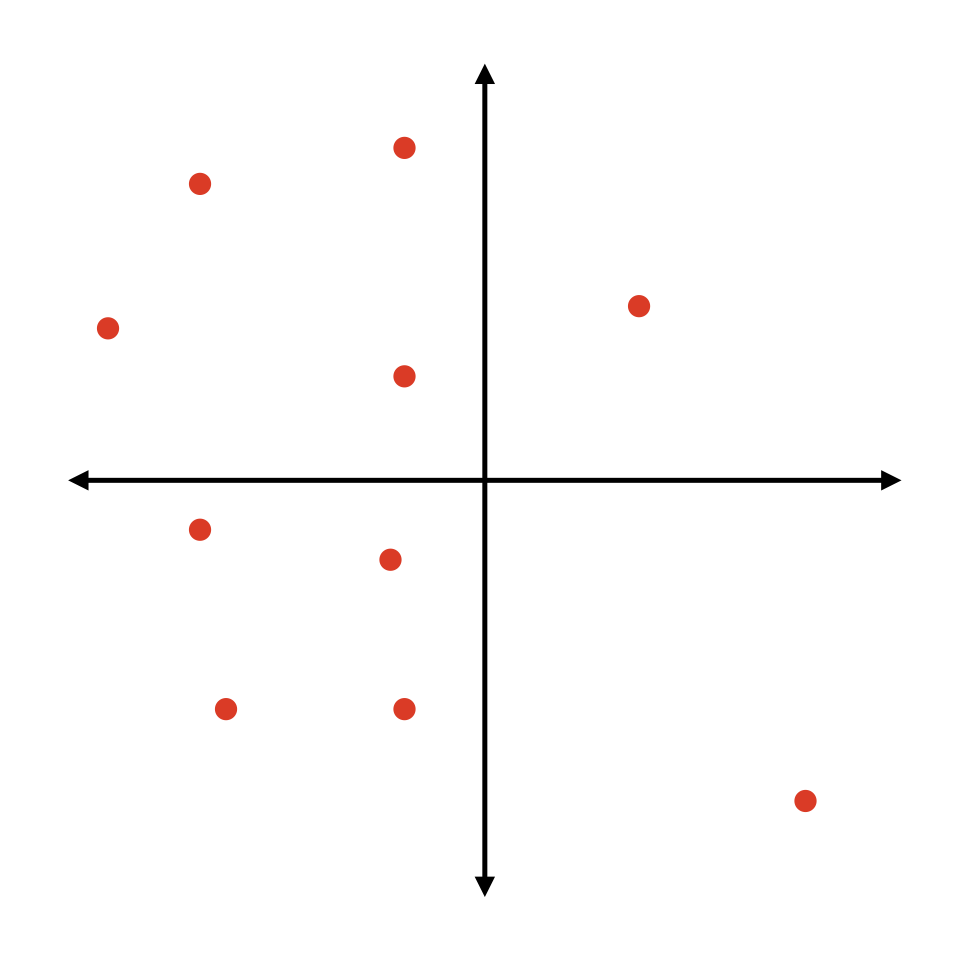}
    \includegraphics[width=0.3\textwidth]{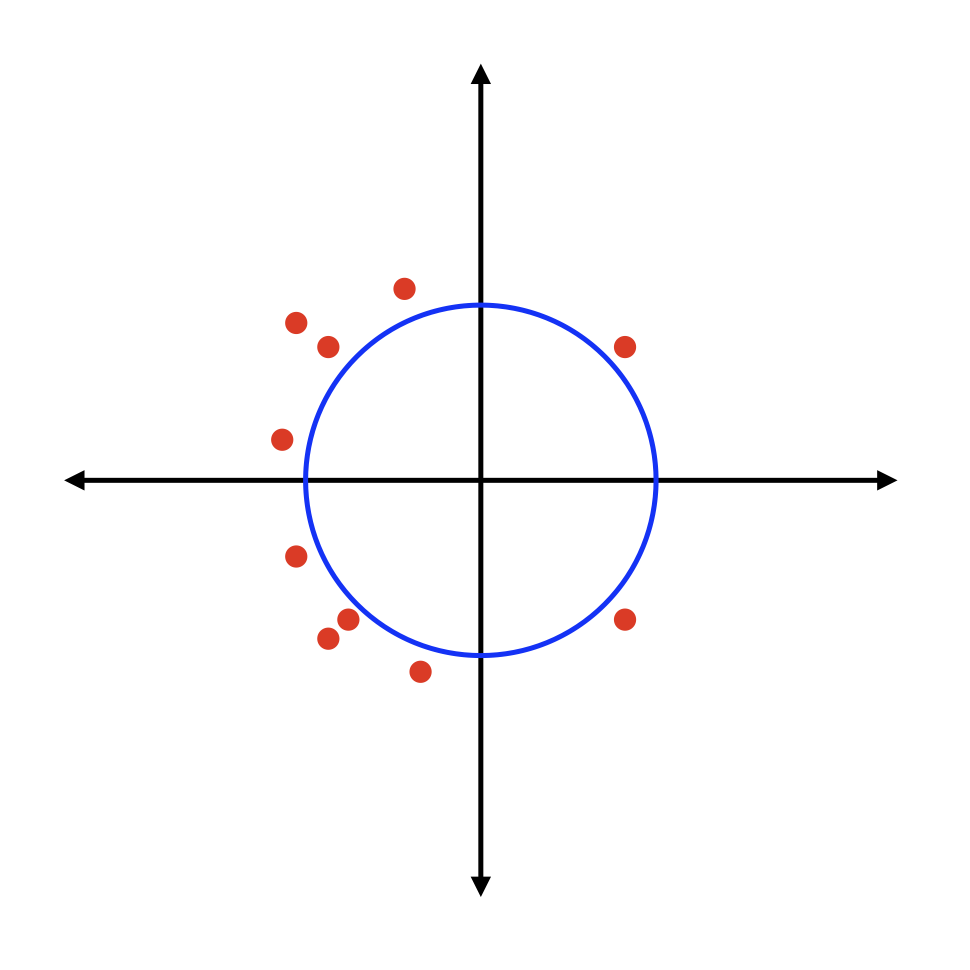}
    \includegraphics[width=0.3\textwidth]{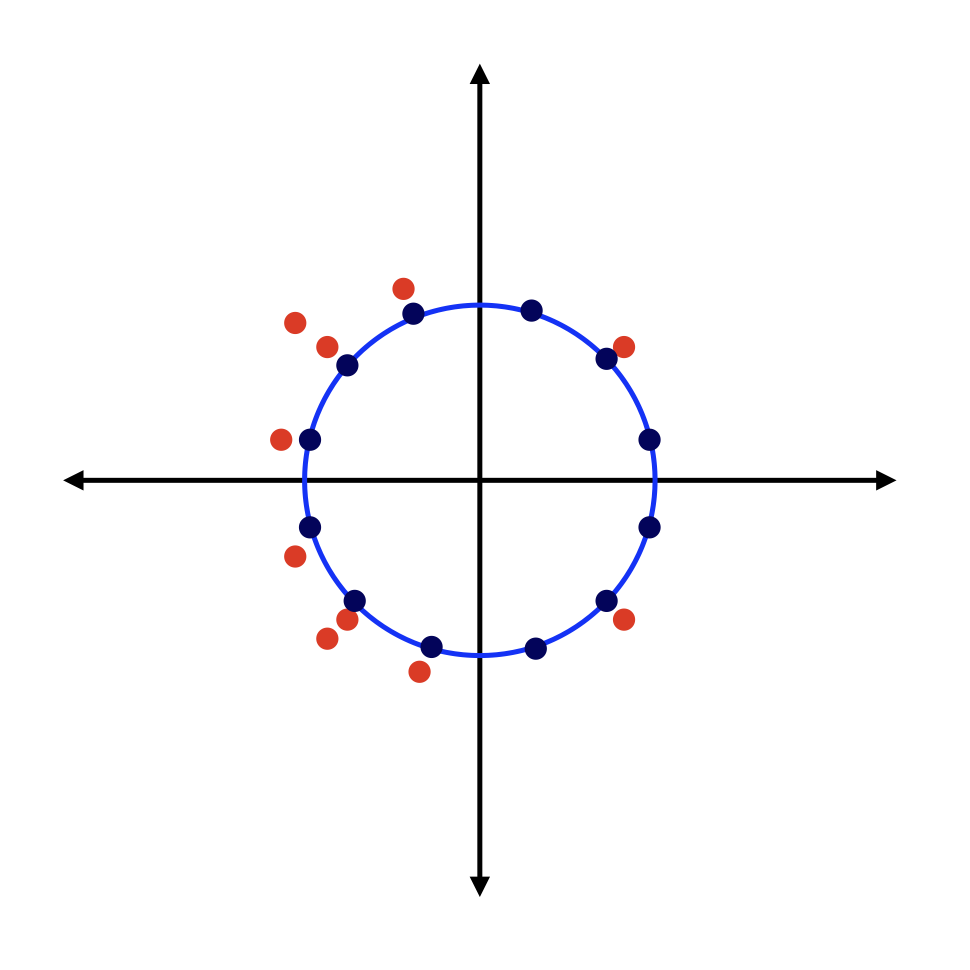}
    \caption[short]{Points in a 2D Coordinate Space shown in red (left) are normalised to lie on the unit sphere (centre) - here where multiple points are in the same location on the surface they appear as stacked. Anchor points, shown in blue, are then uniformly sampled from the surface of the sphere (right)}
\end{subfigure}

\begin{subfigure}{\textwidth}
    \centering
    \includegraphics[width=0.75\textwidth]{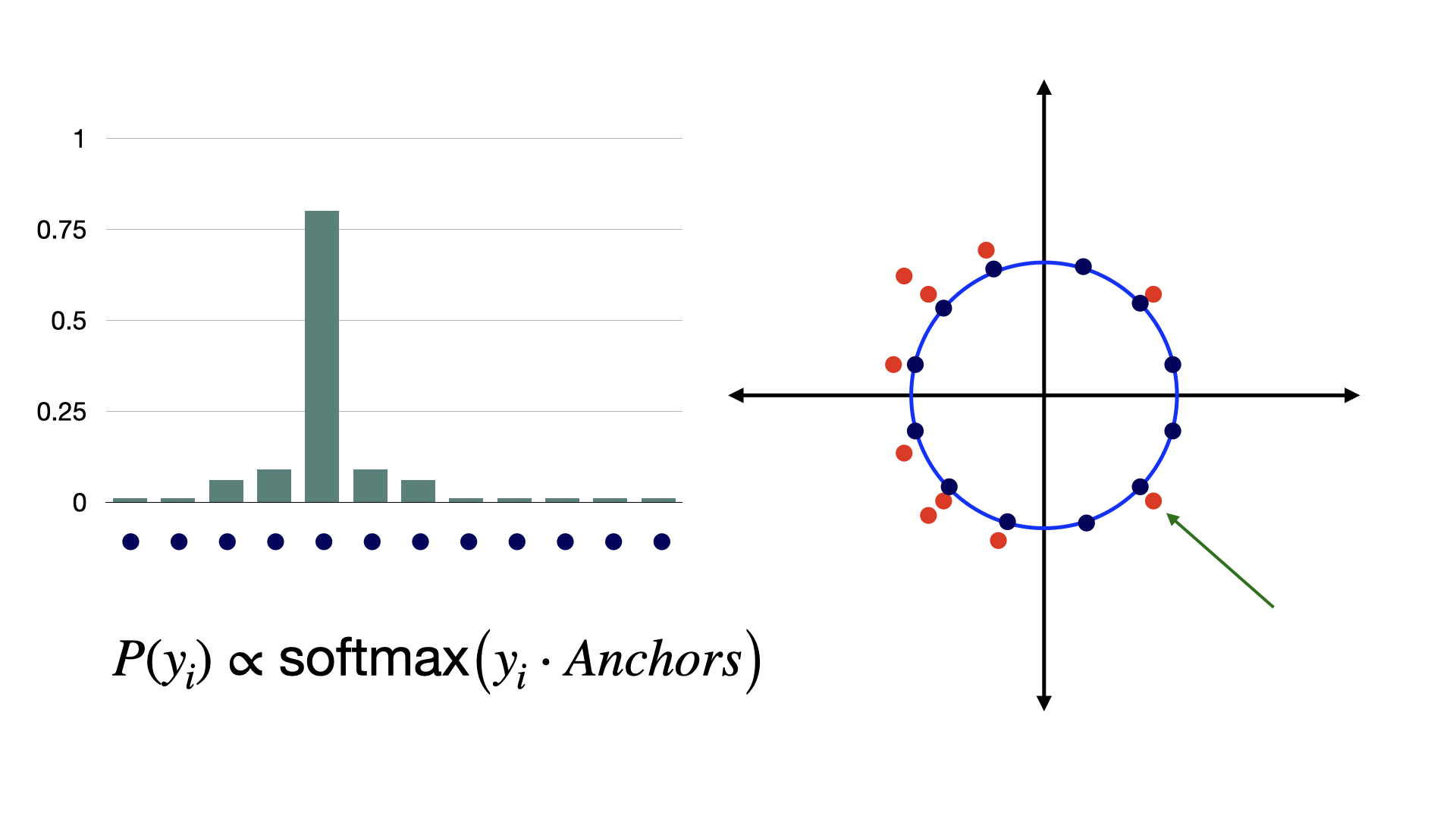}
    \includegraphics[width=0.75\textwidth]{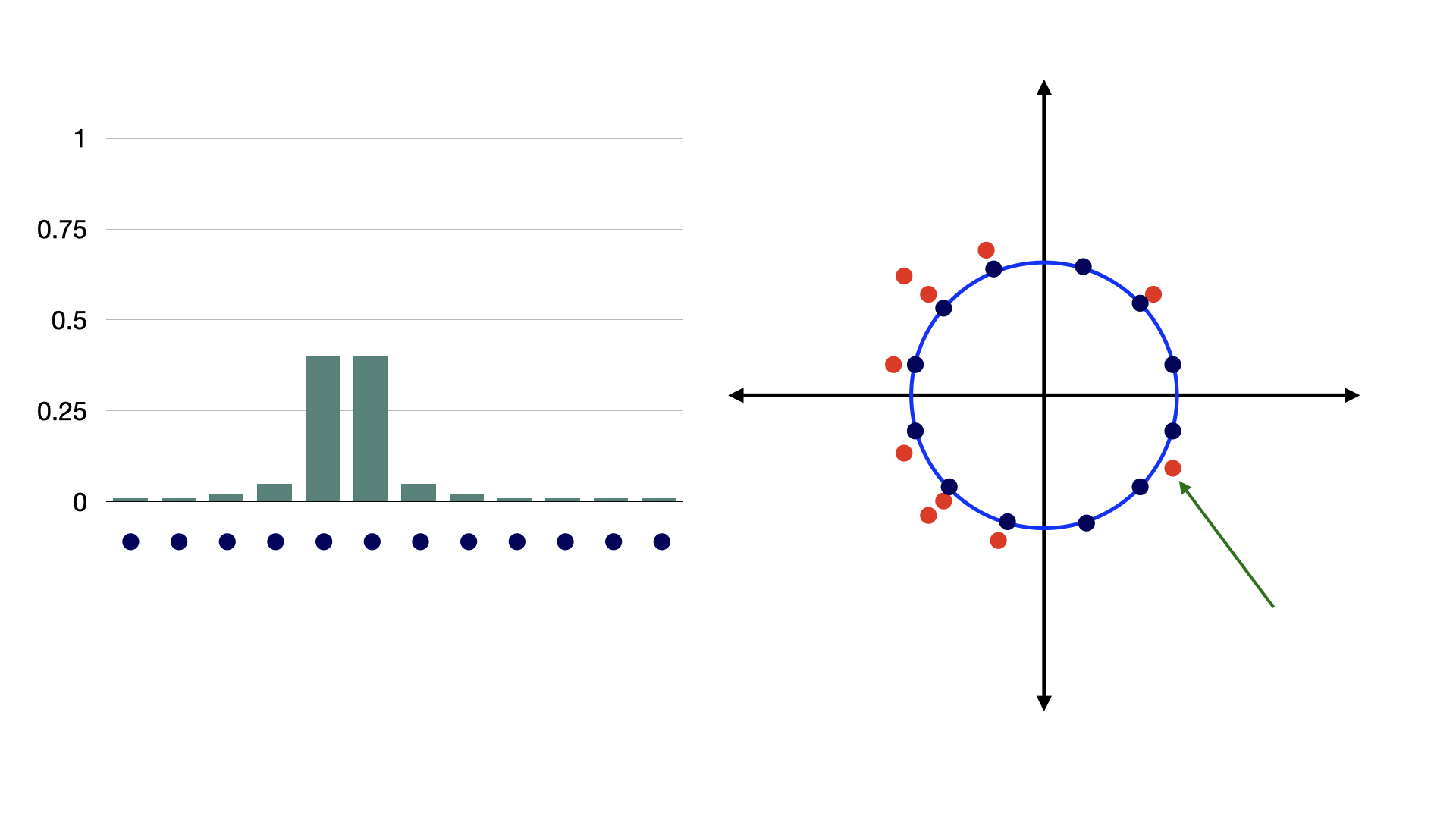}
    \caption[short]{Anchor points are used as events in a categorical distribution. To get a distribution over anchors for each red point we take its distance from each anchor - in terms of a dot product - and pass them through a softmax. Shown at top is the distribution over anchors for the point marked with the green arrow. Below is the same distribution when the point is equidistant between anchors.}
\end{subfigure}
\caption{A visual depiction of the soft entropy estimation process (continued on next page).}
\label{fig:soft_h_sphere_visual}
\end{figure}

\begin{figure}[h!]\ContinuedFloat
	\centering
    \includegraphics[width=0.9\textwidth]{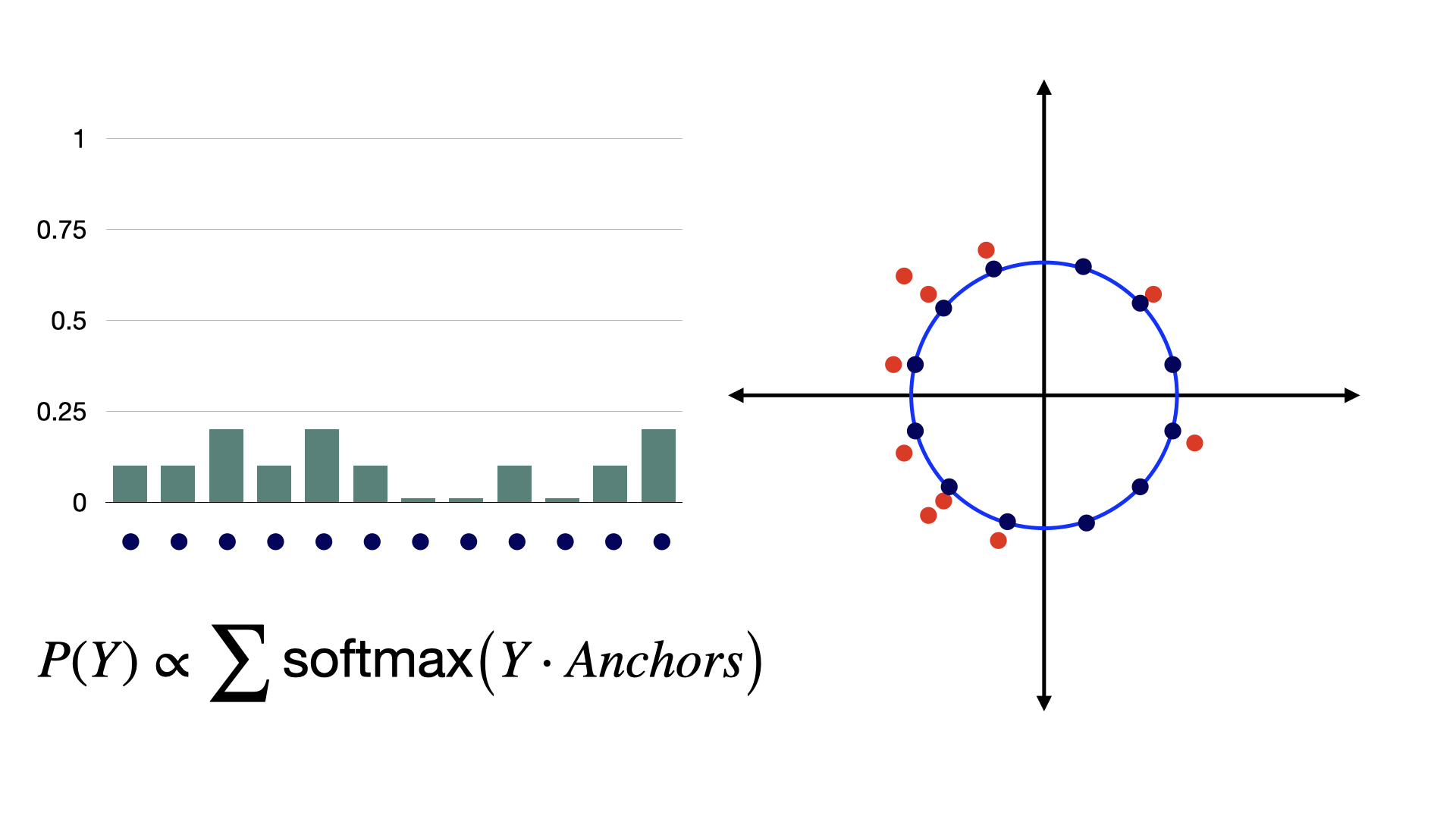}
\caption{(continued) (c) By taking a summation over the distributions for each individual red point and normalising we get a distribution that describes the entire space. Note how the distribution over events in the bar plot at left, is reflective of where red dots are distributed over the surface of the sphere on the right.}
\end{figure}

\noindent Because this gives us a categorical distribution, estimation of information-theoretic quantities is straightforward. Soft Entropy of the space follows the equation for shannon entropy: $H(Y) = -P(Y)\log P(Y)$. We can also quantify entropy in subspaces, as opposed to the entire space by applying the estimator in a multi-headed arrangement.  We reshape the representations from $bs \times hidden $ to $bs \times head \times \frac{hidden}{heads} $ and the points to $\frac{hidden}{heads} \times heads \times bins $. This allows us to estimate entropy per-head and mean across them.

H(Y) reflects how uniformly distributed representations are across angles with respect to the origin. It is maximised when representations are uniformly distributed across all 360 degrees, and approaches 0 as representations cluster across an increasingly small subset of angles. This quantity is related to anisiotropy, where representations lie in a narrow cone relative to the origin, but is dramatically faster to compute than taking pairwise cosine similarities between all representations. We draw a parallel between this measure and clustering based estimates of entropy, where representations are first clustered, then discretised \citep{sajjadi2018assessing}. Here when we project points to the unit sphere we make representations with high cosine similarity, close to each other. To get a clustering estimate we could replace the events in the categorical distribution $P(Y)$ with clusters on the unit sphere rather than uniformly sampled points. In practice sampling points is substantially faster than performing clustering.

\subsection{Parameters \& Computational Efficiency}

In the same way that discretisation methods are sensitive to the number of bins used, soft entropy is sensitive to number of `points' although less so than the discrete case: if two representations are close to each other they can't be split into separate `bins,' given we get a distribution over points for each representation rather than assigning it to a single point. This means that increasing the number of points in S doesn't necessarily have a detrimental effect on mutual information and divergences but can still inflate the estimate. In the experiments presented here we use 50 points unless otherwise noted. Additionally a softmax is not invariant to linear transformations and the distances from the dot product are bounded between -1 and 1, this can mean the default estimate is relatively high. After testing on reference distributions we opt to rescale the distances to lie between -100 and 100. This scaling factor is a parameter, like the bandwith parameter in kernel density estimation \citep{parzen1962estimation}, controlling the spread with respect to each point.

Our methods map representational space to a categorical distribution using a single dot product, softmax, and summation. These operations are differentiable, memory efficient, fast, and parallelisable. This process is non-parametric, requires no clustering, and is substantially more memory efficient than binning based approaches to entropy estimation which usually requires a step where representations are $bs \times seq \times hidden \times bins$ - using 100 bins on a model with 4096 dimensional spaces proves problematic. %

\thinline
\vspace{-10mm}
	{\flushright \makesans \textbf{\textit{\underline{Measure Visualisations}\\[1em]}}}

\begin{figure}[hp]
    \centering
    \includegraphics[width=0.9\textwidth]{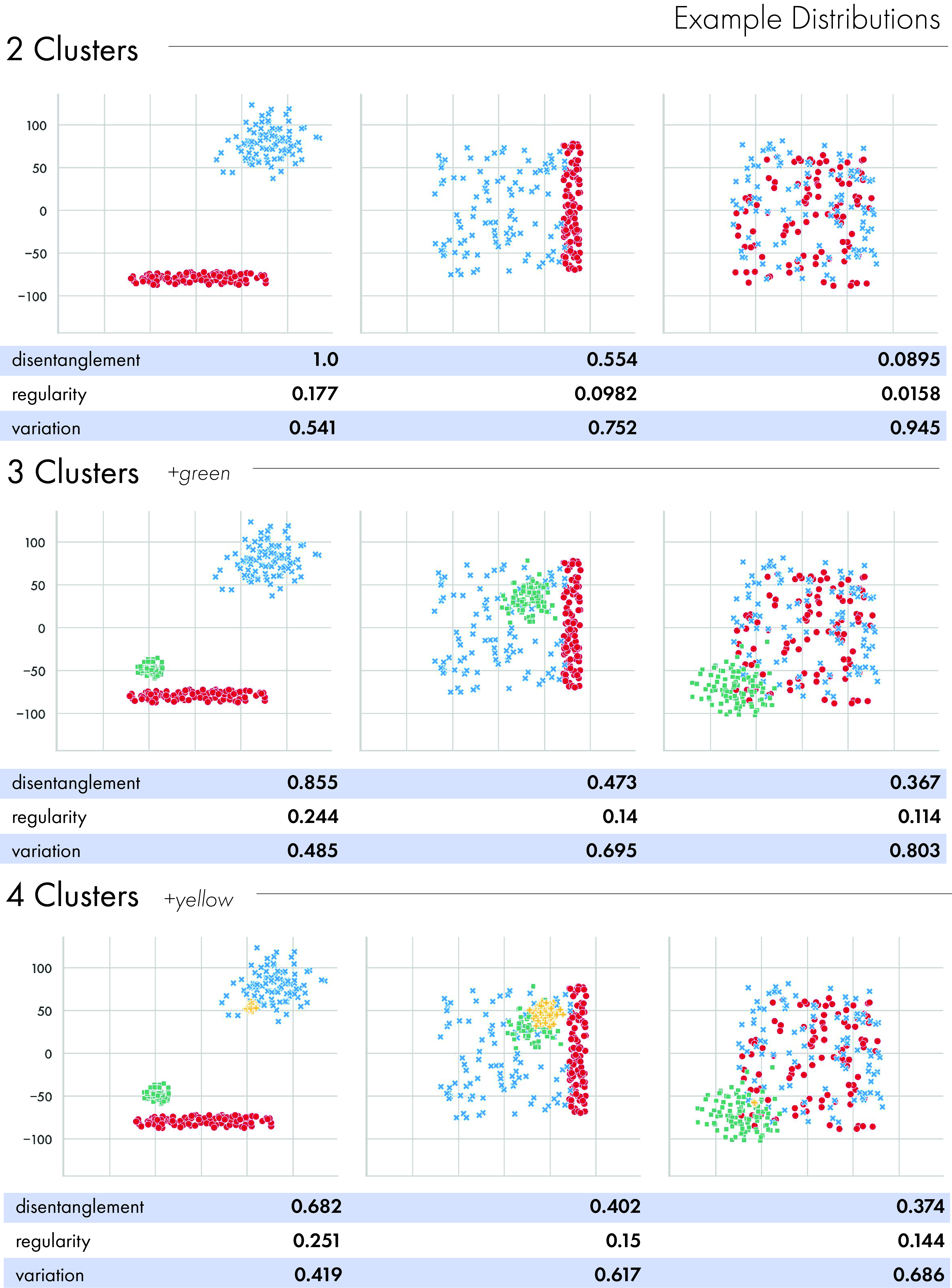}
    \caption{The soft entropy estimator applied to exemplar distributions. Examples shown for 2 to 4 clusters with each column being additive: each facet includes the distributions from above in the column. Disentanglement, regularity and variation scores for each facet are reported beneath it. In the top row disentanglement decreases with each facet from left to right, with the leftmost example being fully separable, and the rightmost fully entangled. Looking down the right column, adding the green cluster increases regularity as it is largely separable from the red and blue. The addition of green also decreases variation, as the average cluster size has decreased.}
    \label{fig:cosine_disentanglement}
\end{figure}

\noindent Figure \ref{fig:cosine_disentanglement} visualises each of the information structure measures using soft entropy. These show the measures here applied to the same example visuals as in the previous chapter (figure \ref{fig:dimension_wise_measures}) but leveraging soft entropy rather than dimension-wise discretisation. Each visualisation contains 2 to 4 different distributions indicated by colour - each corresponding to a `label' in the analysis. For each distribution either a uniform, or multivariate normal distribution is selected at random, then randomly parameterised. 100 samples are drawn from each distribution, and the 4 measures introduced above are applied to these samples. To enable straight-forward visualisation each distribution is 2 dimensional. These visuals help us to link each measure to properties of clusters in real-valued space. Broadly the values here are similar to the dimension-wise approach, but can better account for clusters that overlap on each dimension but are separable in 2D space.

\section{Validation \& Comparison With Existing Methods}

As this represents a novel approach to entropy estimation for vector spaces it is important to relate it to existing approaches to this problem. Doing so gives a sense of how precise the estimator is, and the degree to which the quantity it measures is related to existing notions of entropy. This in and of itself proves somewhat challenging - other approaches to estimating the shannon entropy of continuous spaces are also \emph{estimators}, and so do not provide a ground truth value with which we can compare. In an effort to provide an indication of both how the soft entropy estimate relates to existing estimators, and ground truth estimates, we relate our entropy estimator to differential entropy; differential entropy is the usual continuous analog of shannon entropy. As a reminder the equation for shannon entropy is:

\begin{equation}\label{eq:discrete_h_llm}
	\mathcal{H}(x)=-\sum_{i=1}^{n} p(x_i) \log p(x_i)
\end{equation}

\noindent Differential entropy replaces the summation in the equation above with an integral. For a given density function $f$ differential entropy $\mathcal{D}$ can be expressed as:

\begin{equation}\label{eq:differential_h_llm}
	\mathcal{D}(f)=-\int_{x} f(x) \log f(x) dx
\end{equation}

\noindent for certain density functions, like a gaussian, this has an analytic solution.

 Given that computation of differential entropy requires commitment to a particular density function, and computing an integral, it is common to instead discretise representations and compute an entropy estimate using equation \ref{eq:discrete_h_llm}. This is the approach utilised in the preceding chapter, and the general intuition underpinning the soft-entropy estimator introduced here. The histogram entropy estimator (described in \citep{paninski_estimation_2003}) allows conversion between discrete and differential entropy estimates. To do so a histogram estimator takes the equation for discrete entropy and inside of the logarithm divides the probability of bin $i$, $p(x_i)$ by the width of the bin $w(x_i)$.
 
 \begin{equation}\label{eq:histogram_estimator}
	\mathcal{D}_h(x)=-\sum_{i=1}^{n} p(x_i) \log \left( \frac{p(x_i)}{w(x_i)} \right)
\end{equation}

\noindent This converts the estimate from shannon entropy, to differential entropy, via a method that can be applied to any "binning based" entropy estimate. 

With this in mind we return to the problem at hand - benchmarking the soft entropy estimator against a ground truth. We select a gaussian distribution with a closed-form differential entropy, drawing samples from it, before applying the soft entropy estimator to those samples, and converting the resulting estimate to differential entropy. This gives us a ground-truth entropy value (differential entropy of the underlying gaussian), and an estimate of that value (soft entropy estimate, converted to differential entropy via the histogram method). With these two quantities we can get a sense of the relationship between the estimator introduced here and existing notions of entropy in continuous spaces. I will note that this multi-step procedure is imperfect for at least two reasons worth highlighting. First, the accuracy of the resulting estimate is a product of all steps in the process -  the histogram conversion to differential entropy may introduce error separate from any error in the soft entropy estimator itself. Second, if we want to have a ground-truth with which to compare we are constrained to benchmarking against distributions with a closed form differential entropy. This is a surprisingly short list of density functions, and we only look at gaussians here. While gaussian mixtures would be more interesting - and more representative of the kinds of distributions found in models' hidden states- they offer no analytic solution to their differential entropy and so would require us to also select an estimation procedure for the `ground-truth' against which we are benchmarking. With these caveats in mind, the results that follow still offer a substantive sanity check that the estimation procedure introduced here measures a quantity with clear relation to existing notions of entropy, and existing estimators.

\begin{figure}[hp]
	
	\includegraphics[width=\textwidth]{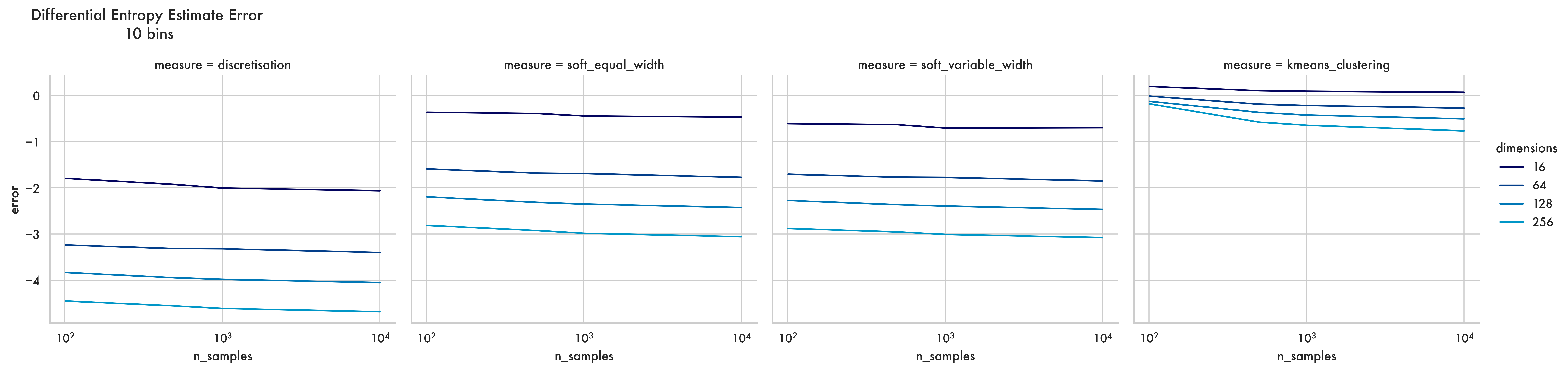}
	\includegraphics[width=\textwidth]{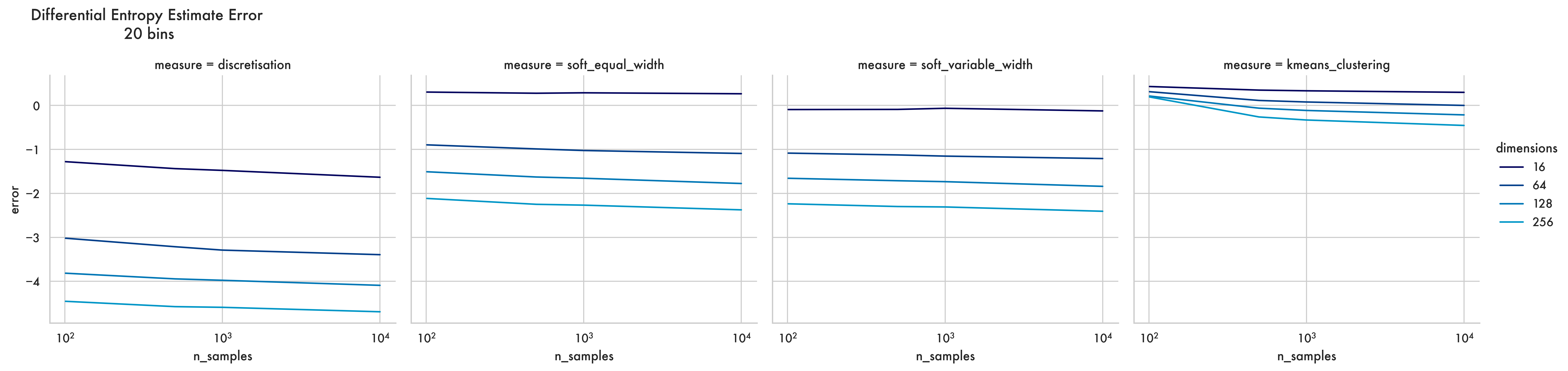}
	\includegraphics[width=\textwidth]{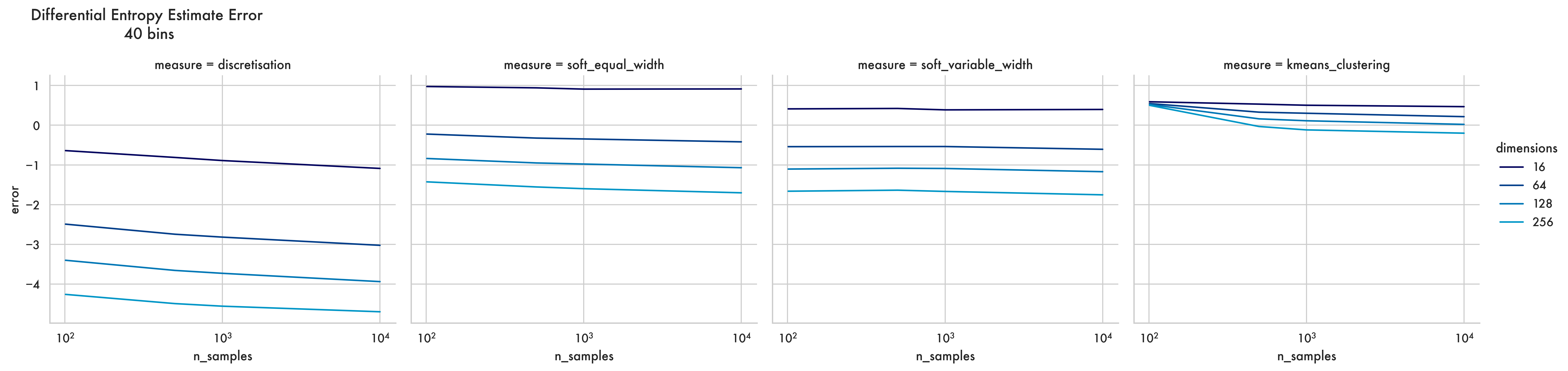}
	\includegraphics[width=\textwidth]{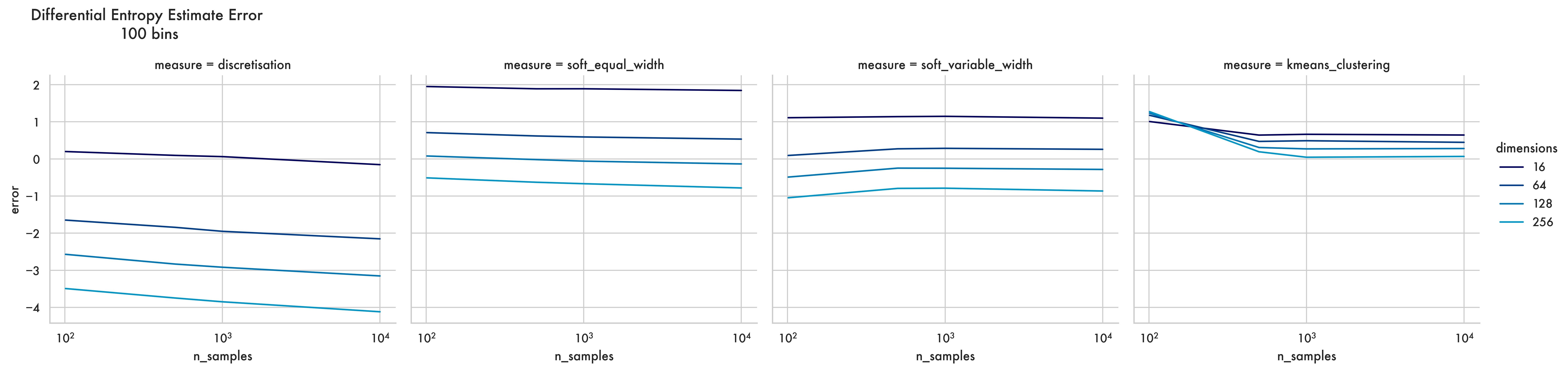}
	\caption{Comparison of Entropy Estimation Methods | In each facet the x-axis reflects number of samples used to compute the estimate, y-axis shows error of the estimator relative to the closed-form differential entropy of the underlying distribution. Each line is the mean error of the estimator applied to samples from 1000 different random multivariate normal distributions. Different lines indicate different dimensionalities for the distribution, ranging from 16 to 256. Columns are different entropy estimation methods, from left to right: full discretisation, soft entropy using equal width bins in the histogram estimator's conversion to differential entropy, soft entropy using variable width bins, k-means clustering. Rows reflect different numbers of bins, or clusters, ranging from 10 to 100.}
	\label{fig:soft_entropy_benchmarking}

\end{figure}

\subsection{Comparing Soft Entropy to Existing Entropy Estimators}

In addition to relating our soft-entropy estimator with differential entropy, we also compare against two other discretisation-based estimators: fully discretising (akin to the preceding chapter), and k-means clustering (as described in \textcite{sajjadi2018assessing}), results are shown in figure \ref{fig:soft_entropy_benchmarking}. In each plot the x axis shows the number of samples used to compute the estimate, the y axis shows the difference between the closed form entropy of the distribution samples are drawn from and the estimator. Positive values indicate the estimator has over-estimated the entropy of the space, while negative values reflect under-estimation. The x-axis gives a notion of sample efficiency, showing how the estimates change as more samples are provided from the underlying distribution. Each column is a different entropy estimation method, from left to right these methods are full discretisation, soft entropy estimation with equal width bins used in the histogram conversion, soft entropy with variable width bins used in the conversion, then on the right k-means clustering. Each row in the visual uses a different number of bins, ranging from 10 to 100 to give an idea of how number of bins affects each estimator but also how sample efficient different estimators are. Each line is the mean of 1000 runs of the simulation each using a different randomly generated normal distribution. Additionally the different lines on each plot reflect different dimensionalities - meaning we perform this benchmarking on representations ranging from 16 dimensions to 256 to affirm that our estimator performs well in a variety of different spaces.

First of all, results show that the soft entropy estimator behaves similarly to the two existing approaches - discretisation and clustering. This is reflected by soft entropy consistently having a small error (the y axis in figure \ref{fig:soft_entropy_benchmarking} is close to 0), meaning the estimator consistently gets close to the ground-truth entropy of the distribution from which samples were drawn. This acts as a confirmation that the quantity we assess with this estimator has clear links to existing formalisations of entropy. Being directly relatable to the closed form differential entropy of a distribution suggests that soft entropy is best seen as a more scalable, performant estimator of existing quantities rather than introducing some new quantity unrelated to existing work.

Additionally the soft entropy estimator appears to be more sample efficient than either discretisation or clustering, with lower error than existing approaches when using only 100 samples from the distribution. In the general case discretisation seems to consistently under-estimate the entropy of the space, especially in higher dimensions. By contrast clustering tends to over-estimate, with soft-entropy ending up in between. Consistently providing more accurate estimates than full discretisation, but slightly less accurate than clustering particularly in higher-dimensional spaces. It is worth noting that the slight increased precision of clustering relative to soft entropy comes at dramatically higher computational complexity: clustering requires the convergence of a clustering algorithm for each estimate, while soft entropy requires only a dot product. Further benchmarking showing the effects of computing subspace entropies on the sample efficiency of the soft entropy estimate can be found in appendix \ref{appendix:soft_h_benchmark}.

\section{Experiments}
We use our measures of structure in a mapping, and soft entropy estimation to analyse properties of large language models in three ways. First we look at the how structure develops over the course of training in an encoder-only transformer, analysing 5 different initialisation of BERT over 2 million training steps in section \ref{sec:timecourse}. In section \ref{sec:size} we look at how model size affects representational space in both encoder and decoder only models, comparing structures inside decoder-only models ranging from 14 million parameters to 12 billion from the Pythia collection of models \citep{biderman2023pythia}. We also look at different sizes of BERT released in \cite{turc2019well}, which allows us to make more precise comparisons varying number of layers, or hidden size independently, rather than just overall parameter count. Finally in section \ref{sec:predict} we look at the relationship between representation structure and downstream task performance. 
We use the Multiberts \citep{sellam_multiberts_2022}, 25 BERT base models that differ only in their initialisation, correlating their representation structure at the end of pretraining with performance 2 million steps of fine-tuning later. 

\subsection{Estimating Entropy To Enable Model Comparisons}
Making fair comparisons between different models is often challenging given differences in number of layers and dimensionalities. Previous information theoretic analyses in deep-learning often report estimates for each layer separately \citep[e.g.][]{shwartz-ziv_opening_2017, voita_language_2021}, which can make overall interpretation and comparison difficult. Instead we look at a model's hidden state as a single random variable distributed across layers. In practice though larger hidden states will have more information, what we want in order to make fair comparisons is a relative entropy estimate, reflecting how much information a representation space encodes proportional to its size.

To this end we report two different quantities, \emph{layer entropy} and \emph{subspace entropy}. For layer entropy we compute an estimate at each layer, then mean across them. This lets us directly compare models of the same dimensionalities but differing depths. For subspace entropy we apply the soft entropy estimator in a multi-headed arrangement as described in section \ref{sec:h_estimation}. This lets us break representation spaces into lower dimensional subspaces; in the results here subspace entropy is computed over 32-dimensional spaces across every layer in the model then aggregated. This lets us compare entropies over the same sizes subspace for models with different overall dimensionalities. While breaking a vector into subspaces may break some cross-dimensional dependencies we believe that this effect is relatively small - results testing this on sample distributions are included with other entropy estimate benchmarking in the appendix \ref{appendix:soft_h_benchmark}.

\begin{figure}[hp]
	\centering

	\includegraphics[width=\textwidth]{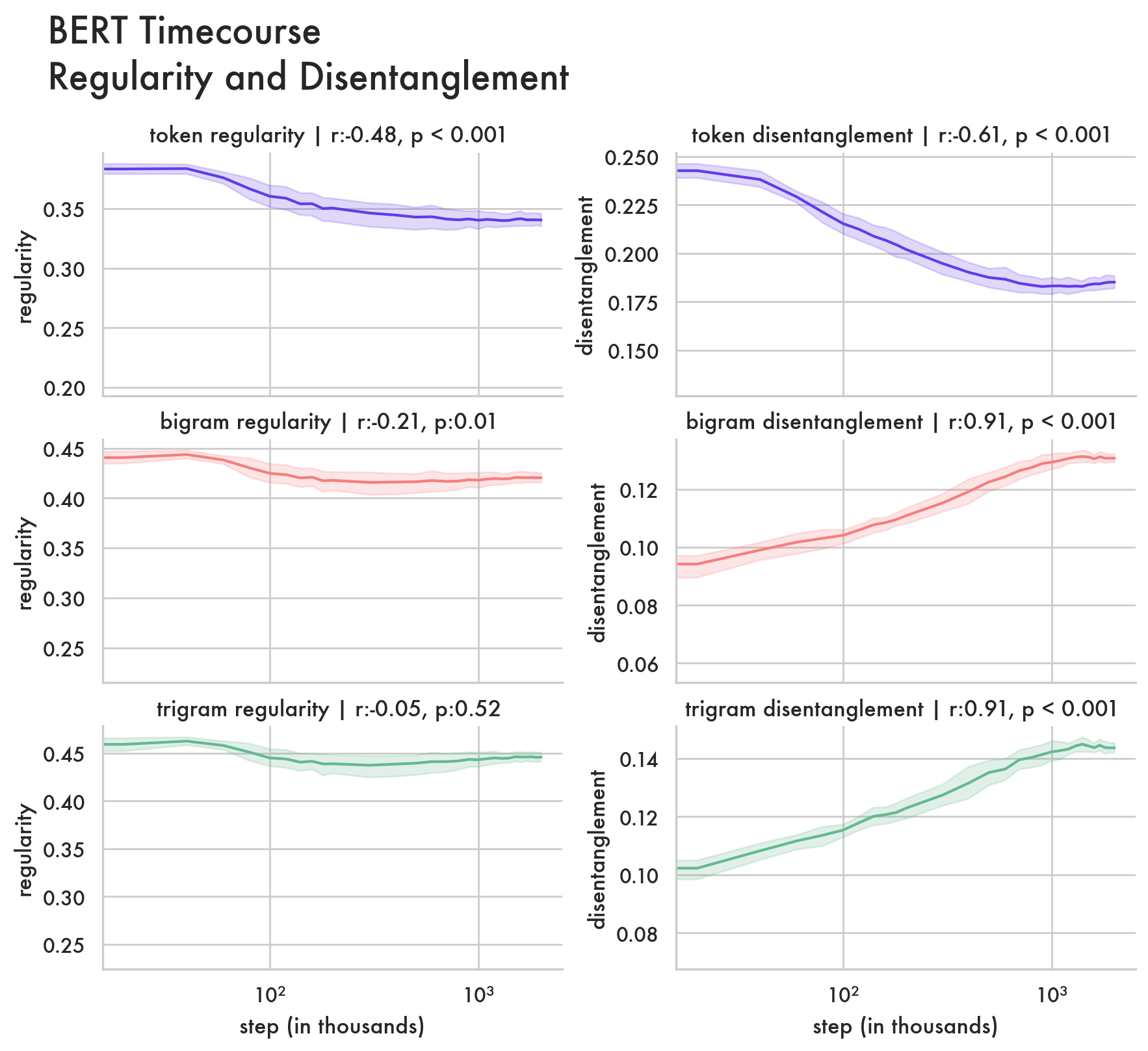}
	\caption{Information Structure with respect to 256,000 sentences from wikipedia over 2 million steps of training. Each line represents the mean of 5 different initialisations of BERT with shading representing 95\% confidence intervals. Also included above each facet is a spearman correlation between x and y. Estimates here use layer entropy, given there's no need to compare different dimensionalities. Shown here are regularity and disentanglement, with variation and information proportion on the following page. Note that checkpoints released by \protect\citet{sellam_multiberts_2022} are only every 20,000 steps of training. Timecourses for decoder-only models visualised in figure \protect\ref{fig:decoder_timecourse} provide visualisations of information structure in early training. (continued on next page)}
	\label{fig:timecourse}
\end{figure}

\begin{figure}[hp]\ContinuedFloat
	\centering

	\includegraphics[width=\textwidth]{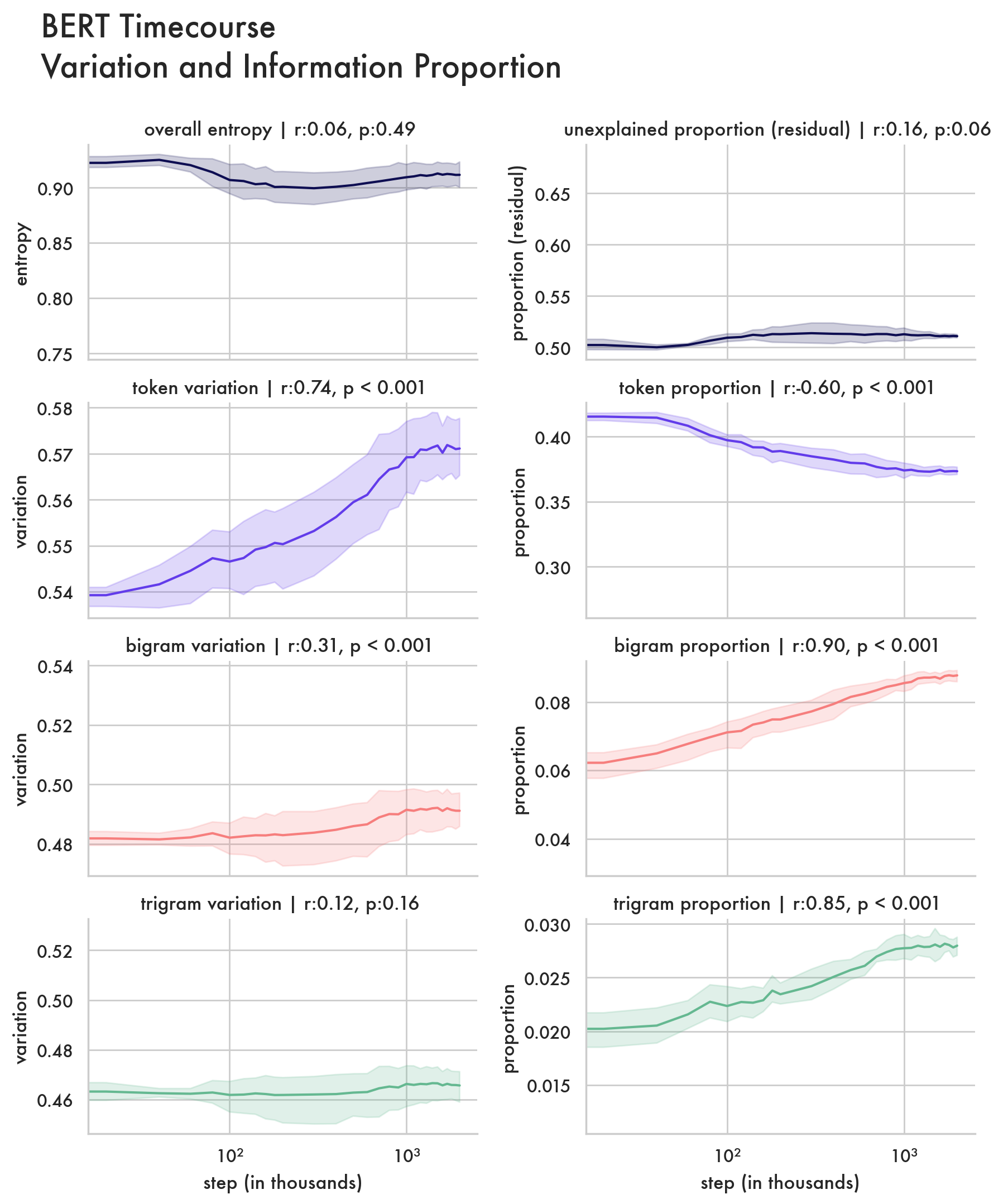}
	\caption{(continued): Shown here are the variation and information proportion timecourses for the same 5 BERT models over training. Also shown in the top two facets are the overall entropy of the space and the unexplained (residual) information proportion over training. Overall entropy exhibits minor fluctuations but no overall pattern of compression - which is present in the decoder-only model timecourse shown in the next figure.}
\end{figure}

\subsection{When Structure Emerges During Training}\label{sec:timecourse}

We look at 5 different initialisations of BERT over the course of 2 million training steps (model checkpoints also released as part of \citet{sellam_multiberts_2022}). At each checkpoint we compute our 3 structure measures with respect to token, bigram and trigram labels from the wikipedia data. We choose to use these labels because they are known for virtually every text dataset that's fed into a model. %

Main findings are shown in figure \ref{fig:timecourse}. Overall trajectories for each measure are remarkably similar to the phases of training described in \citet{conklin2024representations}, which applied a similar analysis to 3 layer encode-decoder transformers trained on a single semantic-parsing task - suggesting some generality to this characterisation of training dynamics in deep-learning. At the start of training ($<100,000$ steps) representations quickly align with token-level information, with distinct tokens becoming represented in distinct, disentangled parts of space. Past this point the dynamic shifts as representations begin to contextualise. Token disentanglement drops significantly, while bigram and trigram disentanglement increase. These likely contrast because in order to better represent lower-level information like bigrams, separate tokens need to spread out (variation increases) and overlap (disentanglement decreases). This process of contextualisation is the defining dynamic of the majority of training. Unlike findings in \citet{conklin2024representations}, later stages of training are not characterised by overall compression of the space (overall entropy decreasing), this may be a difference between single task models and LLMs or may reflect that BERT was substantially undertrained, as noted in \citet{liu2019roberta}.

\begin{figure}[hp]
	\centering

	\includegraphics[width=\textwidth]{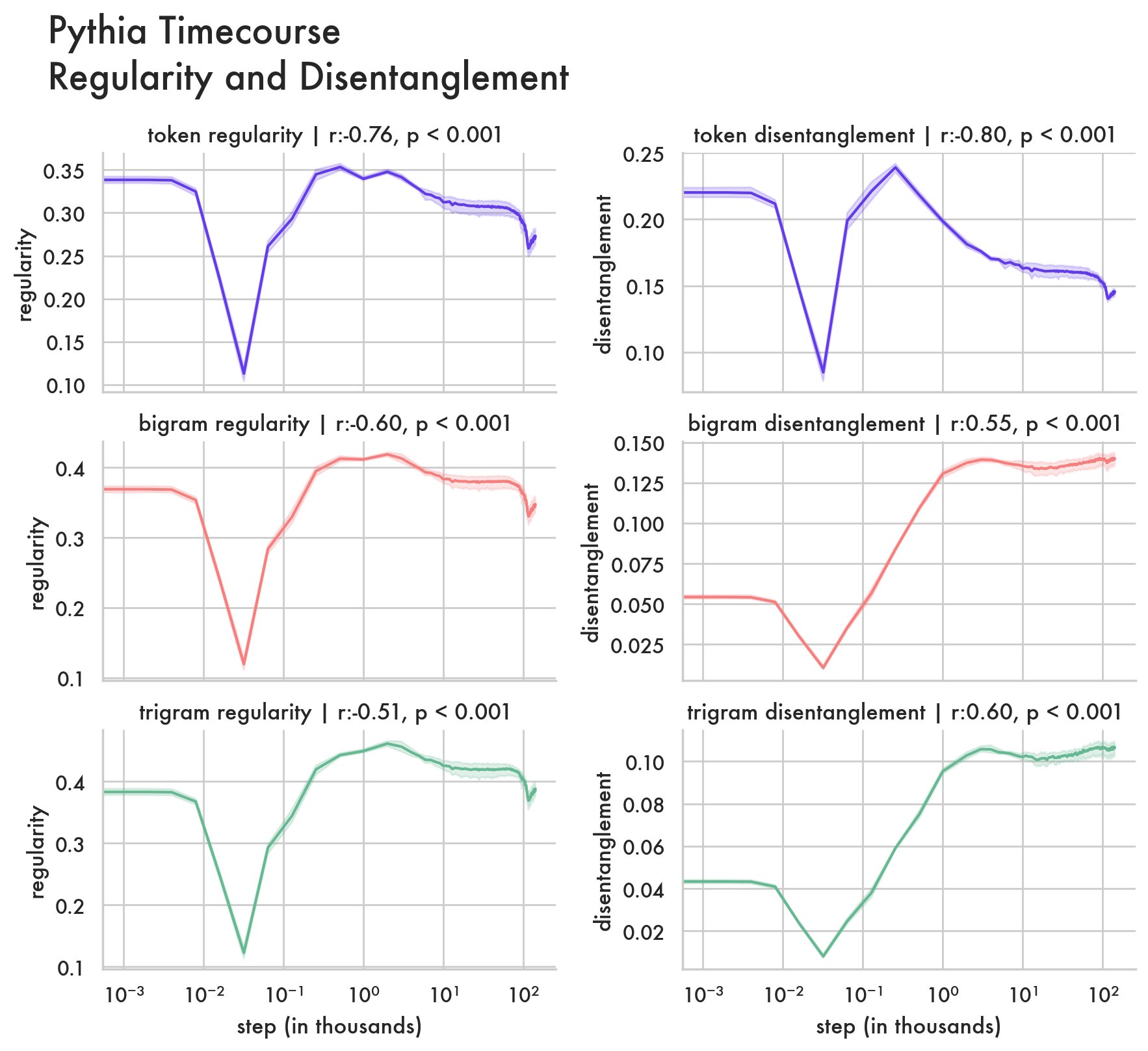}
	\caption{Information Structure with respect to 256,000 sentences from wikipedia over 2 million steps of training. Each line represents the mean of 5 different initialisations of the Pythia 410m parameter model with shading representing 95\% confidence intervals. Also included above each facet is a spearman correlation between x and y. Estimates here use layer entropy. Note that unlike the BERT model timecourse shown in the previous plots these include log-spaced model checkpoints between step 0 and step 20,000. As a result the dramatic spike across all measures at 10e-2.5 steps does not appear on the BERT timecourses - it may still take place, but we lack the checkpoints to verify. (continued on next page)}
	\label{fig:decoder_timecourse}
\end{figure}

\begin{figure}[hp!]\ContinuedFloat
	\centering

	\includegraphics[width=\textwidth]{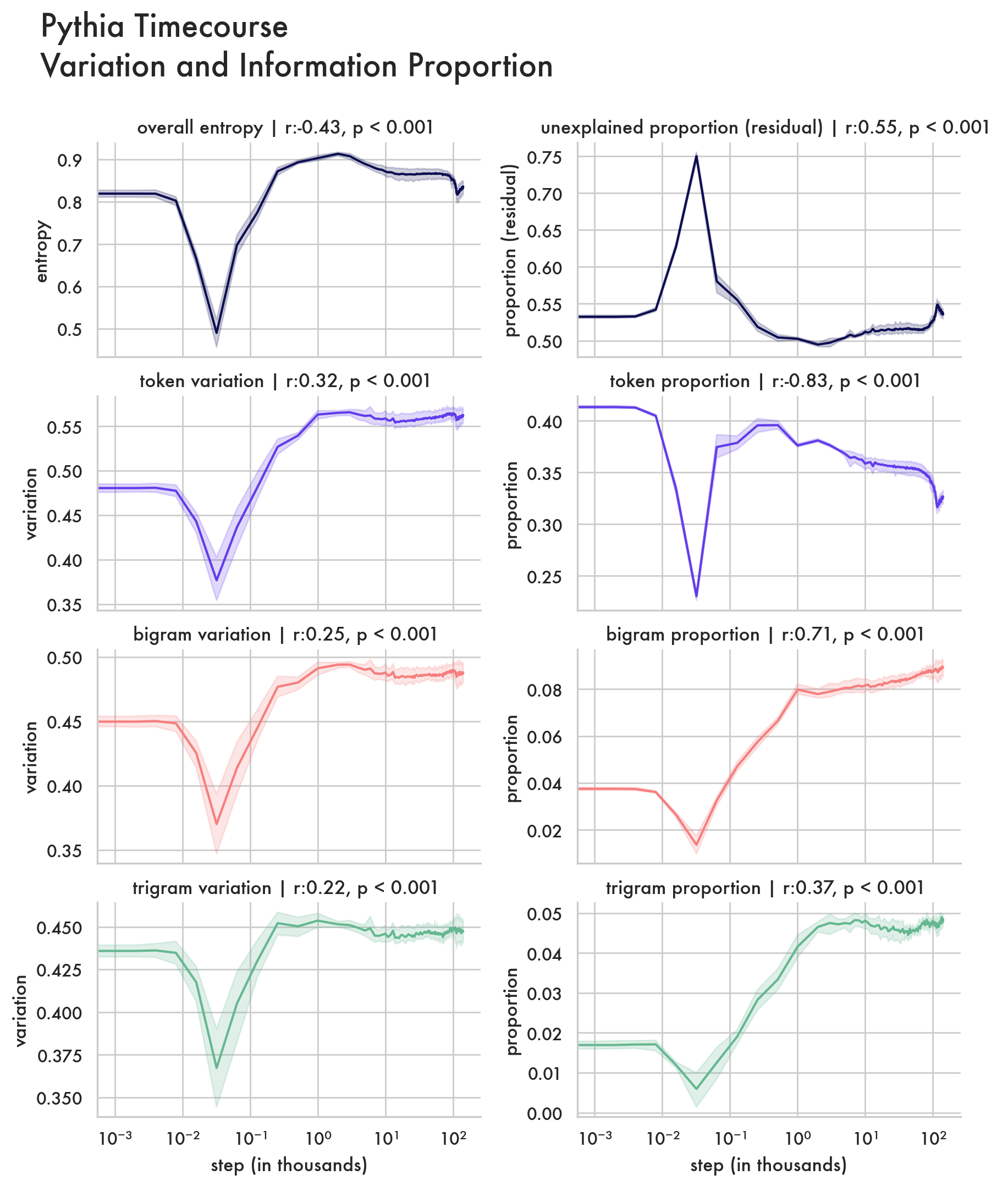}
	\caption{(continued): Shown here are the variation and information proportion timecourses for 5 Pythia 410 million parameter models models over training. Also shown in the top two facets are the overall entropy of the space and the unexplained (residual) information proportion over training. Overall entropy exhibits a significant trend of compression later in training. Note that during the spike early in training, the space compresses and the unexplained proportion accounts for 75\% of the space before the token proportion starts to steadily increase.}
\end{figure}

\subsubsection*{Decoder-Only Model Timecourse}\label{sec:decoder_timecourse}

To compare how structure develops over time in encoder only models (like BERT) and decoder-only models, we also look at the trajectory of 5 different initialisations of the 410 million parameter Pythia model. We select this parameterisation because it is comparable in scale to BERT - differing only in embedding matrix parameters - and uses the same number of layers and attention heads. Main Findings are shown in figure \ref{fig:decoder_timecourse}, it is important to note though that the the BERT models released by \citet{sellam_multiberts_2022} (visualised previously) only provide checkpoints every 20,000 steps early in training. By contrast the Pythia models include log-spaced checkpoints early on, allowing us to look at what happens during the first 1000 steps. This early phase in particular closely resembles the results from \citet{conklin2024representations} for models trained on a single task. With the space quickly compressing, then expanding as representations align with token-level information, before a long contextualisation phase. The pattern is also broadly similar to the timecourse of the encoder-only model, although here the decoder model compresses its overall representation space significantly in the latter phase of training.  This may reflect architectural differences between the models, or differences in their objectives - with encoder-only models predicting a masked word given the surrounding context, and decoder-only models predicting the next word based on preceding context.

Overall the timecourse results are striking given their similarity across different kinds of large-language models, and to models from previous work trained on a single task. This suggests some real generality to the two-phase framing of Deep-Learning training trajectories adopted in \citet{conklin2024representations}. 

\begin{figure}[htp]
	\centering
	\includegraphics[width=\textwidth]{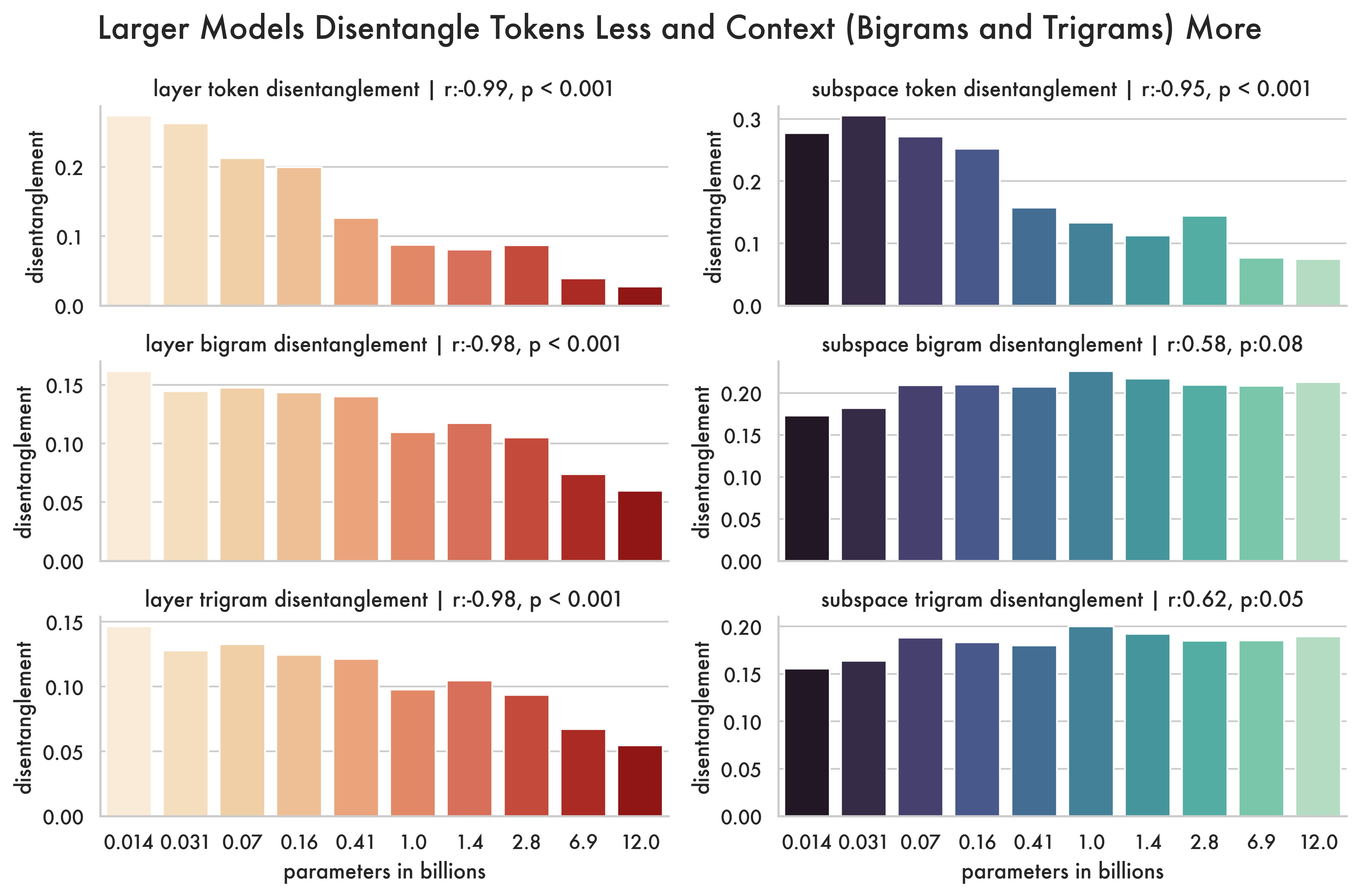}
	\thinline
	\vspace{10mm}
	\includegraphics[width=\textwidth]{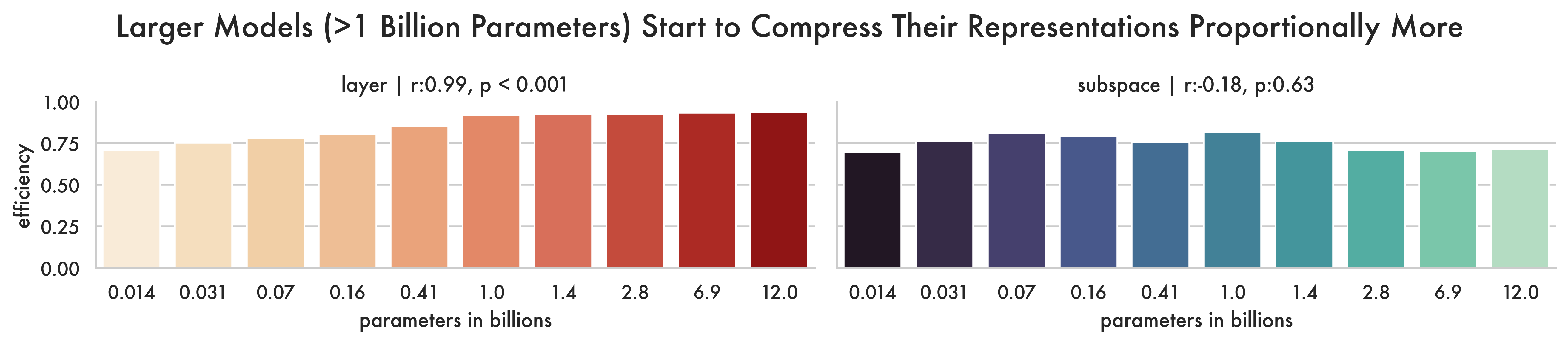}
	\caption{Analyses computed on pythia decoder-only models ranging from 14m parameters to 12 billion. Red/Orange bars show mean layer entropy, blue/green bars show mean subspace entropy. Above each plot is a spearman correlation between x and y. \textbf{ Top:} y-axis shows disentanglement for different model sizes. Subspace entropy shows contextual information (bigrams and trigrams) are more disentangled in larger models, with size significantly correlating with disentanglement \textbf{Bottom:} y-axis shows overall entropy of each model size. While layer entropy increases monotonically with size as expected - subspace entropy begins to compress in larger models.}
	\label{fig:pythia_size}
\end{figure}

\begin{figure}[htp]
	\centering
	\includegraphics[width=\textwidth]{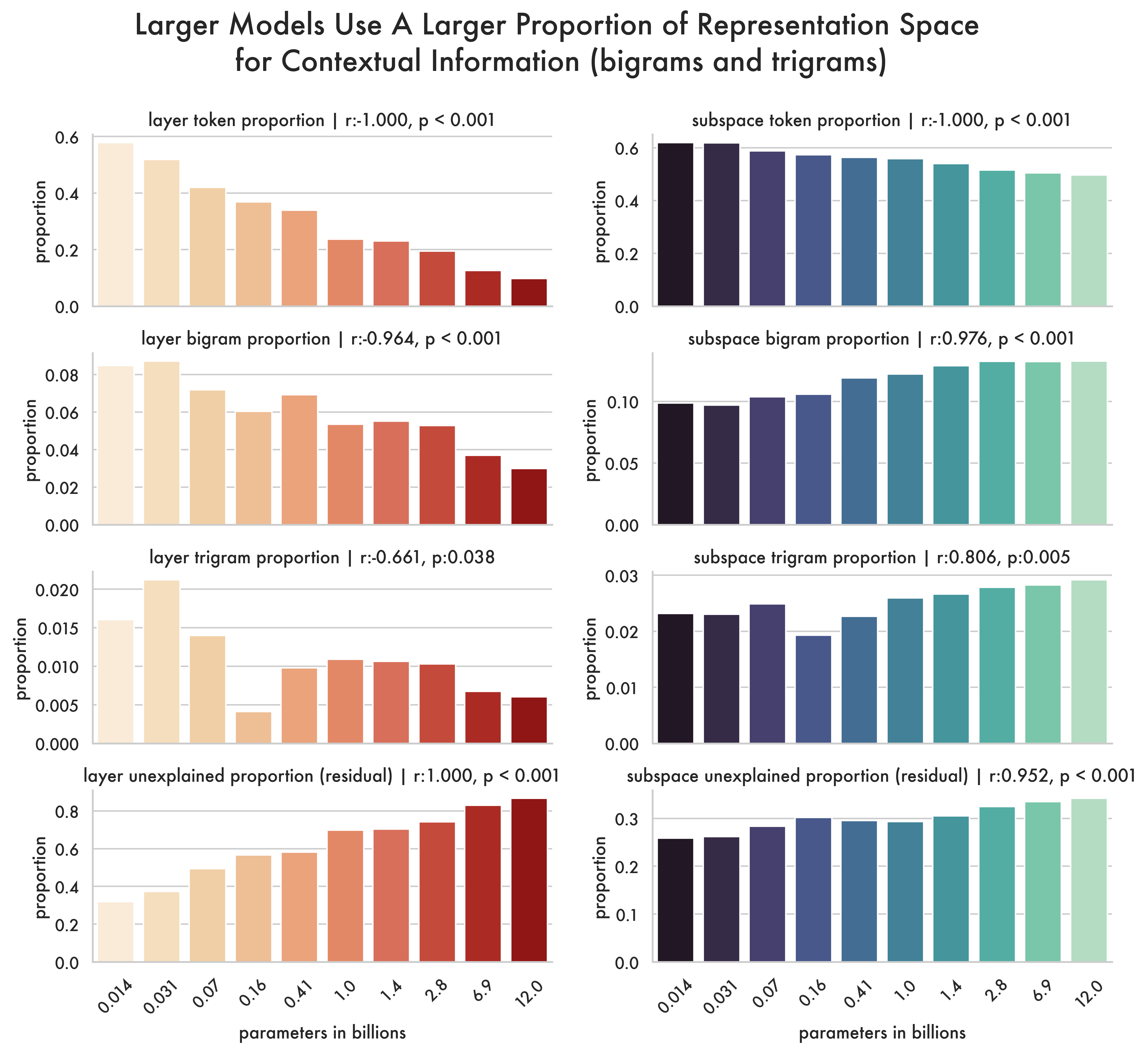}
	
	\caption{Analyses computed on pythia decoder-only models ranging from 14m parameters to 12 billion. Red/Orange bars show mean layer entropy, blue/green bars show mean subspace entropy. Above each facet is a spearman correlation between x and y. y-axis shows the proportions of representation space that encode token/bigram/trigram information for each model size(on the x-axis). Subspace entropy shows larger models use proportionally more space for token and bigram information. }
	\label{fig:pythia_size_proportion}
\end{figure}

\subsection{Model Size Conditions Representational Structure}\label{sec:size}
How does scale affect representational structure? We look at this in both decoder-only and encoder only models, again performing structure estimates using 256,000 sentences from english wikipedia, and labels for token, bigram, and trigram information. Figure \ref{fig:pythia_size} shows results for the decoder-only models, with both layer and subspace entropy reported. Both are reported for reference, and to give an intuition to how they relate - but as discussed above layer entropy does not allow a like-for-like comparison between different dimensionalities. As you would expect larger models have higher layer entropy - each layer of the 12b model has 5120 dimensions compared with 128 in the smallest, it would be surprising if they contained the same amount of information. Subspace entropy - which provides a more directly comparable estimate between model sizes - reveals a different pattern with the largest models beginning to compress their representations more, with the 12b version almost matching the subspace entropy of the smallest model. Because the representation space is larger information can be more distributed across space, meaning each subspace can compress more on a relative basis. We draw an analogy to Shannon's source coding model \citep{Shannon1948AMT} where meanings are mapped to signals; signal space has two key parameters - signal length and alphabet size. A smaller alphabet has less uncertainty, exemplified by morse code's binary alphabet where operators only need to tell the difference between a dot and dash. Smaller alphabets require a longer signal - sentences in morse code are far longer than in english - this is the tradeoff for a more robust encoding. In our subspace entropy analysis larger models have more subspaces, analogous to a longer signal. This can enable compression of each subspace like shrinking the alphabet at each character in a signal, which may help explain their improved performance.

Figure \ref{fig:pythia_size_proportion} plots the proportion of representation space that encodes token, bigram and trigram information and the information we can't explain in terms of any of the labels - the residual. This is estimated by comparing the regularity for each set of labels with the information left over which isn't regular with respect to any label. Looking at the subspace analysis larger models devote more of their representational space to contextual information, and less to token information. They also have more information we can't explain in terms of these labels. That could reflect other information from the training data not explainable in terms of lexical/contextual labels, or it could reflect artefacts not explainable in terms of any label. The middle plot shows disentanglement across model sizes, with larger models subspaces disentangling contextual information more.

\begin{figure}[htp]
	\centering
	\includegraphics[width=\textwidth]{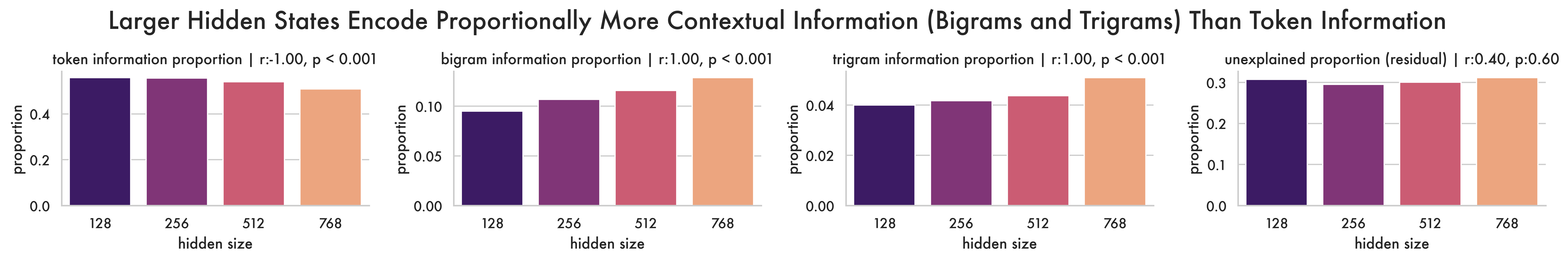}
	\vspace{5mm}
	\includegraphics[width=\textwidth]{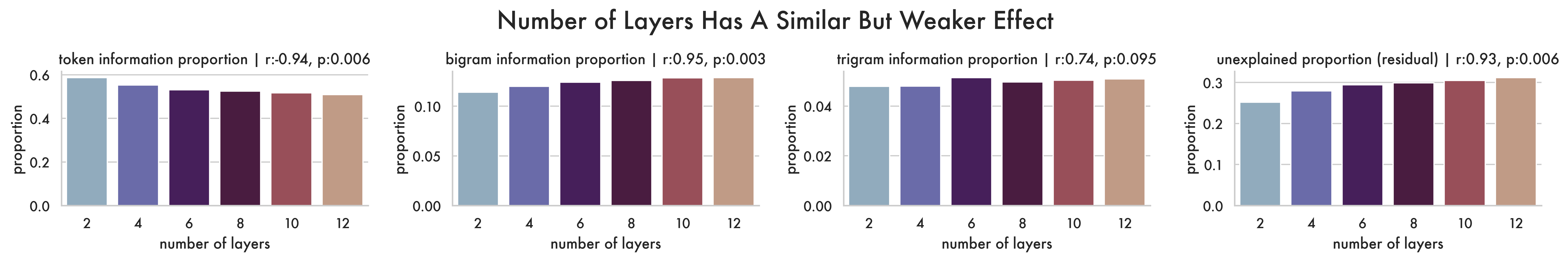}
	\thinline
	\vspace{10mm}
	\includegraphics[width=\textwidth]{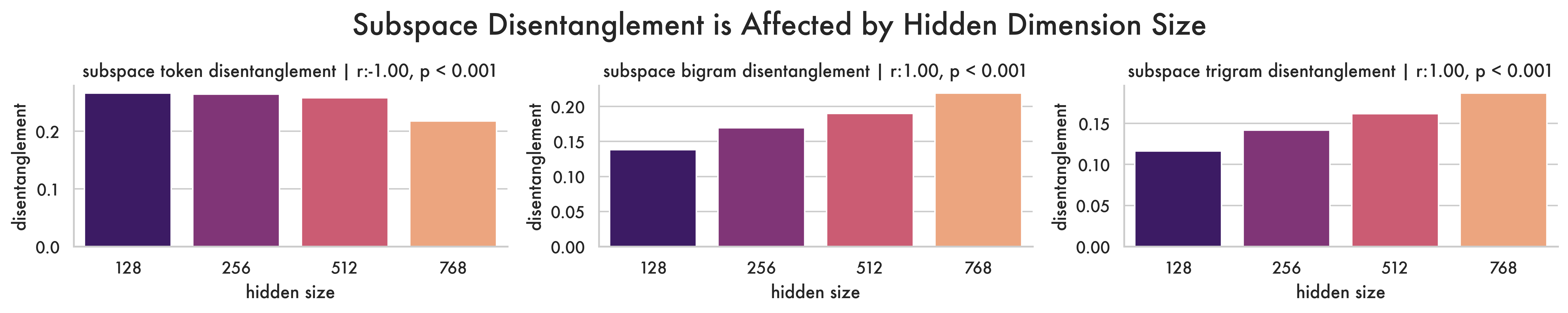}
	\vspace{5mm}
	\includegraphics[width=\textwidth]{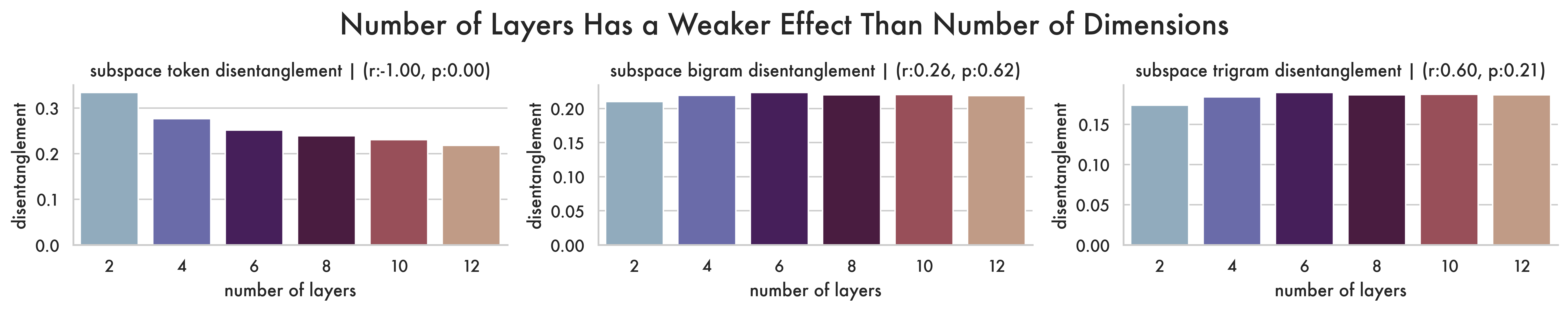}
	\caption{Scaling comparison of depth vs. dimensionalities on BERT models. All plots use subspace entropy, color reflects depth or dimension (both shown on the x axis) -- atop plots is a spearman correlation between x and y \textbf{ Top:} y-axis proportion of representation space that encodes token/bigram/trigram information. Larger models devote a larger proportion of their representation space to contextual information - here captured by bigram and trigram labels - while reducing token-level information. This effect is stronger for increasing size through hidden size, rather than number of layers. However, note that the increased number of layers does much more significantly increase the unexplained (residual) proportion. This could reflect models with greater depth representing more abstract syntactic and semantic information that cannot be captured by the labels used here. \textbf{ Bottom:} disentanglement of label information (y axis) across different sizes. Here increasing model scale reduces token disentanglement. For bigram and trigram disentanglement, it depends on the scaling method used. Increasing hidden size significantly increases disentanglement of bigrams and trigrams, while increasing the number of layers has little effect.}
	\label{fig:bert_size}
\end{figure}

An issue with the pythia suite of models is that while they differ in size, that difference is driven by changes in both depth and dimensionality\footnote{It's also worth noting models also differ in the dimensionality of attention heads. this may have an effect on structure but we lack controlled comparisons to draw conclusions.}. In an effort to isolate the effects of these different kinds of scaling we use sets of BERT models released by \citet{turc2019well}. Figure \ref{fig:bert_size} shows effects on representational structure for models with a dimensionality of 768, but layers ranging from 2 to 12, and models with 12 layers but dimensionalities from 128 to 768. Overall both kinds of scaling have a similar effect, namely increasing contextual information, with dimensionality's effect being much stronger than depth. Although increasing model depth does significantly increase the unexplained (residual) proportion. This could reflect models with greater depth representing more abstract syntactic and semantic information that cannot be captured by the labels used here. By contrast, both scaling methods decrease token disentanglement significantly, but only increasing the size of the hidden dimension affects bigram and trigram disentanglement, significantly increasing both --- increasing model depth has no clear effect on bigram and trigram disentanglement

\subsection{Predicting Downstream Performance}\label{sec:predict}

We look at spearman correlation coefficients between structural properties of representation space and downstream task performance. In order to isolate as many variables as possible we use the Multibert models \citep{sellam_multiberts_2022} which is 25 different initialisations of BERT \citep{devlin_bert_2019}. By comparing performance between models that differ only in terms of the random seed used to initialise them we can have some confidence that effects we measure between representational structure and downstream performance are likely driven by structure rather than model size, training data, or training objective. The Multiberts provide checkpoints at the end of pre-training, and evaluations for fine-tuned versions of each of these across the GLUE benchmarks \citep{wang2018glue}\footnote{For each seed used during pre-training, the multiberts train 5 different seeds during fine tuning. We compute a structure estimate with respect to the 25 different models at the end of pre-training, and correlate this with the performance of all 5 fine-tuned versions of each model.}. We take 10 million sentences sampled randomly from the C4 dataset \citep{raffel2020exploring} and compute our structure measures with respect to token, bigram, and trigram labels. The C4 dataset contains data from a general crawl of the internet - of which we use the english subset. This is a diverse collection of text sources which enables us to get a general structure estimate for each model. We correlate representational structure with respect to C4 at the end of pre-training with performance on GLUE tasks after fine-tuning. It is important to note that this means we are able to predict which of the models will do better on a downstream task before the models are fine-tuned for 2 million steps on data from that task. As far as we're aware this is the first analysis able to predict downstream performance from pre-training. Additionally the structure measures we use in this correlation are not estimated using data from those benchmarks. Despite the estimate using non-task data, on models 2 million steps of fine-tuning removed from evaluation we still find a number of significant correlations. This suggests that representational structure, as captured by our measures, has an effect on generalisation performance.

\begin{figure}[t]
	\centering
	\includegraphics[width=\textwidth]{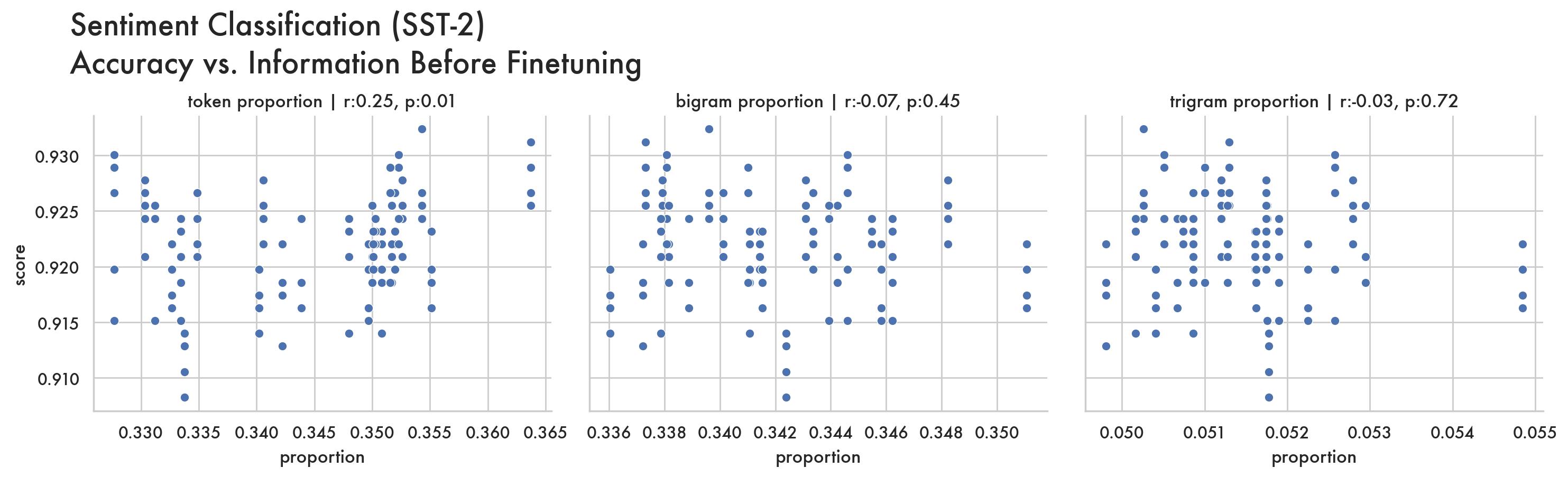}
	\includegraphics[width=\textwidth]{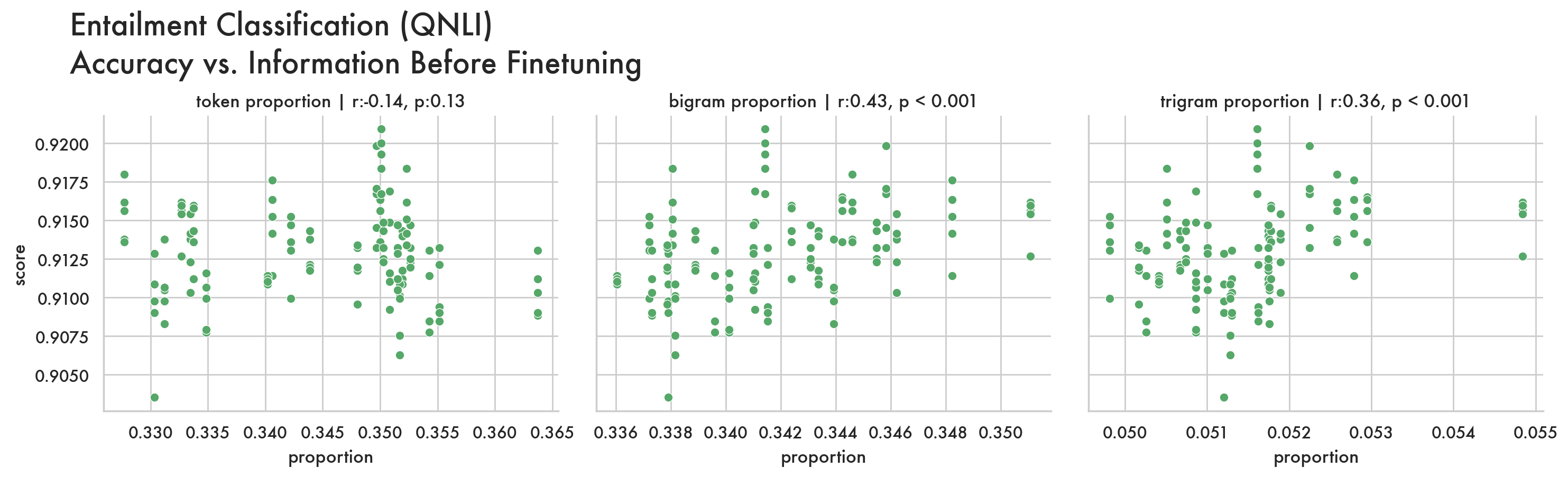}

	\caption{Scatterplots showing model performance on two GLUE benchmarks, sentiment classification (SST-2) and Entailment (QNLI) (y-axis) vs. information proportion at the token bigram and trigram level. Above each facet is a spearman correlation between the x and y axes. Sentiment classification just needs to decide if a sentence is positive or negative, and can rely on token-level information to do so. Entailment classification by contrast requires a model to determine if two sentences have the same meaning. Sentiment classification accuracy correlates with the model having a higher proportion of token information, while Entailment classification correlates with a higher proportion of bigram and trigram information.}
	\label{fig:qnli_scatter}
\end{figure}

\begin{table}[t]
    \centering
    \scriptsize
    \rowcolors{3}{white}{blue!3}
    
    \begin{NiceTabularX}{\textwidth}{XX|Y|YY|YY|YY}                        
    \CodeBefore
	  \rowcolor{blue!5}{1}
	  \rowcolors{2}{blue!5}{white}
	\Body
		\toprule
		 &  & $\mathcal{H}$ & \multicolumn{2}{c}{\nosbold{proportion}} & \multicolumn{2}{c}{\nosbold{disentanglement}} & \multicolumn{2}{c}{\nosbold{variation}} \\
		 &  & {\makesans overall}  & {\makesans token} & {\makesans context} & {\makesans token} & {\makesans context} & {\makesans token} & {\makesans context} \\
		\midrule
		\multirow{2}{*}{\nosbold{QNLI}} & {\makesans r} & 0.039 & -0.138 & 0.410 & -0.016 & 0.161 & 0.136 & -0.020 \\
		 & {\makesans p} & 0.663 & 0.125 & $<0.001$ & 0.855 & 0.072 & 0.131 & 0.827 \\
		 \midrule

		\multirow{2}{*}{\nosbold{SST-2}} & {\makesans r} & -0.152 & 0.247 & -0.048 & -0.044 & -0.196 & -0.242 & -0.197 \\
		 & {\makesans p} & 0.090 & 0.005 & 0.598 & 0.628 & 0.028 & 0.007 & 0.027 \\
		\bottomrule
	\end{NiceTabularX}
\caption{Spearman Correlations between representational structure across 25 different initialisations of BERT at the end of pre-training (before fine-tuning) and downstream task performance on two GLUE benchmarks (after 2M steps of fine-tuning). The same benchmarks are visualised in \protect\ref{fig:qnli_scatter}. For readability bigram and trigram information scores are averaged before performing the correlation to give a general `context' estimate. Positive correlations indicate a higher score on the measure correlates with higher task performance.}
 \label{tab:summary_correlations}
\end{table}

For clarity we focus on two of the glue tasks in particular, a sentiment classification task (SST-2), and an entailment task (QNLI). For sentiment classification, a model only needs to determine if the general sentiment of a sentence is positive or negative. Prior to the advent of large language models, this task was often approached using bag-of-words, or non-contextual word embeddings like word2vec \citep{mikolov2013efficient, barry2017sentiment}. By contrast entailment tasks like QNLI require a model to determine if two sentences entail one another. This requires determining if the meaning of two sentences overlap, which is harder to do with only non-contextual word level information. Figure \ref{fig:qnli_scatter} shows task performance against information proportion at the token, bigram and trigram level, along with spearman correlations between each measure and task performance. Additionally table \ref{tab:summary_correlations} shows spearman correlations between task accuracy and representational disentanglement and variation. Sentiment classification accuracy correlates positively with a higher proportion of representation space being dedicated to token-level information - but does not correlate significantly with bigram or trigram information. Entailment classification inverts this pattern showing strong significant correlations with bigram and trigram information, but no significant correlation with token level information.

\section{Conclusion}

We have introduced a set of measures for thinking about and describing structure in large language models information theoretically. This approach can show how representations become structured over the course of training, how that structure is influenced by model scale, and what structural properties correlate with downstream performance. It's backboned by a new scalable, parallelisable, and differentiable approach to entropy estimation, that can be applied at the subspace level to enable like-to-like comparisons between models of different sizes. We related the structural properties found here to structures in linguistics, and Shannon's model of communication in an effort to contextualise these structures in terms of other areas of science. We think that continued work mapping large language models to spaces and measures about which we have stronger intuitions than vector space is crucial in helping us understand, interpret, and improve models going forward.

\chapter{Biasing Representational Structure with Meta Learning}
\label{chapt:meta}

{\makesans
\emph{
Controlling Information Structure} \\[1em]
}

{\makesans
\begin{quote}
If I when \censor{my wife is sleeping and the baby and Kathleen are sleeping and} the sun is a flame-white disc \censor{in silken mists above shining trees,} if I \censor{in my north room} dance \censor{naked, grotesquely before my mirror} waving my shirt round my head and singing softly to myself
\censor{[..............]}\flushright{-William Carlos Williams}
\end{quote}
}

\drawline

\noindent Given that the last two chapters identified particular structures that predict generalisation performance, can we directly intervene during training to select for those structures? One way of doing this could be to reduce a model's capacity. Chapter \ref{chapt:comp_reg} showed how limiting the capacity of a model in a multi-agent setting can have a regularising effect on on the discrete signals it produces, with models with smaller hidden dimensions converging to languages with less variation. By contrast in chapter \ref{chapt:learning}, we looked at structure inside a model, where models with a larger hidden dimension compressed their representation space more and became more regular with respect to contextual information than their smaller counterparts. At the end of chapter \ref{chapt:learning} we observed that this may be because, when studying model-internal representations, limiting hidden size can have an effect analogous to limiting channel capacity in the discrete case (i.e. restricting signal length). This point was made clearer in the last chapter, where varying a large language model's hidden size has a much stronger effect on representation structure than varying its depth, despite the fact that both affect a model's capacity. In general modifying a model's parameter count can have effects on performance unrelated to capacity; the lottery ticket hypothesis \citep{frankle2018lottery} suggests larger models have a better chance of getting a good initialisation by virtue of having more parameters to be randomly initialised. Larger models could perform better because they luck into a better initialisation rather than for any reason related to their capacity during learning. 

In an ideal case we would be able to isolate capacity as an independent variable without altering other factors that can affect model performance. This chapter uses meta-learning to introduce an inductive bias to the model through the objective it optimises, allowing us to look at the effect of different biases on the same underlying model architecture. The experiments here endeavour to introduce a bias that limits the model's ability to memorise examples in its training data by encouraging update steps on an example that improves performance on similar examples. Results show this approach improves generalisation performance on two different architectures across two different datasets.

\begin{figure}[t]
    \centering
	\includegraphics[width=\textwidth]{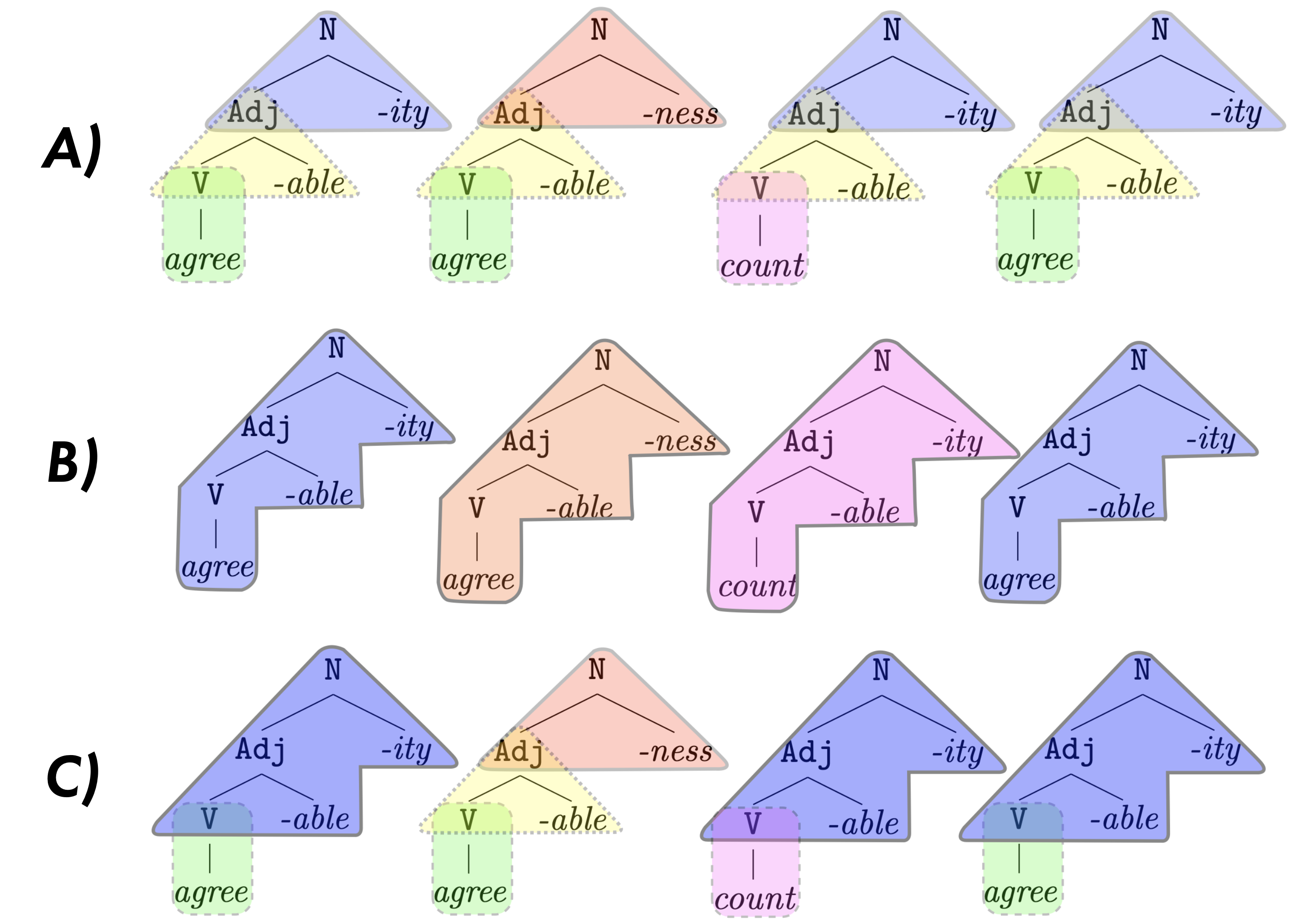}
    \caption{Figures reproduced from \protect\citet{odonnell_productivity_2015} show three different approaches to decomposing a word. Colour in each case indicates lexicon entries. \textbf{A)} decomposes words fully into each individual affix,  \textbf{B)} stores the entire word in memory without decomposition and \textbf{C)} splits the difference, partially decomposing each entry and memoizing frequent subunits
    } 
    \label{fig:odonnel_parsing}
\end{figure}

\subsection{Relating Information Structure to Compositionality, Memorisation \& Generalisation}

As something of a procedural note, the paper this chapter is based on is the first project I worked on during my PhD --- as a result it was written before I adopted the approach to representational structure used throughout this thesis. This chapter talks instead about models' behavioural properties - like generalising or memorising - rather than what those may look like on a representational level.
Given that, it is worth discussing how the two approaches to terminology relate - what compositionality, memorisation and generalisation mean in terms of  regularity, variation, and disentanglement. As discussed in chapter \ref{chapt:intro} regularity reflects how predictable a mapping between spaces is, and compositionality enables predictable mapping by reusing parts across the system which can be composed together. However just because a mapping is compositional doesn't entail its being maximally regular - as discussed at length in chapter \ref{chapt:comp_reg} natural languages manage to be compositional while supporting extensive variation. Given that the models considered in this chapter can generalise to thousands of in-distribution examples it is unlikely that they are incapable of representing information compositionally \citep{brighton_compositional_2002}. \citet{odonnell_productivity_2015} provides a way of thinking about gradations of regularity within a compositional system.  Looking at morphology \citet{odonnell_productivity_2015} considers a context free grammar with a lexicon that stores the lexical items a grammar can combine, figure \ref{fig:odonnel_parsing} shows 3 different approaches to decomposing a word into lexical entries. Colour indicates what portions of each word are stored in the lexicon A) fully decomposes a word into each root and affix, while B) stores the entire word as a single item, C) splits the difference by decomposing the full word, but memoizing frequent subunits. We can look at A) as being compositional and fully regular, B) as being non-compositional with maximal variation, and C) as being compositional with both regularity and variation. In \citet{odonnell_productivity_2015} an approach based on C) ends up providing the best fit to a corpus of natural language data in English. 

Implicitly when we talk about memorising vs. generalising, or more compositional vs. less we talk about how much is stored in the lexicon vs. generated by the grammar. In reality, figure \ref{fig:odonnel_parsing} gives an example of how this distinction need not be binary, with natural language reflecting gradations of both. A point made further by usage-based approaches to language \citep[e.g.][]{goldberg_constructions_1995, croft2001radical, tomasello2005constructing}, which erode the binary distinction between grammar and lexicon in favour of constructions that unify meaning and form and are learned probabilistically on the basis of experience \citep{goldberg_constructions_2003}. In light of this, the approach taken throughout this thesis conceptualises structure probabilistically by using information theory, and quantifies properties of a system that are naturally graded (regularity, variation, and disentanglement) rather than trying to describe a system in terms of binary distinctions. As \citet{russell1912problems} notes, learning from data --- even when learning rules --- is ultimately about probability:

\begin{quote}
\makesans
The man who has fed the chicken every day throughout its life at last wrings its neck instead [...] It must be conceded, to begin with, that the fact that two things have been found often together and never apart does not, by itself, suffice to prove demonstratively that they will be found together in the next case we examine. The most we can hope is that the oftener things are found together, the more probable it becomes that they will be found together another time, and that, if they have been found together often enough, the probability will amount almost to certainty. It can never quite reach certainty, because we know that in spite of frequent repetitions there sometimes is a failure at the last as in the case of the chicken whose neck is wrung. Thus probability is all we ought to seek. \flushright{\textcite{russell1912problems} | p. 30}
\end{quote}

\noindent This is not to say talking about memorisation or generalisation is incorrect (the remainder of this chapter does at some length), but that in models, as in natural language, it is nearly always both --- not so much a question of either/or but a question of degree. To me, questions of degree are best seen as questions of probability.

\thinline
{
\makesans
\noindent The remainder of this chapter is a paper \textbf{Meta-Learning to Compositionally Generalise} that appeared at the International Meeting of the Assosciation of Computational Linguists in 2021. Authors are myself, Bailin Wang, Kenny Smith and Ivan Titov - I wrote the majority of the paper including the theoretical framing and motivation. I contributed to code for the experiments but majority of code was written by Bailin based on a codebase from a previous meta-learning project - Kenny and Ivan supervised the project and gave writing feedback prior to submission to the conference. The paper is presented here minimally changed from the conference version that underwent peer-review. Changes are largely related to formatting to make the content more readable outside of the original conference paper template.
}
\drawline

\section{Meta-Learning to Compositionally Generalise}

Compositionality is the property of human language that allows for the meaning of a sentence to be constructed from the meaning of its parts and the way in which they are combined \citep{cann_formal_1993}. By decomposing phrases into known parts we can generalize to novel sentences despite never having encountered them before. In practice this allows us to produce and interpret a functionally limitless number of sentences given finite means \citep{chomsky_aspects_1965}.

Whether or not neural networks can generalize in this way remains unanswered. Prior work asserts that there exist fundamental differences between cognitive and connectionist architectures that makes compositional generalization by the latter unlikely \citep{fodor_connectionism_1988}. However, recent work has shown these models' capacity for learning some syntactic properties. \citet{hupkes_visualisation_2018} show how some architectures can handle hierarchy in an algebraic context and generalize in a limited way to unseen depths and lengths. Work looking at the latent representations learned by deep machine translation systems show how these models seem to extract constituency and syntactic class information from data \citep{blevins_deep_2018, belinkov_evaluating_2018}. These results, and the more general fact that neural models perform a variety of NLP tasks with high fidelity \citep[eg.][]{vaswani_attention_2017, dong_language_2016}, suggest these models have some sensitivity to syntactic structure and by extension may be able to learn to generalize compositionally.

Recently there have been a number of datasets designed to more formally assess connectionist models' aptitude for compositional generalization \citep{kim_cogs_2020, lake_generalization_2018, hupkes_compositionality_2019}. These datasets frame the problem of compositional generalization as one of out-of-distribution generalization: the model is trained on one distribution and tested on another which differs in ways that would be trivial for a compositional strategy to resolve. A variety of neural network architectures have shown mixed performance across these tasks, failing to show conclusively that connectionist models are reliably capable of generalizing compositionally \cite{keysers_measuring_2020, lake_generalization_2018}. Natural language requires a mixture of memorization and generalization \citep{jiang_characterizing_2020}, memorizing exceptions and atomic concepts with which to generalize. Previous work looking at compositional generalization has suggested that models may memorize large spans of sentences multiple words in length \citep{hupkes_compositionality_2019, keysers_measuring_2020}. This practice may not harm in-domain performance, but if at test time the model encounters a sequence of words it has not encountered before it will be unable to interpret it having not learned the atoms (words) that comprise it. 
\citet{griffiths_understanding_2020} looks at the role of limitations in the development of human cognitive mechanisms. Humans' finite computational ability and limited memory may be central to the emergence of robust generalization strategies like compositionality. A hard upper-bound on the amount we can memorize may be in part what forces us to generalize as we do. Without the same restriction models may prefer a strategy that memorizes large sections of the input potentially inhibiting their ability to compositionally generalize.

In a way the difficulty of these models to generalize out of distribution is unsurprising: supervised learning assumes that training and testing data are drawn from the same distribution, and therefore does not necessarily favour strategies that are robust out of distribution. Data necessarily under-specifies for the generalizations that produced it. Accordingly for a given dataset there may be a large number of generalization strategies that are compatible with the data, only some of which will perform well outside of training \citep{damour_underspecification_2020}. It seems connectionist models do not reliably extract the strategies from their training data that generalize well outside of the training distribution. Here we focus on an approach that tries to to introduce a bias during training such that the model arrives at a more robust strategy.

To do this we implement a variant of the model agnostic meta-learning algorithm \citep[MAML,][] {finn_model-agnostic_2017}. The approach used here follows \citet{wang2020meta} which implements an objective function that explicitly optimizes for out-of-distribution generalization in line with \citet{li2018learning}. \citet{wang2020meta} creates pairs of tasks for each batch (which here we call meta-train and meta-test) by sub-sampling the existing training data. Each meta-train, meta-test task pair is designed to simulate the divergence between training and testing: meta-train is designed to resemble the training distribution, and meta-test to resemble the test distribution. The training objective then requires that update steps taken on meta-train are also beneficial for meta-test. This serves as a kind of regularizer, inhibiting the model from taking update steps that only benefit meta-train. By manipulating the composition of meta-test we can control the nature of the regularization applied. Unlike other meta-learning methods this is not used for few or zero-shot performance. Instead it acts as a kind of meta-augmented supervised learning, that helps the model to generalize robustly outside of its training distribution.

The approach taken by \citet{wang2020meta} relies on the knowledge of the test setting. While it does not assume access to the test distribution, it assumes access to the family of test distributions, from which the actual test distribution will be drawn. While substantially less restrictive than the standard iid setting, it still poses a problem if we do not know the test distribution, or if the model is evaluated in a way that does not lend itself to being represented by discrete pairs of tasks (i.e. if test and train differ in a variety of distinct ways). Here we propose a more general approach that aims to generate meta-train, meta-test pairs which are populated with similar (rather than divergent) examples in an effort to inhibit the model from memorizing its input. Similarity is determined by a string or tree kernel so that for each meta-train task a corresponding meta-test task is created from examples deemed similar.

By selecting for similar examples we design the meta-test task to include examples with many of the same words as meta-train, but in novel combinations. As our training objective encourages gradient steps that are beneficial for both tasks we expect the model to be less likely to memorize large chunks which are unlikely to occur in both tasks, and therefore generalize more compositionally. This generalizes the approach from \citet{wang2020meta}, by using the meta-test task to apply a bias not-strictly related to the test distribution: the design of the meta-test task allows us to design the bias which it applies. It is worth noting that other recent approaches to this problem have leveraged data augmentation to make the training distribution more representative of the test distribution \citep{andreas_good-enough_2020}. We believe this line of work is orthogonal to ours as it does not focus on getting a model to generalize compositionally, but rather making the task simple enough that compositional generalization is not needed. Our method is model agnostic, and does not require prior knowledge of the target distribution. 

We summarise our contributions as follows:
\begin{itemize}[noitemsep,topsep=0pt]

\item We approach the problem of compositional generalization with a meta-learning objective that tries to explicitly reduce input memorization using similarity-driven virtual tasks.

\item We perform experiments on two text-to-semantic compositional datasets: COGS and SCAN.
Our new training objectives lead to significant improvements in accuracy
over a baseline parser trained with conventional supervised
learning.

\end{itemize}

\section{Methods}
We introduce the meta-learning augmented approach to supervised learning from \citet{li2018learning,wang2020meta} that explicitly optimizes for out-of-distribution generalization. Central to this approach is the generation of tasks for meta-learning by sub-sampling training data. We introduce three kinds of similarity metrics used to guide the construction of these tasks.  

\subsection{Problem Definition}

\paragraph*{Compositional Generalization}
  \citet[eg.][]{lake_generalization_2018,kim_cogs_2020} introduce datasets designed to assess compositional generalization. These datasets are created by generating synthetic data with different distributions for testing and training. The differences between the distributions are trivially resolved by a compositional strategy. At their core these tasks tend to assess three key components of compositional ability: systematicity, productivity, and primitive application. Systematicity allows for the use of known parts in novel combinations as in (a). Productivity enables generalization to longer sequences than those seen in training as in (b). Primitive application allows for a word only seen in isolation during training to be applied compositionally at test time as in (c).

\begin{itemize}
    \item[(a)] The cat gives the dog a gift $\rightarrow$ The dog gives the cat a gift
    \item[(b)] The cat gives the dog a gift $\rightarrow$ The cat gives the dog a gift and the bird a gift
    \item[(c)] made $\rightarrow$ The cat made the dog a gift
\end{itemize}
A compositional grammar like the one that generated the data would be able to resolve these three kinds of generalization easily, and therefore performance on these tasks is taken as an indication of a model's compositional ability.

\paragraph*{Conventional Supervised Learning}
The compositional generalization datasets we look at are semantic parsing tasks, mapping between natural language and a formal representation. A usual supervised learning objective for semantic parsing is to minimize the negative log-likelihood of the correct formal representation given a natural language input sentence, i.e. minimising
\vspace{-2mm}
\begin{equation}
    \loss_{\sBatch}(\theta) = - \frac{1}{N} \sum_{i=1}^{N}  \log p_{\param}(y | x) 
\end{equation}
where $N$ is the size of batch $\sBatch$, $y$ is a formal representation and $x$ is a natural language sentence. This approach assumes that the training and testing data are independent and identically distributed.

\paragraph*{Task Distributions}
Following from \citet{wang2020meta}, we utilize a learning algorithm that can enable a parser to benefit from a distribution of virtual tasks, denoted by $p(\tau)$, where $\tau$ refers to an instance of a virtual compositional generalization task that has its own training and test examples.

\subsection{MAML Training}

\begin{algorithm}[t]
\caption{MAML Training Algorithm}
\label{algo:maml}
\begin{algorithmic}[1]
  \REQUIRE Original training set $\origtrain$
  \REQUIRE Learning rate $\alpha$, Batch size $N$
  \FOR{step $\gets 1$ \TO $T$}
    \STATE Sample a random batch from $\origtrain$ as a virtual training set $\mtrainBatch$  \label{algo:line_sample_mtrain}  \\ 
    \STATE Initialize an empty generalization set $\mtestBatch$
    \FOR{$i$ $\gets 1$ \TO $N$}
        \STATE Sample an example from $\tilde p( \cdot \ | \ \mtrainBatch[i])$
        \label{algo:line_sample_mtest}  \\ 
        \STATE Add it to $\mtestBatch$
    \ENDFOR
    \STATE Construct a virtual task $\tau := (\mtrainBatch, \mtestBatch)$
    \STATE Meta-train update: \label{algo:line_meta_train}  \\ 
    \quad $\param'\leftarrow \param - \alpha \nabla_{\param} \loss_{\mtrainBatch} (\param)$ \\ 
    \STATE Compute meta-test objective: \\ \quad $ \loss_{\tau}(\param) = \loss_{\mtrainBatch}  (\param) + \loss_{\mtestBatch}(\param')$ \\
    \STATE Final Update: \label{algo:line_meta_update} \\ 
    \quad  $\param \leftarrow  \update(\param, \nabla_{\param} \loss_{\tau}(\param))$
   \ENDFOR
\end{algorithmic}
\end{algorithm}

Once we have constructed our pairs of virtual tasks we need a training algorithm that encourages compositional generalization in each.
Like \citet{wang2020meta},
we turn to optimization-based
meta-learning algorithms \cite{finn_model-agnostic_2017,li2018learning} 
and apply DG-MAML (Domain Generalization with Model-Agnostic Meta-Learning), 
a variant of MAML \cite{finn_model-agnostic_2017}. 
Intuitively, DG-MAML encourages optimization on meta-training examples to have a positive
effect on the meta-test examples as well. 

During each learning episode of MAML training we randomly sample a task $\tau$
which consists of a training batch $\mtrainBatch$ and a generalization batch $\mtestBatch$ and conduct optimization in two steps, namely \textit{meta-train}
and \textit{meta-test}.  

\paragraph*{Meta-Train}
The meta-train task is sampled at random from the training data. The model performs one stochastic gradient descent step on this batch
\begin{equation}
    \theta'\leftarrow \theta - \alpha \nabla_{\theta} \loss_{\mtrainBatch} (\theta)
    \label{eq:meta_train}
\end{equation}
where $\alpha$ is the meta-train learning rate. 

\paragraph*{Meta-Test}

The fine-tuned parameters $\theta'$ are evaluated on the accompanying generalization task, meta-test, by computing their loss on it
denoted as $\loss_{\mtestBatch}(\theta')$. The final objective for a task $\tau$
is then to jointly optimize the following:
\begin{align}
\begin{split}
     \loss_{\tau}(\theta)&=\loss_{\mtrainBatch}(\theta) + \loss_{\mtestBatch}(\theta') \\
     &=\mathcal{L}_{\mtrainBatch}(\theta) + \mathcal{L}_{\mtestBatch}(\theta - \alpha \nabla_{\theta} \mathcal{L}_{\beta}(\theta) ) 
    \label{eq:maml_obj}
\end{split}
\end{align}
The objective now becomes to reduce the joint loss of both the meta-train and meta-test tasks. Optimizing in this way ensures that updates on meta-train are also beneficial to meta-test. The loss on meta-test acts as a constraint on the loss from meta-train. This is unlike traditional supervised learning ($\loss_{\tau}(\theta)=\loss_{\mtrainBatch}(\theta) + \loss_{\mtestBatch}(\theta)$) where the loss on one batch does not constrain the loss on another. 

With a random $\mtrainBatch$ and $\mtestBatch$, the joint loss function can be seen as a kind of generic regularizer, ensuring that update steps are not overly beneficial to meta-train alone. By constructing $\mtrainBatch$ and $\mtestBatch$ in ways which we expect to be relevant to compositionality, we aim to allow the MAML algorithm to apply specialized regularization during training. Here we design meta-test to be similar to the meta-train task because we believe this highlights the systematicity generalization that is key to compositional ability: selecting for examples comprised of the same atoms but in different arrangements. In constraining each update step with respect to meta-train by performance on similar examples in meta-test we expect the model to dis-prefer a strategy that does not also work for meta-test like memorization of whole phrases or large sections of the input.

\begin{table}[t!]
  \centering
    \resizebox{\columnwidth}{!}{ 
      \begin{tabular}{lr}
    \multicolumn{2}{c}{\textbf{Source Example}: The girl changed a sandwich beside the table .} \\
      \\
      \toprule
      \emph{Neighbours using Tree Kernel} & Similarity\\
      \hline
A sandwich changed . & 0.55 \\
The girl changed . & 0.55 \\
The block was changed by the girl . & 0.39 \\
The girl changed the cake . & 0.39 \\
change & 0.32 \\
       \\
        \emph{Neighbours using String Kernel}\\
       \hline
The girl rolled a drink beside the table . & 0.35 \\
The girl liked a dealer beside the table . & 0.35 \\
The girl cleaned a teacher beside the table . & 0.35 \\
The girl froze a bear beside the table . & 0.35 \\
The girl grew a pencil beside the table . & 0.35 \\
      \\
      \emph{Neighbours using LevDistance} & \\
      \hline
The girl rolled a drink beside the table . & -2.00 \\
The girl liked a dealer beside the table . & -2.00 \\
The girl cleaned a teacher beside the table . & -2.00 \\
The girl froze a bear beside the table . & -2.00 \\
The girl grew a pencil beside the table . & -2.00 \\
      \bottomrule
      \end{tabular}
  }
  \caption{Top scoring examples according to the tree kernel, string kernel and Levenshtein distance for the sentence `The girl changed a sandwich beside the table~.' and accompanying scores.}
  \label{table:kernel-examples}
\vspace{-3mm}
  \end{table}

\begin{figure}[t]
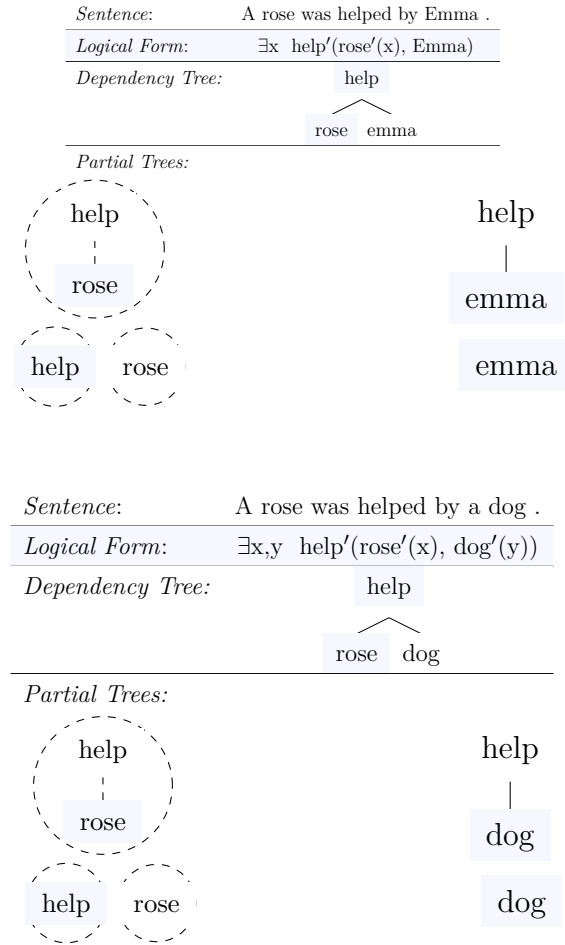

\centering

\resizebox{0.4\textwidth}{!}{%
\begin{tabular}{lc}
    \emph{Sentence}: & A rose was helped by Emma . \\
    \hline
\emph{Logical Form}: & 
$\exists $x \ help$'$(rose$'$(x), Emma) \\
\hline
    \emph{Dependency Tree:} &
\Tree [.help rose emma ] \\
\hline
\emph{Partial Trees:} &
\end{tabular}
}\\
\circled{\Tree [.help rose ]}
\Tree [.help emma ] \\
\circled{\Tree [.help ]}
\circled{\Tree [.rose ]}
\Tree [.emma ] \\[1.5\baselineskip]

\resizebox{0.5\textwidth}{!}{%
\begin{tabular}{lc}
\emph{Sentence}: & A rose was helped by a dog .  \\
\hline
\emph{Logical Form}: & 
$\exists$x,y \ help$'$(rose$'$(x), dog$'$(y)) \\
\hline
\emph{Dependency Tree:} &
\Tree [.help rose dog ] \\
\hline
\emph{Partial Trees:} & 
\end{tabular}
}

\circled{\Tree [.help rose ]}
\Tree [.help dog ] \\
\circled{\Tree [.help ] } \circled{\Tree [.rose ] } \Tree [.dog ] 
\caption{The dependency-tree forms for the logical forms of two sentences.  Shown below each tree are its partial trees. As there are three partial trees shared by the examples their un-normalized tree kernel score is 3.}
\label{figure:partial-trees}
\vspace{-3mm}
\end{figure}

\subsection{Similarity Metrics}

Ideally, the design of virtual tasks should reflect specific generalization cases for each dataset. However, in practice this requires some prior knowledge of the distribution to which the model will be expected to generalize, which is not always available. 
Instead we aim to naively structure the virtual tasks to resemble each other.
To do this we use a number of similarity measures intended to help select examples which highlight the systematicity of natural language.

Inspired by kernel density estimation~\cite{parzen1962estimation}, we define a relevance distribution for each example:
\begin{equation}
   \tilde p ( x', y' | x, y )  \propto \exp \big ( k([x, y], [x', y']) / \eta \big )
   \label{eq:rel_dist}
\end{equation}
where $k$ is the similarity function, 
$[x, y]$ is a training example, $\eta$
is a temperature that controls the sharpness of the distribution.
Based on our extended interpretation of relevance, a high $\tilde p$ implies that $[x, y]$ is systematically relevant to $[x', y']$ - containing many of the same atoms but in a novel combination. We look at three similarity metrics to guide subsampling existing training data into meta-test tasks proportional to each example's $\tilde p$.

\paragraph*{Levenshtein Distance}
First, we consider Levenshtein distance, a 
kind of edit distance widely used to measure the dissimilarity between strings. We compute the negative Levenshtein distance at the word-level between natural language sentences of two examples:
\begin{equation}
    k([x, y], [x', y']) = -1 * \text{LevDistance}(x, x')
\end{equation}
where LevDistance returns the number of edit operations required to transform $x$ into $x'$. See Table~\ref{table:kernel-examples} for examples. %

Another family of %
 similarity metrics for discrete structures are convolution kernels~\cite{haussler1999convolution}.

\paragraph*{String-Kernel Similarity}
We use the string subsequence kernel~\cite{lodhi2002text}: %
\begin{equation}
    k([x, y], [x', y']) = \text{SSK}(x, x')
\end{equation}
where SSK computes the number of common subsequences between natural language sentences at the word-level.
See Table~\ref{table:kernel-examples} for examples.~\footnote{We use the normalized convolution kernels in this work, i.e., $k'(x_1, x_2)=k(x_1, x_2)/ \sqrt{k(x_1, x_1) k(x_2, x_2)}$}

\paragraph*{Tree-Kernel Similarity}
In semantic parsing, the formal representation $y$ usually has a known grammar which can be used to 
represent it as a tree structure.
In light of this we use tree convolution kernels to compute similarity between examples:~\footnote{
Alternatively, we can use tree edit-distance~\cite{zhang1989simple}.}
\begin{equation}
    k([x, y], [x', y']) =  \text{TreeKernel}(y, y')
\end{equation}
where the TreeKernel function is a convolution kernel~\cite{collins_convolution_nodate} applied to trees.
Here we consider a particular case
where $y$ is represented as a dependency structure,
as shown in Figure \ref{figure:partial-trees}.
We use the partial tree kernel \citep{moschitti_making_nodate} which is designed for application to dependency trees. 
For a given dependency tree partial tree kernels generate a series of all possible partial trees: any set of one or more connected nodes. Given two trees the kernel returns the number of partial trees they have in common, interpreted as a similarity score. Compared with string-based similarity, this kernel prefers sentences that share common syntactic sub-structures, some of which are not assigned high scores in string-based similarity metrics, as shown in Table \ref{table:kernel-examples}.

Though tree-structured formal representations are more informative in obtaining relevance, not all logical forms can be represented as tree structures. 
In SCAN~\cite{lake_generalization_2018} $y$ are action sequences without given grammars. As we will show in the experiments, string-based similarity metrics have a broader scope of applications but are less effective than tree kernels in cases where $y$ can be tree-structured. 

\paragraph*{Sampling for Meta-Test}
Using our kernels we compute the relevance distribution in Eq~\ref{eq:rel_dist} to construct virtual tasks for MAML training. 
We show the resulting procedure in Algorithm~\ref{algo:maml}.
In order to construct a virtual task $\tau$,
a meta-train batch is first sampled at random from the training data (line \ref{algo:line_sample_mtrain}), then the accompanying meta-test batch is created by sampling examples similar to those in meta-train (line \ref{algo:line_sample_mtest}). 

We use \emph{Lev-MAML, Str-MAML and Tree-MAML} to denote the meta-training using Levenshtein distance, string-kernel and tree-kernel similarity, respectively.

\section{Experiments}

\subsection{Datasets and Splits}
We evaluate our methods on the following semantic parsing benchmarks
that target compositional generalization.

\paragraph*{SCAN} contains a set of natural language commands and their corresponding 
action sequences~\cite{lake_generalization_2018}. We use the Maximum Compound Divergence (MCD) 
splits~\cite{keysers_measuring_2020}, which are created based on the principle of 
maximizing the divergence between the compound (e.g., patterns of 2 or more action sequences) distributions of
the training and test tests. We apply 
Lev-MAML and Str-MAML to SCAN where similarity measures are applied
to the natural language commands. Tree-MAML (which uses a tree kernel) is not applied as the action sequences do not have an underlying dependency tree-structure. %

\paragraph*{COGS} contains a diverse set of natural language sentences paired with logical 
forms based on lambda calculus~\cite{kim_cogs_2020}. Compared with SCAN, it covers various systematic linguistic abstractions 
(e.g., passive to active) including examples of lexical and structural generalization, and thus better reflects the compositionality of natural language. 
In addition to the standard splits of Train/Dev/Test,
COGS provides a generalization (Gen) set drawn from a different distribution that specifically assesses compositional generalization. We apply Lev-MAML, Str-MAML and Tree-MAML to COGS; Lev-MAML and Str-MAML make use of
the natural language sentences while Tree-MAML uses the dependency structures reconstructed from the logical forms.

\subsection{Baselines}

In general, our method is model-agnostic and can be coupled with any semantic parser to improve its compositional generalization. Additionally Lev-MAML, and Str-MAML are dataset agnostic provided the dataset has a natural language input.
In this work, we apply our methods on two widely used sequence-to-sequences models.~\footnote{
Details of implementations and hyperparameters 
can be found in the Appendix.}

\paragraph*{LSTM-based Seq2Seq} has been the backbone of many neural semantic parsers~\cite{dong_language_2016,jia2016data}.
It utilizes LSTM~\cite{hochreiter1997long} and attention~\cite{bahdanau2014neural} under
an encoder-decoder~\cite{sutskever2014sequence} framework.

\paragraph*{Transformer-based Seq2Seq} also follows the encoder-decoder framework, but 
it uses Transformers~\cite{vaswani_attention_2017}
to replace the LSTM for encoding and decoding. It has proved successful in many NLP
tasks e.g., machine translation. Recently,
it has been adapted for semantic parsing~\cite{wang2019rat} with superior performance.

We try to see whether our  MAML training
can improve the compositional generalization of contemporary semantic parsers, compared with standard supervised learning. Moreover, we include a meta-baseline, referred to as Uni-MAML, 
that constructs meta-train and meta-test splits by uniformly sampling training examples. 
By comparing with this meta-baseline, we show the effect of similarity-driven construction of meta-learning splits.
Note that we do not focus on making comparisons with other methods that feature 
specialized architectures for SCAN datasets (see Section~\ref{sec:related_work}), as these methods do not generalize well to more complex datasets~\cite{furrer_compositional_2021}.

\paragraph*{GECA} We additionally apply the good enough compositional augmentation (GECA) method laid out in \citet{andreas_good-enough_2020} to the SCAN MCD splits. Data augmentation of this kind tries to make the training distribution more representative of the test distribution. This approach is distinct from ours which focuses on the training objective, but the two can be combined with better overall performance as we will show. Specifically, we show the results of GECA applied to the MCD splits as well as GECA combined with our Lev-MAML variant. Note that we elect not to apply GECA to COGS, as the time and space complexity~\footnote{See the original paper for details.} of GECA proves very costly for COGS in our preliminary experiments.

\subsection{Construction of Virtual Tasks}

The similarity-driven sampling distribution $\tilde p$ in Eq~\ref{eq:rel_dist} requires computing 
the similarity between every pair of training examples, which can be very expensive depending on 
the size of of the dataset. As the sampling distributions are fixed during training, we compute 
and cache them beforehand. However, they take an excess of disk space to store as essentially 
we need to store an $N \times N$ matrix where $N$ is the number of training examples. 
To allow efficient storage and sampling, we use the following approximation.
First, we found that usually each example only has a small set of neighbours that are relevant to it.~\footnote{
For example, in COGS, each example only retrieves 3.6\% of the whole training set as its neighbours 
(i.e., have non-zero tree-kernel similarity) on average.}
Motivated by this observation, we only store the top 1000 relevant neighbours for each example sorted by similarity, 
and use it to construct the sampling distribution denoted as $\tilde p_{\text{top}1000}$. 
To allow examples out of top 1000 being sampled, we use a linear interpolation 
between $\tilde p_{\text{top}1000}$ and a uniform distribution. 
Specifically, we end up using the following sampling distribution:
\begin{equation*}
  \tilde p(x', y' | x, y) = \lambda \ \tilde  p_{\text{top}1000} ( x', y' | x, y ) + (1 - \lambda) \frac{1}{N}
\end{equation*}
where $\tilde p_{\text{top}1000}$ assigns 0 probability to out-of top 1000 examples, $N$ is the number of training examples, 
and $\lambda$ is a hyperparameter for interpolation. 
In practice, we set $\lambda$ to $0.5$ in all experiments.
To sample from this distribution, we first decide whether the sample is in the top 1000 by sampling 
from a Bernoulli distribution parameterized by $\lambda$. 
If it is, we use $\tilde p_{\text{top1000}}$ to do the sampling; 
otherwise, we uniformly sample an example from the training set. 

\begin{table}[t]
\centering
\setlength{\tabcolsep}{3pt}
\resizebox{0.65\textwidth}{!}{
\begin{tabular}{lrlrlrl}
\textbf{Model}             &  \mc{2}{MCD1} & \mc{2}{MCD2} & \mc{2}{MCD3}  \\
\toprule
LSTM                        & \acc{4.7}{2.2}   & \acc{7.3}{2.1} & \acc{1.8}{0.7} \\
Transformer                  & \acc{0.4}{0.4}   & \acc{1.8}{0.4} & \acc{0.5}{0.1} \\
T5-base                        & \acc{26.2}{1.7}  & \acc{7.9}{1.6} & \acc{12.1}{0.1}\\
T5-11B                         & \acc{7.9}{}      & \acc{2.4}{}    & \ACC{16.8}{}\\
\hline
LSTM &  \gacc{27.4}{8.2} & \gacc{31.0}{0.4} & \gacc{9.6}{3.7}  \\
\ \emph{w.} Uni-MAML &  \gacc{44.8}{5.4} & \gacc{31.9}{3.4} & \gacc{10.0}{1.4}  \\
\ \emph{w.} Lev-MAML &  \GACC{47.6}{2.3} & \GACC{35.2}{3.9} & \gacc{11.4}{3.0}  \\
\ \emph{w.} Str-MAML &  \gacc{42.2}{2.6} & \gacc{33.6}{4.3} & \gacc{11.4}{2.2}  \\
\hline
Transformer &  \gacc{2.6}{0.8} & \gacc{3.1}{1.0} & \gacc{2.3}{1.3}  \\
\ \emph{w.} Uni-MAML &  \gacc{2.8}{0.7} & \gacc{3.2}{1.0} & \gacc{3.2}{1.6}  \\
\ \emph{w.} Lev-MAML &  \gacc{4.7}{1.8} & \gacc{6.7}{1.4} & \gacc{6.5}{1.2}  \\
\ \emph{w.} Str-MAML &  \gacc{2.8}{0.6} & \gacc{5.6}{1.6} & \gacc{6.7}{1.4}  \\
\hline
GECA + LSTM &  \gacc{51.5}{4.4} & \gacc{30.4}{4.8} & \gacc{12.0}{6.8}  \\
\ \emph{w.} Lev-MAML &  \GACC{58.9}{6.4} & \GACC{34.5}{2.5} & \gacc{12.3}{4.9}  \\
\bottomrule
\end{tabular}
}
\setlength{\tabcolsep}{6pt}
\caption{Main results on SCAN MCD splits.  
We show the mean and variance (95\% confidence interval) of 10 runs. 
Results in the top four rows are from from \protect~\citet{furrer_compositional_2021}, the remainder are results obtained in this paper. }
\label{table:mcd-scan-results}
\end{table}

\subsection{Development Set}
Many tasks that assess out-of-distribution (O.O.D.) generalization (e.g. COGS) do not have an O.O.D. Dev set that is representative of the generalization distribution. This is desirable as a parser in principle should never have knowledge of the Gen set during training. In practice though the lack of an O.O.D. Dev set makes model selection extremely difficult and not reproducible.~\footnote{
We elaborate on this issue in the Appendix.}
In this work, we propose the following strategy to alleviate this issue: 1) we sample a small subset from the Gen set, denoted as `Gen Dev' for tuning meta-learning hyperparmeters, 2) we use two disjoint sets of random seeds for development and testing respectively, i.e., retraining the selected models from scratch before applying them to the final test set. In this way, we make sure that our tuning is not exploiting the models resulting from specific random seeds: we do not perform random seed tuning.
At no point are any of our models trained on the Gen Dev set.

\begin{table}[t]
\centering
\setlength{\tabcolsep}{3pt}
\resizebox{0.65\textwidth}{!}{
    \begin{tabular}{lrlrlrl}
    \textbf{Model}             &  \mc{2}{Gen Dev} & \mc{2}{Test} & \mc{2}{Gen}  \\
    \toprule
    LSTM                        & - &    & \acc{99}{} & \acc{16}{8} \\
    Transformer                  & - &   & \acc{96}{} & \acc{35}{6} \\
    \hline
    LSTM &  \gacc{30.3}{7.3} & \gacc{99.7}{} & \gacc{34.5}{4.5}  \\
    \  \emph{w.} Uni-MAML & \gacc{36.1}{6.7}  & \gacc{99.7}{} & \gacc{36.4}{3.6} \\
    \ \emph{w.} Lev-MAML & \gacc{35.6}{5.3} & \gacc{99.7}{} & \gacc{36.4}{5.2}  \\
    \ \emph{w.} Str-MAML &  \gacc{36.3}{4.2} & \gacc{99.7}{} & \gacc{36.8}{3.5}  \\
    \  \emph{w.} Tree-MAML  & \GACC{41.2}{2.8}  & \gacc{99.7}{} & \GACC{41.0}{4.9} \\
    \hline
    Transformer &  \gacc{54.7}{4.0} & \gacc{99.5}{} & \gacc{58.6}{3.7}  \\
    \  \emph{w.} Uni-MAML & \gacc{60.9}{2.8}  & \gacc{99.6}{} & \gacc{64.4}{4.0} \\
    \  \emph{w.} Lev-MAML & \gacc{62.7}{3.8}  & \gacc{99.7}{} & \gacc{64.9}{6.3} \\
\ \emph{w.} Str-MAML &  \gacc{62.3}{3.0} & \gacc{99.6}{} & \gacc{64.8}{5.5}  \\
    \  \emph{w.} Tree-MAML & \GACC{64.1}{3.2}  & \gacc{99.6}{} & \GACC{66.7}{4.4} \\
    \bottomrule
    \end{tabular}
}
\setlength{\tabcolsep}{6pt}
\caption{Main results on the COGS dataset.  
We show the mean and variance (standard deviation) of 10 runs. 
Results in the top two rows are from \protect~\citet{kim_cogs_2020}, the remainder are results obtained in this paper.}
\label{table:cogs-results}
\end{table}

\subsection{Main Results}
On SCAN, as shown in Table~\ref{table:mcd-scan-results}, Lev-MAML substantially helps both base parsers achieve better performance across three different splits constructed according to the MCD principle.~\footnote{Our base parsers also perform much better than previous methods, likely due to the choice of hyperparameters.} 
Though our models do not utilize pre-training such as T5~\cite{raffel2019exploring}, our best model (Lev-MAML + LSTM) still outperforms T5 based models significantly in MCD1 and MCD2. We show that GECA is also effective for MCD splits (especially in MCD1). More importantly, augmenting GECA with Lev-MAML further boosts the performance substantially in MCD1 and MCD2, signifying that our MAML training is complementary to GECA to some degree.

Table~\ref{table:cogs-results} shows our results on COGS. Tree-MAML boosts the performance of both LSTM and Transformer base parsers by a large margin: 6.5\% and 8.1\% respectively in average accuracy. 
Moreover, Tree-MAML is consistently better than other MAML variants, showing the effectiveness of exploiting tree structures of formal representation to construct virtual tasks.
\footnote{The improvement of all of our MAML variants applied to the Transformer are significant (p < 0.03) compared to the baseline, of our methods applied to LSTMs, Tree-MAML is significant (p < 0.01) compared to the baseline.}

{
\pgfplotsset{height=2.5cm, width=6cm, compat=1.9}
\pgfplotsset{xtick style={draw=none}}
\pgfplotsset{ytick style={draw=none}}

\begin{table*}[ht]
    \centering
    \begin{tabularx}{\textwidth}{XXr}
    \toprule
         Training & Generalization &  Accuracy Distribution  \\ \midrule
         
     \multicolumn{3}{l}{Primitive noun $\rightarrow$ Subject (common noun)} \\ 
     \midrule
         \textbf{shark} & A \textbf{shark} examined the child. &     
             \raisebox{-0.7\totalheight}{\includegraphics{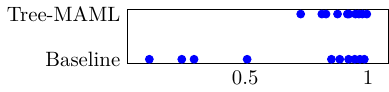}
}  \\
	\midrule        
	\multicolumn{3}{l}{Primitive noun $\rightarrow$ Subject (proper noun)} \\ 
	\midrule 
        \textbf{Paula} & \textbf{Paula} sketched William. &
             \raisebox{-0.7\totalheight}{\includegraphics{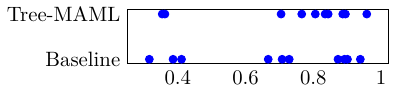}
}  \\
	\midrule        
	\multicolumn{3}{l}{Primitive noun $\rightarrow$ Object (common noun)} \\ 
	\midrule  
         \textbf{shark} & A chief heard the \textbf{shark}.  &
             \raisebox{-0.7\totalheight}{\includegraphics{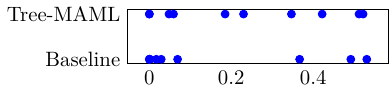}
}  \\
	\midrule       
	\multicolumn{3}{l}{Primitive noun $\rightarrow$ Object (proper noun)} \\ 
	\midrule   
         \textbf{Paula} & The child helped \textbf{Paula}. &
             \raisebox{-0.7\totalheight}{\includegraphics{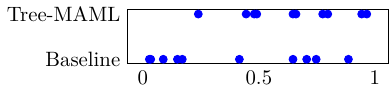}
}  \\ 
             
       \bottomrule
    \end{tabularx}
    \caption{Accuracy on COGS by generalization case. Each dot represents a single run of the model.}
    \label{table:case-results}
\end{table*}

}

\section{Discussion}

\subsection{SCAN Discussion}

The application of our string-similarity driven meta-learning approaches to the SCAN dataset improved the performance of the LSTM baseline parser. Our results are reported on three splits of the dataset generated according to the maximum compound divergence (MCD) principle. 
We report results on the only MCD tasks for SCAN as these tasks explicitly focus on the systematicity of language. As such they assess a model's ability to extract sufficiently atomic concepts from its input, such that it can still recognize those concepts in a new context (i.e. as part of a different compound). To succeed here a model must learn atoms from the training data and apply them compositionally at test time. The improvement in performance our approach achieves on this task suggests that it does disincentivise the model from memorizing large sections - or entire compounds - from its input. %

GECA applied to the SCAN MCD splits does improve performance of the baseline, however not to the same extent as when applied to other SCAN tasks in \citet{andreas_good-enough_2020}. GECA's improvement is comparable to our meta-learning method, despite the fact that our method does not leverage any data augmentation. This means that our method achieves high performance by generalizing robustly outside of its training distribution, rather than by making its training data more representative of the test distribution. The application of our Lev-MAML approach to GECA-augmented data results in further improvements in performance, suggesting that these approaches aid the model in distinct yet complementary ways.

\subsection{COGS Discussion}

All variants of our meta-learning approach improved both the LSTM and Transformer baseline parsers' performance on the COGS dataset. The Tree-MAML method outperforms the Lev-MAML, Str-MAML, and Uni-MAML versions. The only difference between these methods is the similarity metric used, and so differences in performance must be driven by what each metric selects for. For further analysis of the metrics refer to the appendix.

The strong performance of the Uni-MAML variant highlights the usefulness of our approach generally in improving models' generalization performance. Even without a specially designed meta-test task this approach substantially improves on the baseline Transformer model. We see this as evidence that this kind of meta-augmented supervised learning acts as a robust regularizer particularly for tasks requiring out of distribution generalization. 

Although the Uni-MAML, Lev-MAML, and Str-MAML versions perform similarly overall on the COGS dataset they may select for different generalization strategies. The COGS generalization set is comprised of 21 sub-tasks which can be used to better understand the ways in which a model is generalizing (refer to Table~\ref{table:case-results} for examples of subtask performance). Despite having very similar overall performance Uni-MAML and Str-MAML perform distinctly on individual COGS tasks - with their performance appearing to diverge on a number of of them. This would suggest that the design of the meta-test task may have a substantive impact on the kind of generalization strategy that emerges in the model. For further analysis of COGS sub-task performance see the appendix.

Our approaches' strong results on both of these datasets suggest that it aids compositional generalization generally. However it is worth nothing that both datasets shown here are synthetic, and although COGS endeavours to be similar to natural data, the application of our methods outside of synthetic datasets is important future work. 

\section{Related Work}
\label{sec:related_work}

\paragraph*{Compositional Generalization}

A large body of work on compositional generalization provide models with strong compositional bias,
such as specialized neural architectures \citep{li2019compositional,russin2019compositional,gordon_permutation_2020}, or
grammar-based models that accommodate alignments between natural language utterances and programs \citep{shaw_compositional_2020,herzig2020span}.
Another line of work  utilizes data augmentation via fixed rules \citep{andreas_good-enough_2020} 
or a learned network \citep{akyurek_lexicon_nodate} in an effort to transform the out-of-distribution compositional generalization task into an in-distribution one. Our work follows an orthogonal direction, 
injecting compositional bias using a specialized training algorithm.
A related area of research looks at the emergence of compositional languages, often showing that languages which seem to lack natural-language like compositional structure may still be able to generalize to novel concepts \citep{kottur_natural_2017, chaabouni_compositionality_2020}. This may help to explain the ways in which models can generalize robustly on in-distribution data unseen during training while still struggling on tasks specifically targeting compositionality.

\paragraph*{Meta-Learning for NLP}
Meta-learning methods \citep{vinyals2016matching,ravi2016optimization, finn_model-agnostic_2017}
that are widely used for few-shot learning, have been adapted 
for NLP applications like machine translation \citep{gu2018meta} and relation 
classification \citep{obamuyide2019model}.
In this work, we extend the conventional MAML \citep{finn_model-agnostic_2017} algorithm, which was initially 
proposed for few-shot learning, as a tool to inject inductive bias, inspired by \citet{li2018learning} and \citet{wang2020meta}. For compositional generalization, \citet{lake_compositional_nodate} proposes a meta-learning procedure to train a memory-augmented neural model. However, its meta-learning algorithm is specialized for the SCAN dataset~\cite{lake_generalization_2018} and not suitable to more realistic datasets.

\section{Conclusion}

Our work highlights the importance of training objectives that select for robust generalization strategies. The meta-learning augmented approach to supervised learning used here allows for the specification of different constraints on learning through the design of the meta-tasks. Our similarity-driven task design improved on baseline performance on two different compositional generalization datasets, by inhibiting the model's ability to memorize large sections of its input. Importantly though the overall approach used here is model agnostic, with portions of it (Str-MAML, Lev-MAML, and Uni-MAML) proving dataset agnostic as well requiring only that the input be a natural language sentence. Our methods are simple to implement compared with other approaches to improving compositional generalization, and we look forward to their use in combination with other techniques to further improve models' compositional ability.

\chapter{In Conclusion}
\label{chapt:conclusion}

{\makesans
\emph{
Summary and Future Work} \\[1em]
}

{\makesans
\begin{quote}
into the strenuous briefness \censor{[...........................................................................]} 
of yellow \censor{[...............]} coloured twilight i smilingly glide.  I into the big vermilion departure swim,sayingly; (Do you think?)the i do,world is probably made of roses \& hello: \censor{[..............................]}  \\
\flushright{\emph{- e.e. cummings}}
\end{quote}}

\drawline

\noindent This brings us to an end. The past 7 chapters have discussed mappings, and their structure in general terms. Introducing an approach to understanding how they represent, abstract and preserve information, by describing each system in terms of structural primitives. By quantifying the probability of 3 basic kinds of structure we end up being able to better understand how information is structured, how that structure emerges, and what structures drive generalisation in deep learning models. We briefly review how this case was built across chapters before turning to future work.

\section{In Summation}

Chapter \ref{chapt:intro} observed that we lack sufficient tools to understand how neural networks represent information, learn, and generalise. I made a case that one way to address this is to look at models as a member of a more general class: mappings. By drawing parallels with other mappings, like natural language, we can think about the kinds of \emph{system-level structures} that drive their performance. Relating representation spaces in these models to other mappings across the cognitive sciences also allows us to build on existing work for what factors --- like cognitive capacity --- condition or constrain the structures that emerge. The remainder of the thesis builds on these basic ideas working to quantify system-level structure in mappings learned by neural networks, and studying whether model capacity has a similar regularising effect to cognitive capacity seen in related work on humans.

\subsection{Information Structure}

A major goal of this thesis was to introduce flexible, quantitative ways of thinking about structure in a mapping. Chapter \ref{chapt:information} makes this concrete by introducing \emph{information structure} which takes a probabilistic approach - describing a system in terms of the probability of 3 structural primitives: one-to-one, one-to-many, and many-to-one. Each of these structures in a mapping between spaces relates intuitively to basic information theoretic quantities: Mutual Information, Conditional Entropy, and the Jensen-Shannon Divergence. The remainder of the thesis puts milage on these ideas, by leveraging them to study how structure develops in discrete mappings (in chapter \ref{chapt:comp_reg}), continuous mappings learned by deep-learning models trained on a single task (chapter \ref{chapt:learning}), and mappings internal to large language models (chapter \ref{chapt:llm}).

Chapter \ref{chapt:comp_reg} takes initial steps towards the theoretical framework for the thesis as a whole, looking at the discrete $\to$ discrete mappings that emerge in a multi-agent reinforcement learning model.  By quantifying 4 specific kinds of variation found in natural language in terms of conditional entropies, in a model designed to have a high-level analogy to human communication, experiments here build a direct link between information structure and structures in natural language. Chapter \ref{chapt:learning} formalises the information structure framework for transformer models, defining versions of regularity, variation, and disentanglement for vector space. These experiments show how early in training representations become regular with respect to word-level information, then later in training representations contextualise, with different contextualisations for the same token become more disentangled in space. This later phase takes place after in-distribution performance saturates and reflects the period of training where generalisation improves. At the end of training, degree of contextual disentanglement predicts which run of the model will generalise best. Additionally we show how a model's loss on the task correlates strongly with measures of structure early in training, showing the objective the model optimises for selects for specific structural properties in representational space.

Given that the approach in this thesis is built on information theory, we need methods for quickly, and accurately estimating entropy. Entropy estimation in vector spaces has long proved challenging, chapter \ref{chapt:llm} grapples with this problem, discussing limitations of existing approaches and introducing a novel method for estimating the entropy of vector space \emph{soft entropy.} This quantity behaves like discrete entropy but is differentiable, memory efficient, and highly parallelisable - enabling us to apply an information structure analysis to representations of arbitrary size. The remainder of the chapter applies the analysis to large language models ranging in size from 14 million parameters to 12 billion. We show how the timecourse of training larger models looks similar to models trained on a single task, with later steps resulting in disentanglement of different contextualisations of the same token. We also show how larger models use proportionally more of their representation space for contextual information.

\subsection{Capacity}

Across chapters this thesis also studies the effects of capacity on representational structure. In linguistic work cognitive capacity is thought to have a regularising effect, limiting learner's ability to learn lower probability forms \citep{newport_maturational_1990}. More generally, humans' cognitive limitations are thought to drive the robustly generalising strategies we converge to \citep{griffiths_understanding_2020}. In the multi-agent model in Chapter \ref{chapt:comp_reg}, limiting the capacity of each agent had a regularising effect on the signals they produced. When we look at model-internal representations in Chapter \ref{chapt:learning}, this effect inverted with models that have larger hidden representations compressing their representation spaces more per-dimension and converging to more regular representations with respect to contextual information. The pattern of larger models being more regular with respect to contextual information holds true for the large language models studied in chapter \ref{chapt:llm}.

We make sense of these contrasting results by discussing how modifying the hidden-size of a model (the number of dimensions available to it) affects capacity, but also key parameters of the representational space. When encoding meanings in discrete signals, two parameters of the signal space affect structural properties of encodings: maximum length and the size of the alphabet used at each position in the signal. A smaller alphabet is more robust to noise but requires a longer signal - illustrated by morse code where operators only need to differentiate between two possibilities at each position, (dot or dash), but where sentences in morse are far longer than their english counterparts. In our analysis increasing the number of hidden dimensions available to the model is akin to increasing the maximum signal length, which can allow the model to compress more per-dimension, arriving at more robust representations. As a result in Chapter \ref{chapt:comp_reg} where we measure regularity in the discrete signals between models, varying model capacity but not signal capacity shows a regularising effect. Understandably when we modify the number of hidden dimensions while also assessing regularity in hidden representations (chapters \ref{chapt:learning} and \ref{chapt:llm}) we see a different pattern. The final experimental chapter (\ref{chapt:meta}) looks at ways to modify the capacity of models without otherwise altering properties of their representational spaces. Using a meta-learning objective to limit model's ability to memorise training data results in improved generalisation performance. 

\section{Future Work}

There are three broad directions for further work that follow from this thesis, optimising for information structure, accounting for information content, and understanding the effects of a machine learning pipeline. 

\paragraph*{Optimising for Information Structure}
The clearest extension of the work presented here is to directly optimise for representational structure. Chapters \ref{chapt:learning} \& \ref{chapt:llm} identify representational structures that correlate with improved generalisation performance.  Given the soft entropy estimation method is differentiable, and fast to compute, we could directly optimise a model's representations to exhibit the structural properties we believe are desirable. This would mean adding our information structure estimates to the objective function and directly optimising for models to have higher regularity, variation, or disentanglement. While these may prove difficult to directly optimise for, and multi-term objectives can be challenging to work with, there is real appeal to being able to provide this kind of high-level structural supervision to hidden states in neural models.

\paragraph*{Accounting for Information}

In the experiments here the conditioning labels used to assess information structure were most often token, bigram, and trigram labels - because these are always available for text data. In future it would be interesting to try and account for all of the information in a model, reducing the `residual' as much as possible. This would require data with other sets of labels to try and explain other sources of variance in representation space. As an initial direction we could look at multi-linguality, encoding text from a number of different languages, tagged with their language ID. We could then compute what proportion of representation space is devoted to language-specific information. Additionally when conditioning on language ID labels, disentanglement would tell us how separable two languages are in representational space --- we could use this to see if languages that are more phylogenetically similar are more entangled in the model. Or if models which group language families together in space ultimately perform better across multi-lingual tasks.

\paragraph*{Understanding the Machine Learning Pipeline}

When training a deep-learning model in 2024 there are a huge number of hyper-parameters that need to be set, largely using specific values identified as optimal by previous work. We have a limited understanding of why certain training regimes are better or worse than others. For example the AdamW optimiser consistently out performs the Adam optimiser which consistently out performs standard stochastic gradient descent. By analysing the information structure that each of these optimisers selects for we could better understand what mechanisms drive robust generalisation in a model. In the general case, identifying the design choices that seem to have the greatest effect on performance, and analysing them from an information structure perspective could give use a representational account of why these choices matter. A clear extension to the preceeding chapter would be to analyse the information structure selected for by the meta-learning objective to give a representational account of how that objective affects model behaviour.

\section{In Closing}

Mappings relate two different spaces, transforming things of one kind into another; they are ubiquitous across the sciences and the world around us. By working to understand them in the general case, we work to unify understanding from a number of different perspectives across disciplines. Doing so lets us ground new, largely inscrutable systems like neural networks, in existing work on representational structure, how it emerges, and how it evolves. Understanding artificial systems in terms of natural ones also forces us to remember that solutions to complex problems are complex in their own right. Natural language is remarkable not just for the parts of it that are simple, and predictable, but for the way it weaves complexity and simplicity, variation and regularity together.

\printbibliography

@article{akyurek_compositionality_2022,
	title        = {Compositionality as {Lexical} {Symmetry}},
	author       = {Akyürek, Ekin and Andreas, Jacob},
	year         = {2022},
	month        = jan,
	journal      = {arXiv:2201.12926 [cs]},
	url          = {http://arxiv.org/abs/2201.12926},
	urldate      = {2022-04-28},
	note         = {arXiv: 2201.12926},
	language     = {en}
}

@article{akyurek_lexicon_nodate,
	title        = {Lexicon {Learning} for {Few}-{Shot} {Neural} {Sequence} {Modeling}},
	author       = {Akyürek, Ekin and Andreas, Jacob},
	pages        = {13},
	language     = {en}
}

@book{anderson_-morphous_1992,
	title        = {A-{Morphous} {Morphology}},
	author       = {Anderson, Stephen R.},
	year         = {1992},
	publisher    = {Cambridge University Press},
	address      = {Cambridge},
	doi          = {10.1017/CBO9780511586262},
	isbn         = {978-0-511-58626-2},
	url          = {http://ebooks.cambridge.org/ref/id/CBO9780511586262},
	urldate      = {2018-11-14},
	language     = {en}
}

@inproceedings{andreas_good-enough_2020,
	title        = {Good-{Enough} {Compositional} {Data} {Augmentation}},
	author       = {Andreas, Jacob},
	year         = {2020},
	booktitle    = {Proceedings of the 58th {Annual} {Meeting} of the {Association} for {Computational} {Linguistics}},
	publisher    = {Association for Computational Linguistics},
	address      = {Online},
	pages        = {7556--7566},
	doi          = {10.18653/v1/2020.acl-main.676},
	url          = {https://www.aclweb.org/anthology/2020.acl-main.676},
	urldate      = {2021-01-30},
	language     = {en}
}

@article{andreas_measuring_2019,
	title        = {Measuring {Compositionality} in {Representation} {Learning}},
	author       = {Andreas, Jacob},
	year         = {2019},
	month        = apr,
	journal      = {arXiv:1902.07181 [cs, stat]},
	url          = {http://arxiv.org/abs/1902.07181},
	urldate      = {2020-12-18},
	note         = {arXiv: 1902.07181},
	language     = {en}
}

@article{bahdanau2014neural,
	title        = {Neural machine translation by jointly learning to align and translate},
	author       = {Bahdanau, Dzmitry and Cho, Kyunghyun and Bengio, Yoshua},
	year         = {2014},
	journal      = {arXiv preprint arXiv:1409.0473}
}

@inproceedings{barry2017sentiment,
	title        = {Sentiment Analysis of Online Reviews Using Bag-of-Words and LSTM Approaches.},
	author       = {Barry, James},
	year         = {2017},
	booktitle    = {AICS},
	pages        = {272--274}
}

@article{Beckner2009,
	title        = {{Language Is a Complex Adaptive System: Position Paper}},
	author       = {Beckner, Clay and Blythe, Richard and Bybee, Joan and Christiansen, Morten H. and Croft, William and Ellis, Nick C. and Holland, John and Ke, Jinyun and Larsen-Freeman, Diane and Schoenemann, Tom},
	year         = {2009},
	journal      = {Language Learning},
	volume       = {59},
	number       = {March 2007},
	pages        = {1--26},
	doi          = {10.1111/j.1467-9922.2009.00533.x},
	issn         = {00238333},
	url          = {http://doi.wiley.com/10.1111/j.1467-9922.2009.00533.x}
}

@article{beirlant1997nonparametric,
	title        = {Nonparametric entropy estimation: An overview},
	author       = {Beirlant, Jan and Dudewicz, Edward J and Gy{\"o}rfi, L{\'a}szl{\'o} and Van der Meulen, Edward C and others},
	year         = {1997},
	journal      = {International Journal of Mathematical and Statistical Sciences},
	publisher    = {THESAURUS PUBLISHING},
	volume       = {6},
	number       = {1},
	pages        = {17--39}
}

@article{belinkov_evaluating_2018,
	title        = {Evaluating {Layers} of {Representation} in {Neural} {Machine} {Translation} on {Part}-of-{Speech} and {Semantic} {Tagging} {Tasks}},
	author       = {Belinkov, Yonatan and Màrquez, Lluís and Sajjad, Hassan and Durrani, Nadir and Dalvi, Fahim and Glass, James},
	year         = {2018},
	month        = jan,
	journal      = {arXiv:1801.07772 [cs]},
	url          = {http://arxiv.org/abs/1801.07772},
	urldate      = {2020-06-18},
	note         = {arXiv: 1801.07772},
	language     = {en}
}

@inproceedings{belinkov-etal-2017-neural,
	title        = {What do Neural Machine Translation Models Learn about Morphology?},
	author       = {Belinkov, Yonatan  and Durrani, Nadir  and Dalvi, Fahim  and Sajjad, Hassan  and Glass, James},
	year         = {2017},
	month        = jul,
	booktitle    = {Proceedings of the 55th Annual Meeting of the Association for Computational Linguistics (Volume 1: Long Papers)},
	publisher    = {Association for Computational Linguistics},
	address      = {Vancouver, Canada},
	pages        = {861--872},
	doi          = {10.18653/v1/P17-1080},
	url          = {https://aclanthology.org/P17-1080},
	editor       = {Barzilay, Regina  and Kan, Min-Yen}
}

@article{bengio_representation_2013,
	title        = {Representation {Learning}: {A} {Review} and {New} {Perspectives}},
	shorttitle   = {Representation {Learning}},
	author       = {Bengio, Y. and Courville, A. and Vincent, P.},
	year         = {2013},
	month        = aug,
	journal      = {IEEE Transactions on Pattern Analysis and Machine Intelligence},
	publisher    = {IEEE},
	volume       = {35},
	number       = {8},
	pages        = {1798--1828},
	doi          = {10.1109/TPAMI.2013.50},
	issn         = {0162-8828, 2160-9292},
	url          = {http://ieeexplore.ieee.org/document/6472238/},
	urldate      = {2024-01-31},
	language     = {en}
}

@article{futrell2019neural,
  title={Neural language models as psycholinguistic subjects: Representations of syntactic state},
  author={Futrell, Richard and Wilcox, Ethan and Morita, Takashi and Qian, Peng and Ballesteros, Miguel and Levy, Roger},
  journal={arXiv preprint arXiv:1903.03260},
  year={2019}
}

@book{lashley1951problem,
  title={The problem of serial order in behavior},
  author={Lashley, Karl Spencer},
  volume={21},
  year={1951},
  publisher={Bobbs-Merrill Oxford}
}

@book{goodman1955new,
  title={The new riddle of induction},
  author={Goodman, Nelson},
  year={1955},
  publisher={na}
}

@article{miller1951language,
  title={Language and communication},
  author={Miller, George Armitage},
  year={1951},
  publisher={McGraw-Hill}
}

@article{bricken2023towards,
  title={Towards monosemanticity: Decomposing language models with dictionary learning},
  author={Bricken, Trenton and Templeton, Adly and Batson, Joshua and Chen, Brian and Jermyn, Adam and Conerly, Tom and Turner, Nick and Anil, Cem and Denison, Carson and Askell, Amanda and others},
  journal={Transformer Circuits Thread},
  volume={2},
  year={2023}
}

@article{elhage2022toy,
  title={Toy models of superposition},
  author={Elhage, Nelson and Hume, Tristan and Olsson, Catherine and Schiefer, Nicholas and Henighan, Tom and Kravec, Shauna and Hatfield-Dodds, Zac and Lasenby, Robert and Drain, Dawn and Chen, Carol and others},
  journal={arXiv preprint arXiv:2209.10652},
  year={2022}
}

@article{muller2023subspace,
  title={Subspace chronicles: How linguistic information emerges, shifts and interacts during language model training},
  author={M{\"u}ller-Eberstein, Max and Van Der Goot, Rob and Plank, Barbara and Titov, Ivan},
  journal={arXiv preprint arXiv:2310.16484},
  year={2023}
}

@article{bickerton1984language,
	title        = {The language bioprogram hypothesis},
	author       = {Bickerton, Derek},
	year         = {1984},
	journal      = {Behavioral and brain sciences},
	publisher    = {Cambridge University Press},
	volume       = {7},
	number       = {2},
	pages        = {173--188}
}

@inproceedings{biderman2023pythia,
	title        = {Pythia: A suite for analyzing large language models across training and scaling},
	author       = {Biderman, Stella and Schoelkopf, Hailey and Anthony, Quentin Gregory and Bradley, Herbie and O'Brien, Kyle and Hallahan, Eric and Khan, Mohammad Aflah and Purohit, Shivanshu and Prashanth, USVSN Sai and Raff, Edward and others},
	year         = {2023},
	booktitle    = {International Conference on Machine Learning},
	pages        = {2397--2430},
	organization = {PMLR}
}

@book{bishop_pattern_2006,
	title        = {Pattern recognition and machine learning},
	author       = {Bishop, Christopher M.},
	year         = {2006},
	publisher    = {Springer},
	address      = {New York},
	series       = {Information science and statistics},
	isbn         = {978-0-387-31073-2},
	language     = {en}
}

@article{blevins_deep_2018,
	title        = {Deep {RNNs} {Encode} {Soft} {Hierarchical} {Syntax}},
	author       = {Blevins, Terra and Levy, Omer and Zettlemoyer, Luke},
	year         = {2018},
	month        = may,
	journal      = {arXiv:1805.04218 [cs]},
	url          = {http://arxiv.org/abs/1805.04218},
	urldate      = {2020-06-18},
	note         = {arXiv: 1805.04218},
	language     = {en}
}

@article{brighton_compositional_2002,
	title        = {Compositional {Syntax} {From} {Cultural} {Transmission}},
	author       = {Brighton, Henry},
	year         = {2002},
	month        = jan,
	journal      = {Artificial Life},
	publisher    = {MIT Press  238 Main St., Suite 500, Cambridge, MA 02142-1046 USA journals-info@mit.edu},
	volume       = {8},
	number       = {1},
	pages        = {25--54},
	doi          = {10.1162/106454602753694756},
	issn         = {1064-5462, 1530-9185},
	url          = {https://direct.mit.edu/artl/article/8/1/25-54/2396},
	urldate      = {2022-06-21},
	language     = {en}
}

@article{brighton_language_2005,
	title        = {Language as an evolutionary system},
	author       = {Brighton, H and Smith, K and Kirby, S},
	year         = {2005},
	month        = sep,
	journal      = {Physics of Life Reviews},
	publisher    = {Elsevier},
	volume       = {2},
	number       = {3},
	pages        = {177--226},
	doi          = {10.1016/j.plrev.2005.06.001},
	issn         = {15710645},
	url          = {https://linkinghub.elsevier.com/retrieve/pii/S1571064505000229},
	urldate      = {2022-01-29},
	language     = {en}
}

@article{brighton2006understanding,
	title        = {Understanding linguistic evolution by visualizing the emergence of topographic mappings},
	author       = {Brighton, Henry and Kirby, Simon},
	year         = {2006},
	journal      = {Artificial life},
	publisher    = {MIT Press},
	volume       = {12},
	number       = {2},
	pages        = {229--242}
}

@article{brown2020language,
	title        = {Language models are few-shot learners},
	author       = {Brown, Tom B},
	year         = {2020},
	journal      = {arXiv preprint arXiv:2005.14165}
}

@book{cann_formal_1993,
	title        = {Formal semantics an introduction},
	author       = {Cann, Ronnie},
	year         = {1993},
	publisher    = {Cambridge University Press},
	address      = {Cambridge [etc.},
	isbn         = {978-1-139-16631-7},
	url          = {http://0-www.ebooks.cambridge.org.cataleg.uoc.edu/ebook.jsf?bid=CBO9781139166317},
	urldate      = {2020-06-23},
	note         = {OCLC: 1120437841},
	language     = {English}
}

@book{carver1981love,
	title        = {What we talk about when we talk about love},
	author       = {Carver, Raymond},
	year         = {1981},
	publisher    = {Vintage}
}

@article{chaabouni_compositionality_2020,
	title        = {Compositionality and {Generalization} {In} {Emergent} {Languages}},
	author       = {Chaabouni, Rahma and Kharitonov, Eugene and Bouchacourt, Diane and Dupoux, Emmanuel and Baroni, Marco},
	year         = {2020},
	pages        = {16},
	language     = {en}
}

@article{chalmers1993connectionism,
	title        = {Connectionism and compositionality: Why Fodor and Pylyshyn were wrong},
	author       = {Chalmers, David J},
	year         = {1993},
	publisher    = {Taylor \& Francis}
}

@article{chater_simplicity_2003,
	title        = {Simplicity: a unifying principle in cognitive science?},
	shorttitle   = {Simplicity},
	author       = {Chater, Nick and Vitányi, Paul},
	year         = {2003},
	month        = jan,
	journal      = {Trends in Cognitive Sciences},
	publisher    = {Elsevier},
	volume       = {7},
	number       = {1},
	pages        = {19--22},
	doi          = {10.1016/S1364-6613(02)00005-0},
	issn         = {13646613},
	url          = {https://linkinghub.elsevier.com/retrieve/pii/S1364661302000050},
	urldate      = {2020-06-16},
	language     = {en}
}

@article{chater2009restrictions,
	title        = {Restrictions on biological adaptation in language evolution},
	author       = {Chater, Nick and Reali, Florencia and Christiansen, Morten H},
	year         = {2009},
	journal      = {Proceedings of the National Academy of Sciences},
	publisher    = {National Acad Sciences},
	volume       = {106},
	number       = {4},
	pages        = {1015--1020}
}

@inproceedings{chen_isolating_2018,
	title        = {Isolating {Sources} of {Disentanglement} in {Variational} {Autoencoders}},
	author       = {Chen, Ricky T. Q. and Li, Xuechen and Grosse, Roger B and Duvenaud, David K},
	year         = {2018},
	journal      = {Advances in neural information processing systems},
	booktitle    = {Advances in {Neural} {Information} {Processing} {Systems}},
	publisher    = {Curran Associates, Inc.},
	volume       = {31},
	url          = {https://proceedings.neurips.cc/paper_files/paper/2018/hash/1ee3dfcd8a0645a25a35977997223d22-Abstract.html},
	urldate      = {2024-01-31}
}

@article{cho2014learning,
	title        = {Learning phrase representations using RNN encoder-decoder for statistical machine translation},
	author       = {Cho, Kyunghyun and Van Merri{\"e}nboer, Bart and Gulcehre, Caglar and Bahdanau, Dzmitry and Bougares, Fethi and Schwenk, Holger and Bengio, Yoshua},
	year         = {2014},
	journal      = {arXiv preprint arXiv:1406.1078}
}

@inproceedings{Cho2014OnTP,
	title        = {On the Properties of Neural Machine Translation: Encoder–Decoder Approaches},
	author       = {Kyunghyun Cho and Bart van Merrienboer and Dzmitry Bahdanau and Yoshua Bengio},
	year         = {2014},
	booktitle    = {SSST@EMNLP}
}

@article{choi_compositional_2018,
	title        = {Compositional {Obverter} {Communication} {Learning} {From} {Raw} {Visual} {Input}},
	author       = {Choi, Edward and Lazaridou, Angeliki and de Freitas, Nando},
	year         = {2018},
	month        = apr,
	journal      = {arXiv:1804.02341 [cs]},
	pages        = {18},
	url          = {http://arxiv.org/abs/1804.02341},
	urldate      = {2022-01-17},
	note         = {arXiv: 1804.02341},
	language     = {en}
}

@book{chomsky_aspects_1965,
	title        = {Aspects of the theory of syntax},
	author       = {Chomsky, Noam},
	year         = {1965},
	publisher    = {The MIT Press},
	address      = {Cambridge, Massachusetts},
	series       = {Massachusetts {Institute} of {Technology}. {Research} {Laboratory} of {Electronics}. {Special} technical report},
	number       = {no. 11},
	isbn         = {978-0-262-52740-8},
	edition      = {50th Anniversary Edition}
}

@article{chomsky1957logical,
	title        = {Logical Structure in Language},
	author       = {Chomsky, Noam},
	year         = {1957},
	journal      = {Journal of the American Society for Information Science},
	publisher    = {American Documentation Institute},
	volume       = {8},
	number       = {4},
	pages        = {284}
}

@incollection{chomsky1969quine,
	title        = {Quine's empirical assumptions},
	author       = {Chomsky, Noam},
	year         = {1969},
	booktitle    = {Words and objections: Essays on the work of WV Quine},
	publisher    = {Springer},
	pages        = {53--68}
}

@article{chomsky1995language,
	title        = {Language and nature},
	author       = {Chomsky, Noam},
	year         = {1995},
	journal      = {Mind},
	publisher    = {JSTOR},
	volume       = {104},
	number       = {413},
	pages        = {1--61}
}

@book{chomsky2014minimalist,
	title        = {The minimalist program},
	author       = {Chomsky, Noam},
	year         = {2014},
	publisher    = {MIT press}
}

@inproceedings{Christiansen1994InniteLF,
	title        = {Innite Languages, Finite Minds Connectionism, Learning and Linguistic Structure},
	author       = {Morten H. Christiansen},
	year         = {1994},
	url          = {https://api.semanticscholar.org/CorpusID:15139134}
}

@article{collins_convolution_nodate,
	title        = {Convolution {Kernels} for {Natural} {Language}},
	author       = {Collins, Michael and Duffy, Nigel},
	year         = {2001},
	booktitle    = {Advances in neural information processing systems},
	pages        = {8},
	language     = {en}
}

@article{conklin_meta-learning_2021,
	title        = {Meta-{Learning} to {Compositionally} {Generalize}},
	author       = {Conklin, Henry and Wang, Bailin and Smith, Kenny and Titov, Ivan},
	year         = {2021},
	month        = jun,
	journal      = {arXiv:2106.04252 [cs]},
	url          = {http://arxiv.org/abs/2106.04252},
	urldate      = {2022-05-04},
	note         = {arXiv: 2106.04252},
	language     = {en}
}

@inproceedings{conklin2022compositionality,
	title        = {Compositionality with Variation Reliably Emerges in Neural Networks},
	author       = {Conklin, Henry and Smith, Kenny},
	year         = {2022},
	booktitle    = {The Eleventh International Conference on Learning Representations}
}

@article{conklin2024representations,
	title        = {Representations as Language: An Information-Theoretic Framework for Interpretability},
	author       = {Conklin, Henry and Smith, Kenny},
	year         = {2024},
	journal      = {arXiv preprint arXiv:2406.02449}
}

@book{croft2001radical,
	title        = {Radical construction grammar: Syntactic theory in typological perspective},
	author       = {Croft, William},
	year         = {2001},
	publisher    = {Oxford University Press, USA}
}

@article{csordas_devil_2021,
	title        = {The {Devil} is in the {Detail}: {Simple} {Tricks} {Improve} {Systematic} {Generalization} of {Transformers}},
	shorttitle   = {The {Devil} is in the {Detail}},
	author       = {Csordás, Róbert and Irie, Kazuki and Schmidhuber, Jürgen},
	year         = {2021},
	month        = sep,
	journal      = {arXiv:2108.12284 [cs]},
	url          = {http://arxiv.org/abs/2108.12284},
	urldate      = {2021-10-14},
	note         = {arXiv: 2108.12284},
	language     = {en}
}

@article{culbertson_simplicity_2016,
	title        = {Simplicity and {Specificity} in {Language}: {Domain}-{General} {Biases} {Have} {Domain}-{Specific} {Effects}},
	shorttitle   = {Simplicity and {Specificity} in {Language}},
	author       = {Culbertson, Jennifer and Kirby, Simon},
	year         = {2016},
	month        = jan,
	journal      = {Frontiers in Psychology},
	publisher    = {Frontiers Media SA},
	volume       = {6},
	pages        = {1964},
	doi          = {10.3389/fpsyg.2015.01964},
	issn         = {1664-1078},
	url          = {http://journal.frontiersin.org/Article/10.3389/fpsyg.2015.01964/abstract},
	urldate      = {2020-06-16},
	language     = {en}
}

@article{dagaev_too-good--be-true_2021,
	title        = {A {Too}-{Good}-to-be-{True} {Prior} to {Reduce} {Shortcut} {Reliance}},
	author       = {Dagaev, Nikolay and Roads, Brett D. and Luo, Xiaoliang and Barry, Daniel N. and Patil, Kaustubh R. and Love, Bradley C.},
	year         = {2021},
	month        = jun,
	journal      = {arXiv:2102.06406 [cs]},
	url          = {http://arxiv.org/abs/2102.06406},
	urldate      = {2021-09-23},
	note         = {arXiv: 2102.06406},
	language     = {en}
}

@article{dale_understanding_2012,
	title        = {{UNDERSTANDING} {THE} {ORIGINS} {OF} {MORPHOLOGICAL} {DIVERSITY}: {THE} {LINGUISTIC} {NICHE} {HYPOTHESIS}},
	shorttitle   = {{UNDERSTANDING} {THE} {ORIGINS} {OF} {MORPHOLOGICAL} {DIVERSITY}},
	author       = {Dale, Rick and Lupyan, Gary},
	year         = {2012},
	month        = may,
	journal      = {Advances in Complex Systems},
	publisher    = {World Scientific},
	volume       = {15},
	number       = {03n04},
	pages        = {1150017},
	doi          = {10.1142/S0219525911500172},
	issn         = {0219-5259, 1793-6802},
	url          = {https://www.worldscientific.com/doi/abs/10.1142/S0219525911500172},
	urldate      = {2022-01-17},
	language     = {en}
}

@article{damour_underspecification_2020,
	title        = {Underspecification {Presents} {Challenges} for {Credibility} in {Modern} {Machine} {Learning}},
	author       = {D'Amour, Alexander and Heller, Katherine and Moldovan, Dan and Adlam, Ben and Alipanahi, Babak and Beutel, Alex and Chen, Christina and Deaton, Jonathan and Eisenstein, Jacob and Hoffman, Matthew D. and Hormozdiari, Farhad and Houlsby, Neil and Hou, Shaobo and Jerfel, Ghassen and Karthikesalingam, Alan and Lucic, Mario and Ma, Yian and McLean, Cory and Mincu, Diana and Mitani, Akinori and Montanari, Andrea and Nado, Zachary and Natarajan, Vivek and Nielson, Christopher and Osborne, Thomas F. and Raman, Rajiv and Ramasamy, Kim and Sayres, Rory and Schrouff, Jessica and Seneviratne, Martin and Sequeira, Shannon and Suresh, Harini and Veitch, Victor and Vladymyrov, Max and Wang, Xuezhi and Webster, Kellie and Yadlowsky, Steve and Yun, Taedong and Zhai, Xiaohua and Sculley, D.},
	year         = {2020},
	month        = nov,
	journal      = {arXiv:2011.03395 [cs, stat]},
	url          = {http://arxiv.org/abs/2011.03395},
	urldate      = {2021-01-24},
	note         = {arXiv: 2011.03395},
	language     = {en}
}

@article{devlin_bert_2019,
	title        = {{BERT}: {Pre}-training of {Deep} {Bidirectional} {Transformers} for {Language} {Understanding}},
	shorttitle   = {{BERT}},
	author       = {Devlin, Jacob and Chang, Ming-Wei and Lee, Kenton and Toutanova, Kristina},
	year         = {2019},
	month        = may,
	journal      = {arXiv:1810.04805 [cs]},
	url          = {http://arxiv.org/abs/1810.04805},
	urldate      = {2020-06-19},
	note         = {arXiv: 1810.04805},
	language     = {en}
}

@misc{dong_language_2016,
	title        = {Language to {Logical} {Form} with {Neural} {Attention}},
	author       = {Dong, Li and Lapata, Mirella},
	year         = {2016},
	month        = jun,
	journal      = {arXiv preprint arXiv:1601.01280},
	publisher    = {arXiv},
	url          = {http://arxiv.org/abs/1601.01280},
	urldate      = {2022-09-10},
	note         = {arXiv:1601.01280 [cs]},
	language     = {en}
}

@article{dubey2024llama,
	title        = {The llama 3 herd of models},
	author       = {Dubey, Abhimanyu and Jauhri, Abhinav and Pandey, Abhinav and Kadian, Abhishek and Al-Dahle, Ahmad and Letman, Aiesha and Mathur, Akhil and Schelten, Alan and Yang, Amy and Fan, Angela and others},
	year         = {2024},
	journal      = {arXiv preprint arXiv:2407.21783}
}

@article{dziri2024faith,
	title        = {Faith and fate: Limits of transformers on compositionality},
	author       = {Dziri, Nouha and Lu, Ximing and Sclar, Melanie and Li, Xiang Lorraine and Jiang, Liwei and Lin, Bill Yuchen and Welleck, Sean and West, Peter and Bhagavatula, Chandra and Le Bras, Ronan and others},
	year         = {2024},
	journal      = {Advances in Neural Information Processing Systems},
	volume       = {36}
}

@article{elhage2021mathematical,
	title        = {A mathematical framework for transformer circuits},
	author       = {Elhage, Nelson and Nanda, Neel and Olsson, Catherine and Henighan, Tom and Joseph, Nicholas and Mann, Ben and Askell, Amanda and Bai, Yuntao and Chen, Anna and Conerly, Tom and others},
	year         = {2021},
	journal      = {Transformer Circuits Thread},
	volume       = {1},
	pages        = {1}
}

@article{elman1990finding,
	title        = {Finding structure in time},
	author       = {Elman, Jeffrey L},
	year         = {1990},
	journal      = {Cognitive science},
	publisher    = {Wiley Online Library},
	volume       = {14},
	number       = {2},
	pages        = {179--211}
}

@article{fedorenko2014role,
	title        = {The role of domain-general cognitive control in language comprehension},
	author       = {Fedorenko, Evelina},
	year         = {2014},
	journal      = {Frontiers in psychology},
	publisher    = {Frontiers Media SA},
	volume       = {5},
	pages        = {335}
}

@article{ferdinand_cognitive_2019,
	title        = {The cognitive roots of regularization in language},
	author       = {Ferdinand, Vanessa and Kirby, Simon and Smith, Kenny},
	year         = {2019},
	month        = mar,
	journal      = {Cognition},
	volume       = {184},
	pages        = {53--68},
	doi          = {10.1016/j.cognition.2018.12.002},
	issn         = {00100277},
	url          = {https://linkinghub.elsevier.com/retrieve/pii/S0010027718303135},
	urldate      = {2022-11-13},
	language     = {en}
}

@article{finn_model-agnostic_2017,
	title        = {Model-{Agnostic} {Meta}-{Learning} for {Fast} {Adaptation} of {Deep} {Networks}},
	author       = {Finn, Chelsea and Abbeel, Pieter and Levine, Sergey},
	year         = {2017},
	month        = jul,
	journal      = {arXiv:1703.03400 [cs]},
	booktitle    = {Proceedings of the 34th International Conference on Machine Learning-Volume 70},
	pages        = {10},
	url          = {http://arxiv.org/abs/1703.03400},
	urldate      = {2020-05-25},
	note         = {arXiv: 1703.03400},
	language     = {en},
	organization = {JMLR. org}
}

@article{fodor_connectionism_1988,
	title        = {Connectionism and cognitive architecture: {A} critical analysis},
	shorttitle   = {Connectionism and cognitive architecture},
	author       = {Fodor, Jerry A. and Pylyshyn, Zenon W.},
	year         = {1988},
	month        = mar,
	journal      = {Cognition},
	volume       = {28},
	number       = {1-2},
	pages        = {3--71},
	doi          = {10.1016/0010-0277(88)90031-5},
	issn         = {00100277},
	url          = {https://linkinghub.elsevier.com/retrieve/pii/0010027788900315},
	urldate      = {2021-01-25},
	language     = {en}
}

@misc{fodor1975language,
	title        = {The language of thought},
	author       = {Fodor, Jerry A},
	year         = {1975},
	publisher    = {Harvard University Press},
	volume       = {5}
}

@article{frankle2018lottery,
	title        = {The lottery ticket hypothesis: Finding sparse, trainable neural networks},
	author       = {Frankle, Jonathan and Carbin, Michael},
	year         = {2018},
	journal      = {arXiv preprint arXiv:1803.03635}
}

@misc{furrer_compositional_2021,
	title        = {Compositional {Generalization} in {Semantic} {Parsing}: {Pre}-training vs. {Specialized} {Architectures}},
	shorttitle   = {Compositional {Generalization} in {Semantic} {Parsing}},
	author       = {Furrer, Daniel and van Zee, Marc and Scales, Nathan and Schärli, Nathanael},
	year         = {2021},
	month        = sep,
	journal      = {arXiv preprint arXiv:2007.08970},
	publisher    = {arXiv},
	url          = {http://arxiv.org/abs/2007.08970},
	urldate      = {2023-10-23},
	note         = {arXiv:2007.08970 [cs]}
}

@article{futrell2018rnns,
	title        = {RNNs as psycholinguistic subjects: Syntactic state and grammatical dependency},
	author       = {Futrell, Richard and Wilcox, Ethan and Morita, Takashi and Levy, Roger},
	year         = {2018},
	journal      = {arXiv preprint arXiv:1809.01329}
}

@book{goldberg_constructions_1995,
	title        = {Constructions: a construction grammar approach to argument structure},
	shorttitle   = {Constructions},
	author       = {Goldberg, Adele E.},
	year         = {1995},
	publisher    = {University of Chicago Press},
	address      = {Chicago},
	series       = {Cognitive theory of language and culture}
}

@article{goldberg_constructions_2003,
	title        = {Constructions: a new theoretical approach to language},
	shorttitle   = {Constructions},
	author       = {Goldberg, Adele E},
	year         = {2003},
	month        = may,
	journal      = {Trends in Cognitive Sciences},
	publisher    = {Elsevier},
	volume       = {7},
	number       = {5},
	pages        = {219--224},
	doi          = {10.1016/S1364-6613(03)00080-9},
	issn         = {13646613},
	url          = {https://linkinghub.elsevier.com/retrieve/pii/S1364661303000809},
	urldate      = {2020-06-23},
	language     = {en}
}

@book{goldberg_constructions_2006,
	title        = {Constructions at work: the nature of generalization in language},
	shorttitle   = {Constructions at work},
	author       = {Goldberg, Adele E},
	year         = {2006},
	publisher    = {Oxford University Press},
	address      = {Oxford; New York},
	url          = {http://public.ebookcentral.proquest.com/choice/publicfullrecord.aspx?p=3052348},
	urldate      = {2020-06-23},
	note         = {OCLC: 193697889},
	language     = {English.}
}

@article{goldfarb-tarrant_intrinsic_2021,
	title        = {Intrinsic {Bias} {Metrics} {Do} {Not} {Correlate} with {Application} {Bias}},
	author       = {Goldfarb-Tarrant, Seraphina and Marchant, Rebecca and Sanchez, Ricardo Muñoz and Pandya, Mugdha and Lopez, Adam},
	year         = {2021},
	month        = jun,
	journal      = {arXiv:2012.15859 [cs]},
	url          = {http://arxiv.org/abs/2012.15859},
	urldate      = {2021-08-05},
	note         = {arXiv: 2012.15859},
	language     = {en}
}

@article{goldfeld2018estimating,
	title        = {Estimating information flow in deep neural networks},
	author       = {Goldfeld, Ziv and Berg, Ewout van den and Greenewald, Kristjan and Melnyk, Igor and Nguyen, Nam and Kingsbury, Brian and Polyanskiy, Yury},
	year         = {2018},
	journal      = {arXiv preprint arXiv:1810.05728}
}

@article{goodfellow2014explaining,
	title        = {Explaining and harnessing adversarial examples},
	author       = {Goodfellow, Ian J and Shlens, Jonathon and Szegedy, Christian},
	year         = {2014},
	journal      = {arXiv preprint arXiv:1412.6572}
}

@article{gordon_permutation_2020,
	title        = {{PERMUTATION} {EQUIVARIANT} {MODELS} {FOR} {COMPOSITIONAL} {GENERALIZATION} {IN} {LANGUAGE}},
	author       = {Gordon, Jonathan and Lopez-Paz, David and Baroni, Marco and Bouchacourt, Diane},
	year         = {2020},
	booktitle    = {International Conference on Learning Representations},
	pages        = {12},
	language     = {en}
}

@article{griffiths_understanding_2020,
	title        = {Understanding {Human} {Intelligence} through {Human} {Limitations}},
	author       = {Griffiths, Thomas L.},
	year         = {2020},
	month        = nov,
	journal      = {Trends in Cognitive Sciences},
	publisher    = {Elsevier},
	volume       = {24},
	number       = {11},
	pages        = {873--883},
	doi          = {10.1016/j.tics.2020.09.001},
	issn         = {13646613},
	url          = {https://linkinghub.elsevier.com/retrieve/pii/S1364661320302151},
	urldate      = {2024-01-30},
	language     = {en}
}

@article{Griffiths2020UnderstandingHI,
	title        = {Understanding Human Intelligence through Human Limitations},
	author       = {Thomas L. Griffiths},
	year         = {2020},
	journal      = {Trends in Cognitive Sciences},
	volume       = {24},
	pages        = {873--883},
	url          = {https://api.semanticscholar.org/CorpusID:221996148}
}

@article{guo_expressivity_2021,
	title        = {Expressivity of {Emergent} {Language} is a {Trade}-off between {Contextual} {Complexity} and {Unpredictability}},
	author       = {Guo, Shangmin and Ren, Yi and Mathewson, Kory and Kirby, Simon and Albrecht, Stefano V. and Smith, Kenny},
	year         = {2021},
	month        = jun,
	journal      = {arXiv:2106.03982 [cs]},
	url          = {http://arxiv.org/abs/2106.03982},
	urldate      = {2021-10-21},
	note         = {arXiv: 2106.03982},
	language     = {en}
}

@article{hahn_resource-rational_2022,
	title        = {A resource-rational model of human processing of recursive linguistic structure},
	author       = {Hahn, Michael and Futrell, Richard and Levy, Roger and Gibson, Edward},
	year         = {2022},
	month        = oct,
	journal      = {Proceedings of the National Academy of Sciences},
	volume       = {119},
	number       = {43},
	pages        = {e2122602119},
	doi          = {10.1073/pnas.2122602119},
	issn         = {0027-8424, 1091-6490},
	url          = {https://pnas.org/doi/10.1073/pnas.2122602119},
	urldate      = {2022-11-02},
	language     = {en}
}

@techreport{haussler1999convolution,
	title        = {Convolution kernels on discrete structures},
	author       = {Haussler, David},
	year         = {1999},
	institution  = {Technical report, Department of Computer Science, University of California~…}
}

@article{havrylov_emergence_2017,
	title        = {Emergence of {Language} with {Multi}-agent {Games}: {Learning} to {Communicate} with {Sequences} of {Symbols}},
	shorttitle   = {Emergence of {Language} with {Multi}-agent {Games}},
	author       = {Havrylov, Serhii and Titov, Ivan},
	year         = {2017},
	month        = nov,
	journal      = {arXiv:1705.11192 [cs]},
	url          = {http://arxiv.org/abs/1705.11192},
	urldate      = {2020-01-21},
	note         = {arXiv: 1705.11192},
	language     = {en}
}

@article{herzig2020span,
	title        = {Span-based semantic parsing for compositional generalization},
	author       = {Herzig, Jonathan and Berant, Jonathan},
	year         = {2020},
	journal      = {arXiv preprint arXiv:2009.06040}
}

@article{hochreiter1997long,
	title        = {Long short-term memory},
	author       = {Hochreiter, Sepp and Schmidhuber, J{\"u}rgen},
	year         = {1997},
	journal      = {Neural computation},
	publisher    = {MIT Press},
	volume       = {9},
	number       = {8},
	pages        = {1735--1780}
}

@article{Hockett1960,
	title        = {{The Origin of Speech}},
	author       = {Hockett, Charles F.},
	year         = {1960},
	journal      = {Scientific American},
	publisher    = {JSTOR},
	volume       = {203},
	number       = {3},
	pages        = {88--97},
	doi          = {10.2307/24940617}
}

@article{hu_systematic_2020,
	title        = {A {Systematic} {Assessment} of {Syntactic} {Generalization} in {Neural} {Language} {Models}},
	author       = {Hu, Jennifer and Gauthier, Jon and Qian, Peng and Wilcox, Ethan and Levy, Roger P.},
	year         = {2020},
	month        = may,
	journal      = {arXiv:2005.03692 [cs]},
	url          = {http://arxiv.org/abs/2005.03692},
	urldate      = {2020-06-23},
	note         = {arXiv: 2005.03692},
	language     = {en}
}

@article{hudson_kam_investigating_2009,
	title        = {Investigating the cause of language regularization in adults: {Memory} constraints or learning effects?},
	shorttitle   = {Investigating the cause of language regularization in adults},
	author       = {Hudson Kam, Carla L. and Chang, Ann},
	year         = {2009},
	journal      = {Journal of Experimental Psychology: Learning, Memory, and Cognition},
	volume       = {35},
	number       = {3},
	pages        = {815--821},
	doi          = {10.1037/a0015097},
	issn         = {1939-1285, 0278-7393},
	url          = {http://doi.apa.org/getdoi.cfm?doi=10.1037/a0015097},
	urldate      = {2022-09-15},
	language     = {en}
}

@article{hudson_kam_regularizing_2005,
	title        = {Regularizing {Unpredictable} {Variation}: {The} {Roles} of {Adult} and {Child} {Learners} in {Language} {Formation} and {Change}},
	shorttitle   = {Regularizing {Unpredictable} {Variation}},
	author       = {Hudson Kam, Carla L. and Newport, Elissa L.},
	year         = {2005},
	month        = apr,
	journal      = {Language Learning and Development},
	volume       = {1},
	number       = {2},
	pages        = {151--195},
	doi          = {10.1080/15475441.2005.9684215},
	issn         = {1547-5441, 1547-3341},
	url          = {http://www.tandfonline.com/doi/abs/10.1080/15475441.2005.9684215},
	urldate      = {2023-01-27},
	language     = {en}
}

@article{hupkes_compositionality_2019,
	title        = {The compositionality of neural networks: integrating symbolism and connectionism},
	shorttitle   = {The compositionality of neural networks},
	author       = {Hupkes, Dieuwke and Dankers, Verna and Mul, Mathijs and Bruni, Elia},
	year         = {2019},
	month        = aug,
	journal      = {arXiv:1908.08351 [cs, stat]},
	url          = {http://arxiv.org/abs/1908.08351},
	urldate      = {2020-01-13},
	note         = {arXiv: 1908.08351},
	language     = {en}
}

@article{hupkes_visualisation_2018,
	title        = {Visualisation and '{Diagnostic} {Classifiers}' {Reveal} {How} {Recurrent} and {Recursive} {Neural} {Networks} {Process} {Hierarchical} {Structure}},
	author       = {Hupkes, Dieuwke and Veldhoen, Sara and Zuidema, Willem},
	year         = {2018},
	month        = apr,
	journal      = {Journal of Artificial Intelligence Research},
	volume       = {61},
	pages        = {907--926},
	doi          = {10.1613/jair.1.11196},
	issn         = {1076-9757},
	url          = {http://jair.org/index.php/jair/article/view/11196},
	urldate      = {2021-06-03},
	language     = {en}
}

@inproceedings{hurford2003synonymy,
	title        = {Why synonymy is rare: Fitness is in the speaker},
	author       = {Hurford, James R},
	year         = {2003},
	booktitle    = {European conference on artificial life},
	pages        = {442--451},
	organization = {Springer}
}

@article{Jaynes1957InformationTA,
	title        = {Information Theory and Statistical Mechanics},
	author       = {Edwin T. Jaynes},
	year         = {1957},
	journal      = {Physical Review},
	volume       = {106},
	pages        = {620--630},
	url          = {https://api.semanticscholar.org/CorpusID:17870175}
}

@article{jiang_characterizing_2020,
	title        = {Characterizing {Structural} {Regularities} of {Labeled} {Data} in {Overparameterized} {Models}},
	author       = {Jiang, Ziheng and Zhang, Chiyuan and Talwar, Kunal and Mozer, Michael C.},
	year         = {2020},
	month        = jun,
	journal      = {arXiv:2002.03206 [cs, stat]},
	url          = {http://arxiv.org/abs/2002.03206},
	urldate      = {2021-01-25},
	note         = {arXiv: 2002.03206},
	language     = {en}
}

@inproceedings{kalchbrenner2013recurrent,
	title        = {Recurrent continuous translation models},
	author       = {Kalchbrenner, Nal and Blunsom, Phil},
	year         = {2013},
	booktitle    = {Proceedings of the 2013 conference on empirical methods in natural language processing},
	pages        = {1700--1709}
}

@article{kemp2018semantic,
	title        = {Semantic typology and efficient communication},
	author       = {Kemp, Charles and Xu, Yang and Regier, Terry},
	year         = {2018},
	journal      = {Annual Review of Linguistics},
	publisher    = {Annual Reviews},
	volume       = {4},
	number       = {1},
	pages        = {109--128}
}

@article{kennedy2023public,
	title        = {Public awareness of artificial intelligence in everyday activities},
	author       = {Kennedy, Brian and Tyson, Alec and Saks, Emily},
	year         = {2023},
	publisher    = {Pew Research Center}
}

@article{keysers_measuring_2020,
	title        = {Measuring {Compositional} {Generalization}: {A} {Comprehensive} {Method} on {Realistic} {Data}},
	author       = {Keysers, Daniel and Schärli, Nathanael and Scales, Nathan and Buisman, Hylke and Furrer, Daniel and Kashubin, Sergii and Momchev, Nikola and Sinopalnikov, Danila and Staﬁniak, Lukasz and Tihon, Tibor and Tsarkov, Dmitry and Wang, Xiao and van Zee, Marc and Bousquet, Olivier},
	year         = {2020},
	journal      = {arXiv preprint arXiv:1912.09713},
	booktitle    = {International Conference on Learning Representations},
	pages        = {38},
	url          = {https://openreview.net/forum?id=SygcCnNKwr},
	language     = {en}
}

@article{kharitonov_compo_2020,
	title        = {Emergent Language Generalization and Acquisition Speed are not tied to Compositionality},
	author       = {Kharitonov, Eugene and Baroni, Marco},
	year         = {2020},
	month        = apr,
	journal      = {arXiv:2004.03420 [cs]},
	url          = {http://arxiv.org/abs/2004.03420},
	urldate      = {2022-01-29},
	note         = {arXiv: 2004.03420},
	language     = {en}
}

@article{kharitonov_egg_2019,
	title        = {{EGG}: a toolkit for research on {Emergence} of {lanGuage} in {Games}},
	shorttitle   = {{EGG}},
	author       = {Kharitonov, Eugene and Chaabouni, Rahma and Bouchacourt, Diane and Baroni, Marco},
	year         = {2019},
	month        = oct,
	journal      = {arXiv:1907.00852 [cs]},
	url          = {http://arxiv.org/abs/1907.00852},
	urldate      = {2022-01-14},
	note         = {arXiv: 1907.00852},
	language     = {en}
}

@article{kim_cogs_2020,
	title        = {{COGS}: {A} {Compositional} {Generalization} {Challenge} {Based} on {Semantic} {Interpretation}},
	shorttitle   = {{COGS}},
	author       = {Kim, Najoung and Linzen, Tal},
	year         = {2020},
	month        = oct,
	journal      = {arXiv:2010.05465 [cs]},
	booktitle    = {Proceedings of the 2020 conference on empirical methods in natural language processing (emnlp)},
	pages        = {9087--9105},
	url          = {http://arxiv.org/abs/2010.05465},
	urldate      = {2021-01-25},
	note         = {arXiv: 2010.05465},
	language     = {en}
}

@article{kingma2014adam,
	title        = {Adam: A method for stochastic optimization},
	author       = {Kingma, Diederik P and Ba, Jimmy},
	year         = {2014},
	journal      = {arXiv preprint arXiv:1412.6980}
}

@article{kirby_compression_2015,
	title        = {Compression and communication in the cultural evolution of linguistic structure},
	author       = {Kirby, Simon and Tamariz, Monica and Cornish, Hannah and Smith, Kenny},
	year         = {2015},
	month        = aug,
	journal      = {Cognition},
	publisher    = {Elsevier B.V.},
	volume       = {141},
	pages        = {87--102},
	doi          = {10.1016/j.cognition.2015.03.016},
	isbn         = {0393320782},
	issn         = {00100277},
	url          = {https://linkinghub.elsevier.com/retrieve/pii/S0010027715000815},
	urldate      = {2020-01-29},
	language     = {en},
	pmid         = {25966840}
}

@article{kirby_cumulative_2008,
	title        = {Cumulative cultural evolution in the laboratory: {An} experimental approach to the origins of structure in human language},
	shorttitle   = {Cumulative cultural evolution in the laboratory},
	author       = {Kirby, S. and Cornish, H. and Smith, K.},
	year         = {2008},
	month        = aug,
	journal      = {Proceedings of the National Academy of Sciences},
	publisher    = {National Academy of Sciences},
	volume       = {105},
	number       = {31},
	pages        = {10681--10686},
	doi          = {10.1073/pnas.0707835105},
	issn         = {0027-8424, 1091-6490},
	url          = {http://www.pnas.org/cgi/doi/10.1073/pnas.0707835105},
	urldate      = {2020-01-29},
	language     = {en},
	pmid         = {18667697}
}

@article{kirby_iterated_2014,
	title        = {Iterated learning and the evolution of language},
	author       = {Kirby, Simon and Griffiths, Tom and Smith, Kenny},
	year         = {2014},
	month        = oct,
	journal      = {Current Opinion in Neurobiology},
	publisher    = {Elsevier Ltd},
	volume       = {28},
	pages        = {108--114},
	doi          = {10.1016/j.conb.2014.07.014},
	issn         = {09594388},
	url          = {https://linkinghub.elsevier.com/retrieve/pii/S0959438814001421},
	urldate      = {2020-01-29},
	language     = {en},
	pmid         = {25062470}
}

@article{kirby_spontaneous_2001,
	title        = {Spontaneous evolution of linguistic structure-an iterated learning model of the emergence of regularity and irregularity},
	author       = {Kirby, S.},
	year         = {2001},
	month        = apr,
	journal      = {IEEE Transactions on Evolutionary Computation},
	volume       = {5},
	number       = {2},
	pages        = {102--110},
	doi          = {10.1109/4235.918430},
	issn         = {1089778X},
	url          = {http://ieeexplore.ieee.org/document/918430/},
	urldate      = {2020-01-29},
	language     = {en}
}

@article{kottur_natural_2017,
	title        = {Natural {Language} {Does} {Not} {Emerge} '{Naturally}' in {Multi}-{Agent} {Dialog}},
	author       = {Kottur, Satwik and Moura, José M. F. and Lee, Stefan and Batra, Dhruv},
	year         = {2017},
	month        = aug,
	journal      = {arXiv:1706.08502 [cs]},
	booktitle    = {Proceedings of the 2017 Conference on Empirical Methods in Natural Language Processing},
	publisher    = {Association for Computational Linguistics},
	address      = {Stroudsburg, PA, USA},
	pages        = {2962--2967},
	doi          = {10.18653/v1/D17-1321},
	url          = {http://arxiv.org/abs/1706.08502},
	urldate      = {2020-06-23},
	note         = {arXiv: 1706.08502},
	language     = {en}
}

@book{laka1996brief,
	title        = {A brief grammar of Euskara, the Basque language},
	author       = {Laka Mugarza, Itziar},
	year         = {1996},
	publisher    = {Universidad del Pa{\'\i}s Vasco, Euskal Herriko Unibertsitatea, Euskarazko~…}
}

@article{lake_compositional_nodate,
	title        = {Compositional generalization through meta sequence-to-sequence learning},
	author       = {Lake, Brenden M},
	year         = {2019},
	journal      = {arXiv preprint arXiv:1906.05381},
	pages        = {12},
	language     = {en}
}

@article{lake_generalization_2018,
	title        = {Generalization without systematicity: {On} the compositional skills of sequence-to-sequence recurrent networks},
	shorttitle   = {Generalization without systematicity},
	author       = {Lake, Brenden M. and Baroni, Marco},
	year         = {2018},
	month        = jun,
	journal      = {arXiv:1711.00350 [cs]},
	booktitle    = {International conference on machine learning},
	pages        = {10},
	url          = {http://arxiv.org/abs/1711.00350},
	urldate      = {2020-06-19},
	note         = {arXiv: 1711.00350},
	language     = {en},
	organization = {PMLR}
}

@article{lazaridou_emergence_2018,
	title        = {Emergence of {Linguistic} {Communication} from {Referential} {Games} with {Symbolic} and {Pixel} {Input}},
	author       = {Lazaridou, Angeliki and Hermann, Karl Moritz and Tuyls, Karl and Clark, Stephen},
	year         = {2018},
	month        = apr,
	journal      = {arXiv:1804.03984 [cs]},
	url          = {http://arxiv.org/abs/1804.03984},
	urldate      = {2022-01-17},
	note         = {arXiv: 1804.03984},
	language     = {en}
}

@article{lazaridou_emergent_2020,
	title        = {Emergent {Multi}-{Agent} {Communication} in the {Deep} {Learning} {Era}},
	author       = {Lazaridou, Angeliki and Baroni, Marco},
	year         = {2020},
	month        = jul,
	journal      = {arXiv:2006.02419 [cs]},
	url          = {http://arxiv.org/abs/2006.02419},
	urldate      = {2022-01-17},
	note         = {arXiv: 2006.02419},
	language     = {en}
}

@article{lazaridou_multi-agent_2017,
	title        = {{MULTI}-{AGENT} {COOPERATION} {AND} {THE} {EMERGENCE} {OF} ({NATURAL}) {LANGUAGE}},
	author       = {Lazaridou, Angeliki and Peysakhovich, Alexander and Baroni, Marco},
	year         = {2017},
	pages        = {11},
	language     = {en}
}

@article{lecun1989backpropagation,
	title        = {Backpropagation applied to handwritten zip code recognition},
	author       = {LeCun, Yann and Boser, Bernhard and Denker, John S and Henderson, Donnie and Howard, Richard E and Hubbard, Wayne and Jackel, Lawrence D},
	year         = {1989},
	journal      = {Neural computation},
	publisher    = {MIT Press},
	volume       = {1},
	number       = {4},
	pages        = {541--551}
}

@article{levina2004maximum,
	title        = {Maximum likelihood estimation of intrinsic dimension},
	author       = {Levina, Elizaveta and Bickel, Peter},
	year         = {2004},
	journal      = {Advances in neural information processing systems},
	volume       = {17}
}

@book{lewis2008convention,
	title        = {Convention: A philosophical study},
	author       = {Lewis, David},
	year         = {1970},
	publisher    = {John Wiley \& Sons}
}

@article{lezon2006using,
	title        = {Using the principle of entropy maximization to infer genetic interaction networks from gene expression patterns},
	author       = {Lezon, Timothy R and Banavar, Jayanth R and Cieplak, Marek and Maritan, Amos and Fedoroff, Nina V},
	year         = {2006},
	journal      = {Proceedings of the National Academy of Sciences},
	publisher    = {National Acad Sciences},
	volume       = {103},
	number       = {50},
	pages        = {19033--19038}
}

@misc{li_slog_2023,
	title        = {{SLOG}: {A} {Structural} {Generalization} {Benchmark} for {Semantic} {Parsing}},
	shorttitle   = {{SLOG}},
	author       = {Li, Bingzhi and Donatelli, Lucia and Koller, Alexander and Linzen, Tal and Yao, Yuekun and Kim, Najoung},
	year         = {2023},
	month        = oct,
	publisher    = {arXiv},
	url          = {http://arxiv.org/abs/2310.15040},
	urldate      = {2024-01-28},
	note         = {arXiv:2310.15040 [cs]}
}

@article{li2019compositional,
	title        = {Compositional generalization for primitive substitutions},
	author       = {Li, Yuanpeng and Zhao, Liang and Wang, Jianyu and Hestness, Joel},
	year         = {2019},
	journal      = {arXiv preprint arXiv:1910.02612}
}

@article{lieder_resource-rational_2020,
	title        = {Resource-rational analysis: {Understanding} human cognition as the optimal use of limited computational resources},
	shorttitle   = {Resource-rational analysis},
	author       = {Lieder, Falk and Griffiths, Thomas L.},
	year         = {2020},
	journal      = {Behavioral and Brain Sciences},
	publisher    = {Cambridge University Press},
	volume       = {43},
	pages        = {e1},
	doi          = {10.1017/S0140525X1900061X},
	issn         = {0140-525X, 1469-1825},
	url          = {https://www.cambridge.org/core/product/identifier/S0140525X1900061X/type/journal_article},
	urldate      = {2022-09-07},
	language     = {en}
}

@article{liu2019roberta,
	title        = {Roberta: A robustly optimized bert pretraining approach},
	author       = {Liu, Yinhan},
	year         = {2019},
	journal      = {arXiv preprint arXiv:1907.11692}
}

@inproceedings{locatello_challenging_2019,
	title        = {Challenging {Common} {Assumptions} in the {Unsupervised} {Learning} of {Disentangled} {Representations}},
	author       = {Locatello, Francesco and Bauer, Stefan and Lucic, Mario and Raetsch, Gunnar and Gelly, Sylvain and Schölkopf, Bernhard and Bachem, Olivier},
	year         = {2019},
	month        = may,
	booktitle    = {Proceedings of the 36th {International} {Conference} on {Machine} {Learning}},
	publisher    = {PMLR},
	pages        = {4114--4124},
	url          = {https://proceedings.mlr.press/v97/locatello19a.html},
	urldate      = {2024-01-31},
	note         = {ISSN: 2640-3498},
	language     = {en},
	organization = {PMLR}
}

@article{lodhi2002text,
	title        = {Text classification using string kernels},
	author       = {Lodhi, Huma and Saunders, Craig and Shawe-Taylor, John and Cristianini, Nello and Watkins, Chris},
	year         = {2002},
	journal      = {Journal of Machine Learning Research},
	volume       = {2},
	number       = {Feb},
	pages        = {419--444}
}

@misc{lu_expressiveness_2022,
	title        = {Expressiveness and {Learnability}: {A} {Unifying} {View} for {Evaluating} {Self}-{Supervised} {Learning}},
	shorttitle   = {Expressiveness and {Learnability}},
	author       = {Lu, Yuchen and Liu, Zhen and Baratin, Aristide and Laroche, Romain and Courville, Aaron and Sordoni, Alessandro},
	year         = {2022},
	month        = jun,
	publisher    = {arXiv},
	url          = {http://arxiv.org/abs/2206.01251},
	urldate      = {2023-02-14},
	note         = {arXiv:2206.01251 [cs]},
	language     = {en}
}

@article{lupyan2010language,
	title        = {Language structure is partly determined by social structure},
	author       = {Lupyan, Gary and Dale, Rick},
	year         = {2010},
	journal      = {PloS one},
	publisher    = {Public Library of Science San Francisco, USA},
	volume       = {5},
	number       = {1},
	pages        = {e8559}
}

@article{MacKay2004InformationTI,
	title        = {Information Theory, Inference, and Learning Algorithms},
	author       = {David John Cameron MacKay},
	year         = {2004},
	journal      = {IEEE Transactions on Information Theory},
	volume       = {50},
	pages        = {2544--2545}
}

@article{martins2019language,
	title        = {Language evolution and complexity considerations: The no half-Merge fallacy},
	author       = {Martins, Pedro Tiago and Boeckx, Cedric},
	year         = {2019},
	journal      = {PLoS biology},
	publisher    = {Public Library of Science San Francisco, CA USA},
	volume       = {17},
	number       = {11},
	pages        = {e3000389}
}

@article{marvin_targeted_2018,
	title        = {Targeted {Syntactic} {Evaluation} of {Language} {Models}},
	author       = {Marvin, Rebecca and Linzen, Tal},
	year         = {2018},
	month        = aug,
	journal      = {arXiv:1808.09031 [cs]},
	booktitle    = {Proceedings of the 2018 {Conference} on {Empirical} {Methods} in {Natural} {Language} {Processing}},
	publisher    = {Association for Computational Linguistics},
	address      = {Brussels, Belgium},
	pages        = {1192--1202},
	doi          = {10.18653/v1/D18-1151},
	url          = {http://arxiv.org/abs/1808.09031},
	urldate      = {2020-06-22},
	note         = {arXiv: 1808.09031},
	language     = {en}
}

@article{mccoy2019right,
	title        = {Right for the Wrong Reasons: Diagnosing Syntactic Heuristics in Natural Language Inference},
	author       = {McCoy, RT},
	year         = {2019},
	journal      = {arXiv preprint arXiv:1902.01007}
}

@article{merrill2023tale,
	title        = {A tale of two circuits: Grokking as competition of sparse and dense subnetworks},
	author       = {Merrill, William and Tsilivis, Nikolaos and Shukla, Aman},
	year         = {2023},
	journal      = {arXiv preprint arXiv:2303.11873}
}

@article{mikolov2013efficient,
	title        = {Efficient estimation of word representations in vector space},
	author       = {Mikolov, Tomas},
	year         = {2013},
	journal      = {arXiv preprint arXiv:1301.3781}
}

@article{Miller1955NoteOT,
  title={Note on the bias of information estimates},
  author={Miller, George},
  journal={Information theory in psychology: Problems and methods},
  year={1955},
  publisher={Free Press}
}

@article{mordatch_emergence_2018,
	title        = {Emergence of {Grounded} {Compositional} {Language} in {Multi}-{Agent} {Populations}},
	author       = {Mordatch, Igor and Abbeel, Pieter},
	year         = {2018},
	month        = jul,
	journal      = {arXiv:1703.04908 [cs]},
	doi          = {10.1007/BF00341314},
	issn         = {0029-8549},
	url          = {http://arxiv.org/abs/1703.04908},
	urldate      = {2020-01-29},
	note         = {arXiv: 1703.04908},
	language     = {en},
	archiveprefix = {arXiv},
	arxivid      = {1703.04908},
	eprint       = {1703.04908}
}

@article{moschitti_making_nodate,
	title        = {Making {Tree} {Kernels} practical for {Natural} {Language} {Learning}},
	author       = {Moschitti, Alessandro},
	pages        = {8},
	language     = {en}
}

@article{nakkiran2021deep,
	title        = {Deep double descent: Where bigger models and more data hurt},
	author       = {Nakkiran, Preetum and Kaplun, Gal and Bansal, Yamini and Yang, Tristan and Barak, Boaz and Sutskever, Ilya},
	year         = {2021},
	journal      = {Journal of Statistical Mechanics: Theory and Experiment},
	publisher    = {IOP Publishing},
	volume       = {2021},
	number       = {12},
	pages        = {124003}
}

@article{newport_maturational_1990,
	title        = {Maturational constraints on language learning},
	author       = {Newport, Elissa L.},
	year         = {1990},
	journal      = {Cognitive Science},
	volume       = {14},
	number       = {1},
	pages        = {11--28},
	doi          = {https://doi.org/10.1016/0364-0213(90)90024-Q},
	issn         = {0364-0213},
	url          = {https://www.sciencedirect.com/science/article/pii/036402139090024Q}
}

@article{nowak_evolution_2000,
	title        = {The evolution of syntactic communication},
	author       = {Nowak, Martin A. and Plotkin, Joshua B. and Jansen, Vincent A. A.},
	year         = {2000},
	month        = mar,
	journal      = {Nature},
	volume       = {404},
	number       = {6777},
	pages        = {495--498},
	doi          = {10.1038/35006635},
	issn         = {0028-0836, 1476-4687},
	url          = {http://www.nature.com/articles/35006635},
	urldate      = {2020-01-29},
	language     = {en}
}

@book{odonnell_productivity_2015,
	title        = {Productivity and {Reuse} in {Language}: {A} {Theory} of {Linguistic} {Computation} and {Storage}},
	shorttitle   = {Productivity and {Reuse} in {Language}},
	author       = {O'Donnell, Timothy J.},
	year         = {2015},
	publisher    = {The MIT Press},
	doi          = {10.7551/mitpress/9780262028844.001.0001},
	isbn         = {978-0-262-32680-3},
	url          = {https://direct.mit.edu/books/book/4024},
	urldate      = {2022-09-09},
	language     = {en}
}

@article{paninski_estimation_2003,
	title        = {Estimation of {Entropy} and {Mutual} {Information}},
	author       = {Paninski, Liam},
	year         = {2003},
	month        = jun,
	journal      = {Neural Computation},
	volume       = {15},
	number       = {6},
	pages        = {1191--1253},
	doi          = {10.1162/089976603321780272},
	issn         = {0899-7667, 1530-888X},
	url          = {https://direct.mit.edu/neco/article/15/6/1191-1253/6731},
	urldate      = {2024-01-23},
	language     = {en}
}

@article{partee1995lexical,
	title        = {Lexical semantics and compositionality},
	author       = {Partee, Barbara and others},
	year         = {1995},
	journal      = {An invitation to cognitive science: Language},
	publisher    = {MIT Press Cambridge, MA:},
	volume       = {1},
	pages        = {311--360}
}

@article{parzen1962estimation,
	title        = {On estimation of a probability density function and mode},
	author       = {Parzen, Emanuel},
	year         = {1962},
	journal      = {The annals of mathematical statistics},
	publisher    = {JSTOR},
	volume       = {33},
	number       = {3},
	pages        = {1065--1076}
}

@article{paszke2019pytorch,
	title        = {Pytorch: An imperative style, high-performance deep learning library},
	author       = {Paszke, Adam and Gross, Sam and Massa, Francisco and Lerer, Adam and Bradbury, James and Chanan, Gregory and Killeen, Trevor and Lin, Zeming and Gimelshein, Natalia and Antiga, Luca and others},
	year         = {2019},
	journal      = {Advances in neural information processing systems},
	volume       = {32}
}

@article{piantadosi_communicative_2012,
	title        = {The communicative function of ambiguity in language},
	author       = {Piantadosi, Steven T. and Tily, Harry and Gibson, Edward},
	year         = {2012},
	month        = mar,
	journal      = {Cognition},
	volume       = {122},
	number       = {3},
	pages        = {280--291},
	doi          = {10.1016/j.cognition.2011.10.004},
	issn         = {00100277},
	url          = {https://linkinghub.elsevier.com/retrieve/pii/S0010027711002496},
	urldate      = {2023-01-29},
	language     = {en}
}

@article{pimentel2020information,
	title        = {Information-theoretic probing for linguistic structure},
	author       = {Pimentel, Tiago and Valvoda, Josef and Maudslay, Rowan Hall and Zmigrod, Ran and Williams, Adina and Cotterell, Ryan},
	year         = {2020},
	journal      = {arXiv preprint arXiv:2004.03061}
}

@article{power2022grokking,
	title        = {Grokking: Generalization beyond overfitting on small algorithmic datasets},
	author       = {Power, Alethea and Burda, Yuri and Edwards, Harri and Babuschkin, Igor and Misra, Vedant},
	year         = {2022},
	journal      = {arXiv preprint arXiv:2201.02177}
}

@article{raffel2019exploring,
	title        = {Exploring the limits of transfer learning with a unified text-to-text transformer},
	author       = {Raffel, Colin and Shazeer, Noam and Roberts, Adam and Lee, Katherine and Narang, Sharan and Matena, Michael and Zhou, Yanqi and Li, Wei and Liu, Peter J},
	year         = {2019},
	journal      = {arXiv preprint arXiv:1910.10683}
}

@article{raffel2020exploring,
	title        = {Exploring the limits of transfer learning with a unified text-to-text transformer},
	author       = {Raffel, Colin and Shazeer, Noam and Roberts, Adam and Lee, Katherine and Narang, Sharan and Matena, Michael and Zhou, Yanqi and Li, Wei and Liu, Peter J},
	year         = {2020},
	journal      = {Journal of machine learning research},
	volume       = {21},
	number       = {140},
	pages        = {1--67}
}

@article{ravi2016optimization,
	title        = {Optimization as a model for few-shot learning},
	author       = {Ravi, Sachin and Larochelle, Hugo},
	year         = {2016}
}

@article{Reali2009,
	title        = {{The evolution of frequency distributions: Relating regularization to inductive biases through iterated learning}},
	author       = {Reali, Florencia and Griffiths, Thomas L.},
	year         = {2009},
	journal      = {Cognition},
	publisher    = {Elsevier B.V.},
	volume       = {111},
	number       = {3},
	pages        = {317--328},
	doi          = {10.1016/j.cognition.2009.02.012},
	issn         = {00100277},
	url          = {http://dx.doi.org/10.1016/j.cognition.2009.02.012}
}

@article{resnick_capacity_2020,
	title        = {Capacity, {Bandwidth}, and {Compositionality} in {Emergent} {Language} {Learning}},
	author       = {Resnick, Cinjon and Gupta, Abhinav and Foerster, Jakob and Dai, Andrew M. and Cho, Kyunghyun},
	year         = {2020},
	month        = apr,
	journal      = {arXiv:1910.11424 [cs, stat]},
	url          = {http://arxiv.org/abs/1910.11424},
	urldate      = {2022-01-29},
	note         = {arXiv: 1910.11424},
	language     = {en}
}

@article{rumelhart1986learning,
	title        = {Learning representations by back-propagating errors},
	author       = {Rumelhart, David E and Hinton, Geoffrey E and Williams, Ronald J},
	year         = {1986},
	journal      = {nature},
	publisher    = {Nature Publishing Group UK London},
	volume       = {323},
	number       = {6088},
	pages        = {533--536}
}

@book{russell1912problems,
	title        = {The problems of philosophy},
	author       = {Russell, Bertrand},
	year         = {1912},
	publisher    = {OUP Oxford}
}

@article{russin2019compositional,
	title        = {Compositional generalization in a deep seq2seq model by separating syntax and semantics},
	author       = {Russin, Jake and Jo, Jason and O'Reilly, Randall C and Bengio, Yoshua},
	year         = {2019},
	journal      = {arXiv preprint arXiv:1904.09708}
}

@article{sajjadi2018assessing,
	title        = {Assessing generative models via precision and recall},
	author       = {Sajjadi, Mehdi SM and Bachem, Olivier and Lucic, Mario and Bousquet, Olivier and Gelly, Sylvain},
	year         = {2018},
	journal      = {Advances in neural information processing systems},
	volume       = {31}
}

@article{saussure_course_1916,
	title        = {Course in general linguistics},
	author       = {de Saussure, Ferdinand},
	year         = {1916},
	pages        = {264},
	language     = {en}
}

@article{saxe2019information,
	title        = {On the information bottleneck theory of deep learning},
	author       = {Saxe, Andrew M and Bansal, Yamini and Dapello, Joel and Advani, Madhu and Kolchinsky, Artemy and Tracey, Brendan D and Cox, David D},
	year         = {2019},
	journal      = {Journal of Statistical Mechanics: Theory and Experiment},
	publisher    = {IOP Publishing},
	volume       = {2019},
	number       = {12},
	pages        = {124020}
}

@article{schneider2010brief,
	title        = {A brief review of molecular information theory},
	author       = {Schneider, Thomas D},
	year         = {2010},
	journal      = {Nano communication networks},
	publisher    = {Elsevier},
	volume       = {1},
	number       = {3},
	pages        = {173--180}
}

@article{sellam_multiberts_2022,
	title        = {{THE} {MULTIBERTS}: {BERT} {REPRODUCTIONS} {FOR} {ROBUSTNESS} {ANALYSIS}},
	author       = {Sellam, Thibault and Yadlowsky, Steve and Tenney, Ian and Wei, Jason and Saphra, Naomi and D’Amour, Alexander and Linzen, Tal and Bastings, Jasmijn and Turc, Iulia and Eisenstein, Jacob and Das, Dipanjan and Pavlick, Ellie},
	year         = {2022},
	journal      = {arXiv preprint arXiv:2106.16163},
	language     = {en}
}

@inproceedings{senghas1997argument,
	title        = {Argument structure in Nicaraguan Sign Language: The emergence of grammatical devices},
	author       = {Senghas, Ann and Coppola, Marie and Newport, Elissa L and Supalla, Ted},
	year         = {1997},
	booktitle    = {Proceedings of the Boston university conference on language development},
	volume       = {21},
	number       = {2},
	pages        = {550--61},
	organization = {Cascadilla Press Somerville, MA}
}

@article{shannon_communication_1949,
	title        = {Communication in the {Presence} of {Noise}},
	author       = {Shannon, C.E.},
	year         = {1949},
	month        = jan,
	journal      = {Proceedings of the IRE},
	volume       = {37},
	number       = {1},
	pages        = {10--21},
	doi          = {10.1109/JRPROC.1949.232969},
	issn         = {0096-8390},
	url          = {http://ieeexplore.ieee.org/document/1697831/},
	urldate      = {2024-01-28},
	language     = {en}
}

@article{Shannon1948AMT,
	title        = {A mathematical theory of communication},
	author       = {Shannon, C.E.},
	year         = {1948},
	journal      = {Bell Syst. Tech. J.},
	volume       = {27},
	pages        = {623--656},
	url          = {https://api.semanticscholar.org/CorpusID:55379485}
}

@article{shaw_compositional_2020,
	title        = {Compositional {Generalization} and {Natural} {Language} {Variation}: {Can} a {Semantic} {Parsing} {Approach} {Handle} {Both}?},
	shorttitle   = {Compositional {Generalization} and {Natural} {Language} {Variation}},
	author       = {Shaw, Peter and Chang, Ming-Wei and Pasupat, Panupong and Toutanova, Kristina},
	year         = {2020},
	month        = oct,
	journal      = {arXiv:2010.12725 [cs]},
	url          = {http://arxiv.org/abs/2010.12725},
	urldate      = {2020-12-18},
	note         = {arXiv: 2010.12725},
	language     = {en}
}

@inproceedings{li2018learning,
  title={Learning to generalize: Meta-learning for domain generalization},
  author={Li, Da and Yang, Yongxin and Song, Yi-Zhe and Hospedales, Timothy},
  booktitle={Proceedings of the AAAI conference on artificial intelligence},
  volume={32},
  number={1},
  year={2018}
}

@inproceedings{shi-etal-2016-string,
	title        = {Does String-Based Neural {MT} Learn Source Syntax?},
	author       = {Shi, Xing  and Padhi, Inkit  and Knight, Kevin},
	year         = {2016},
	month        = nov,
	booktitle    = {Proceedings of the 2016 Conference on Empirical Methods in Natural Language Processing},
	publisher    = {Association for Computational Linguistics},
	address      = {Austin, Texas},
	pages        = {1526--1534},
	doi          = {10.18653/v1/D16-1159},
	url          = {https://aclanthology.org/D16-1159},
	editor       = {Su, Jian  and Duh, Kevin  and Carreras, Xavier}
}

@misc{shwartz-ziv_opening_2017,
	title        = {Opening the {Black} {Box} of {Deep} {Neural} {Networks} via {Information}},
	author       = {Shwartz-Ziv, Ravid and Tishby, Naftali},
	year         = {2017},
	month        = apr,
	journal      = {arXiv preprint arXiv:1703.00810},
	publisher    = {arXiv},
	url          = {http://arxiv.org/abs/1703.00810},
	urldate      = {2023-01-29},
	note         = {arXiv:1703.00810 [cs]},
	language     = {en}
}

@article{smith_cultural_2008,
	title        = {Cultural evolution: implications for understanding the human language faculty and its evolution},
	shorttitle   = {Cultural evolution},
	author       = {Smith, Kenny and Kirby, Simon},
	year         = {2008},
	journal      = {Philosophical Transactions of the Royal Society B: Biological Sciences},
	publisher    = {The Royal Society London},
	volume       = {363},
	number       = {1509},
	pages        = {3591--3603},
	doi          = {10.1098/rstb.2008.0145},
	url          = {https://royalsocietypublishing.org/doi/10.1098/rstb.2008.0145},
	urldate      = {2022-01-17},
	language     = {en},
	archiveprefix = {arXiv},
	arxivid      = {arXiv:1011.1669v3},
	eprint       = {arXiv:1011.1669v3},
	pmid         = {18801718}
}

@article{smith_eliminating_2010,
	title        = {Eliminating unpredictable variation through iterated learning},
	author       = {Smith, Kenny and Wonnacott, Elizabeth},
	year         = {2010},
	month        = sep,
	journal      = {Cognition},
	publisher    = {Elsevier},
	volume       = {116},
	number       = {3},
	pages        = {444--449},
	doi          = {10.1016/j.cognition.2010.06.004},
	issn         = {00100277},
	url          = {https://linkinghub.elsevier.com/retrieve/pii/S0010027710001320},
	urldate      = {2020-01-29},
	language     = {en}
}

@article{smith_learning_2011,
	title        = {Learning {Bias}, {Cultural} {Evolution} of {Language}, and the {Biological} {Evolution} of the {Language} {Faculty}},
	author       = {Smith, Kenny},
	year         = {2011},
	month        = apr,
	journal      = {Human Biology},
	volume       = {83},
	number       = {2},
	pages        = {261--278},
	doi          = {10.3378/027.083.0207},
	issn         = {0018-7143, 1534-6617},
	url          = {http://www.bioone.org/doi/abs/10.3378/027.083.0207},
	urldate      = {2018-05-02},
	language     = {en}
}

@article{smith2003complex,
	title        = {Complex systems in language evolution: the cultural emergence of compositional structure},
	author       = {Smith, Kenny and Brighton, Henry and Kirby, Simon},
	year         = {2003},
	journal      = {Advances in complex systems},
	publisher    = {World Scientific},
	volume       = {6},
	number       = {04},
	pages        = {537--558}
}

@article{smolensky_tensor_1990,
	title        = {Tensor product variable binding and the representation of symbolic structures in connectionist systems},
	author       = {Smolensky, Paul},
	year         = {1990},
	month        = nov,
	journal      = {Artificial Intelligence},
	publisher    = {Elsevier},
	volume       = {46},
	number       = {1-2},
	pages        = {159--216},
	doi          = {10.1016/0004-3702(90)90007-M},
	issn         = {00043702},
	url          = {https://linkinghub.elsevier.com/retrieve/pii/000437029090007M},
	urldate      = {2022-02-10},
	language     = {en}
}

@book{stevenson2010oxford,
	title        = {Oxford dictionary of English},
	author       = {Stevenson, Angus},
	year         = {2010},
	publisher    = {Oxford University Press, USA}
}

@article{sutskever2014sequence,
	title        = {Sequence to sequence learning with neural networks},
	author       = {Sutskever, Ilya and Vinyals, Oriol and Le, Quoc V},
	year         = {2014},
	journal      = {arXiv preprint arXiv:1409.3215}
}

@article{symons2014systematicity,
	title        = {Systematicity: an overview},
	author       = {Symons, John and Calvo, Paco},
	year         = {2014}
}

@misc{tishby_deep_2015,
	title        = {Deep {Learning} and the {Information} {Bottleneck} {Principle}},
	author       = {Tishby, Naftali and Zaslavsky, Noga},
	year         = {2015},
	month        = mar,
	booktitle    = {2015 ieee information theory workshop (itw)},
	publisher    = {arXiv},
	pages        = {1--5},
	url          = {http://arxiv.org/abs/1503.02406},
	urldate      = {2023-01-29},
	note         = {arXiv:1503.02406 [cs]},
	language     = {en},
	organization = {IEEE}
}

@book{tomasello2005constructing,
	title        = {Constructing a language: A usage-based theory of language acquisition},
	author       = {Tomasello, Michael},
	year         = {2005},
	publisher    = {Harvard university press}
}

@article{turc2019well,
	title        = {Well-read students learn better: On the importance of pre-training compact models},
	author       = {Turc, Iulia and Chang, Ming-Wei and Lee, Kenton and Toutanova, Kristina},
	year         = {2019},
	journal      = {arXiv preprint arXiv:1908.08962}
}

@article{vaswani_attention_2017,
	title        = {Attention is {All} you {Need}},
	author       = {Vaswani, Ashish and Shazeer, Noam and Parmar, Niki and Uszkoreit, Jakob and Jones, Llion and Gomez, Aidan N and Kaiser, Łukasz and Polosukhin, Illia},
	year         = {2017},
	journal      = {Advances in neural information processing systems},
	volume       = {30},
	pages        = {11},
	language     = {en}
}

@article{veldhoen_diagnostic_2016,
	title        = {Diagnostic classiﬁers: revealing how neural networks process hierarchical structure},
	author       = {Veldhoen, Sara and Hupkes, Dieuwke and Zuidema, Willem},
	year         = {2016},
	pages        = {10},
	language     = {en}
}

@article{vinga2014information,
	title        = {Information theory applications for biological sequence analysis},
	author       = {Vinga, Susana},
	year         = {2014},
	journal      = {Briefings in bioinformatics},
	publisher    = {Oxford University Press},
	volume       = {15},
	number       = {3},
	pages        = {376--389}
}

@inproceedings{vinyals2016matching,
	title        = {Matching networks for one shot learning},
	author       = {Vinyals, Oriol and Blundell, Charles and Lillicrap, Timothy and Wierstra, Daan and others},
	year         = {2016},
	booktitle    = {Advances in neural information processing systems},
	pages        = {3630--3638}
}

@misc{voita_information-theoretic_2020,
	title        = {Information-{Theoretic} {Probing} with {Minimum} {Description} {Length}},
	author       = {Voita, Elena and Titov, Ivan},
	year         = {2020},
	month        = mar,
	journal      = {arXiv preprint arXiv:2003.12298},
	publisher    = {arXiv},
	url          = {http://arxiv.org/abs/2003.12298},
	urldate      = {2023-11-21},
	note         = {arXiv:2003.12298 [cs]}
}

@misc{voita_language_2021,
	title        = {Language {Modeling}, {Lexical} {Translation}, {Reordering}: {The} {Training} {Process} of {NMT} through the {Lens} of {Classical} {SMT}},
	shorttitle   = {Language {Modeling}, {Lexical} {Translation}, {Reordering}},
	author       = {Voita, Elena and Sennrich, Rico and Titov, Ivan},
	year         = {2021},
	month        = sep,
	publisher    = {arXiv},
	url          = {http://arxiv.org/abs/2109.01396},
	urldate      = {2023-10-30},
	note         = {arXiv:2109.01396 [cs]}
}

@book{von1863humboldt,
	title        = {Humboldt:'On language': On the diversity of human language construction and its influence on the mental development of the human species},
	author       = {Von Humboldt, Wilhelm},
	year         = {1863},
	publisher    = {Cambridge University Press}
}

@article{luong2015effective,
  title={Effective approaches to attention-based neural machine translation},
  author={Luong, Minh-Thang and Pham, Hieu and Manning, Christopher D},
  journal={arXiv preprint arXiv:1508.04025},
  year={2015}
}

@inproceedings{obamuyide2019model,
  title={Model-agnostic meta-learning for relation classification with limited supervision},
  author={Obamuyide, Abiola and Vlachos, Andreas},
  year={2019},
  organization={Association for Computational Linguistics}
}

@article{gu2018meta,
  title={Meta-learning for low-resource neural machine translation},
  author={Gu, Jiatao and Wang, Yong and Chen, Yun and Cho, Kyunghyun and Li, Victor OK},
  journal={arXiv preprint arXiv:1808.08437},
  year={2018}
}

@article{jia2016data,
  title={Data recombination for neural semantic parsing},
  author={Jia, Robin and Liang, Percy},
  journal={arXiv preprint arXiv:1606.03622},
  year={2016}
}

@article{wang2019rat,
  title={Rat-sql: Relation-aware schema encoding and linking for text-to-sql parsers},
  author={Wang, Bailin and Shin, Richard and Liu, Xiaodong and Polozov, Oleksandr and Richardson, Matthew},
  journal={arXiv preprint arXiv:1911.04942},
  year={2019}
}

@article{wang2018glue,
	title        = {Glue: A multi-task benchmark and analysis platform for natural language understanding},
	author       = {Wang, Alex},
	year         = {2018},
	journal      = {arXiv preprint arXiv:1804.07461}
}

@article{wang2020meta,
	title        = {Meta-Learning for Domain Generalization in Semantic Parsing},
	author       = {Wang, Bailin and Lapata, Mirella and Titov, Ivan},
	year         = {2020},
	journal      = {arXiv preprint arXiv:2010.11988}
}

@article{warstadt_blimp_2019,
	title        = {{BLiMP}: {A} {Benchmark} of {Linguistic} {Minimal} {Pairs} for {English}},
	shorttitle   = {{BLiMP}},
	author       = {Warstadt, Alex and Parrish, Alicia and Liu, Haokun and Mohananey, Anhad and Peng, Wei and Wang, Sheng-Fu and Bowman, Samuel R.},
	year         = {2019},
	month        = dec,
	journal      = {arXiv:1912.00582 [cs]},
	url          = {http://arxiv.org/abs/1912.00582},
	urldate      = {2020-01-13},
	note         = {arXiv: 1912.00582},
	language     = {en}
}

@article{wehbe2021incremental,
	title        = {Incremental language comprehension difficulty predicts activity in the language network but not the multiple demand network},
	author       = {Wehbe, Leila and Blank, Idan Asher and Shain, Cory and Futrell, Richard and Levy, Roger and von der Malsburg, Titus and Smith, Nathaniel and Gibson, Edward and Fedorenko, Evelina},
	year         = {2021},
	journal      = {Cerebral Cortex},
	publisher    = {Oxford University Press},
	volume       = {31},
	number       = {9},
	pages        = {4006--4023}
}

@article{weinreich_empirical_1968,
	title        = {Empirical foundations for a theory of language change},
	author       = {Weinreich, Uriel and Labov, William and Herzog, Marvin},
	year         = {1968},
	journal      = {University of Texas Press},
	pages        = {100},
	language     = {en}
}

@article{williams1992simple,
	title        = {Simple statistical gradient-following algorithms for connectionist reinforcement learning},
	author       = {Williams, Ronald J},
	year         = {1992},
	journal      = {Machine learning},
	publisher    = {Springer},
	volume       = {8},
	number       = {3},
	pages        = {229--256}
}

@article{winters2018contextual,
	title        = {Contextual predictability shapes signal autonomy},
	author       = {Winters, James and Kirby, Simon and Smith, Kenny},
	year         = {2018},
	journal      = {Cognition},
	publisher    = {Elsevier},
	volume       = {176},
	pages        = {15--30}
}

@article{yurtsever2020survey,
	title        = {A survey of autonomous driving: Common practices and emerging technologies},
	author       = {Yurtsever, Ekim and Lambert, Jacob and Carballo, Alexander and Takeda, Kazuya},
	year         = {2020},
	journal      = {IEEE access},
	publisher    = {IEEE},
	volume       = {8},
	pages        = {58443--58469}
}

@article{zaslavsky_efficient_2018,
	title        = {Efficient compression in color naming and its evolution},
	author       = {Zaslavsky, Noga and Kemp, Charles and Regier, Terry and Tishby, Naftali},
	year         = {2018},
	month        = jul,
	journal      = {Proceedings of the National Academy of Sciences},
	volume       = {115},
	number       = {31},
	pages        = {7937--7942},
	doi          = {10.1073/pnas.1800521115},
	issn         = {0027-8424, 1091-6490},
	url          = {https://pnas.org/doi/full/10.1073/pnas.1800521115},
	urldate      = {2023-02-05},
	language     = {en}
}

@article{zech2018variable,
	title        = {Variable generalization performance of a deep learning model to detect pneumonia in chest radiographs: a cross-sectional study},
	author       = {Zech, John R and Badgeley, Marcus A and Liu, Manway and Costa, Anthony B and Titano, Joseph J and Oermann, Eric Karl},
	year         = {2018},
	journal      = {PLoS medicine},
	publisher    = {Public Library of Science San Francisco, CA USA},
	volume       = {15},
	number       = {11},
	pages        = {e1002683}
}

@article{zhang1989simple,
	title        = {Simple fast algorithms for the editing distance between trees and related problems},
	author       = {Zhang, Kaizhong and Shasha, Dennis},
	year         = {1989},
	journal      = {SIAM journal on computing},
	publisher    = {SIAM},
	volume       = {18},
	number       = {6},
	pages        = {1245--1262}
}

@inproceedings{zuidema_2002,
	title        = {How the Poverty of the Stimulus Solves the Poverty of the Stimulus},
	author       = {Zuidema, Willem},
	year         = {2002},
	booktitle    = {Advances in Neural Information Processing Systems},
	publisher    = {MIT Press},
	volume       = {15},
	pages        = {},
	url          = {https://proceedings.neurips.cc/paper_files/paper/2002/file/04ad5632029cbfbed8e136e5f6f7ddfa-Paper.pdf},
	editor       = {S. Becker and S. Thrun and K. Obermayer},
	language     = {en}
}
\appendix

\chapter{What We Talk About When We Talk about Compositionality}

\section{Full Formalisations}

For readability body text includes simple definitions for each measure, without aggregation to the language level. Below are more detailed equations that include aggreagation steps:

\begin{equation}
    Synonymy(\mathcal{L}) = \frac{1}{|R|}\sum_{r=1}^{|R|}\frac{1}{|A_r|}\sum_{i=1}^{|A_r|}\frac{min \Bigl[ \mathcal{H}(char_p|atom_{r,i}) : \forall p \in P \Bigl] }{\log (n_{chars})}
\end{equation}

\begin{equation}
    Homonomy(\mathcal{L})=\frac{1}{|R|}\sum_{r=1}^{|R|}\frac{1}{|C|}\sum_{j=1}^{|C|}\frac{ min \Bigl[\mathcal{H}(atom_{r} | char_{p,j}):\forall p \in P\Bigl]}{\log (n_{atoms})}
\end{equation}

\section{Rolling Mixed Effects Model Implementation}\label{appendix:mixed-effects}

We implement a rolling mixed effects model using the python statsmodels package. We fit a separate model for each of our 6 independent variables: synonymy, entanglement, freedom, homonymy, topsim, posids. The dependent variable across all of them is o.o.d. generalization performance.  For each model we include two random effects, random intercepts based on the seed used to initialize that run of the model, and random slopes for the epochs of training. This allows the model to account for variation between different models, given that some initializations outperform others.

The model is fit to a window of 100 epochs of training data at a time, at each step it fits a regression to 100 epochs of data for 20 initializations of the model. It then moves forward one epoch at a time (e.g. the first fit of the model is on epochs 0-100, the second on 1-101, the third 2-102, etc.). We plot the resulting regression coefficient (b value) obtained from each mixed effects model fit to each IV, at each window of data.

For reference the corresponding command to run this model with the LMER package in R (a standard method for fitting these kinds of models) is: lmer(ood\_acc $\sim$ IV + (1 + Epoch $|$ Seed)) where IV is one of the variation measures.

\section{O.O.D. Accuracy Vs. Variation Slices}
In addition to the regression analysis presented in the results section we show relational plots for two different epochs in training: one from mid way through and one from late in training near convergence. In line with the regression analysis a more linear relationship between o.o.d. performance and variation is visible earlier in training before the language becomes regular enough for the task. Entanglement in particular shows a steep negative relationship in the 100 epoch plot but is totally scattered by epoch 500. Were we to only assess the relationship between generalization and variation at the end of training we would could easily conclude in line with previous work that they were not meaningfully related. 

It's worth noting that the pattern here may not appear as salient as it appears in the rolling mixed effects model presented earlier, there are two major reasons for this: first the rolling model considers 100 epochs at a time, rather than a single slice with only 20 data points, providing it with 100 times the data visualized here by which to assess the relationship between variation and generalization. Secondly the rolling model has a random intercept based on the seed used in each run of the model. This is important because in line with other work on o.o.d. generalization we see a substantial effect of initialization on generalization performance, by including it as a random effect the rolling model can look at each seed separately to see if each seed's generalization performance over the run is related to its language's variation. So while in the visualizations below we may see some seeds which appear like outliers, the rolling model accounts for this, fitting a separate intercept for each run. 
\begin{figure}[H]
\centering
    \includegraphics[width=0.98\textwidth]{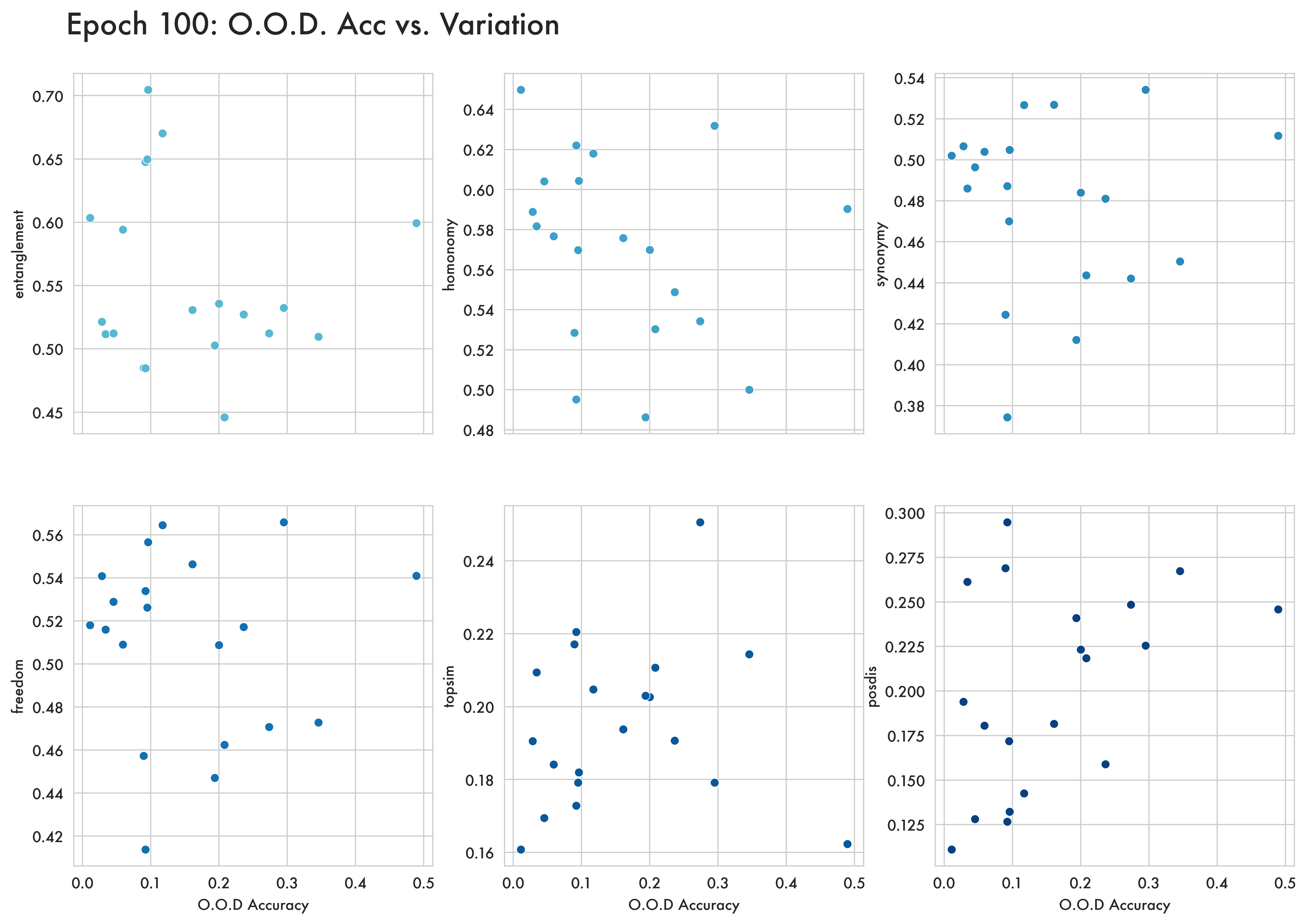}
        \caption{Plots show each of the variation measures plotted against o.o.d. accuracy at the 100th epoch of training.}
    \label{fig:epoch_relations}
\end{figure}
\begin{figure}[H]
\centering
    \includegraphics[width=0.98\textwidth]{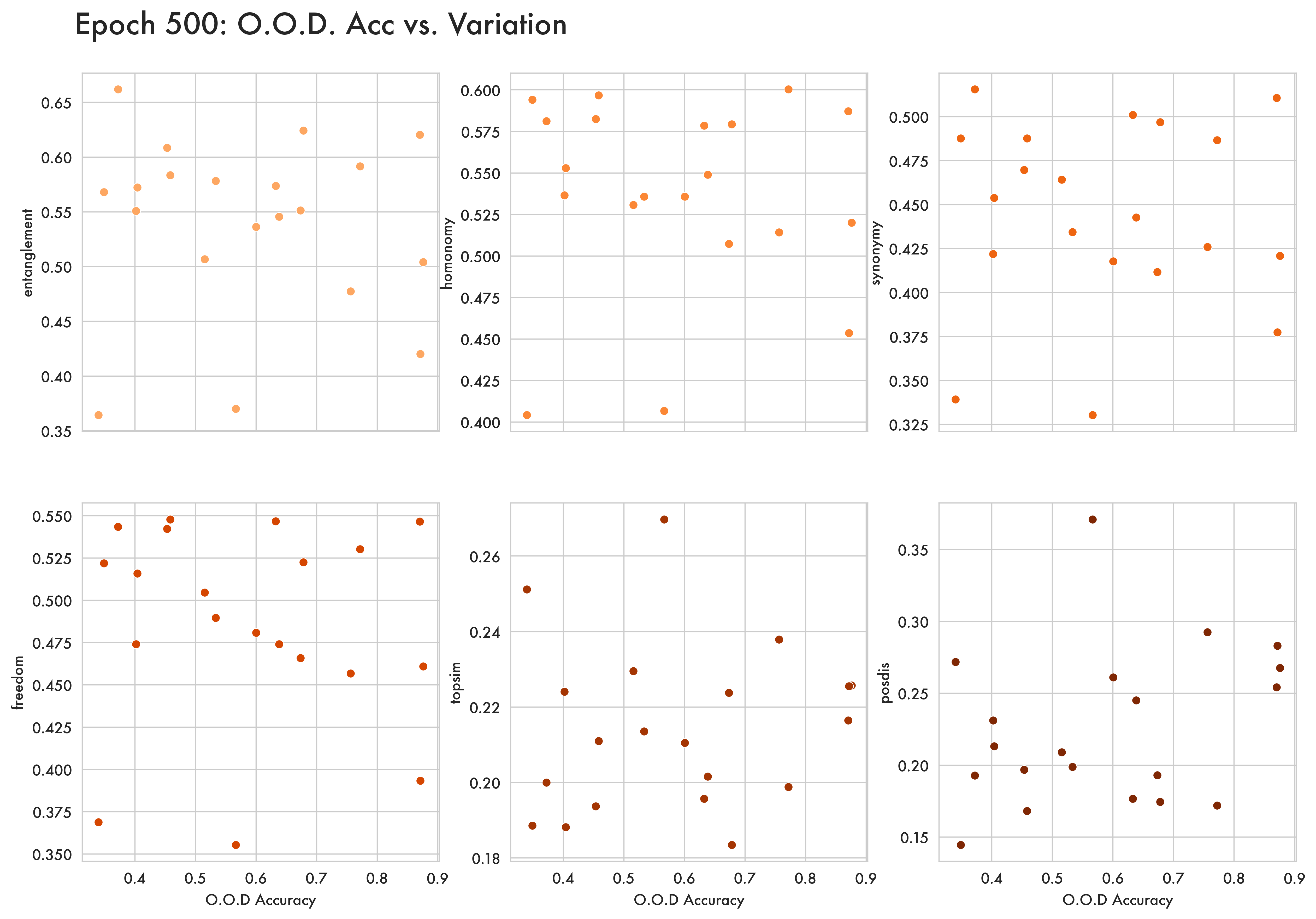}
    \caption{Plots show each of the variation measures plotted against o.o.d. accuracy at the 500th epoch of training.}
    \label{fig:epoch_relations_500}
\end{figure}

\section{O.O.D. Accuracy Vs. Variation Discussion}\label{appendix:oodregression-discussion}
As noted in section \ref{sec:results} we see a strong relationship between regularity and o.o.d. performance early in training but this effect goes away as the model converges. We attribute this to each run of the model decreasing the degree of variation in its language over time, resulting in a language sufficiently regular to succeed at the task. Where because all languages are sufficiently regular, whether one is slightly more regular than another doesn't necessarily result in better generalization performance. This dovetails with the overarching argument here that as seen in natural language even a high-degree of variation doesn't necessarily undermine a language's ability to generalize.

However it is worth noting that the relationship between regularity and generalization early on could be driven in part by more regular languages being easier for the listener to learn. \citet{chaabouni_compositionality_2020} observes that higher topsim languages are easier to acquire. By taking emergent languages from the end of training, and separately training a model to map between signals and meanings using supervised-learning they show higher topsim languages require less training to converge. Here it could be the case that early in training more regular languages are easier for the listener to learn, improving generalization performance early on and explaining in part the early correlation between regularity and generalisation. However, this is an emergent model and languages are not static throughout training meaning what the listener tries to learn is a `moving-target' changing at each step. For this reason framing emergent results in terms of results that look at the learnability of static languages would potentially seem to draw a false equivalency. It's unclear if the learnability of a language at timestep $n$ matters at $n+1$ when the language has changed. Given this, and the fact that we see all conditions decrease variation in the emergent language over time, we believe the best interpretation of our results is that regularity matters for generalization until the language becomes sufficiently regular for the task. Once sufficient regularity is reached greater regularity doesn't necessarily improve generalization performance so we see no correlation. Although it is possible learnability has some effect - further study of the role learnability plays in an emergent context, where what is learned changes, is needed to understand the full picture.

\section{Use of Equation 1 for all 4 measures of variation} \label{appendix:eq-usage}

We use equation \ref{eq:co_prob} (copied below in simplified notation) to calculate the conditional probabilities used in all 4 measures of variation:

\begin{equation}\label{eq:co_prob_appendix_1}
    \mathbb{P}(char | position, atom, role) = \frac{count(char, position, atom, role)}{count(position, atom, role)}
\end{equation}

This gives us a distribution over characters for each position, for each atom, in each role. This distribution is intuitively useful for estimating synonymy which can be seen as the entropy over characters in a position. 

\subsection{Freedom and Entanglement}
However it's natural to wonder why we use this distribution again when calculating measures of word order freedom and entanglement. Both of these measures refer to how likely it is that a given role is encoded in a position in the signal. Freedom looks at how consistently the atoms in a role are encoded in a position, while entanglement looks at how consistently any two roles are encoded in the same position. Intuitively we might want to calculate a distribution over positions given roles instead, like:

\begin{equation}\label{eq:co_prob_appendix_2}
    \mathbb{P}(position, char, atom | role) = \frac{count(position, char, atom, role)}{count(role)}
\end{equation}

then marginalize over characters and atoms so that we could directly estimate the probability of a position given a role $\mathbb{P}(position | role)$. The problem with this is that every signal has a character in every position, and every meaning has more than one role (i.e. subject, verb, and object, rather than just having a subject) meaning that the distribution over positions is always uniform. If we were to only marginalize over atoms, to get a distribution over characters and positions $\mathbb{P}(position, char | role)$ this is also nearly uniform, because different atoms are encoded using different characters. So marginalizing over atoms combines distinct distributions for each atom into a near-uniform one. Similarly if we only marginalize over characters to get a distribution over positions and atoms  $\mathbb{P}(position, atoms | role)$ because every signal has a character in every position the resulting distribution is also uniform.

Fundamentally the only relevant probability distribution which is consistently non-uniform is the one described in equation 1: $\mathbb{P}(char | position, atom, role)$. Even though every signal has a character in every position every character is not equally likely given a specific $position, atom, role$ combination. As a result this is the distribution we use to calculate measures of word order freedom and entanglement. In order to do so we first observe that in signal positions where an $atom, role$ combination is not encoded $\mathbb{P}(char | position, atom, role)$ is uniform as the distribution is not conditioned by the selected atom and role. Accordingly we take lower conditional entropy $\mathcal{H}(\mathbb{P}(char | position, atom, role))$ (used in equation \ref{eq:synonymy_pt_2}) as an indication that an $atom, role$ combination is more likely to be encoded in a position. By taking a mean of this conditional entropy across all atoms in a given role (described in equation \ref{eq:freedom_pt_1}) we can see if it is consistently low in the same position(s) of the signal for all atoms in that role - indicating adherence to a single word order. Equation \ref{eq:freedom_pt_2} then aggregates this across all roles.

Entanglement looks to see if there is consistent encoding of multiple roles in a single position. Seeing as we take low conditional entropy as an indication that an $atom, role$ combination is encoded in a given position we compare the mean from equation \ref{eq:freedom_pt_1} with means from other roles to see if they are consistently low in the same parts of the signal.

\subsection{Homonymy}

Given that homonymy assesses the probability that a letter in a position encodes each atom in a role, it is possible to look at this by estimating the distribution $\mathbb{P}(atom | char, position, role)$ directly. The same distribution can be calculated by instead taking the $\mathbb{P}(char | position, atom, role)$ distribution and re-normalizing it along the atom axis:

\begin{equation}\label{eq:appendix_homonymy}
    \mathbb{H}(char_{p,j}, r) =  \frac{\Bigl\{ \mathbb{P}(char_{p,j}|atom_{r,i}): \forall i \in A_r\Bigl\}}{ \sum_{i=1}^{|A_r|}\mathbb{P}(char_{p,j}|atom_{r,i})}
\end{equation}

We find empirically that this is equivalent to computing $\mathcal{H}(\mathbb{P}(atom | char, position, role))$ (see results in table \ref{table:homonomy}) with the only differences between the two resulting from small rounding errors. In table \ref{table:homonomy} we report results for both approaches to computing homonomy across model sizes to show their equivalency. Note these results are the means of 6 seeds so differ slightly from figures in the core results.
When introducing the measures in section 2.3 of these two approaches we opt for the re-normalization of $\mathbb{P}(char | position, atom, role)$ rather than computing a new probability distribution because we believe this makes the formulation of the homonymy measure more intuitively related to the others, and makes the visualizations in figure \ref{fig:synonymy} a direct reflection of how the measures are computed while producing equivalent results. 

\begin{table}[H]
\centering
\setlength{\tabcolsep}{6pt}
\resizebox{0.7\textwidth}{!}{
\begin{tabular}{lllllll}
\toprule
 \textbf{epoch} & ideal & random &  small & medium &   large  &\\
\toprule
\emph{homonymy} & $0.12$ & $0.99$ & $0.56\pm 0.14$ & $0.62\pm0.15$ & $0.72\pm 0.05$  \\
\emph{direct homonymy} & $0.12$ & $0.99$ & $0.56\pm 0.14$ & $0.62\pm 0.15$ & $0.72\pm 0.05$ \\
\bottomrule
\end{tabular}
}
\setlength{\tabcolsep}{6pt}
\caption{Homonymy refers to the method of computing homonymy used in the core results and described in equation \ref{eq:appendix_homonymy} while direct homonymy instead directly estimates the distribution $\mathbb{P}(atom | char, position, role)$. Results are the mean of 6 initializations at the best generalizing epoch, so values differ slightly from those in the main results which are the mean of 20.
}
\label{table:homonomy}
\end{table}

\section{Residual Entropy}\label{appendix:residual-entropy}
In addition to topsim and posdis reported in the main results we also report results from one other measure from previous work, residual entropy \citep{resnick_capacity_2020}. The results here follow the same pattern as the other measures of variation with larger models arriving and more irregular languages. Additionally all conditions increase the regularity of the emergent language over the course of training. Also shown is the correlation analysis between Residual entropy and O.O.D. performance, showing like other measures of variation residual entropy is a strong predictor early in training but that this effect goes away later on.
\begin{table}[H]
\centering
\setlength{\tabcolsep}{6pt}
\resizebox{0.7\textwidth}{!}{
\begin{tabular}{lllllll}
\toprule
 \textbf{epoch} & ideal & random &  small & medium &   large  &\\
\toprule
\emph{best} & \ac{0.061} &     \ac{0.625} & \ac{0.299}$\pm 0.16$ & \ac{0.378}$\pm0.12$ & \ac{0.465}$\pm 0.08$  \\
\emph{$\Delta$ o.o.d.} &     &     & \ac{0.523}$\pm 0.05$ & \ac{0.446}$\pm 0.08$ & \ac{0.237}$\pm 0.20$ \\
\bottomrule
\end{tabular}
}
\setlength{\tabcolsep}{6pt}
\caption{Residual entropy scores at the best generalizing epoch and the difference between the best generalizing epoch and one drawn from early in training. Results are the mean of 6 initializations.}
\label{table:resent}
\end{table}

\begin{figure}[h]
    \centering
    \includegraphics[width=\textwidth]{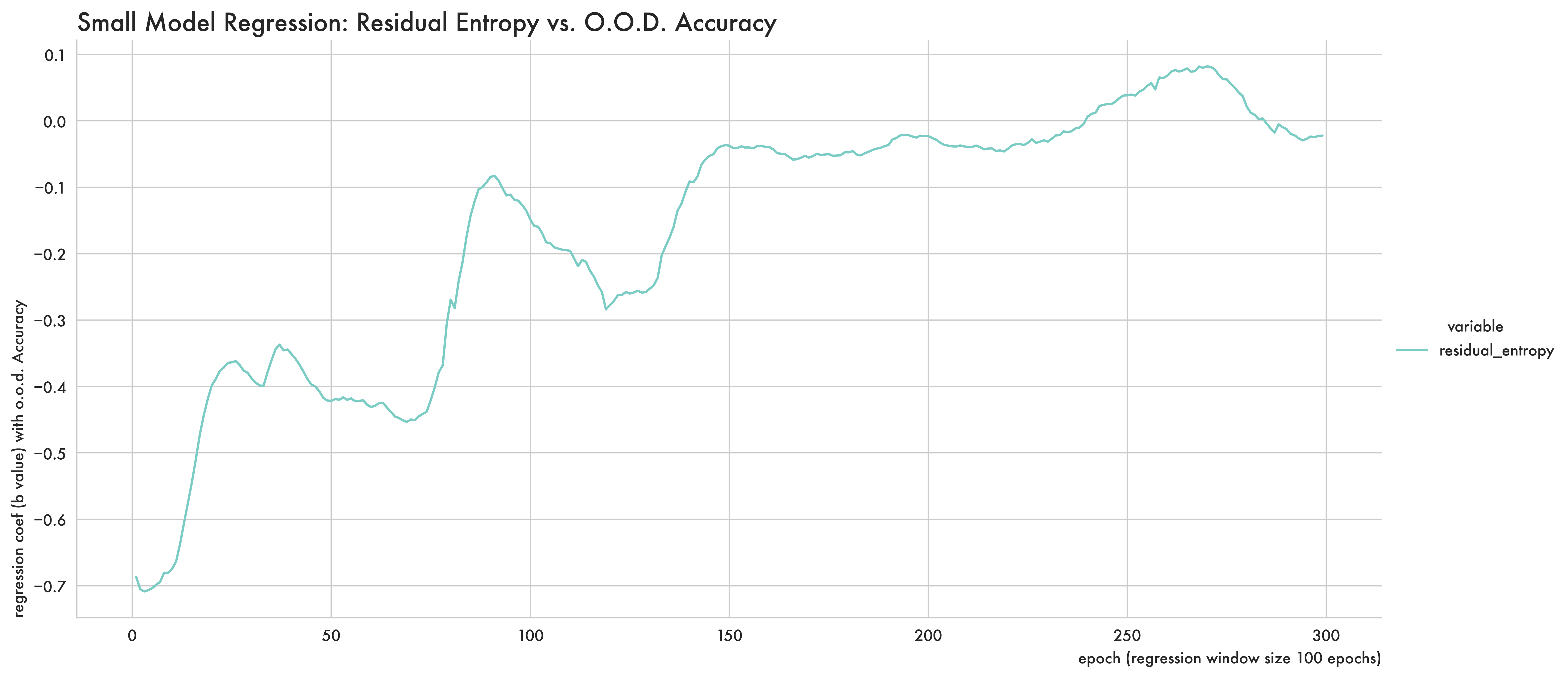}
    \caption{The rolling mixed effects model coefficients between Residual Entropy and o.o.d. generalization accuracy for the \msmall model for each window. 
    }
    \label{fig:resentcorrelations}
\end{figure}

\section{I.I.D. Correlation Results}\label{appendix:iid-correlation}
We also include the correlation results between the measures of variation and in-distribution generalization. The results follow a similar pattern with degree of regularity being a strong predictor of generalization performance early in training but this effect goes away as the emergent language becomes regular enough to generalize well.
Interestingly in-distribution and out-of-distribution correlations align almost exactly. This is reassuring in that it shows degree of regularity is important for generalization in general whether it is in or out of distribution.    

\begin{figure}[H]
    \centering
    \includegraphics[width=\textwidth]{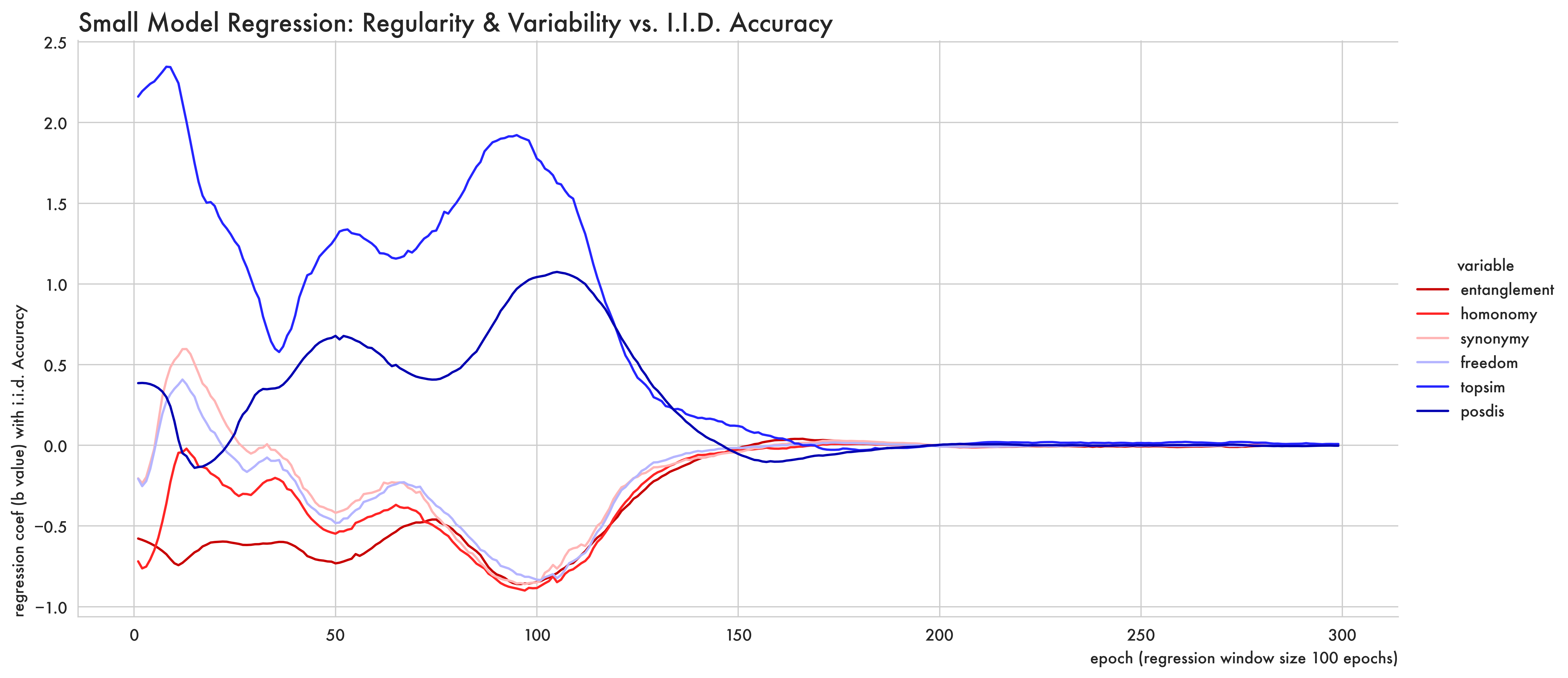}
    \caption{The model is fit to a window of data from 100 epochs at a time across 20 initializations. 
    The window slides forward one epoch at a time (i.e. epochs 0-100, 1-101 ...) and fits a different model between i.i.d. accuracy and each measure of variation for each window. Shown are the regression coefficients (b values) of our four measures of variation, and two previous measures of regularity (topsim and posdis) with o.o.d. generalization accuracy for the \msmall model for each window. 
    }
    \label{fig:iidcorrelations}
\end{figure}

\section{Significance Testing for Variation Differences}

\begin{table}[H]
\centering
\setlength{\tabcolsep}{6pt}
\resizebox{0.8\textwidth}{!}{
\begin{tabular}{llllllll}
\toprule
 \textbf{params} & synonymy & entanglement &  freedom & homonomy &   topsim &   posdis  &\\
\toprule
\emph{250 vs 500} & \ac{0.027519} &     \ac{0.104873} & \ac{0.057428} & \ac{0.062946} & \ac{0.256488} & \ac{0.099342} \\
\emph{250 vs 800} & $<0.0001$ &     $<0.0001$ & $<0.0001$ & $<0.0001$ & $<0.0001$ & $<0.0001$ \\
\emph{500 vs 800} & \ac{0.000098} &     $<0.0001$ & \ac{0.000177} & \ac{0.000136} & \ac{0.000482} & \ac{0.000080} \\
\bottomrule
\end{tabular}
}
\setlength{\tabcolsep}{6pt}
\caption{P Values obtained from a t-test comparing variation measures from different sized initialization. The difference between \mbig and \msmall, and \mbig and \mmid are significant. Of differences between \msmall and \mmid only synonymy and posdis are significant}
\label{table:significance}
\end{table}

\subsection{Hyperparameters}\label{appendix:hyperparams}
\begin{itemize}
    \item Recurrent Unit: GRU
    \item Hidden Size: 250, 500, 800
    \item Entropy Regularization Coefficient: (sender 0.5, receiver 0.0)
    \item Batch Size: 5000
    \item Learning Rate: 1e-3
    \item Signal Length: 6
    \item Character Inventory: 26
    \item Training Epochs: 800
    \item Embedding Size: 52
    \item Roles: 3
    \item Atoms: 25
    \item Optimizer: (Sender: Reinforce, Receiver: ADAM)
    
\end{itemize}

\chapter{Information Structure in LLMs}

\section{Benchmarking: Effect of Number of Heads, Mean and Scale}\label{appendix:soft_h_benchmark}

We compare versions of our estimator across different levels of subspacing. Angular entropy is the version that appears in the paper, discrete follows the methods of soft entropy estimation but then argmaxes to assign representations to a single point on the sphere. We also include a version that uses euclidean distances instead of the cosine-sphere comparison used in the paper. In principle this is nice because models also represent information topographically, encoding meaning in magnitude as well as angle in representational space. In practice euclidean distances end up being dramatically less memory efficient (and a factor of 4 slower to compute) than cosine similarities when using built-in pytorch methods. This means for scalability reasons we elected to only focus on the cosine case.

\includegraphics[width=\textwidth]{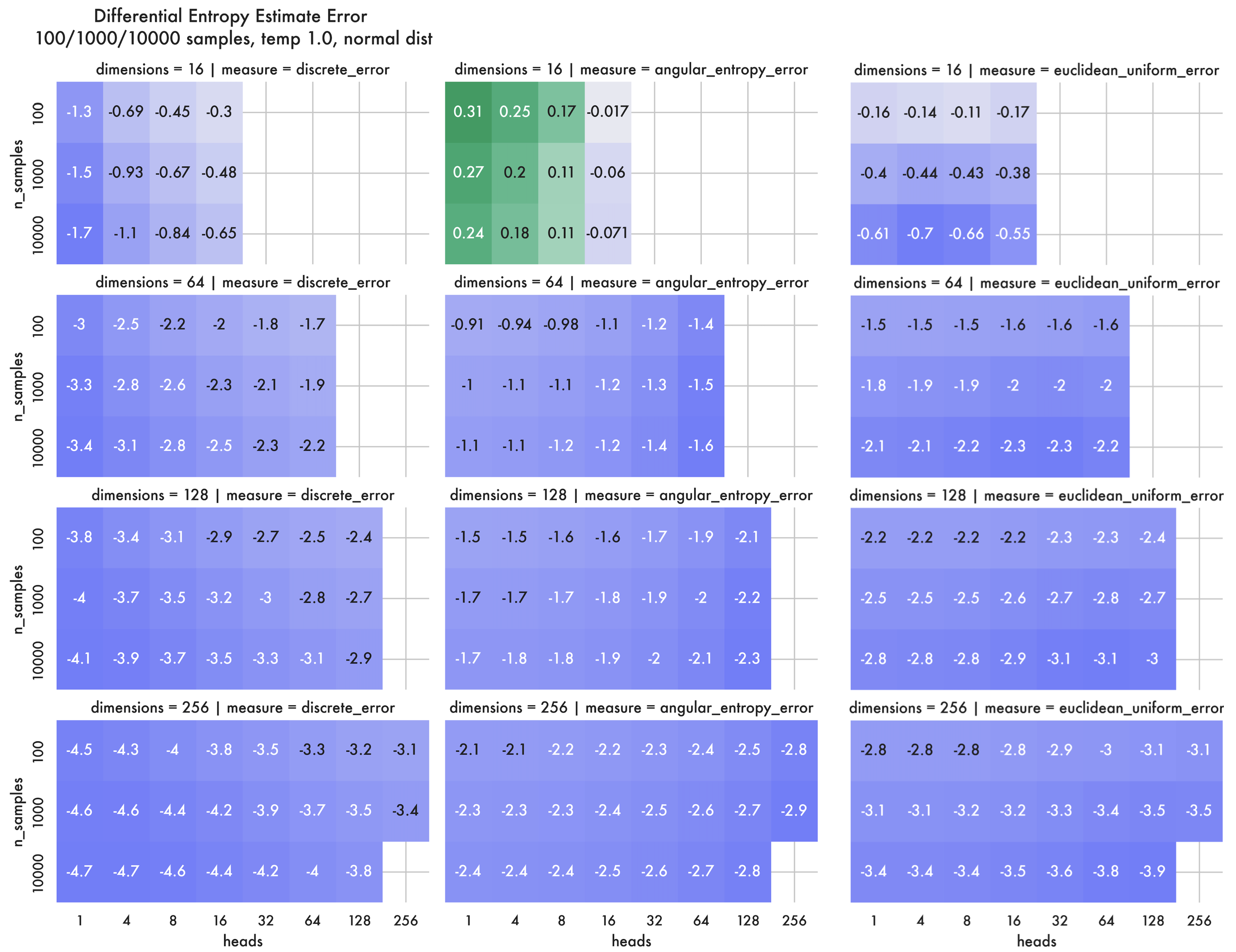}

\includegraphics[width=\textwidth]{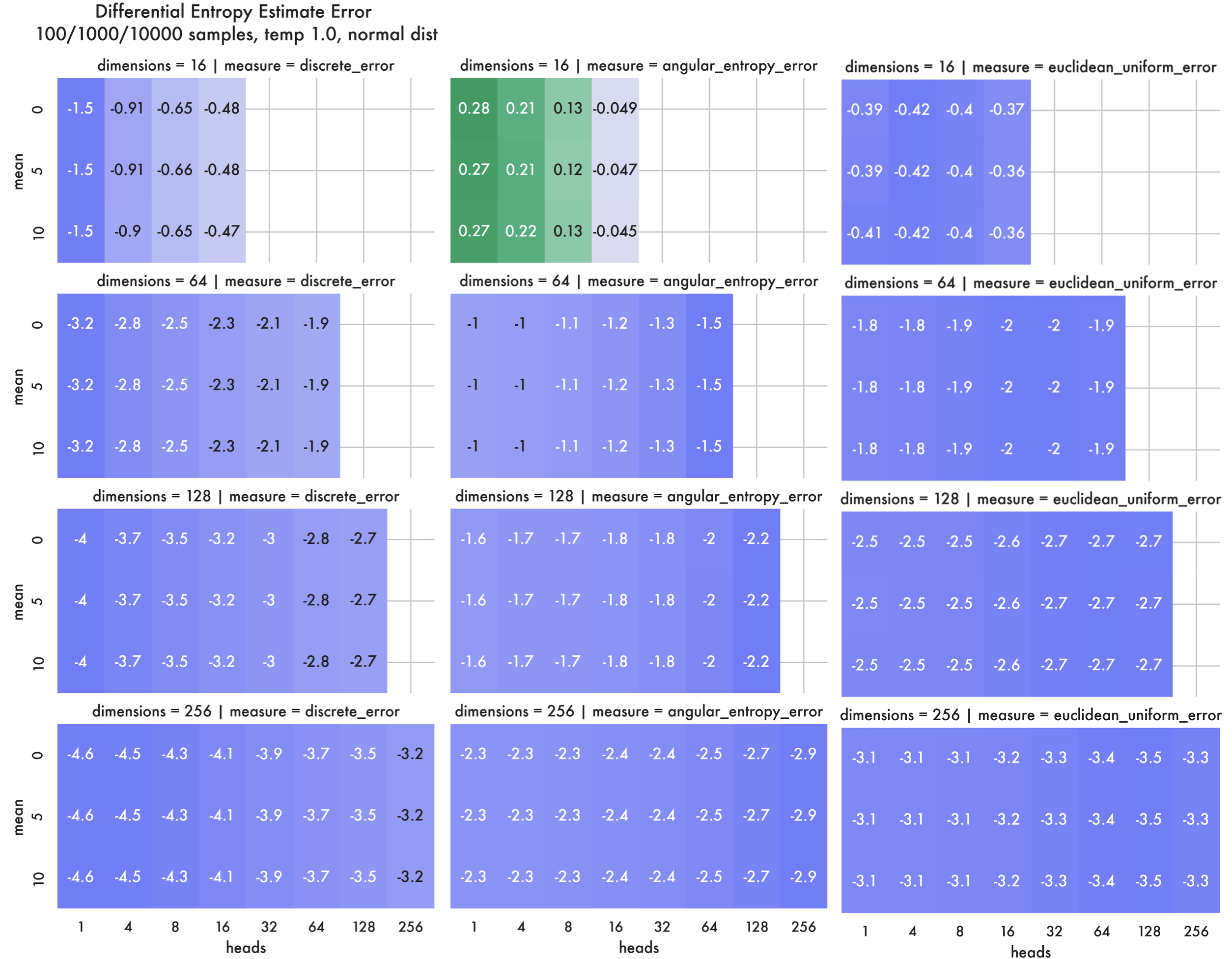}

\includegraphics[width=\textwidth]{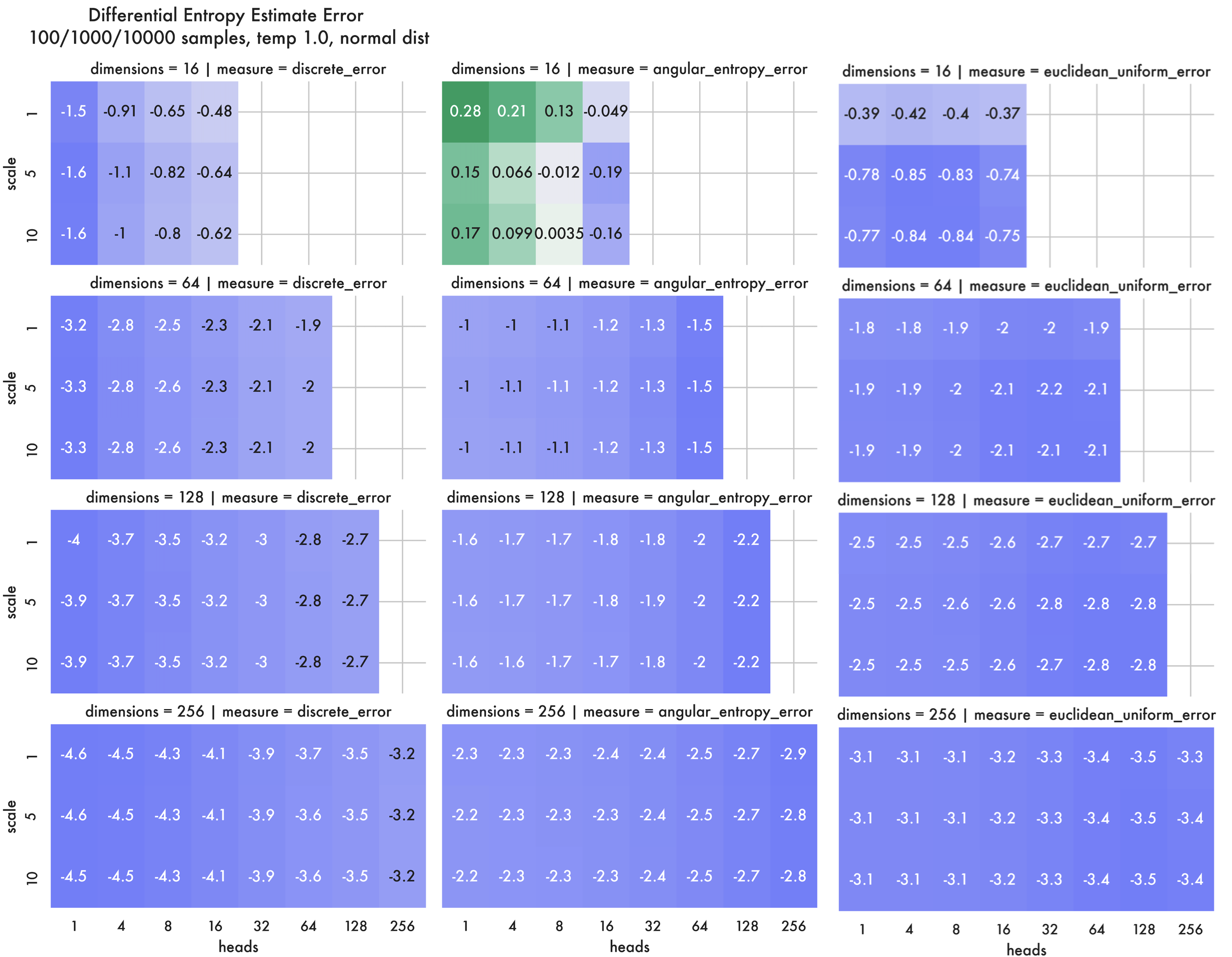}

\section{All GLUE Correlations by Task}\label{sec:appendix_glue_correlations}

There are a huge volume of correlations for which I apologise

\begin{figure}[hp]
\includegraphics[width=\textwidth]{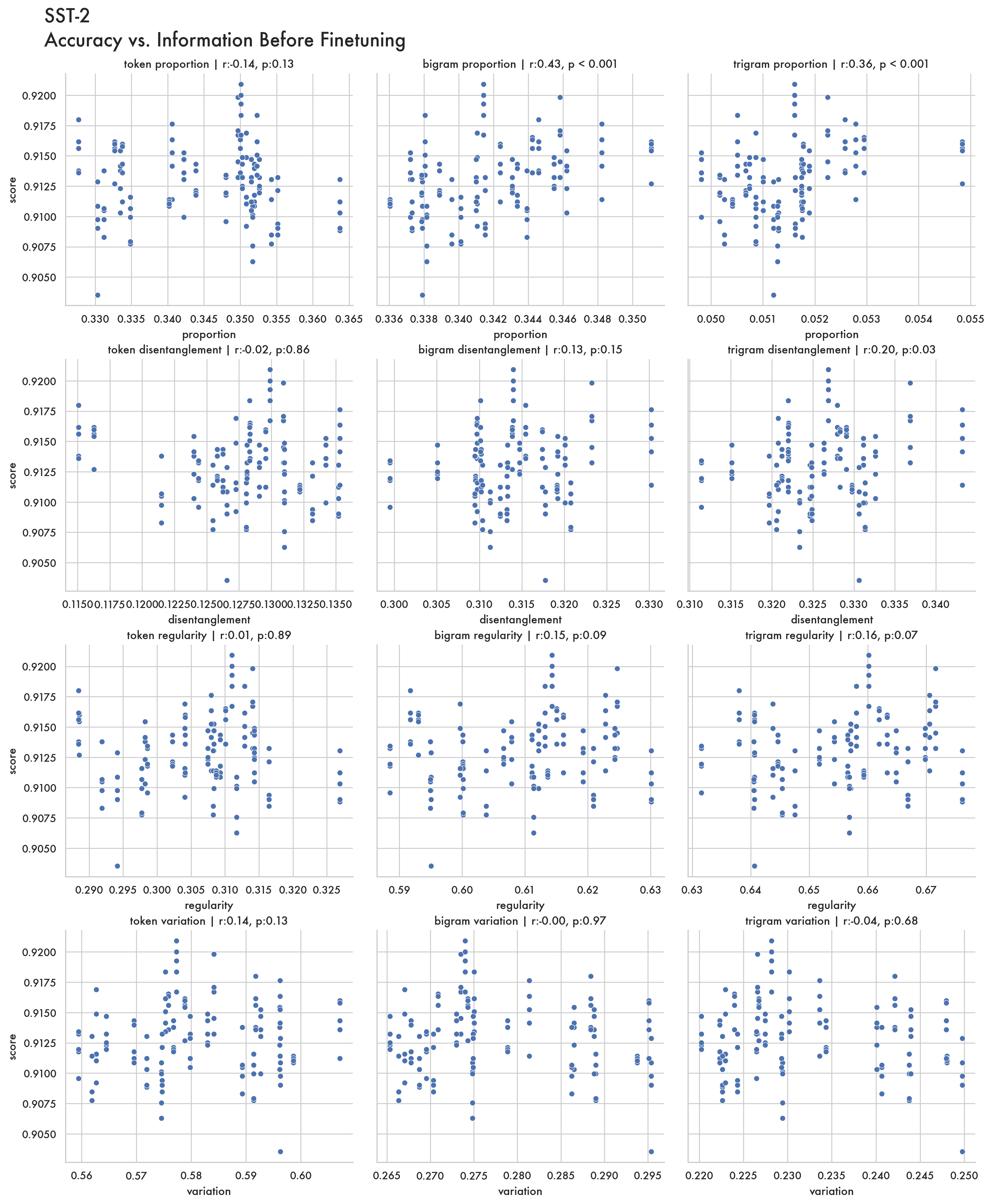}
\end{figure}

\pagebreak
\begin{figure}[hp]
\includegraphics[width=\textwidth]{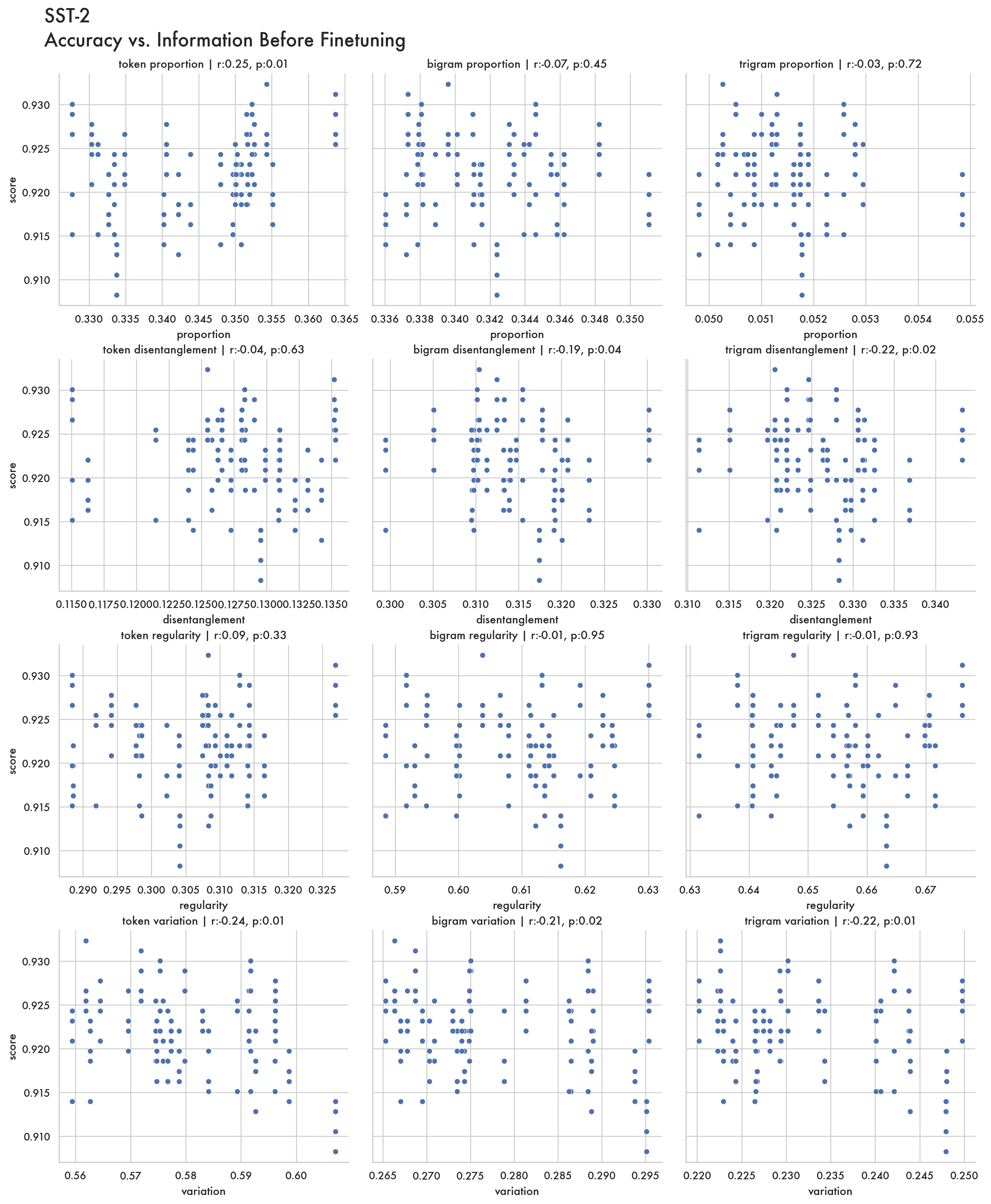}
\end{figure}

\pagebreak
\begin{figure}[hp]
\includegraphics[width=\textwidth]{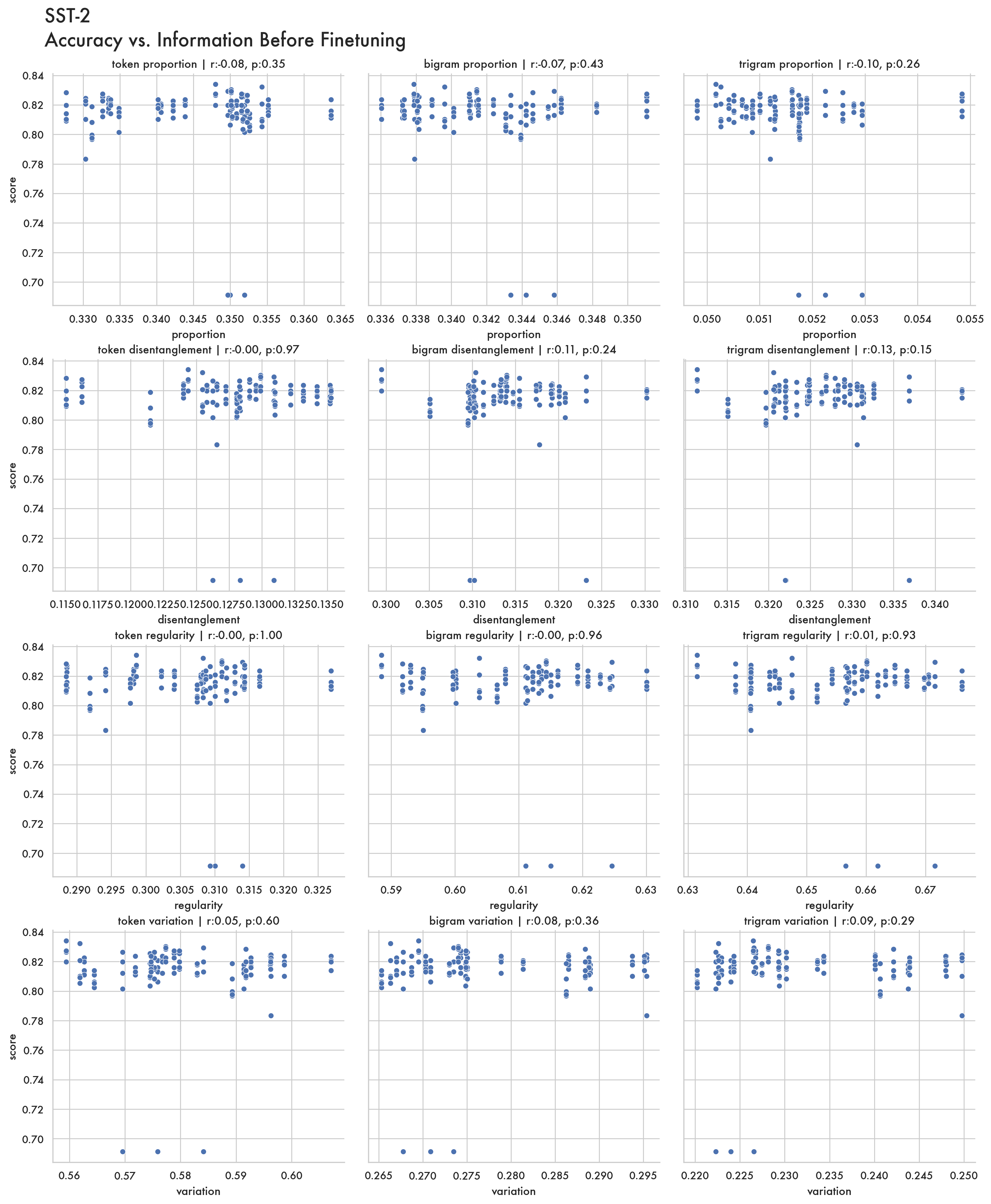}
\end{figure}

\pagebreak

\begin{figure}[hp]
\includegraphics[width=\textwidth]{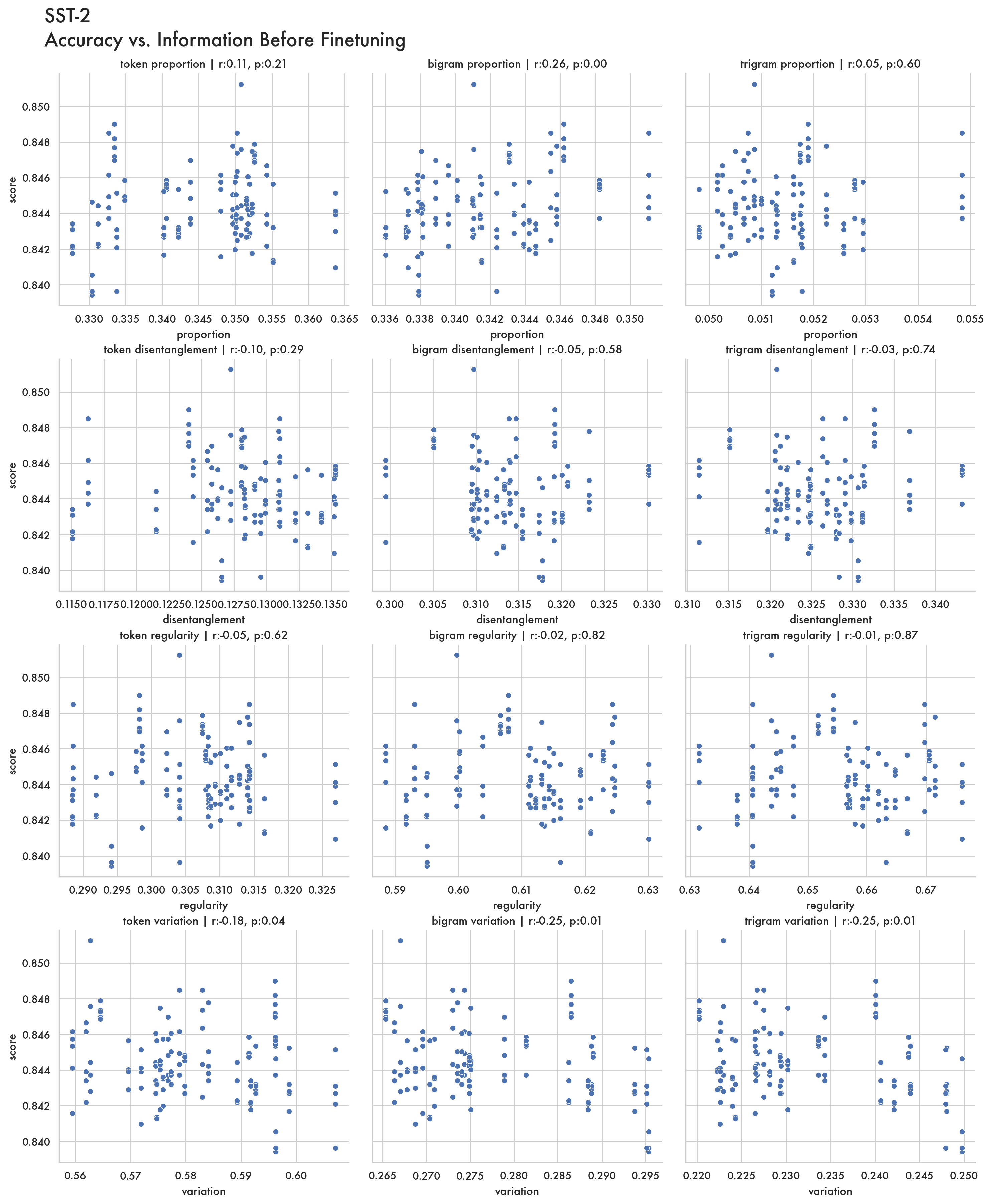}
\end{figure}

\pagebreak

\begin{figure}[hp]
\includegraphics[width=\textwidth]{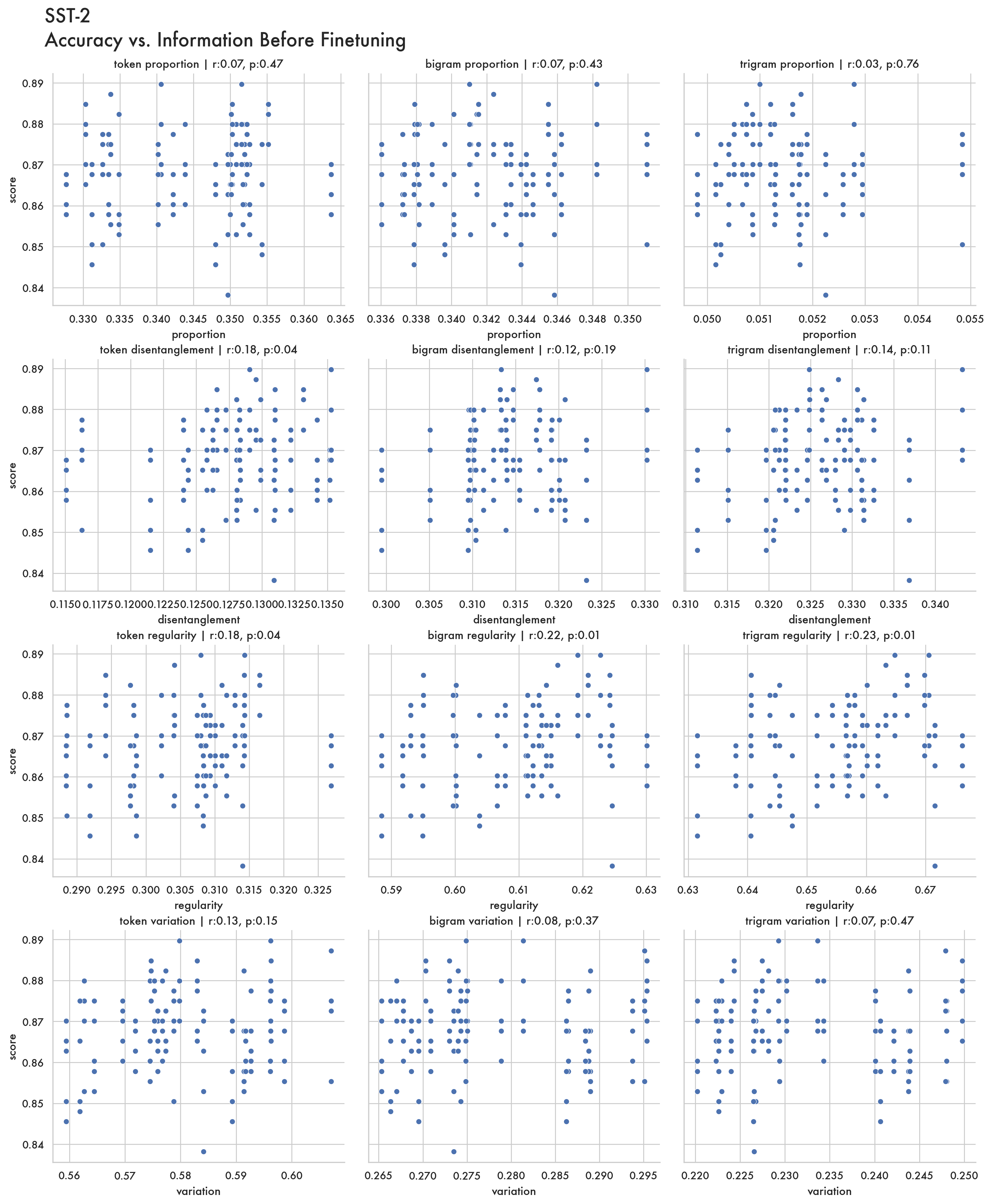}
\end{figure}

\pagebreak

\begin{figure}[hp]
\includegraphics[width=\textwidth]{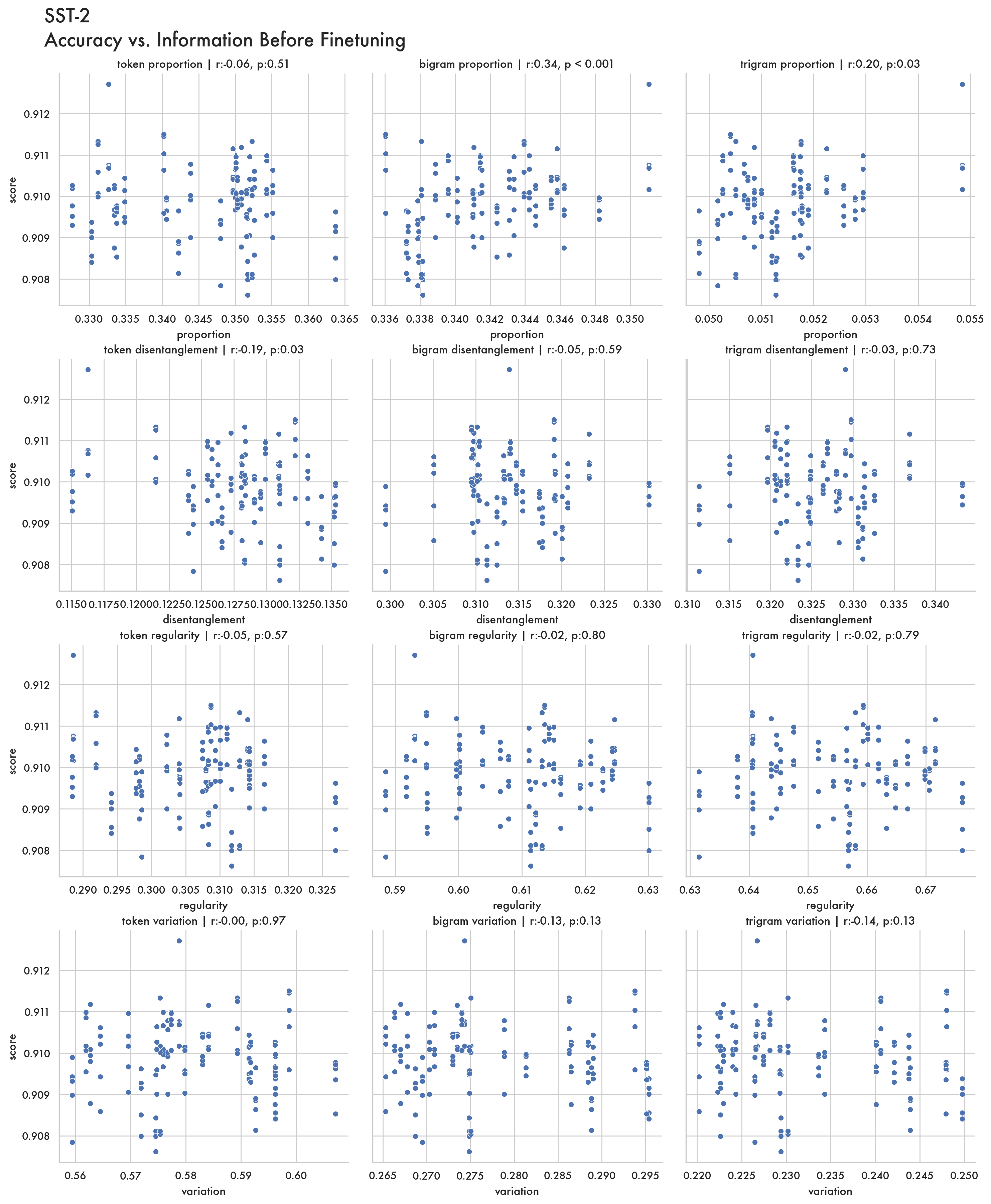}
\end{figure}

\pagebreak

\begin{figure}[hp]
\includegraphics[width=\textwidth]{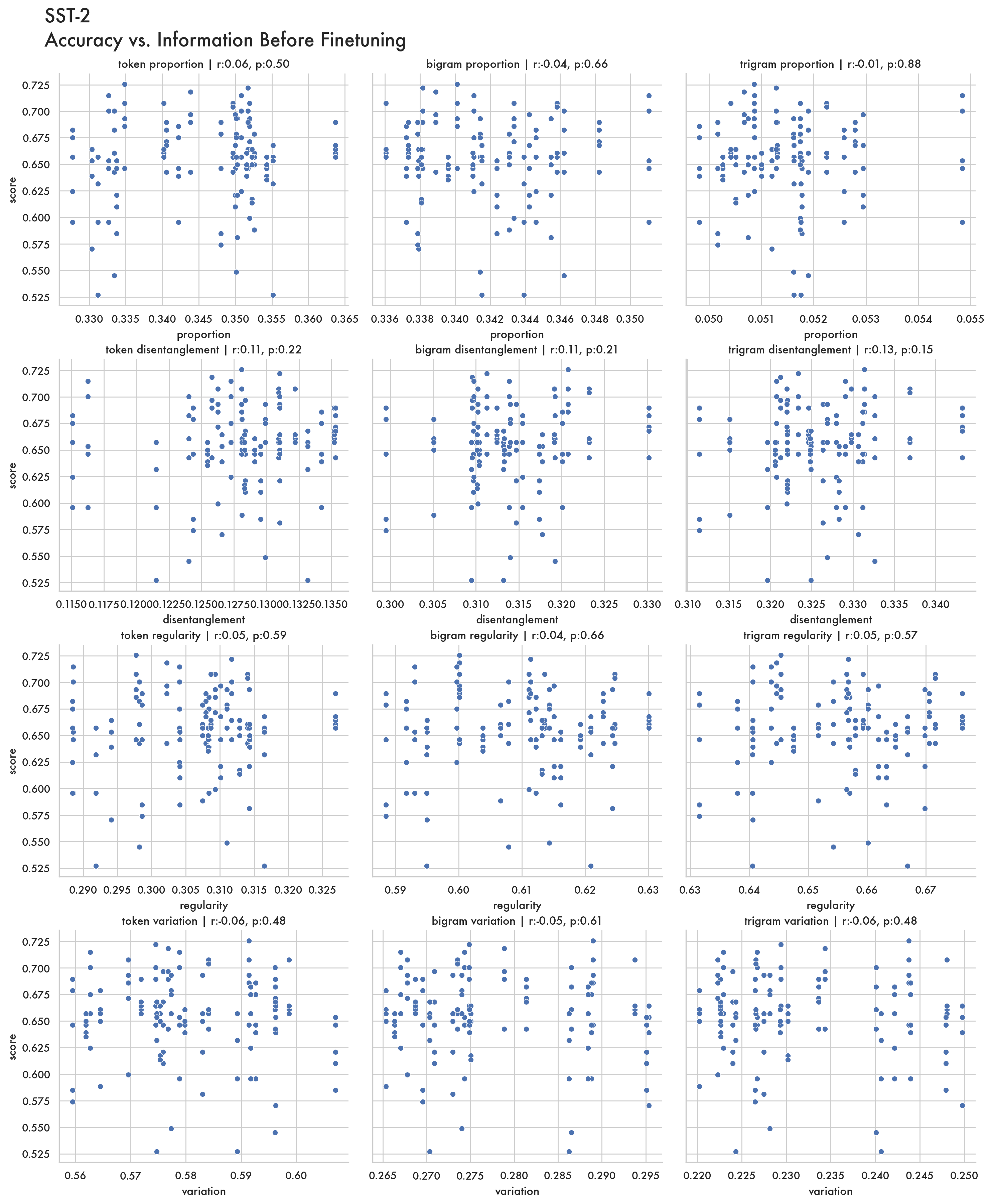}
\end{figure}

\pagebreak

\begin{figure}[hp]
\includegraphics[width=\textwidth]{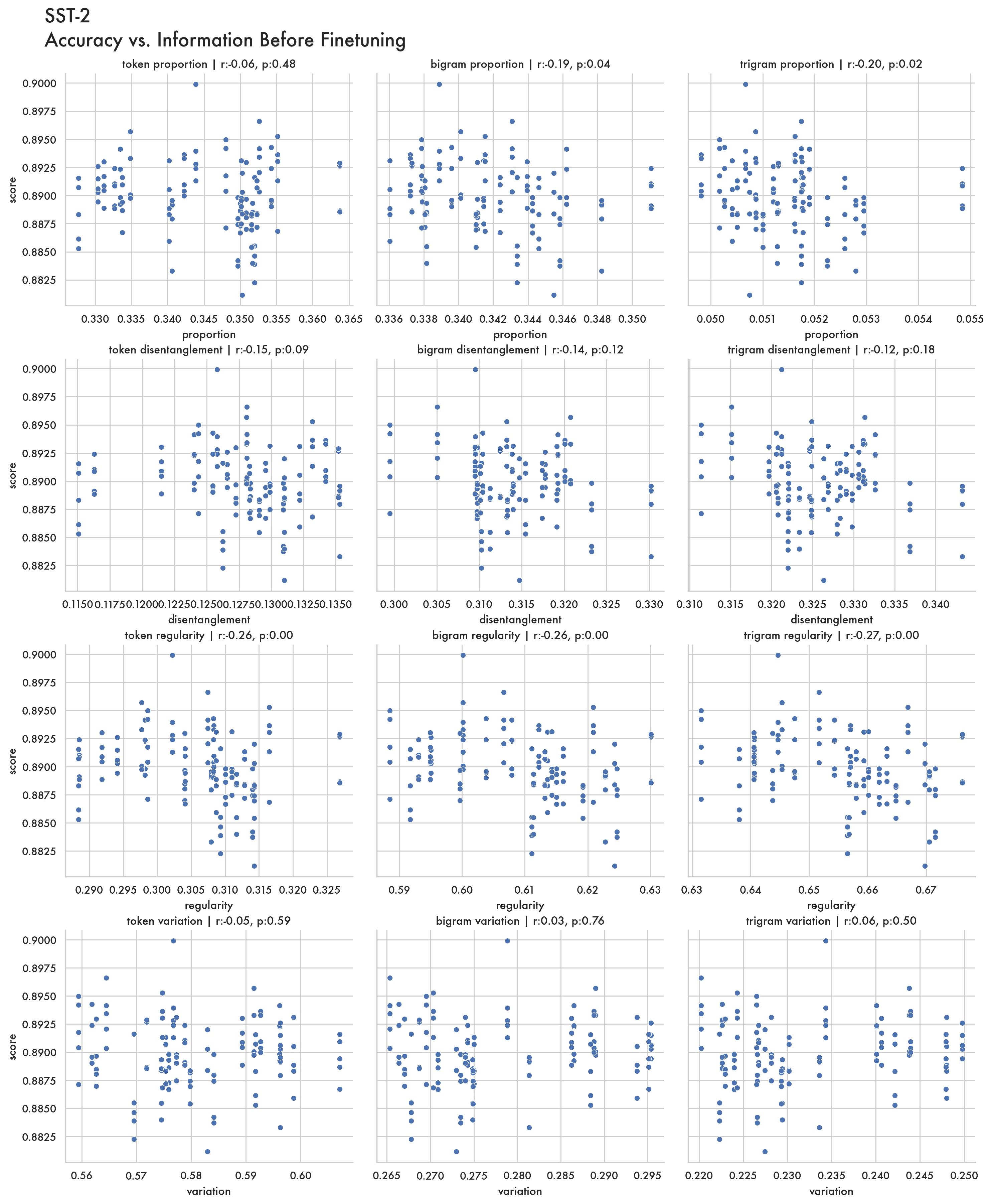}
\end{figure}

\pagebreak

\begin{figure}[hp]
\includegraphics[width=\textwidth]{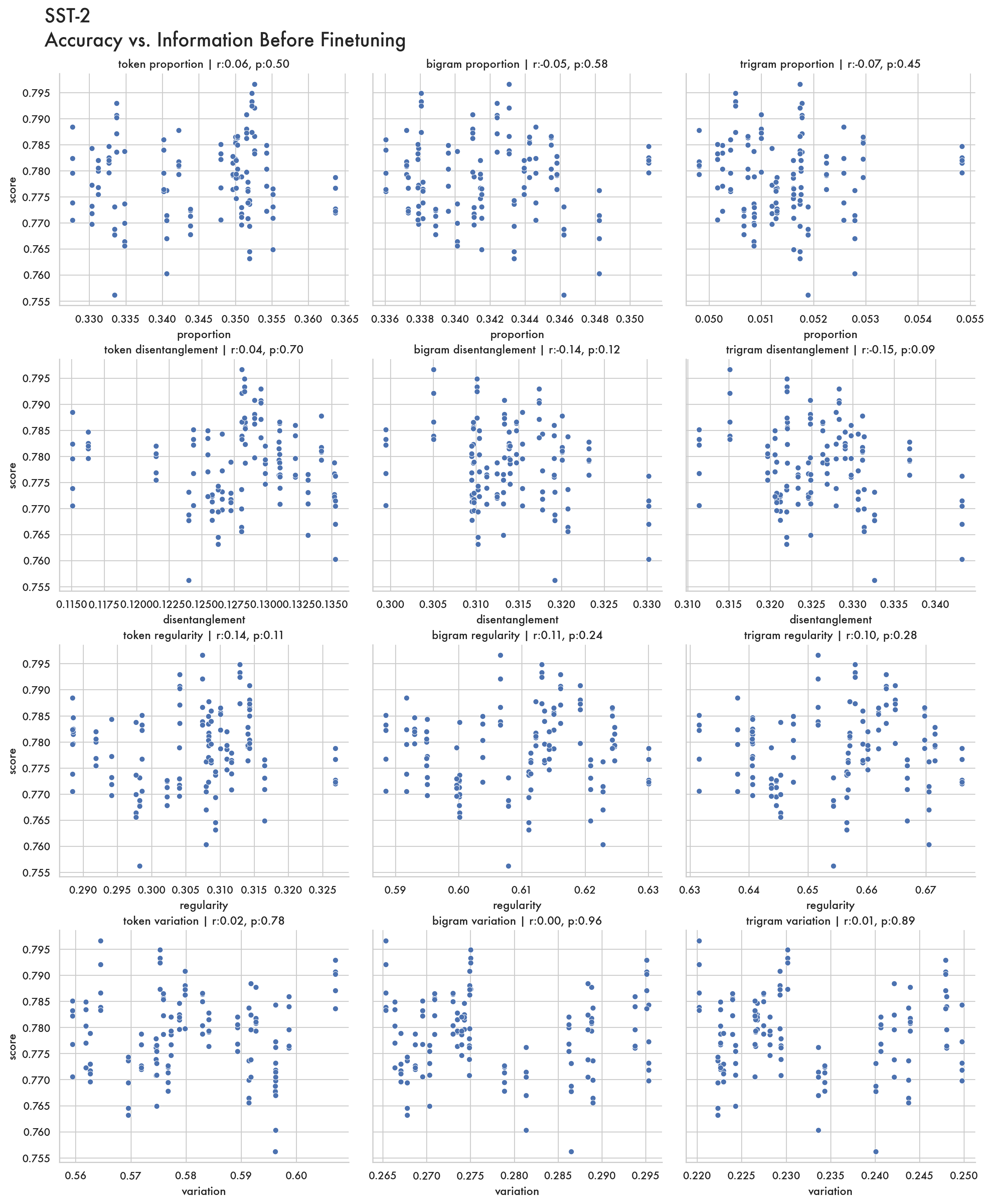}
\end{figure}
\pagebreak

\chapter{Meta-Learning To Compositionally Generalise}
\section{Details of Base Parsers}

We implemented all models with Pytorch~\cite{paszke2019pytorch}. 
For the LSTM parsers, we use a two-layer encoder and one-layer decoder with attention~\cite{bahdanau2014neural} and input-feeding~\cite{luong2015effective}.
We only test bidirectional LSTM encoders, as unidirectional LSTM models do not perform very well in our preliminary experiments. 
For Transformer parsers, we use 2 encoder and decoder layers, 4 attention heads, and a feed-foward dimension of 1024. The hidden size for both LSTM and Transformer models are 256. 
The hyparameters of base parsers are mostly 
borrowed from related work and not tuned, as the primary goal of this work is the MAML training algorithm.
To experiment with a wide variety of possible Seq2Seq models, we also try a Transformer encoder + LSTM decoder and find that this variant actually performs slightly better than both vanilla Transformer and LSTM models.
Further exploration of this combination in pursuit of a better neural architecture for compositional generalization might be interesting for future work.

\section{Details of Sampling for Meta-Test}

The similarity-driven sampling distribution $\tilde p$ (~Equation 4 of the main paper) requires computing the similarity between every pair of training examples, which can be very expensive depending on the size of of the dataset. As the sampling distributions are fixed during training, we compute and cache them beforehand. However, they take an excess of disk space to store as essentially we need to store an $N \times N$ matrix where $N$ is the number of training examples. To allow efficient storage and sampling, we use the following approximation.
First, we found that usually each example only has a small set of neighbours that are relevant to it. For example, in COGS, each example only retrieves 3.6\% of the whole training set as its neighbours (i.e., have non-zero tree-kernel similarity) on average. Motivated by this observation, we only store the top 1000 relevant neighbours for each example sorted by similarity, and use it to construct the sampling distribution denoted as $\tilde p_{top1000}$. To allow examples out of top 1000 being sampled, we use a linear interpolation between $\tilde p_{top1000}$ and a uniform distribution. Specifically, we end up using the following sampling distribution:
\begin{equation}
  \tilde p(x', y' | x, y) = \lambda \ \tilde  p_{top1000} ( x', y' | x, y ) + (1 - \lambda) \frac{1}{N}
\end{equation}
where $\tilde p_{top1000}$ assigns 0 probability to out-of top-1000 examples, $N$ is the number of training examples, and $\lambda$ is a hyperparameter for interpolation. 
In practice, we set $\lambda$ to $0.5$ in all experiments.

To sample from this distribution, we first decide whether the sample is in the top 1000 by sampling from a Bernoulli distribution parameterized by $\lambda$. If it is, we use $\tilde p_{top1000}$ to do the sampling; if not, we just uniformly sample an example from the training set.

\section{Model Selection Protocol}

In our preliminary experiments on COGS, we find almost all the Seq2Seq models achieve $> 99\%$ in accuracy on the original Dev set. However, their performance on the Gen set diverge dramatically, ranging from $10\%$ to $70\%$. The lack of an informative Dev set makes model selection extremely difficult and difficult to reproduce. This issue might also be one of the factors that results in the large variance of performance reported in previous work.
Meanwhile, we found that some random seeds~\footnote{Random seeds control the initialization of parameters and the order of training batches.} yield consistently better performance than others across different conditions. For example, among the ten random seeds used for Lev-MAML + Transformer on COGS, the best performing seed obtains $73\%$ whereas the lowest performing seed obtains $54\%$.  Thus, it is important to compare different models using the same set of random seeds, and not to tune the random seeds in any model. To alleviate these two concerns, we choose the protocol that is mentioned in the main paper. This protocol helps to make the results reported in our paper reproducible. 

\section{Details of Training and Evaluation}

Following~\cite{kim_cogs_2020},
we train all models from scratch using randomly initialized embeddings.
For SCAN, models are trained for 1,000 steps with batch size 128.
We choose model checkpoints based on their performance on the Dev set.
For COGS, models are trained for 6,000 steps with batch size of 128. 
We choose the meta-train learning rate ($\alpha$ in Equation 2 of the main paper), temperature ($\eta$ in Equation 4 of the main paper) based on the performance on the Gen Dev set. Finally we use the chosen $\alpha$, $\eta$ to train models with new random seeds, and only the last checkpoints (at step 6,000) are used for evaluation on the Test and Gen set.

\section{Other Splits of SCAN}
The SCAN dataset contains many splits, such as Add-Jump, Around Right, and Length split, each assessing a particular case of compositional generalization. We think that MCD splits are more representative of compositional generalization due to the nature of the principle of maximum compound divergence. Moreover, it is more challenging than other splits (except the Length split) according to \citet{furrer_compositional_2021}. That GECA, which obtains $82\%$ in accuracy on JUMP and Around Right splits, only obtains $< 52\%$ in accuracy on MCD splits in our experiments confirms that MCD splits are more challenging.

\section{Kernel Analysis}
 The primary difference between the tree-kernel and string-kernel methods is in the diversity of the examples they select for the meta-test task. The tree kernel selects a broader range of lengths, often including atomic examples, a single word in length, matching a word in the original example from meta-train (see table~\ref{table:discussion-kernels}). By design the partial tree kernel will always assign a non-zero value to an example that is an atom contained in the original sentence. We believe the diversity of the sentences selected by the tree kernel accounts for the superior performance of Tree-MAML compared with the other MAML conditions. The selection of a variety of lengths for meta-test constrains model updates on the meta-train task such that they must also accommodate the diverse and often atomic examples selected for meta-test. This constraint would seem to better inhibit memorizing large spans of the input unlikely to be present in meta-test. 
 
 \begin{table*}
\centering
\resizebox{\textwidth}{!}{%
\begin{tabular}{ cc } 
    \begin{tabular}{lrrr}
    \textbf{Partial Tree Kernel}             &  \textbf{top 10} & \textbf{100} & \textbf{1000}  \\
    \toprule
    Mean Example Length (chars) & 26.71 & 26.59 & 29.87 \\
    Std dev & $\pm$ 6.80 & $\pm$ 7.61  & $\pm$ 8.85 \\
    \hline
    Mean No. of Atoms & 0.46 & 0.81 & 1.13\\
    Std dev & $\pm$ 0.67 & $\pm$ 1.05 & $\pm$ 0.81\\
    \bottomrule
    \end{tabular}

    \begin{tabular}{lrrr}
    \textbf{LevDistance}             &  \textbf{top 10} & \textbf{100} & \textbf{1000}  \\
    \toprule
    Mean Example Length (chars) & 31.04 &  30.45  & 29.28 \\
    Std dev & $\pm$ 2.80 &  $\pm$ 3.77 & $\pm$ 4.78 \\
    \hline
    Mean No. of Atoms & 0.00 & 0.00 & 0.02\\
    Std dev & $\pm$ 0.00 & $\pm$ 0.02 & $\pm$ 0.17\\
    \bottomrule
    \end{tabular}

\end{tabular}
}
\caption{Analyses of kernel diversity. Reporting mean example length and number of atoms for the top k highest scoring examples for each kernel. Note that atoms are only counted that also occur in the original example.}
\label{table:discussion-kernels}
\end{table*}

\section{Meta-Test Examples}
In Table~\ref{table:kernel-examples-appendix}, we show top scoring examples retrieved by the similarity metrics for two sentences. We found that in some cases (e.g., the right part of Table~\ref{table:kernel-examples-appendix}), the tree-kernel can retrieve examples that diverge in length but are still semantically relevant. In contrast, string-based similarity metrics, especially LevDistance, tends to choose examples with similar lengths.

\begin{table*}[t!]
  \centering
    \resizebox{\textwidth}{!}{ 
\begin{tabular}{cc}
      \begin{tabular}{lr}
    \multicolumn{2}{c}{\textbf{Source Example}: Emma lended the donut to the dog .} \\
      \\
      \toprule
      \emph{Neighbours using Tree Kernel} & Similarity\\
      \hline
Emma was lended the donut . & 0.74 \\
The donut was lended to Emma . & 0.62 \\
Emma lended the donut to a dog . & 0.55 \\
Emma lended Liam the donut . & 0.55 \\
Emma lended a girl the donut . & 0.55 \\
       \\
        \emph{Neighbours using String Kernel}\\
       \hline
Emma lended the donut to a dog . & 0.61 \\
Emma lended the box to a dog . & 0.36 \\
Emma gave the cake to the dog . & 0.33 \\
Emma lended the cake to the girl . & 0.33 \\
Emma lended the liver to the girl . & 0.33 \\
      \\
      \emph{Neighbours using LevDistance} & \\
      \hline
Emma lended the donut to a dog . & -1.00 \\
Emma loaned the donut to the teacher . & -2.00 \\
Emma forwarded the donut to the monster . & -2.00 \\
Emma gave the cake to the dog . & -2.00 \\
Charlotte lended the donut to the fish . & -2.00 \\
      \bottomrule
      \end{tabular}
    &
      \begin{tabular}{lr}
    \multicolumn{2}{c}{\textbf{Source Example}:  The crocodile valued that a girl snapped .} \\
      \\
      \toprule
      \emph{Neighbours using Tree Kernel} & Similarity\\
      \hline
A girl snapped . & 0.55 \\
A rose was snapped by a girl . & 0.39 \\
The cookie was snapped by a girl . & 0.39 \\
girl & 0.32 \\
value & 0.32 \\
       \\
        \emph{Neighbours using String Kernel}\\
       \hline
The crocodile liked a girl . & 0.28 \\
The girl snapped . & 0.27 \\
The crocodile hoped that a boy observed a girl . & 0.26 \\
The boy hoped that a girl juggled . & 0.15 \\
The cat hoped that a girl sketched . & 0.15 \\
      \\
      \emph{Neighbours using LevDistance} & \\
      \hline
The crocodile liked a girl . & -3.00 \\
The boy hoped that a girl juggled . & -3.00 \\
The cat hoped that a girl sketched . & -3.00 \\
The cat hoped that a girl smiled . & -3.00 \\
Emma liked that a girl saw . & -4.00 \\
      \bottomrule
      \end{tabular}
\end{tabular}
  }
  \caption{Top scoring examples according to the tree kernel, string kernel and Levenshtein distance for two sentences and accompanying scores.}
  \label{table:kernel-examples-appendix}
  \end{table*}

\section{COGS Subtask Analysis}

\begin{figure}
    \centering
\includegraphics[scale=0.85]{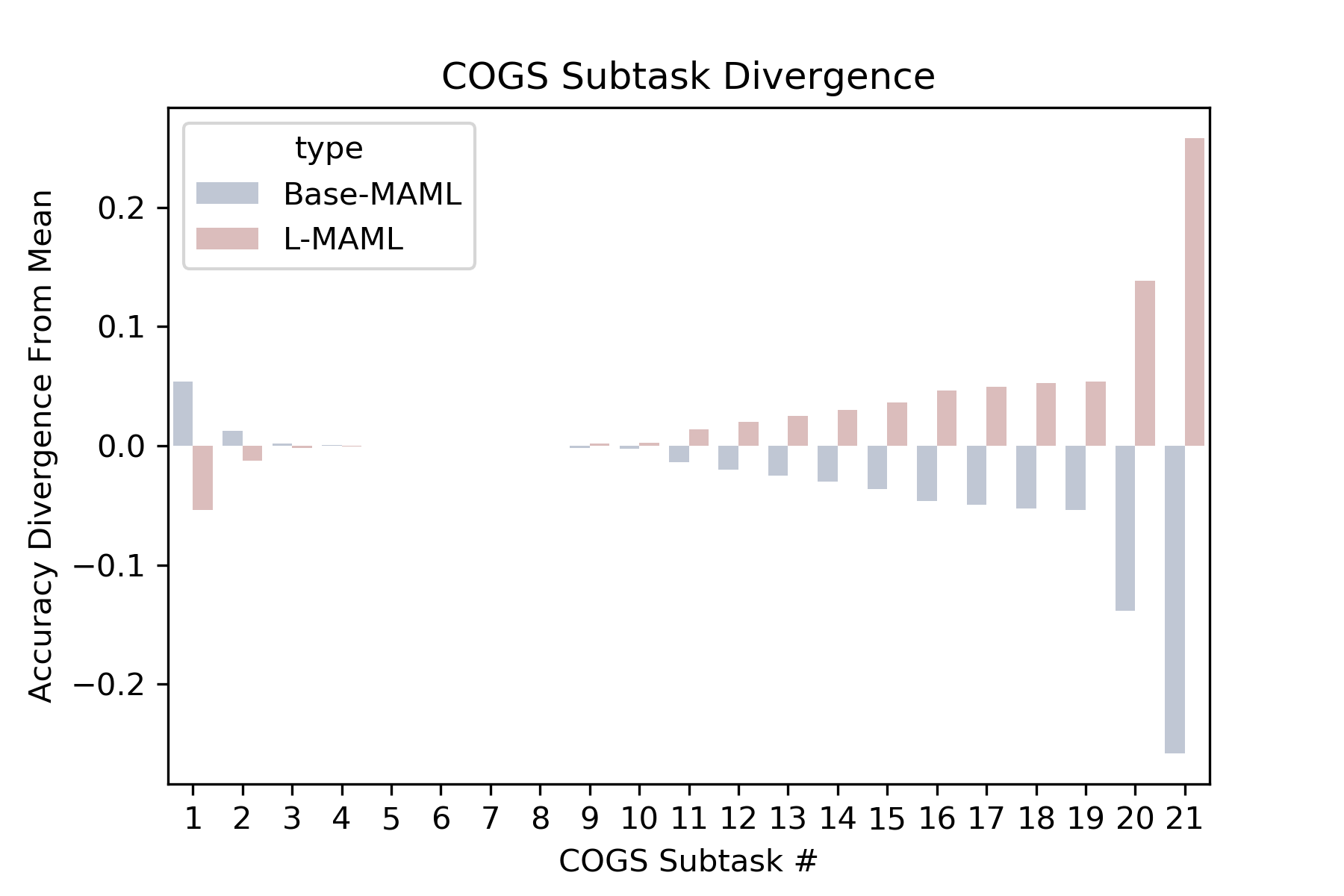}
    \caption{Performance for the Uni-MAML and Lev-MAML conditions compared to the mean of those two conditions. \\
    (1) prim$\rightarrow$subj proper, \\
(2) active$\rightarrow$passive,\\
(3) only seen as unacc subj $\rightarrow$ unerg subj,\\
(4) subj$\rightarrow$obj proper,\\
(5) only seen as unacc subj $\rightarrow$ obj omitted transitive subj,\\
(6) pp recursion,\\
(7) cp recursion,\\
(8) obj pp$\rightarrow$subj pp,\\
(9) obj$\rightarrow$subj common,\\
(10) do dative$\rightarrow$pp dative,\\
(11) passive$\rightarrow$active,\\
(12) only seen as transitive subj $\rightarrow$ unacc subj,\\
(13) obj omitted transitive$\rightarrow$transitive, \\
(14) subj$\rightarrow$obj common, \\
(15) prim$\rightarrow$obj proper, \\
(16) obj$\rightarrow$subj proper, \\
(17) pp dative$\rightarrow$do dative, \\
(18) unacc$\rightarrow$transitive, \\
(19) prim$\rightarrow$subj common, \\
(20) prim$\rightarrow$obj common, \\
(21) prim$\rightarrow$inf arg.} 
    \label{fig:Cogssubtask}
\end{figure}

We notice distinct performance for different conditions on the different subtasks from the COGS dataset. In Figure \ref{fig:Cogssubtask} we show the performance of the Uni-MAML and Str-MAML conditions compared with the mean of those conditions. Where the bars are equal to zero the models' performance on that task is roughly equal.

\end{document}